%% file: neurips_2025.tex
\documentclass{article}

    \PassOptionsToPackage{numbers, compress}{natbib}

 \usepackage[dandb, final]{neurips_2025}

\usepackage{amssymb}
\usepackage{amsmath}
\usepackage{pifont}
\usepackage{xspace}
\usepackage{enumitem}
\usepackage{adjustbox}
\newcommand{\benchmark}{CXReasonBench\xspace}
\newcommand{\extractor}{CheXStruct\xspace}
\newcommand{\revision}[1]{\textcolor{black}{#1}}

\usepackage[utf8]{inputenc} 
\usepackage[T1]{fontenc}    
\usepackage{hyperref}       
\usepackage{url}            
\usepackage{booktabs}       
\usepackage{amsfonts}       
\usepackage{nicefrac}       
\usepackage{microtype}      
\usepackage{xcolor}         

\title{\benchmark: A Benchmark for Evaluating Structured Diagnostic Reasoning in Chest X-rays}

\author{
    Hyungyung Lee$^{1}$,
    Geon Choi$^{1}$,
    Jung-Oh Lee$^{2}$, \\
    \textbf{Hangyul Yoon$^{1}$},
    \textbf{Hyuk Gi Hong$^{3}$},
    \textbf{Edward Choi$^{1}$} \\
    $^{1}$KAIST \;
    $^{2}$Seoul National University Hospital\;
    $^{3}$Seoul Medical Center \; \\
    \texttt{\{ttumyche, edwardchoi\}@kaist.ac.kr}}

\begin{document}

\maketitle

\begin{abstract}
\input{MainSections/1.Abstract}
\end{abstract}

\section{Introduction}\label{section:introduction}
\input{MainSections/2.Introduction}

\section{Related Works}
\input{MainSections/3.RelatedWorks}

\input{MainSections/4.CheXStruct}\label{section:chexstruct}

\input{MainSections/5.CXReasonBench}

\section{Experiments}
\input{MainSections/6.Experiments}

\section{Discussion}\label{section:discussion}
\input{MainSections/7.Discussion}

\section*{Acknowledgments and Disclosure of Funding}
This work was supported by Microsoft Research Asia, the Institute of Information \& Communications Technology Planning \& Evaluation (IITP) grants (No.RS-2019-II190075, No.RS-2024-00436680), the Korea Health Industry Development Institute (KHIDI) grant (No.RS-2025-02213750), the National Research Foundation of Korea (NRF) grants (NRF-2020H1D3A2A03100945), the Medical Scientist Training Program funded by the Korea government (MSIT, MOHW).


\newpage
\bibliographystyle{plain}
\bibliography{neurips_2025}






\newpage

\appendix
\section*{Supplementary Material}

\input{Supple/1.CheXStruct_Extractor}

\input{Supple/1.CheXStruct_QC}

\input{Supple/2.CXReaonBench}

\input{Supple/3.Experiments}

\input{Supple/4.Results}

\end{document}

%% file: MainSections/1.Abstract.tex
Recent progress in Large Vision-Language Models (LVLMs) has enabled promising applications in medical tasks, such as report generation and visual question answering. However, existing benchmarks focus mainly on the final diagnostic answer, offering limited insight into whether models engage in clinically meaningful reasoning. To address this, we present \textbf{CheXStruct} and \textbf{CXReasonBench}, a structured pipeline and benchmark built on the publicly available MIMIC-CXR-JPG dataset. CheXStruct automatically derives a sequence of intermediate reasoning steps directly from chest X-rays, such as segmenting anatomical regions, deriving anatomical landmarks and diagnostic measurements, computing diagnostic indices, and applying clinical thresholds. CXReasonBench leverages this pipeline to evaluate whether models can perform clinically valid reasoning steps and to what extent they can learn from structured guidance, enabling fine-grained and transparent assessment of diagnostic reasoning.
The benchmark comprises 18,988 QA pairs across 12 diagnostic tasks and 1,200 cases, each paired with up to 4 visual inputs, and supports multi-path, multi-stage evaluation including visual grounding via anatomical region selection and diagnostic measurements.
Even the strongest of 12 evaluated LVLMs struggle with structured reasoning and generalization, often failing to link abstract knowledge with anatomically grounded visual interpretation.
The code is available at \texttt{https://github.com/ttumyche/CXReasonBench}


%% file: MainSections/2.Introduction.tex
Large Vision-Language Models (LVLMs) have recently been applied to medical tasks such as report generation and visual question answering (VQA) ~\cite{chen2024chexagent, deperrois2025radvlm, lin2025healthgpt}.
To evaluate their performance, chest X-rays have become a standard benchmark due to their clinical relevance and accessibility.

However, existing benchmarks~\cite{ben2021overview, he2020pathvqa, lau2018dataset, zhang2023pmc} primarily assess the correctness of the final diagnostic answer, offering limited insight into the underlying reasoning process, a critical omission in clinical decision-making.
Some recent efforts~\cite{bae2023ehrxqa, hu2023expert, liu2024gemex, zuo2025medxpertqa} incorporate explanations or visual grounding, but these still more emphasize outputs than the intermediate steps that lead to them.
For example, when asked \textit{``Which region shows abnormalities?''}, a model might respond \textit{``Cardiac region''} and add \textit{``The heart appears enlarged.''}
While the answer may seem plausible, it remains unclear whether the model truly recognized relevant anatomical structures, performed appropriate measurements, and applied clinical rules such as the cardiothoracic ratio~\cite{truszkiewicz2021radiological}, all of which are key components of trustworthy diagnostic reasoning.
Without evaluating these intermediate steps, we cannot tell whether the model truly understands the image or relies on superficial patterns.

To bridge this critical gap, we present \textit{\extractor} and \textit{\benchmark}, two complementary contributions designed to evaluate the intermediate steps of diagnostic reasoning from chest X-rays.
\extractor (Figure~\ref{figure:overview_chexstruct}) is a fully automated pipeline that extracts clinically relevant reasoning steps directly from chest X-ray images.
While prior structured datasets~\cite{bannur2024maira, castro2024padchest, wu2021chest} provide bounding box annotations linking report-derived findings to image regions, they primarily focus on high-level findings and offer limited supervision for intermediate reasoning steps.
\extractor addresses this limitation by systematically modeling the full reasoning process: it begins with anatomical segmentation, then derives diagnostic measurements (\textit{e.g.}, cardiac width, thoracic width), computes diagnostic indices (\textit{e.g.}, cardiothoracic ratio), and applies clinical thresholds, following expert-defined guidelines.
Recognizing the importance of the quality of the anatomical segmentation, we implemented task-specific quality control (QC) processes to ensure that the measurements and indices derived from the segmented regions are clinically reliable and consistent with expert-defined criteria.
Images that fail to meet quality thresholds are automatically filtered out, ensuring that only reliable data is used for evaluation.
\extractor performs the entire pipeline, segmentation, measurement, QC filtering, and structured information extraction automatically, producing multi-step reference answers for evaluating a model’s diagnostic reasoning.

Building on this structured pipeline, we introduce \benchmark (Figure~\ref{figure:overview_cxreasonbench}), a novel benchmark designed to evaluate whether a model’s diagnostic reasoning process aligns with clinically grounded steps, not just diagnostic decisions.
It compares model responses to the reference answers generated by \extractor across multiple intermediate stages, such as selecting diagnostic criteria, identifying anatomical structures, interpreting anatomical changes, computing quantitative measurements, and applying clinical thresholds, enabling fine-grained analysis of diagnostic reasoning.
\benchmark includes two complementary evaluation paths.
Path 1, \textit{Direct Reasoning Process Evaluation}, evaluates a model’s ability to reconstruct its diagnostic reasoning through multiple intermediate stages and adhere to standardized or expert-defined criteria.
Path 2, \textit{Guided Reasoning and Re-evaluation}, tests whether a model can internalize and apply structured diagnostic reasoning when provided with medically accurate intermediate steps.
These two paths support robust and transparent assessment of model reasoning, facilitate detailed failure analysis, and ensure that performance is consistent with genuine clinical understanding, not merely pattern-based shortcuts.

Built using the MIMIC-CXR-JPG dataset~\cite{johnson2019mimic}, our benchmark includes 18,988 QA pairs covering 12 diagnostic tasks and 1,200 cases, supporting multi-path, multi-stage evaluation, including visual grounding through anatomical region selection and diagnostic measurements.
Each QA pair is accompanied by up to 4 visual inputs (\textit{e.g.}, the original X-ray and overlaid segmentation masks).
Based on this benchmark, in our evaluation of 12 LVLMs, we find that even the most capable models struggle with valid structured diagnostic reasoning, frequently failing to connect abstract diagnostic knowledge with anatomically grounded visual interpretation.
While structured guidance aids diagnostic reasoning, most models fail to generalize this reasoning to new cases, largely due to persistent challenges in visual grounding, such as accurately identifying anatomical regions, despite occasional improvement in non-visual aspects like diagnostic criterion selection.



%% file: MainSections/3.RelatedWorks.tex
\input{Tables/Main_Comparison_CheXstruct}

\textbf{Structuring Clinical Information from Chest X-rays}
Prior studies on structuring clinical information from chest X-rays~\cite{bannur2024maira, castro2024padchest, wu2021chest} typically link radiology report sentences to specific regions using bounding boxes to align findings with X-rays.
However, these approaches focus on coarse-grained annotations and rely on supervision from existing reports, offering limited insight into the underlying diagnostic reasoning.
In contrast, \extractor is an automated framework that extracts clinically relevant information directly from chest X-rays by segmenting anatomical regions, deriving anatomical landmarks and diagnostic measurements, computing diagnostic indices, and applying clinical thresholds.
This approach captures fine-grained reasoning steps, enabling more detailed and interpretable diagnostic information.
See Table~\ref{table:comparison_chexstruct} for a comparison.
 
\textbf{Medical VQA Benchmarks}
Existing benchmarks focus on evaluating whether a model arrives at the correct diagnostic answer, offering limited insight into its reasoning process.
Earlier studies~\cite{ben2021overview, he2020pathvqa, lau2018dataset, zhang2023pmc} focus on visual perception, such as identifying abnormalities.
Recent works~\cite{bae2023ehrxqa, hu2023expert, liu2024gemex, zuo2025medxpertqa} introduce longer contexts and visual grounding, aiming to reflect implicit reasoning.
However, these still emphasize answer correctness than evaluate intermediate diagnostic reasoning.
\benchmark bridges this gap by explicitly evaluating a model’s ability to perform and follow diagnostic reasoning through multi-stage, multi-path evaluations, including visual grounding via anatomical structure selection and diagnostic measurements.
See Table~\ref{table:comparison_cxreasonbench} for a comparison.

\input{Tables/Main_Comparison_CXReasonBench}

%% file: Tables/Main_Comparison_CheXstruct.tex
\begin{table}[htb!]
\vspace{-1mm}
  \caption{
  Comparison of chest X-ray structuring frameworks.
  ``Multi.'' indicates segmentation masks, anatomical landmarks, diagnostic measurements, and diagnostic indices.
  ``Model-assisted'' indicates the use of a segmentation model.
  }
  \label{table:comparison_chexstruct}
  \centering
  {\begin{adjustbox}{width=\linewidth}
  \begin{tabular}{cccccc}
    \toprule
    & \textbf{Structuring} & \multicolumn{4}{c}{\textbf{Image Annotation}} \\
    \cmidrule(r){3-6}
    \textbf{Framework} & \textbf{Target} & \textbf{Source} & \textbf{Type} & \textbf{Level}  & \textbf{Method} \\
    \midrule
    Chest ImaGenome~\cite{wu2021chest} & Chest X-rays & Reports & Bounding box &  Coarse-grained & Model-based \\
    PadChest-GR~\cite{castro2024padchest} & Chest X-rays & Reports & Bounding box &  Coarse-grained & Manual + Model-based\\
    GR-Bench~\cite{bannur2024maira} & Chest X-rays & Reports & Bounding box &  Coarse-grained & Manual + Model-based\\
    \midrule
    \textbf{\extractor (Ours)} & Chest X-rays & Chest X-rays & Multi. &  Fine-grained & Rule-based (Model-assisted)
\\
    
    \bottomrule
  \end{tabular}
  \end{adjustbox}
  }
  \vspace{-0.5mm}
\end{table}

%% file: Tables/Main_Comparison_CXReasonBench.tex
\begin{table}[htb!]
  \caption{Comparison of medical VQA benchmarks.
  For visual grounding, \benchmark utilizes anatomical structure selection and diagnostic measurements, while GEMeX relies on bounding box.
  }
  \label{table:comparison_cxreasonbench}
  \centering
  {\begin{adjustbox}{width=\linewidth}
  \begin{tabular}{ccccc}
    \toprule
    \textbf{Benchmark} & \textbf{QA Pairs} & \textbf{Visual Reasoning Type} & \textbf{Evaluation Format} & \textbf{Visual Grounding} \\ 
    \midrule
    VQA-RAD~\cite{lau2018dataset} & 3.5k  & Perception & Single-stage & \ding{55} \\
    VQA-Med~\cite{ben2021overview} & 5.5k & Perception & Single-stage & \ding{55} \\
    PathVQA~\cite{he2020pathvqa} & 32.7k & Perception & Single-stage & \ding{55} \\
    PMC-VQA~\cite{zhang2023pmc} & 227k & Perception & Single-stage & \ding{55} \\
    MIMIC-CXR-VQA~\cite{bae2023ehrxqa} & 337k & Implicit Reasoning  & Single-stage & \ding{55} \\
    MIMIC-Diff-VQA~\cite{hu2023expert} & 70k & Implicit Reasoning  & Single-stage & \ding{55} \\
    GEMeX~\cite{liu2024gemex} & 1.6M  & Implicit Reasoning  & Single-stage & \ding{51} \\
    MedXpertQA~\cite{zuo2025medxpertqa} & 2k & Implicit Reasoning  & Single-stage & \ding{55} \\

    \midrule
    \textbf{\benchmark (Ours)} & 18.9k & Explicit Reasoning & Multi-path, Multi-stage & \ding{51} \\
    
    \bottomrule
  \end{tabular}
  \end{adjustbox}
  }
  \vspace{-3mm}
\end{table}

  

%% file: MainSections/4.CheXStruct.tex
\section{\extractor}\label{sec:chexstruct}

\begin{figure}[htb!]
    \centering
\includegraphics[width=\linewidth]{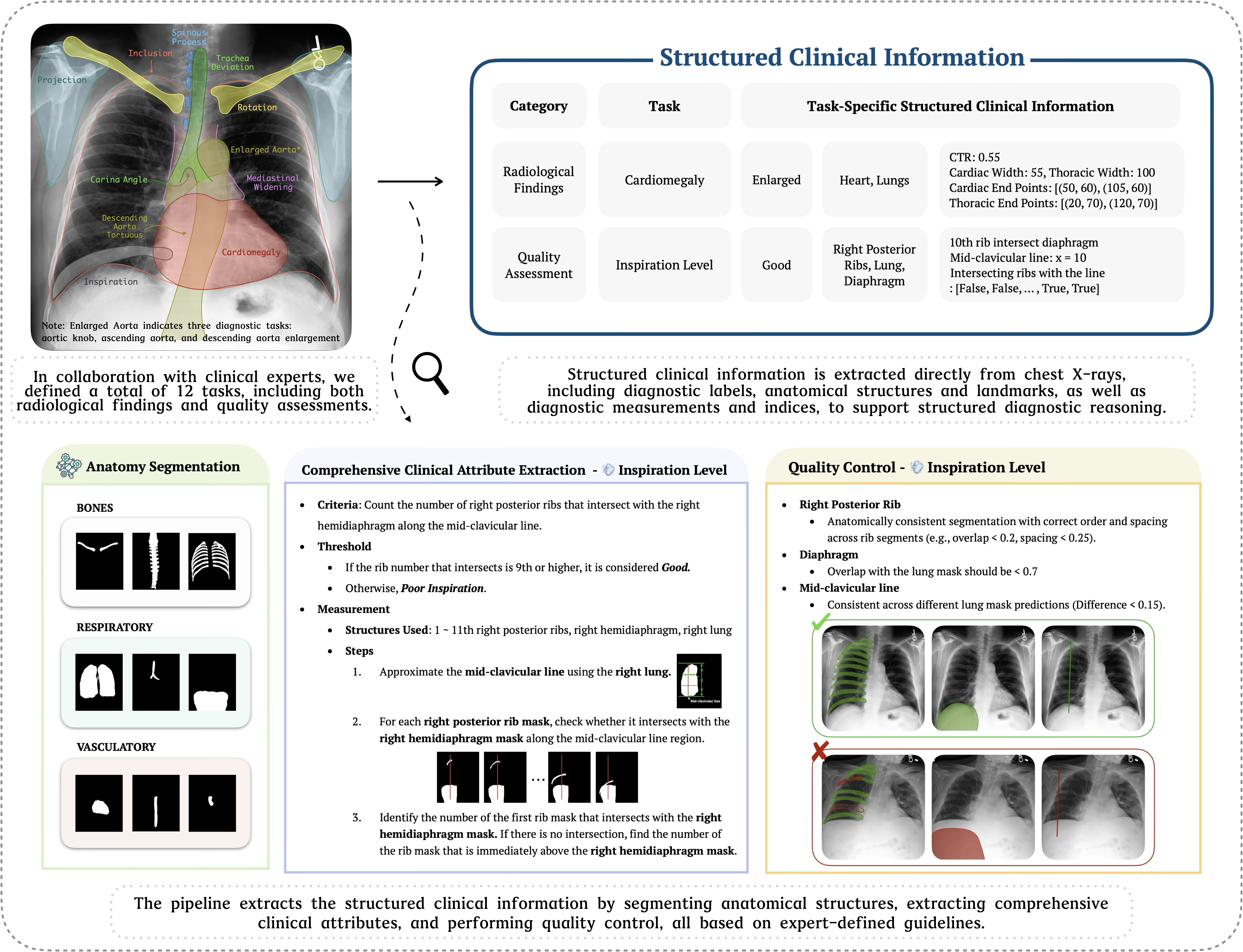}
    \caption{
    Overview of the \extractor pipeline. Given a chest X-ray, \extractor performs anatomical segmentation, derives anatomical landmarks and diagnostic measurements, computes diagnostic indices, and applies clinical thresholds, followed by task-specific quality control.
    }
\label{figure:overview_chexstruct}
\end{figure}

As shown in Figure~\ref{figure:overview_chexstruct}, \extractor is a fully automated pipeline designed to extract structured clinical information directly from chest X-rays.
The pipeline performs anatomical segmentation, derives anatomical landmarks and diagnostic measurements, computes diagnostic indices, and applies clinical thresholds, followed by task-specific quality control based on expert-defined guidelines.

Since \extractor relies on a segmentation model trained to identify anatomical structures~\cite{seibold2023accurate}, it is not equipped to detect pathology-specific pixel patterns such as opacities or air-fluid levels.
Therefore, CheXStruct currently supports only findings that can be derived through structural reasoning.
To ensure both clinical relevance and technical feasibility, we collaborated with clinical experts to define 12 radiological findings and quality assessment tasks that align with this structural pipeline.
Please refer to Appendix~\ref{apd:chexstruct_details} for detailed descriptions of the diagnostic tasks and the entire pipeline.

\subsection{Structured Clinical Information Extraction}
To enable structured diagnostic reasoning, \extractor extracts clinically meaningful information from raw chest X-ray images through a multi-step process.

\textbf{Defining Task-Specific Criteria}
Each of the 12 diagnostic tasks is grounded in clinical measurement guidelines, which fall into one of two categories:
\begin{itemize}[leftmargin=*,noitemsep,nolistsep]
    \item \textbf{Standardized, Quantifiable Criteria}
    For tasks with well-established criteria that do not depend on imaging metadata (\textit{e.g.}, pixel spacing that represents the real-world distance of each pixel), we adopt standard clinical measurement rules.
    For instance, \textit{Cardiomegaly} is assessed using the cardiothoracic ratio (CTR), defined as the ratio of maximum horizontal cardiac width to thoracic width, which does not rely on absolute measurements (\textit{e.g.}, centimeters) that cannot be calculated based only on X-ray images.

    \item \textbf{Expert-Defined Criteria}\label{sec:chexstruct_ex_criteria}
    To ensure reliable evaluation in tasks lacking standardized criteria or involving measurement ambiguity (\textit{e.g.}, when diagnosis depends on absolute measurements derived from imaging metadata), we collaborated with clinical experts to define quantifiable, clinically meaningful standards that guide structured diagnostic reasoning. For instance, \textit{Mediastinal Widening} is typically assessed based on whether the mediastinal width exceeds 8cm, but such absolute measurements cannot be obtained directly from X-ray images without pixel spacing. To address this, we measure the mediastinal width and the lung width at the same axial level, and assess mediastinal widening based on their ratio, enabling consistent evaluation across images.
\end{itemize}

\textbf{Segmenting Anatomical Structures}  
Each diagnostic task requires segmentation of specific anatomical regions.  
We use CXAS~\cite{seibold2023accurate}, a chest X-ray segmentation model trained on expert-curated data, to obtain the necessary segmentation masks.  
For example, \textit{Cardiomegaly} requires heart and lung masks.

\textbf{Comprehensive Clinical Attribute Extraction}
From the segmented masks, we extract anatomical landmarks (\textit{e.g.}, cardiac and thoracic end points) and diagnostic measurements (\textit{e.g.}, cardiac and thoracic width), which are then used to compute diagnostic indices (\textit{e.g.}, CTR) and apply clinical thresholds according to the defined criteria (\textit{e.g.}, cardiomegaly if the CTR exceeds 0.5 in PA view).

\subsection{Quality Control}
\extractor enforces task-specific quality control (QC) to ensure that the extracted structured information is both anatomically valid and clinically reliable.
These QC rules, developed in collaboration with clinical experts, automatically exclude low-quality samples from the benchmark.
Full descriptions of the QC rules can be found in the Appendix~\ref{apd:chexstruct_QC}

\textbf{Quality Criteria Definition}
We define QC rules specific to each task’s anatomical structures and associated measurement criteria.
For instance, for \textit{Inspiration Level}, we verify that the right posterior ribs are segmented in anatomically consistent positions, maintaining proper spatial order and spacing.

\textbf{Threshold-Based Filtering}
Using the defined QC rules, \extractor filters out samples that fail to meet task-specific standards.
For instance, we discard samples with overlapping or disordered rib masks (\textit{Inspiration Level}).
Diagnostic labels are assigned only to samples passing QC, based on expert-defined thresholds.

\subsection{Automation and Data Scalability}
\extractor is fully automated and scales efficiently to large datasets.
While clinical experts were involved in defining task-specific criteria and thresholds used in structured information extraction and quality control, the execution process itself requires no human intervention.
Only image-task pairs that pass all relevant QC rules and yield clinically valid structured outputs are used in the final benchmark, ensuring both reliability and scalability.

%% file: MainSections/5.CXReasonBench.tex
\section{\benchmark}

\begin{figure}[htb!]
    \centering
\includegraphics[width=\linewidth]{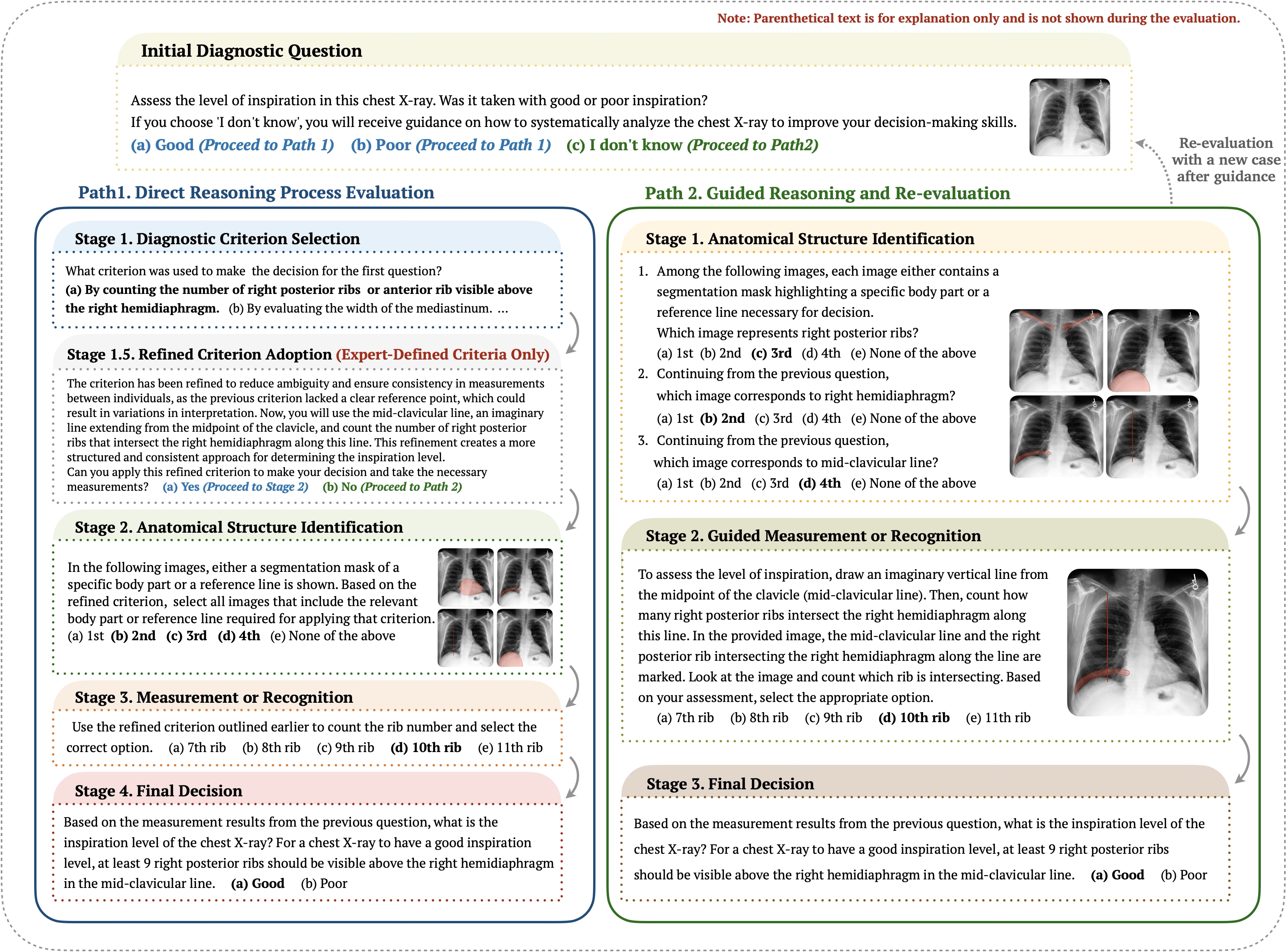}
    \caption{
    Overview of the \benchmark evaluation pipeline.
    The evaluation begins with a direct diagnostic question, followed by two possible paths depending on the model’s response.
    Path 1 evaluates the model’s ability to reconstruct its reasoning through intermediate steps or to apply expert-defined criteria when necessary.
    Path 2 provides structured guidance to develop the reasoning process when the model expresses lack of confidence.
    }
\label{figure:overview_cxreasonbench}
\end{figure}

Building upon the structured information extracted in \extractor, \benchmark is a multi-path, multi-stage evaluation framework designed to assess a model’s ability to perform structured diagnostic reasoning using chest X-rays.
Figure~\ref{figure:overview_cxreasonbench} describes the overall evaluation pipeline.

\subsection{Evaluation Pipeline} \label{subsection:becnhmark_evaluation_pipeline}

\textbf{Initial Diagnostic Decision and Path Assignment}
Each evaluation begins with a binary diagnostic question (\textit{e.g.}, \textit{Does this patient have cardiomegaly?}).
The model is given three options: A binary answer (\textit{e.g.}, \textit{Yes / No}), or \textit{I don’t know}.
If the model selects a binary answer, it proceeds to Path 1 (Section ~\ref{sec:cxreasonbench_path1}). If it selects \textit{I don’t know}, it is directed to Path 2 (Section ~\ref{sec:cxreasonbench_path2}).

\subsubsection{Path 1: Direct Reasoning Process Evaluation}\label{sec:cxreasonbench_path1}
Path 1 evaluates whether the model’s diagnostic decision is supported by a sequence of clinically coherent reasoning steps as below.
At each stage, an incorrect response terminates the evaluation, under the assumption that clinically valid reasoning cannot be inferred beyond that point.
Refer to Appendix~\ref{apd:cxreasonbench_path1} for stage-specific reference answers and detailed definitions for each task.

\begin{itemize}[leftmargin=*,noitemsep,nolistsep]
    \item \textbf{Stage 1. Diagnostic Criterion Selection}
    The model identifies the criterion used for the initial diagnostic question, allowing us to determine if its decision was based on clinically accepted knowledge.
    Failure to select a valid criterion prevents assessment of the diagnostic reasoning applied, rendering the rest of the reasoning process irrelevant.

    \item \textbf{Stage 1.5. Refined Criterion Adoption (Expert-Defined Criteria Only)}
    For diagnostic tasks where the original criteria are difficult to apply to images (\textit{e.g.}, reliance on imaging metadata) or are inherently ambiguous, the model is provided expert-defined criteria from Section~\ref{sec:chexstruct_ex_criteria}.
    If the model accepts it, it proceeds to Stage 2;
    otherwise, the model advances to Path 2.
    
    \item \textbf{Stage 2. Anatomical Structure Identification}
    The model selects all relevant anatomical regions or reference lines from highlighted chest X-ray images related to the criterion.
    If the model actually used the criterion for the initial diagnosis or genuinely accepted and understood an expert-defined criterion, it should have visually assessed the associated anatomical structures to apply the criterion correctly.
    This stage tests whether such visual grounding truly occurred.
    
    \item \textbf{Stage 3. Measurement or Recognition}
    The model applies the diagnostic criterion.
    In measurement-type tasks, it performs arithmetic computations based on anatomical measurements (\textit{e.g.}, calculating the CTR) and selects a value range (\textit{e.g.}, [0.50–0.52]).
    In recognition-type tasks, it interprets anatomical changes (\textit{e.g.}, identifying tracheal deviation) and selects the appropriate label (\textit{e.g.}, ``shifted to the left'').
    This stage ensures that the model explicitly applies the criterion to make a decision.
    Failure at this stage indicates a misinterpretation of the criterion or an inability to apply it, suggesting that the model may not be able to effectively use the criterion as intended.
    
    \item \textbf{Stage 4. Final Decision}
    The model makes a final decision based on Stage 3.
    The evaluation of this decision depends on whether standardized diagnostic thresholds are available for the task.  
    For tasks with standardized thresholds (\textit{e.g.}, CTR of 0.5 for cardiomegaly in PA view), no threshold is provided to the model.
    For other tasks, an expert-defined threshold from Section~\ref{sec:chexstruct_ex_criteria} is provided.  
\end{itemize}

\subsubsection{Path 2: Guided Reasoning and Re-evaluation}\label{sec:cxreasonbench_path2}
Path 2 is triggered when the model either responds with \textit{``I don’t know''} to the initial diagnostic question or rejects the expert-defined criterion in Stage 1.5.
The goal here is to evaluate whether the model can follow structured guidance to develop its reasoning process and later apply this learned reasoning to a new case within the same diagnostic task.
Evaluation is terminated upon failure at any stage, as each step is inter-dependent; failure at an early stage indicates insufficient foundational knowledge for diagnostic reasoning, which impacts subsequent stages outlined below.
Refer to Appendix~\ref{apd:cxreasonbench_path2} for reference answers and definitions for each task and stage.

\begin{itemize}[leftmargin=*,noitemsep,nolistsep]
    \item \textbf{Stage 1. Anatomical Structures Identification}
    The model is shown highlighted chest X-rays and asked to identify specific anatomical structures (\textit{e.g.}, \textit{“Which image shows the heart?”}).
    This stage evaluates whether the model can visually recognize and distinguish relevant anatomical regions, an essential prerequisite for accurate diagnostic reasoning.

    \item \textbf{Stage 2. Guided Measurement or Recognition}
    The model receives detailed visual annotations (\textit{e.g.}, labeled landmarks, coordinates) along with instructions for a guided diagnostic assessment.
    For instance, in evaluating cardiomegaly, heart and thoracic widths are labeled, coordinates are given, and the cardiothoracic ratio calculation is explained.
    This stage tests the model’s ability to follow structured guidance and derive clinically meaningful conclusions from annotated images.

    \item \textbf{Stage 3. Final Decision}
    The model is provided with a diagnostic threshold and asked to make a final decision based on the result from Stage 2.
    This stage evaluates whether the model can apply the threshold appropriately to reach an accurate diagnostic conclusion.
\end{itemize}

\textbf{Re-evaluated Path 1 after Guidance}
If the model successfully completes guided reasoning, it is considered to have acquired the correct reasoning process.
This path evaluates whether the model can internalize and independently apply that reasoning to a new case within the same diagnostic task.
The model is re-evaluated using the same structure as Path 1, except Stage 1.5 is omitted, as the criterion has already been introduced during guidance.
\begin{itemize}[leftmargin=*,noitemsep,nolistsep]
    \item \textbf{Random Re-selection}
    A new case is randomly chosen from the set of cases that the model previously could not reach the final stage in Path 1, or initially responded to with \textit{``I don’t know''}, excluding the one just used for guidance in Path 2.
    \item \textbf{Re-evaluation Pipeline}
    The model is asked the diagnostic question in the same format as the initial diagnostic decision.
    If the model responds incorrectly or with \textit{“I don’t know”}, the evaluation ends, indicating that the model has not internalized the reasoning process or cannot generalize it to a new case.
    If the model answers correctly, it re-enters Path 1 and resumes the evaluation.
\end{itemize}
Note that, given a test case, a model (\textit{i.e.} vision-language model) proceeds all stages while storing all previous questions and answers in its context.
This ensures that the model can complete Path 1, or it can potentially internalize the lessons from Path 2 and complete Path 1 on the second attempt.

\subsection{Benchmark Structure}\label{sec:benchmark_structure}

\textbf{Benchmark Format}
The benchmark uses a multiple-choice format with both single-choice (\textit{e.g.}, one diagnostic criterion) and multi-choice questions (\textit{e.g.}, all relevant anatomical structures) depending on the stage.
Refer to Appendix ~\ref{apd:cxreasonbench_structure} for details.
\begin{itemize}[leftmargin=*,noitemsep,nolistsep]
    \item \textbf{Two-round format with \textit{``Need new option''}}
    In certain stages, a two-round format is used, though not always. Initially, the correct answer is intentionally excluded, and the model is presented with a \textit{``Need new option''} choice.
    If the model selects this option, a second round is triggered, where the correct answer is revealed. 
    This format assesses whether the model can recognize insufficient reasoning paths and appropriately defer its decision to a subsequent, more complete option set.
    
    \item \textbf{\textit{``None of the above''} option}
    When the \textit{“Need new option”} is not included, a \textit{“None of the above”} option is provided instead.
    In this case, the correct answer is always included in the initial options.
    Selecting \textit{“None of the above”} results in failure, and the model is expected to explain its reasoning.
\end{itemize}

\textbf{Benchmark Scale and Coverage}
\benchmark is constructed by applying \extractor to MIMIC-CXR-JPG~\cite{johnson2019mimic}.
To ensure manual review feasibility, we randomly sampled 100 cases for each of the 12 diagnostic tasks (1,200 in total), and a clinical expert manually reviewed all corresponding segmentation masks for quality assurance (see Appendix~\ref{apd:chexstruct_clinician_review}).
From each case, we generate QA pairs across three distinct evaluation settings (Path 1, Path 2, and Re-evaluated Path 1), supporting multi-stage evaluation and resulting in 18,988 QA pairs in total: 8,044 in Path 1, 3,600 in Path 2, and 7,344 in Re-evaluated Path 1.
Each QA pair is associated with up to 4 visual inputs (\textit{e.g.}, the original X-ray and overlaid segmentation masks).
While our evaluation focuses on these clinically validated cases, \extractor supports scalable benchmark construction, enabling large-scale evaluation.
The number of cases extracted by \extractor for each diagnostic task is summarized in Table~\ref{table:chexstruct_label_distribution}.




%% file: MainSections/6.Experiments.tex
\subsection{Experimental Setup} \label{sec:experimental_setup}

\textbf{Models}
We evaluate 12 LVLMs, including 3 closed-source and 9 open-source LVLMs.
\revision{
The closed-source models are specified as follows: Gemini-2.5-Pro (gemini-2.5-pro-preview-03-25), Gemini-2.5-Flash (gemini-2.5-flash-preview-04-17), and GPT-4.1 (GPT-4.1, 2025-04-14).
The open-source models include: Pixtral-Large, Llama-3.2-90B-Vision, Qwen2.5-VL-72B, Pixtral 12B, Qwen2.5-VL-7B, MedGemma 27B, HealthGPT-L14, RadVLM, and MedGemma 4B. To comply with the MIMIC Data Use Agreement, Gemini models are run on Vertex AI, the GPT model is deployed on Azure’s HIPAA-compliant platform, and open-source LVLMs are evaluated locally.
}

\textbf{Sampling Methods}
We use greedy decoding for deterministic responses and stochastic decoding for generating three diverse responses, with majority voting requiring at least two matches to determine correctness. The stochastic decoding uses a temperature of 0.7 and a top-$p$ of 0.95.

\textbf{Evaluation Metric}
We evaluate the model’s performance as follows.
Except for \textit{Average Reasoning Depth}, all metrics are computed using the Wilson score~\cite{wilson1927probable}, which accounts for the number of attempts rather than relying on simple proportions, allowing for fairer comparison across models with varying numbers of attempts at each stage.

\begin{itemize}[leftmargin=*,noitemsep,nolistsep]
    \item \textbf{Final Stage Completion (reported as Completion)} measures how often the model successfully completes the reasoning process through to the final stage across the benchmark.

    \item \textbf{Average Reasoning Depth (Depth)} indicates the average number of stages the model reaches throughout the benchmark, reflecting its ability to progress through the reasoning pipeline.

    \item \textbf{Decision Alignment (Alignment)} applies only to Path 1 and Re-evaluated Path 1, measuring the agreement between the model’s initial and final decisions, and counting them as aligned only when both are correct.
    In Path 1, alignment can be reliably assessed for tasks with standardized diagnostic thresholds, where consistent application is expected across stages. 
    For other tasks, the final decision is based on an expert-defined threshold provided at Stage 4, while the initial decision may rely on an implicit threshold, making alignment less definitive, though high scores may still reflect consistent reasoning.
    In Re-evaluated Path 1, since the expert-defined threshold has already been provided during Path 2, alignment can be assessed reliably across all tasks.

    \item \textbf{Measurement Consistency (Consistency)}
    For measurement-type tasks, the model is instructed to return the calculated value in the preceding stage along with its final decision in the final stage (\textit{i.e.}, Stages 3 and 4 for Path 1, and Stages 2 and 3 for Path 2).
    We evaluate consistency by checking whether this value falls within the range selected in the preceding stage.
    If it does, the earlier response is considered correct; otherwise, it is considered incorrect.
    Based on this, we report the consistency rate and further refine the other metrics to incorporate correctness in the preceding stage based on this consistency.
    These refined scores, shown in parentheses, reflect the model’s ability to maintain coherent and accurate reasoning across stages.
    
    
\end{itemize}

Further details on the experimental setup can be found in Appendix~\ref{apd:experiment_setup_details}.

\subsection{Results}\label{sec:results}

\textbf{Results from Path 1: Direct Reasoning Process Evaluation}
Table~\ref{table:reasoning_greedy_overall} shows that current models struggle with valid structured diagnostic reasoning, primarily due to difficulties in bridging abstract diagnostic knowledge with visually grounded interpretation.
Even when models correctly identify diagnostic criteria, they often fail to localize the relevant anatomical structures, revealing a disconnect between conceptual understanding and visual grounding.
These limitations are reflected in overall trends: closed-source models generally outperform open-source ones, yet even the strongest model (\textit{i.e.}, Gemini-2.5-Pro) typically only reaches Stage 2 and rarely completes the full reasoning process.

Notably, even at Stage 2, the model’s performance varies by task (Tables~\ref{table:reasoning_greedy_all_recognition} and~\ref{table:reasoning_greedy_all_arithmetic}).
It performs well on tasks involving a single structure, regardless of saliency. For instance, inclusion (lungs; 89\%) and carina angle (carina; 79\%).
Tasks involving multiple structures also show strong performance when the structures are salient and the reference line is clear, as in cardiomegaly (heart and thoracic width; 75\%).
In contrast, performance drops when tasks involve less salient structures or rely on abstract reference lines, such as inspiration level (diaphragm, rib, and mid-clavicular line; 47\%) and tracheal deviation (trachea and midline; 48\%).
These results suggest model performance is not determined solely by the number of structures involved, but by the combined challenges of structural saliency and clarity of the reference line.
Effective diagnostic reasoning thus requires models to integrate fine-grained visual grounding with spatial abstraction across inter-dependent anatomical cues.

Building on the challenges observed in earlier stages, the performance differences at Stage 3 can be further explained by training data composition (Table~\ref{table:reasoning_greedy_all}).
Medical-specialized models (\textit{e.g.}, HealthGPT, RadVLM) are primarily trained on medical imaging and clinical reports~\cite{he2020pathvqa, liu2021slake, zhang2023large}, which emphasize anatomical descriptions and visual interpretation.
This supports relatively better performance on recognition-type tasks but limits capability on measurement-type tasks requiring arithmetic computation.
In contrast, six of eight general-domain models show the opposite trend, which may be attributed to broader exposure to numerical reasoning and arithmetic operations in large-scale corpora~\cite{amini2019mathqa, gao2020pile, jia2021scaling}.
These findings underscore the need for balanced training data to improve diagnostic reasoning across both recognition-type and measurement-type tasks.

\input{Tables/1.Reasoning/Result_Main_Greedy_Overall}

\input{Tables/2.Guidance/Result_Main_Greedy_Overall}

\textbf{Results from Path 2: Guided Reasoning and Re-evaluation}
As shown in Tables~\ref{table:guidance_greedy_overall} and~\ref{table:guidance_greedy_all}, Gemini 2.5-Pro maintains strong and consistent performance across tasks, excelling at following structured guidance to reach diagnostic decisions.
Notably, open-source models such as Qwen2.5-VL-72B and Pixtral-Large achieve reasoning depths comparable to GPT-4.1, reflecting growing potential for guided diagnostic reasoning.

In measurement-type tasks (Table~\ref{table:guidance_greedy_all_arithmetic}), model performance varies with computational complexity.
For simpler anatomical ratios, such as cardiomegaly, some models reliably reach an average depth close to 3, suggesting that clear visual cues and structured prompts facilitate multi-stage reasoning.
In contrast, tasks requiring more complex spatial reasoning, like curvature estimation in descending aorta tortuous, remain difficult, with all models failing to exceed an average depth of 2.
This reveals a persistent limitation in complex visual reasoning, even with explicit guidance.

Moreover, most models struggle to transfer guided reasoning to new cases (Table~\ref{table:reasoning_greedy_overall}), indicating that contextual exposure alone is insufficient for internalizing diagnostic reasoning skills.
While some improvement is observed in non-visual knowledge, such as selecting diagnostic criteria, visual grounding, especially in identifying anatomical structures, remains a major challenge.
This persistent bottleneck highlights the need for training paradigms that explicitly align visual grounding with reasoning supervision, beyond what in-context learning alone can offer.

\textbf{Measurement Consistency Results}
Although this metric was originally designed to assess whether the model maintains coherent and accurate reasoning across stages, the results reveal a deeper issue: current LVLMs often rely on visual cues or shortcuts to make diagnostic decisions, rather than applying diagnostic criteria through structured, measurement-based reasoning.
As shown in Tables~\ref{table:reasoning_greedy_overall} and ~\ref{table:guidance_greedy_overall}, most models exhibit lower consistency under Path 1 than Path 2 (Refer to Tables~\ref{table:reasoning_greedy_all_arithmetic} and~\ref{table:guidance_greedy_all_arithmetic} for details).
Since the same models perform better under Path 2, this suggests that the primary limitation lies not in maintaining coherence across reasoning stages, but in performing the necessary computations to apply diagnostic criteria under Path 1.
In Path 2, where visual landmarks and explicit computational guidance are provided, models follow instructions and perform calculations in the earlier stage, and as a result, the values returned in the final stage are consistent with the selected ranges.
In contrast, under Path 1, models tend to approximate using heuristics, resulting in mismatches between the returned values and the selected ranges.
This points to a key opportunity for future development: while current LVLMs are capable of computation and instruction following, they need an approach to internalize and independently apply diagnostic criteria through structured reasoning.
See Appendix~\ref{apd:additional_results} for qualitative analyses supporting the measurement consistency results, along with additional evaluation results.

%% file: Tables/1.Reasoning/Result_Main_Greedy_Overall.tex
\begin{table} 
  \caption{
  Path 1 and Re-evaluated Path 1 results with greedy sampling for overall tasks.
  Completion: Percentage of cases completing all reasoning stages; 
  Depth: Average number of reasoning stages reached;
  Consistency: Percentage of cases where the value returned at Stage 4 matches the Stage 3 response;
  Alignment: Percentage of agreement between initial and final decisions;
  Refined scores incorporating measurement consistency are shown in parentheses.
  `N/A' indicates that the model did not reach the required stage to compute the corresponding metric.
  }
  \label{table:reasoning_greedy_overall}
  \centering
  {\begin{adjustbox}{width=0.9\linewidth}
  \begin{tabular}{cccccc}
    \toprule
    \textbf{Path} & \textbf{Model} & \textbf{Completion (0-100) $\uparrow$} & \textbf{Depth (0-4) $\uparrow$} & \textbf{Consistency (0-100) $\uparrow$} & \textbf{Alignment (0-100)$\uparrow$} \\
    \midrule

    & Gemini-2.5-Pro  & \textbf{17.03 (16.24)} & \textbf{1.96 (1.95)} & \textbf{68.4}  & \textbf{60.88 (58.61)} \\
    & Gemini-2.5-Flash  & 12.83 (8.56) & 1.4 (1.31) & 43.76  & 50.29 (37.67) \\
    & GPT-4.1  & 8.32 (8.32) & 1.15 (1.15) & 61.22  & 39.8 (39.8) \\
    \cmidrule(r){2-6}
    & Pixtral-Large  & 3.73 (2.31) & 1.0 (0.96) & 28.5  & 36.74 (25.3) \\
    Path 1 & Llama-3.2-90B-Vision  & 0.38 (0.38) & 0.53 (0.53) & 61.27  & 23.32 (23.32) \\
    & Qwen2.5-VL-72B  & 2.34 (2.12) & 0.67 (0.67) & 34.67  & 38.45 (34.36) \\
    & Pixtral 12B  & 0.0 (0.0) & 0.3 (0.3) & 36.39  & 0.0 (0.0) \\
    & Qwen2.5-VL-7B  & 1.21 (0.87) & 0.49 (0.48) & 14.86  & 45.44 (28.35) \\
    & MedGemma 27B  & 3.31 (2.34) & 0.69 (0.68) & 17.06  & 47.19 (31.2) \\
    & HealthGPT-L14  & 1.27 (1.27) & 0.32 (0.32) & 36.39  & 33.87 (33.87) \\
    & RadVLM  & 0.64 (0.39) & 0.28 (0.28) & 25.07  & 32.5 (20.32) \\
    & MedGemma 4B & 1.13 (0.98) & 0.32 (0.32) & 14.86  & 39.12 (31.61) \\

    \midrule

    & Gemini-2.5-Pro  & 0.0 (0.0) & \textbf{1.78 (1.77)} & \textbf{49.71}  & 0.0 (0.0) \\
& Gemini-2.5-Flash  & \textbf{6.03 (5.16)} & 1.41 (1.4) & 46.98  & 19.58 (15.68) \\
& GPT-4.1  & 0.0 (0.0) & 1.11 (1.11) & 36.39  & 0.0 (0.0) \\
\cmidrule(r){2-6}
& Pixtral-Large  & 0.0 (0.0) & 1.13 (1.13) & N/A  & N/A (N/A) \\
Re-evaluated & Llama-3.2-90B-Vision  & 0.0 (0.0) & 1.17 (1.17) & N/A  & N/A (N/A) \\
Path 1 & Qwen2.5-VL-72B  & 2.74 (2.74) & 1.08 (1.08) & 36.39  & \textbf{22.52 (22.52)} \\
& Pixtral 12B  & 0.0 (0.0) & 1.0 (1.0) & N/A  & N/A (N/A) \\
& Qwen2.5-VL-7B  & 0.0 (0.0) & 1.0 (1.0) & N/A  & N/A (N/A) \\
& MedGemma 27B  & 0.0 (0.0) & 1.0 (1.0) & 36.39  & 0.0 (0.0) \\
& HealthGPT-L14  & 0.0 (0.0) & 0.0 (0.0) & N/A  & N/A (N/A) \\
& RadVLM  & N/A (N/A) & N/A (N/A) & N/A  & N/A (N/A) \\
& MedGemma 4B  & 0.0 (0.0) & 0.0 (0.0) & N/A  & N/A (N/A) \\

    \bottomrule
  \end{tabular}
  \end{adjustbox}
  }
\end{table}

%% file: Tables/2.Guidance/Result_Main_Greedy_Overall.tex
\begin{table} 
\vspace{-3mm}
  \caption{
  Path 2 results with greedy sampling for overall tasks.
  Values in parentheses denote refined scores incorporating measurement consistency.
  `N/A' indicates that the model did not reach the required stage to compute the corresponding metric.
  }
  \label{table:guidance_greedy_overall}
  \centering
  {\begin{adjustbox}{width=0.7\linewidth}
  \begin{tabular}{cccc}
    \toprule
    \textbf{Model} & \textbf{Completion (0-100) $\uparrow$} & \textbf{Depth (0-3) $\uparrow$} & \textbf{Consistency (0-100) $\uparrow$} \\
    \midrule
Gemini-2.5-Pro  & \textbf{55.81 (54.29)} & \textbf{2.62 (2.58)} & \textbf{69.93}  \\
Gemini-2.5-Flash  & 34.65 (30.11) & 1.75 (1.64) & 57.73  \\
GPT-4.1  & 19.94 (19.15) & 0.99 (0.97) & 68.7  \\
\midrule
Pixtral-Large  & 19.04 (16.16) & 1.21 (1.16) & 51.89  \\
Llama-3.2-90B-Vision  & 9.55 (8.03) & 0.37 (0.34) & 38.18  \\
Qwen2.5-VL-72B  & 28.45 (25.48) & 1.2 (1.14) & 52.81  \\
Pixtral 12B  & 4.22 (4.22) & 0.45 (0.45) & 43.26  \\
Qwen2.5-VL-7B  & 5.38 (3.6) & 0.46 (0.44) & 11.26  \\
MedGemma 27B  & 17.81 (4.1) & 0.84 (0.5) & 18.83  \\
HealthGPT-L14  & 2.89 (2.46) & 0.21 (0.2) & 37.98  \\
RadVLM  & 0.0 (0.0) & 0.05 (0.05) & N/A  \\
MedGemma 4B & 1.59 (1.59) & 0.15 (0.14) & 27.15  \\

    \bottomrule
  \end{tabular}
  \end{adjustbox}
  }
\vspace{-3mm}
\end{table}

%% file: MainSections/7.Discussion.tex
\textbf{Limitations}
\revision{
In this study, we focus on 12 diagnostic tasks that are structurally inferable using anatomical segmentation. This design reflects both technical and clinical considerations. Technically, as noted in Section~\ref{sec:chexstruct}, CheXStruct relies on anatomical segmentation masks and structural measurements, and therefore cannot capture pixel-level pathologies such as opacities or pleural effusion. Consequently, tasks are limited to those derivable from spatial relationships (e.g., cardiothoracic ratio), while pixel-level pathologies requiring pixel-level texture recognition are not compatible with the current reasoning pipeline.
Clinically, pixel-level pathologies often lack standardized, reproducible criteria and are subject to high inter-reader variability, making them difficult to integrate into interpretable reasoning chains.
While this excludes some important findings, focusing on structurally inferable tasks enables automated, interpretable, and reproducible diagnostic reasoning. It also allows scalable benchmark creation without extensive manual annotations, filling a complementary niche unaddressed by prior benchmarks that focus solely on final predictions.
}

\textbf{Future Directions}
\revision{Future efforts will expand \extractor to include additional diagnostic tasks and datasets, enabling \benchmark to cover a broader and more clinically nuanced range of findings, potentially leveraging multi-modal information such as other imaging modalities, patient history, and lab tests or finer-grained reasoning processes.}
We also plan to develop \extractor further for automated radiology report generation and clinical quality assurance by verifying model-generated reports. 
Additionally, it will improve dataset quality by filtering training data through consistency checks between image-derived reasoning and human-written reports.
Finally, we aim to create instruction-tuning datasets derived from structured reasoning to train models with interpretable and clinically grounded diagnostic capabilities.

%% file: Supple/1.CheXStruct_Extractor.tex
\section{Details of \extractor} \label{apd:chexstruct_details}

\subsection{Diagnostic Tasks}\label{apd:chexstruct_diagnostictasks}
We defined a set of 12 diagnostic tasks in collaboration with clinical experts.
These tasks are categorized into two groups: radiological findings and image quality assessments.

\begin{itemize}[leftmargin=*]
    \item \textbf{Radiological Finding}  
    The selected findings are diagnosable from chest X-rays alone, without requiring additional patient information such as clinical history or symptoms (\textit{e.g.}, pneumonia).  

    \textit{Tasks: Cardiomegaly, Mediastinal Widening, Carina Angle, Trachea Deviation, Aortic Knob Enlargement, Ascending Aorta Enlargement, Descending Aorta Enlargement, and Descending Aorta Tortuous}

    \item \textbf{Image Quality Assessment}  
    These tasks evaluate the technical adequacy of image acquisition, ensuring that the radiograph meets basic quality standards for accurate interpretation.  

    \textit{Tasks: Inclusion, Inspiration Level, Rotation, Projection}
\end{itemize} 

\subsection{Structured Clinical Information Extraction}\label{apd:chexstruct_extraction}
To enable structured diagnostic reasoning, \extractor extracts clinically meaningful information from raw chest X-ray images through a multi-step process.

\subsubsection{Task-Specific Criteria Definition and Clinical Attribute Extraction}\label{apd:chexstruct_criteria}
This section presents detailed descriptions of each diagnostic task’s criteria and how corresponding clinical attributes are extracted.
For each task, we define clinically grounded measurement criteria and illustrate the extraction process using visual examples based on segmentation masks and anatomical landmarks.
These examples demonstrate how quantitative criteria are applied to derive consistent diagnostic evaluation across images.

\begin{figure}[htb!]
    \centering
    \includegraphics[width=\linewidth]{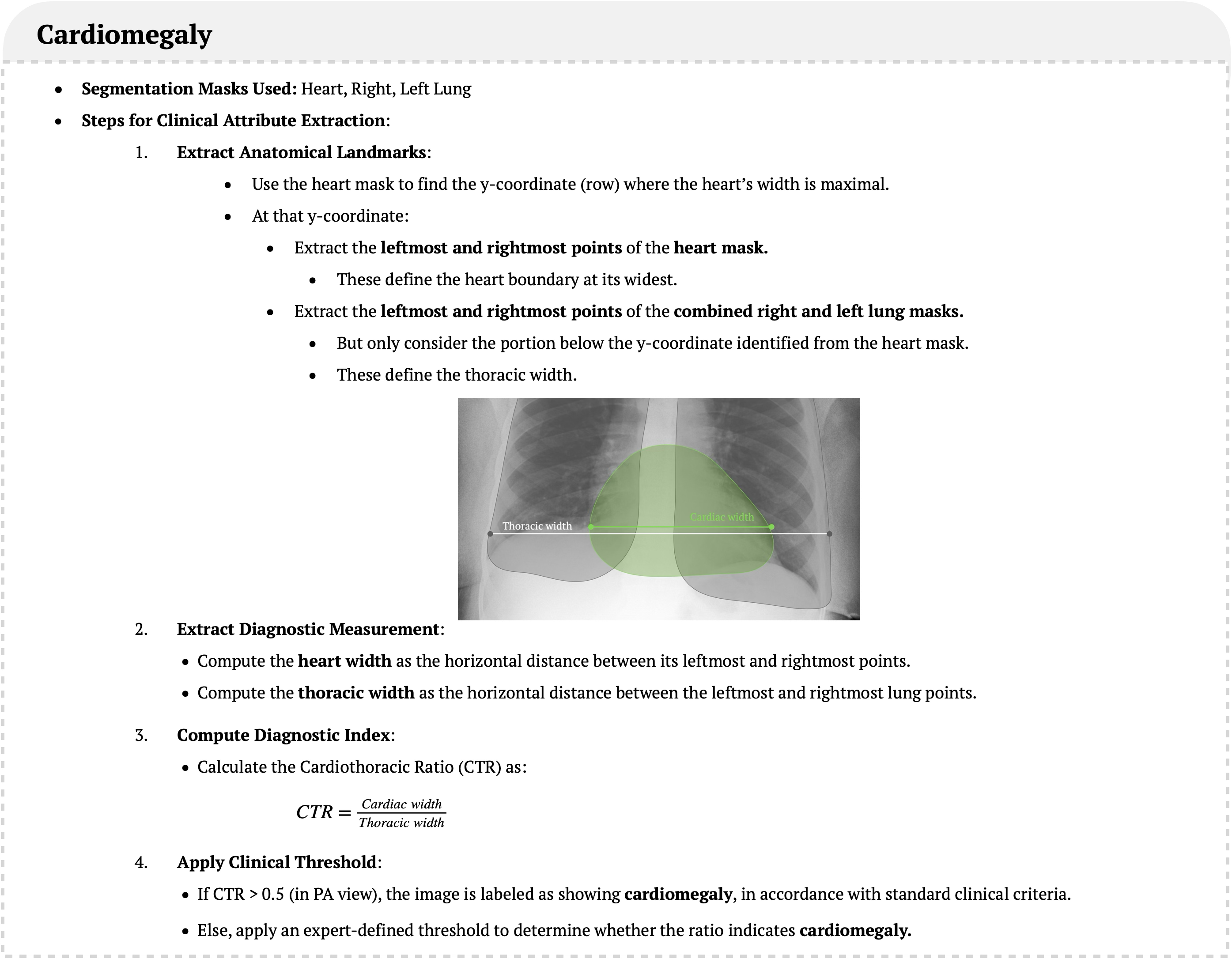}
    \caption{
    Measurement of cardiomegaly using the cardiothoracic ratio (CTR).
    }
\label{figure:chexstruct_cardiomegaly}
\end{figure}

\begin{figure}[htb!]
    \centering
    \includegraphics[width=\linewidth]{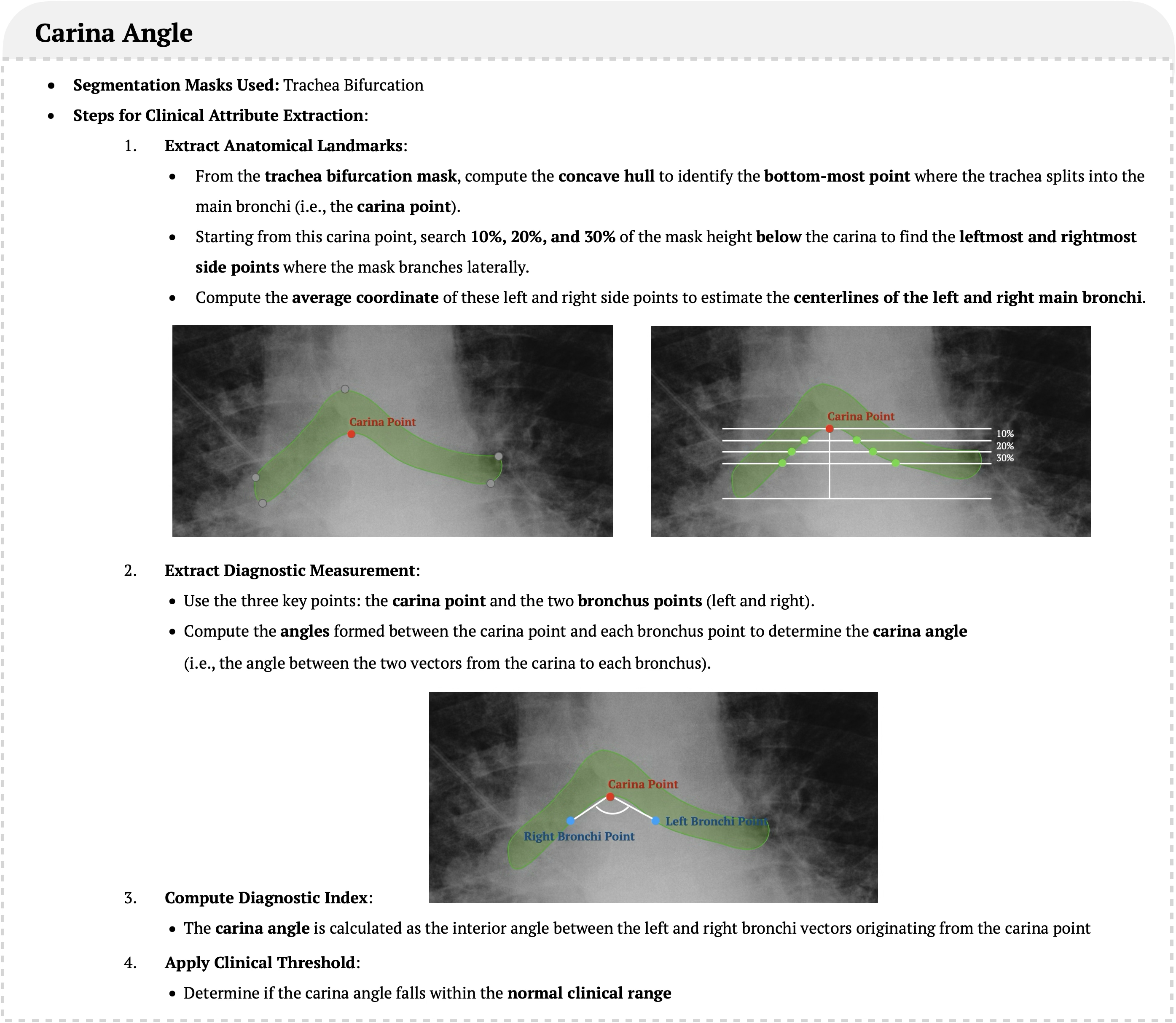}
    \caption{
    Measurement of the carina angle at the tracheal bifurcation.
    }
\label{figure:chexstruct_carina}
\end{figure}

\begin{figure}[htb!]
    \centering
    \includegraphics[width=\linewidth]{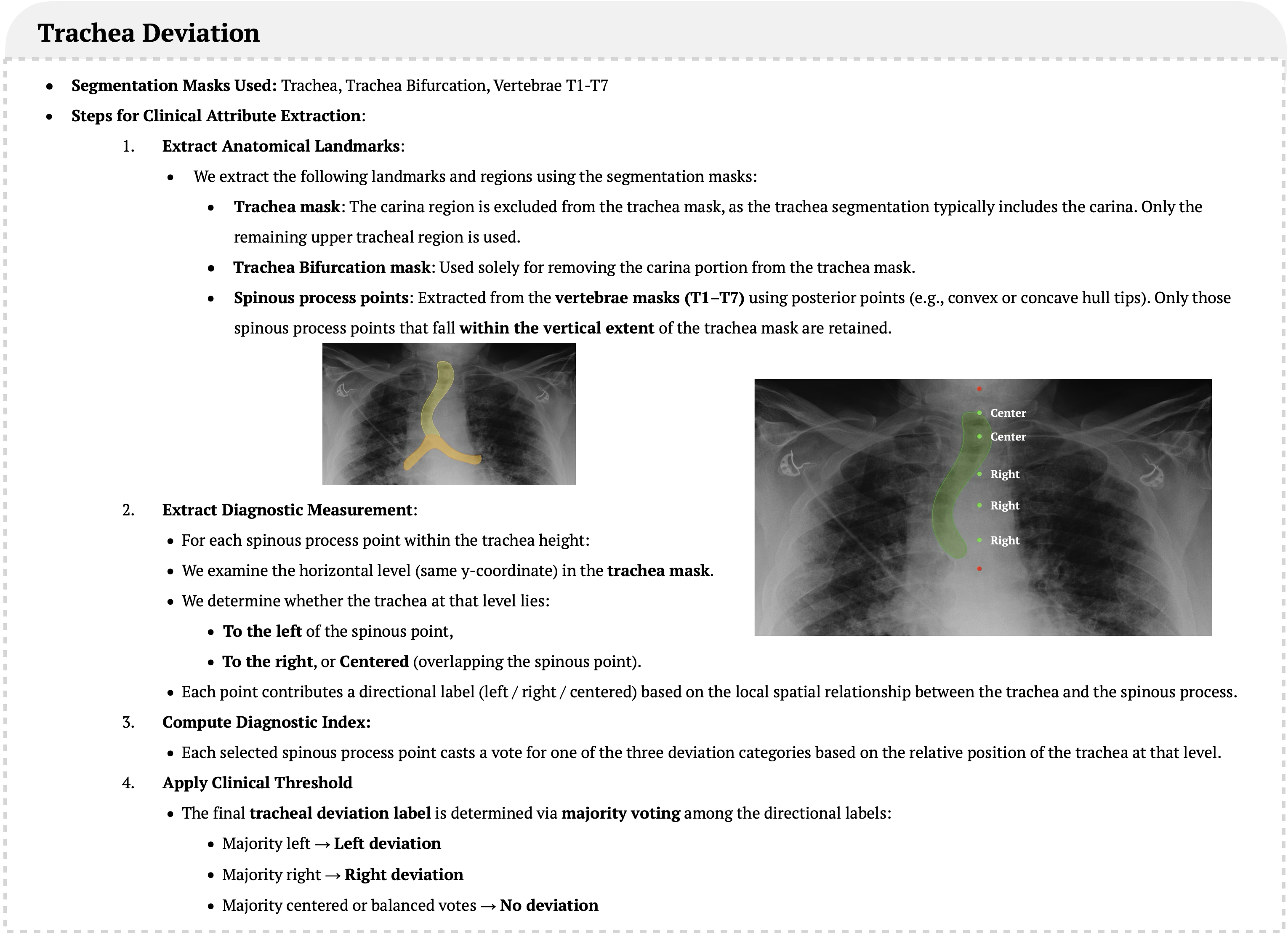}
    \caption{
    Evaluation of tracheal deviation based on lateral displacement from the spinal midline.
    }
\label{figure:chexstruct_tracheadev}
\end{figure}

\begin{figure}[htb!]
    \centering
    \includegraphics[width=\linewidth]{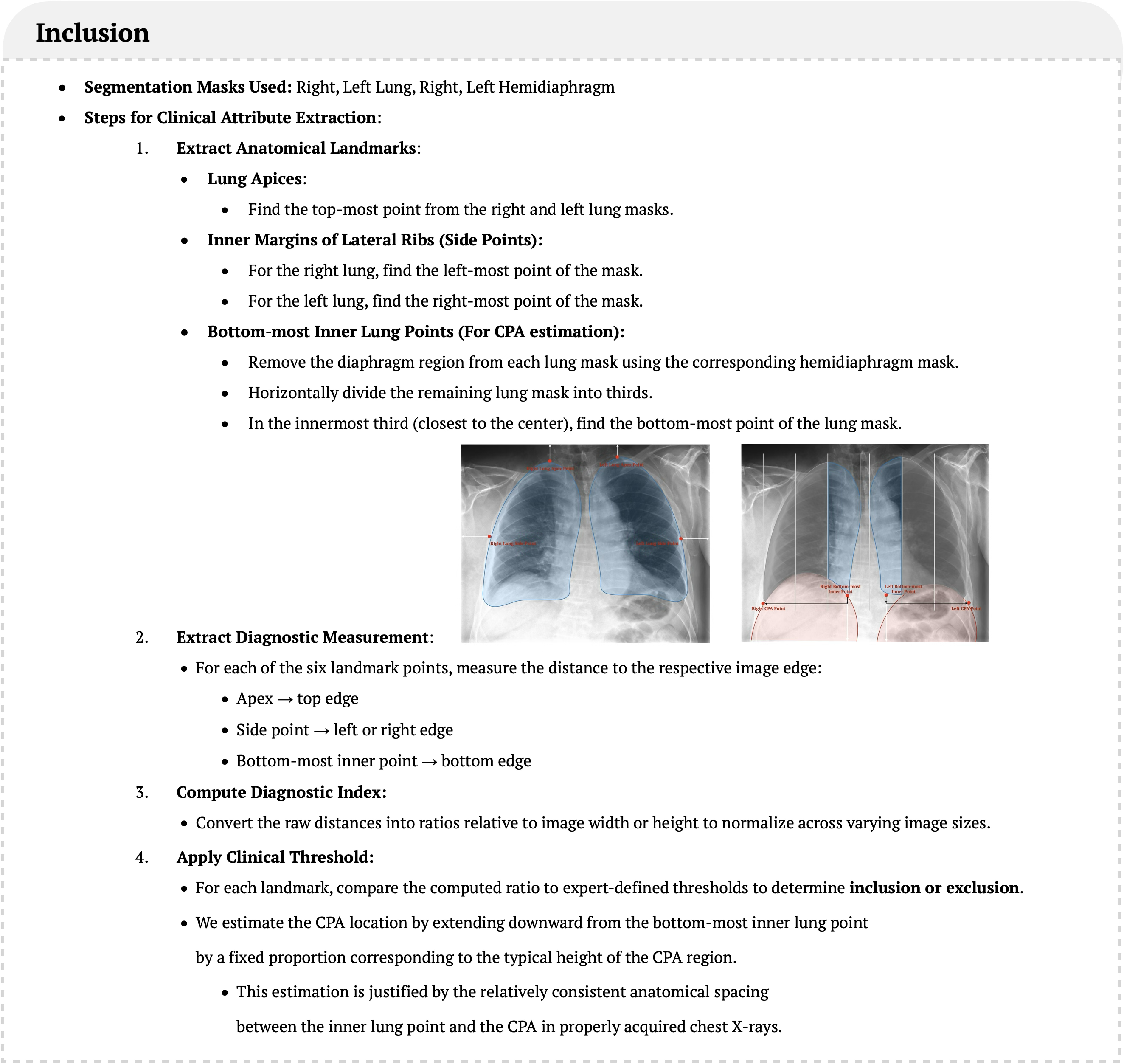}
    \caption{
    Determination of anatomical inclusion by verifying visibility of key thoracic landmarks.
    }
\label{figure:chexstruct_inclusion}
\end{figure}

\begin{figure}[htb!]
    \centering
    \includegraphics[width=\linewidth]{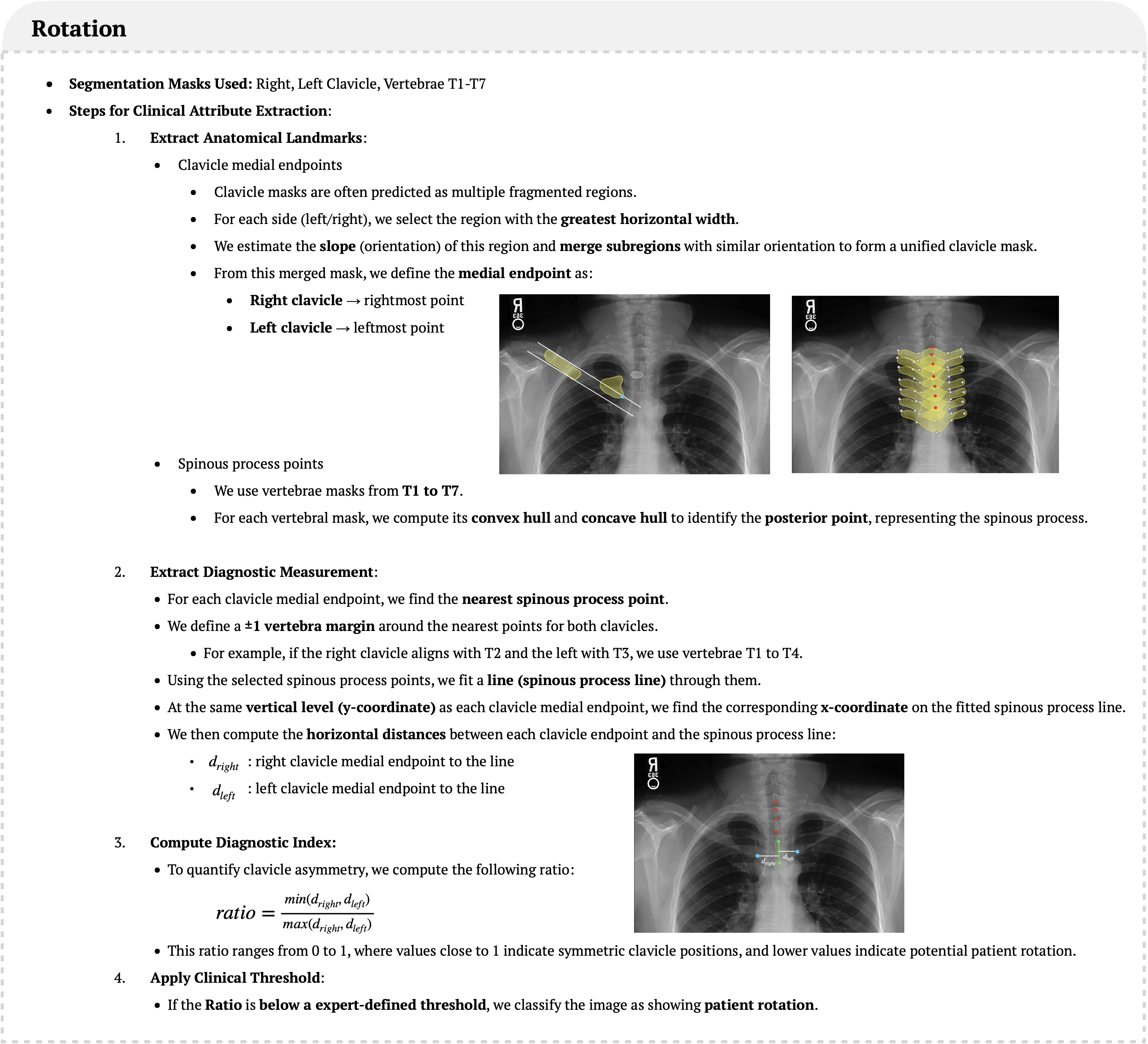}
    \caption{
    Assessment of patient rotation using clavicle symmetry relative to spinal midline.
    }
\label{figure:chexstruct_rotation}
\end{figure}

\begin{figure}[htb!]
    \centering
    \includegraphics[width=\linewidth]{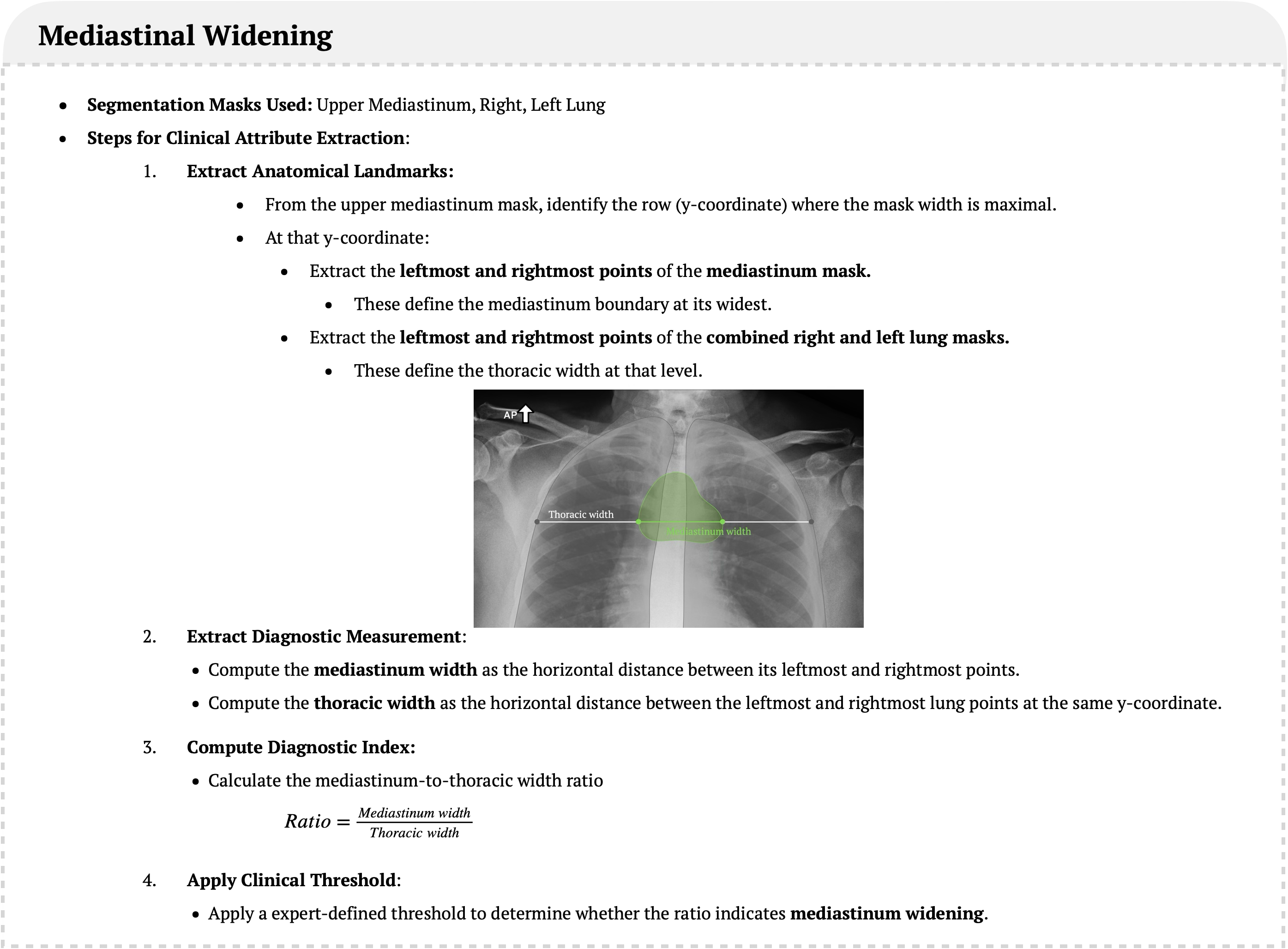}
    \caption{
    Measurement of mediastinal widening using the mediastinum-to-thoracic width ratio.
    }
\label{figure:chexstruct_mw}
\end{figure}

\begin{figure}[htb!]
    \centering
    \includegraphics[width=\linewidth]{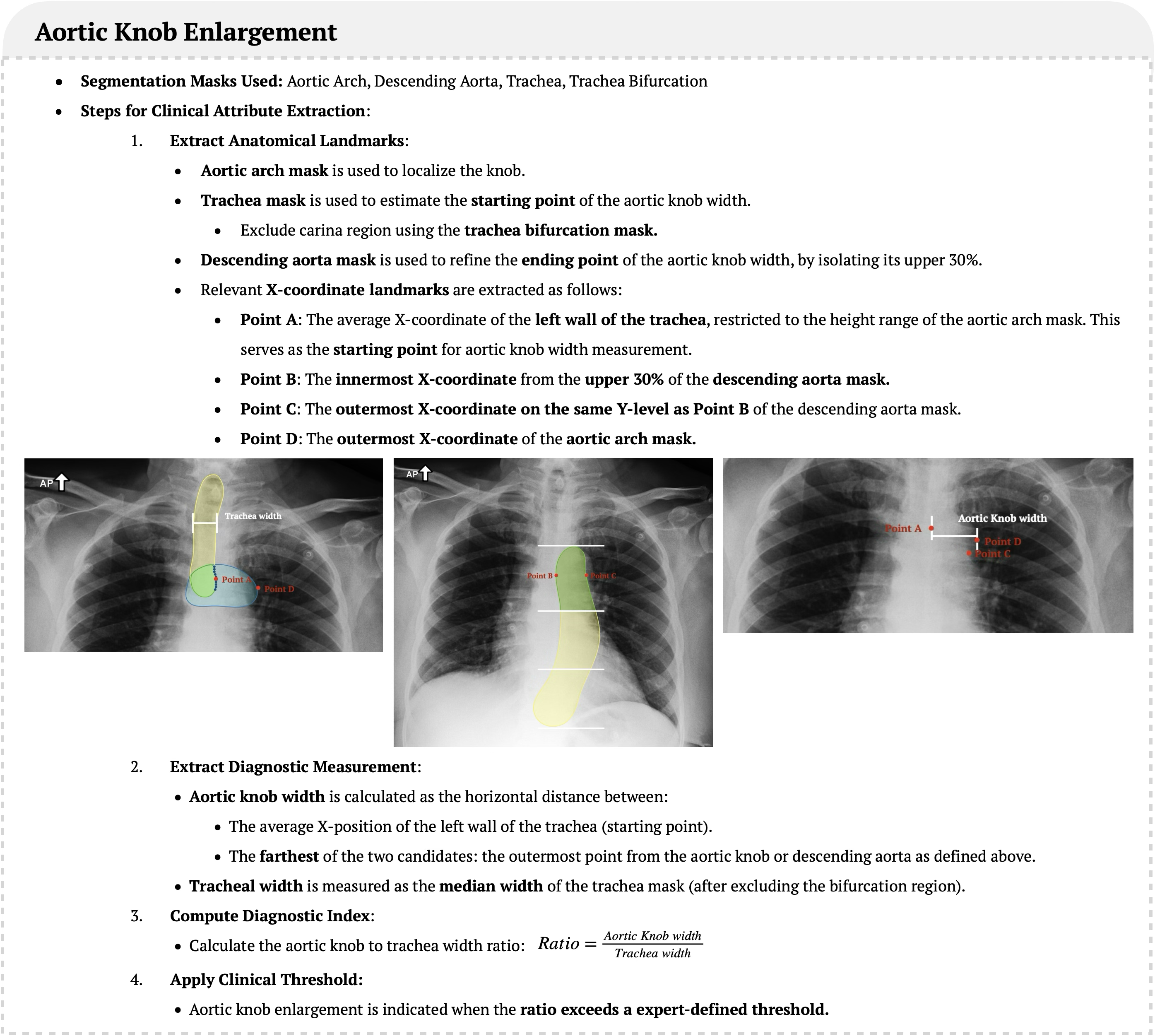}
    \caption{
    Quantification of aortic knob enlargement using the ratio of aortic knob width to tracheal width.
    }
\label{figure:chexstruct_aorticknob}
\end{figure}

\begin{figure}[htb!]
    \centering
    \includegraphics[width=\linewidth]{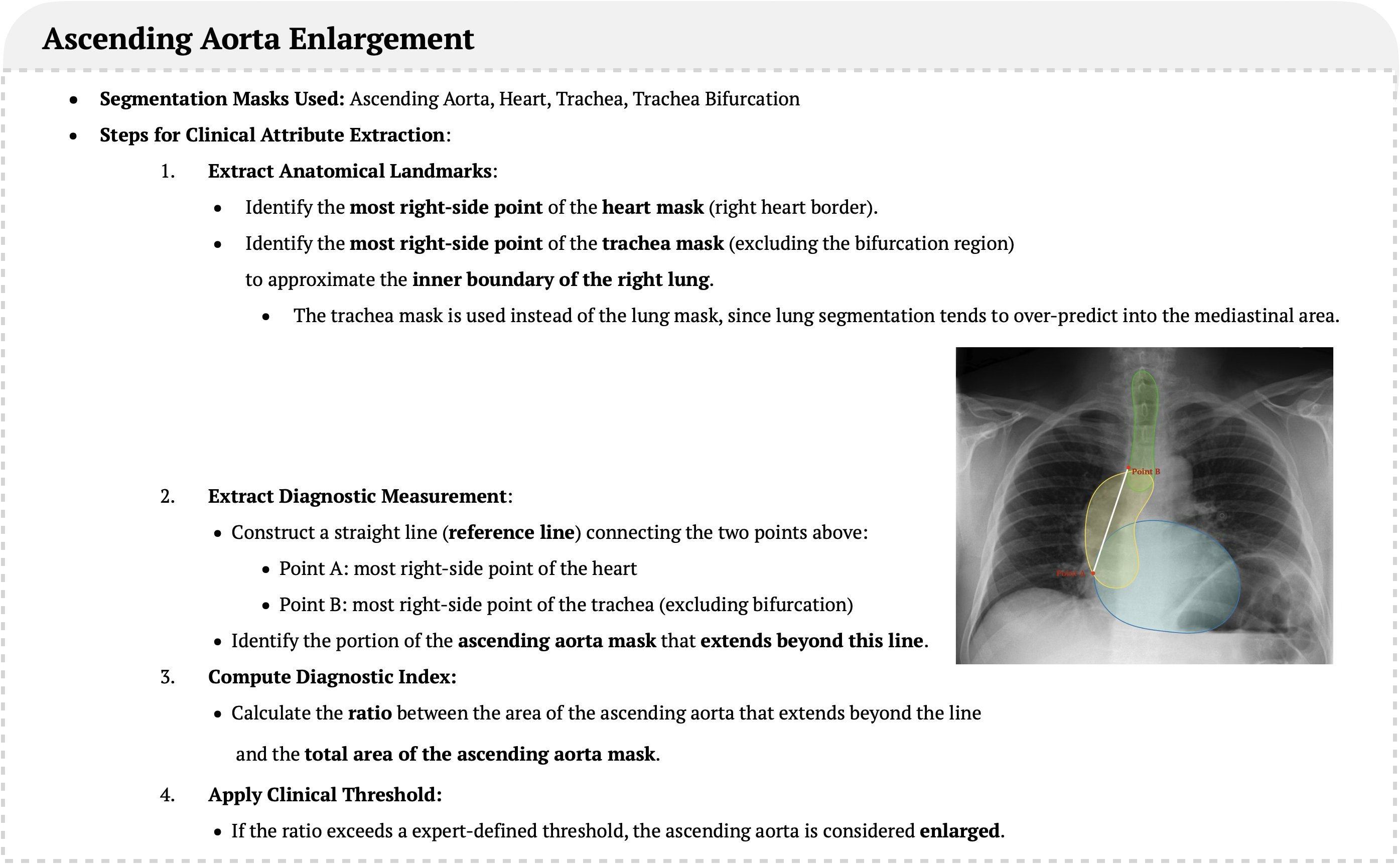}
    \caption{
    Assessment of ascending aorta enlargement based on its extension beyond a defined boundary.
    }
\label{figure:chexstruct_ascaorta}
\end{figure}

\begin{figure}[htb!]
    \centering
    \includegraphics[width=\linewidth]{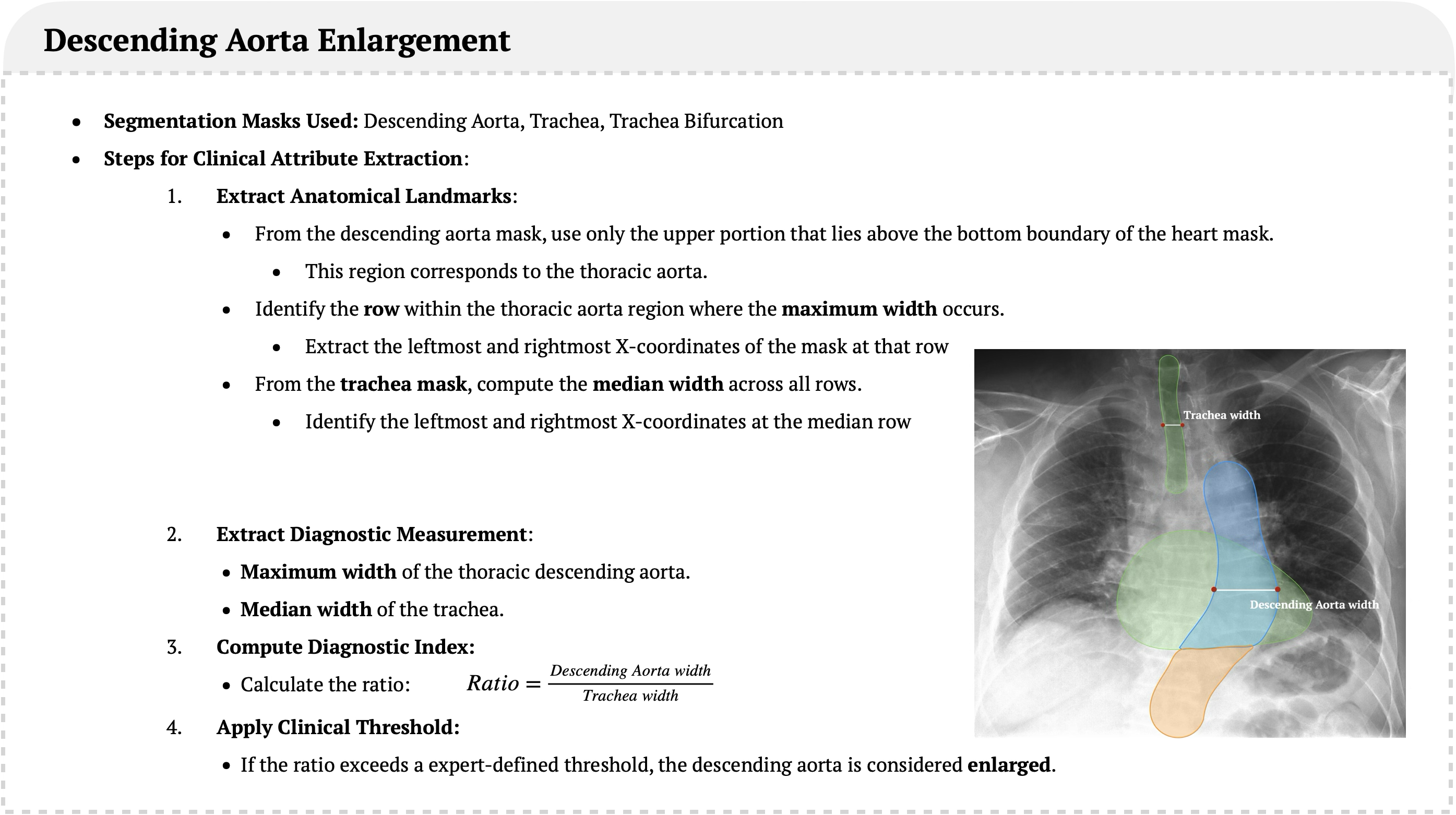}
    \caption{
    Measurement of descending aorta enlargement via the descending aorta width-to-tracheal-width ratio.
    }
\label{figure:chexstruct_dscaorta}
\end{figure}

\begin{figure}[htb!]
    \centering
    \includegraphics[width=\linewidth]{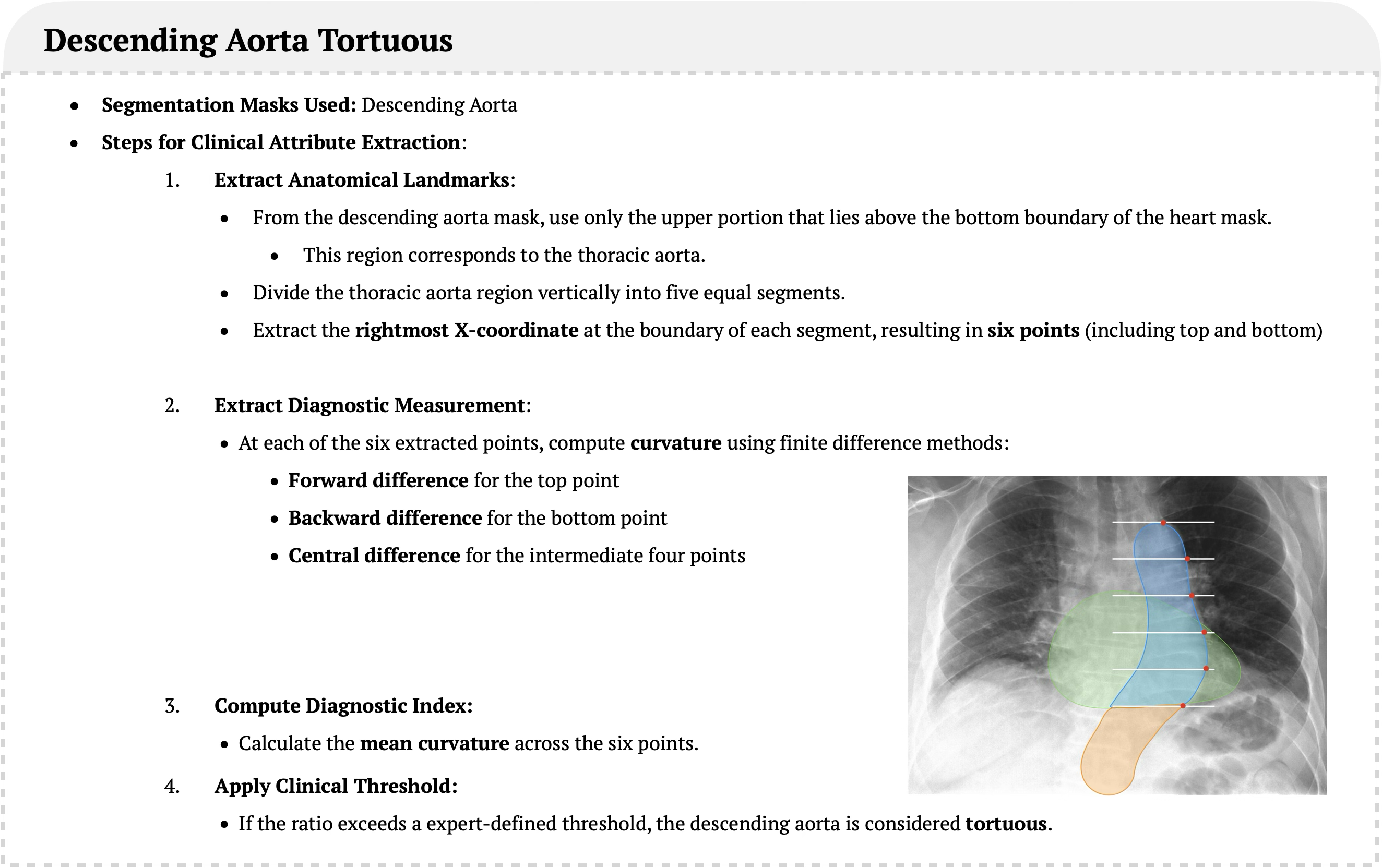}
    \caption{
    Quantitative evaluation of descending aorta tortuous using curvature derived from coordinates.
    }
\label{figure:chexstruct_tortuous}
\end{figure}

\begin{figure}[htb!]
    \centering
    \includegraphics[width=\linewidth]{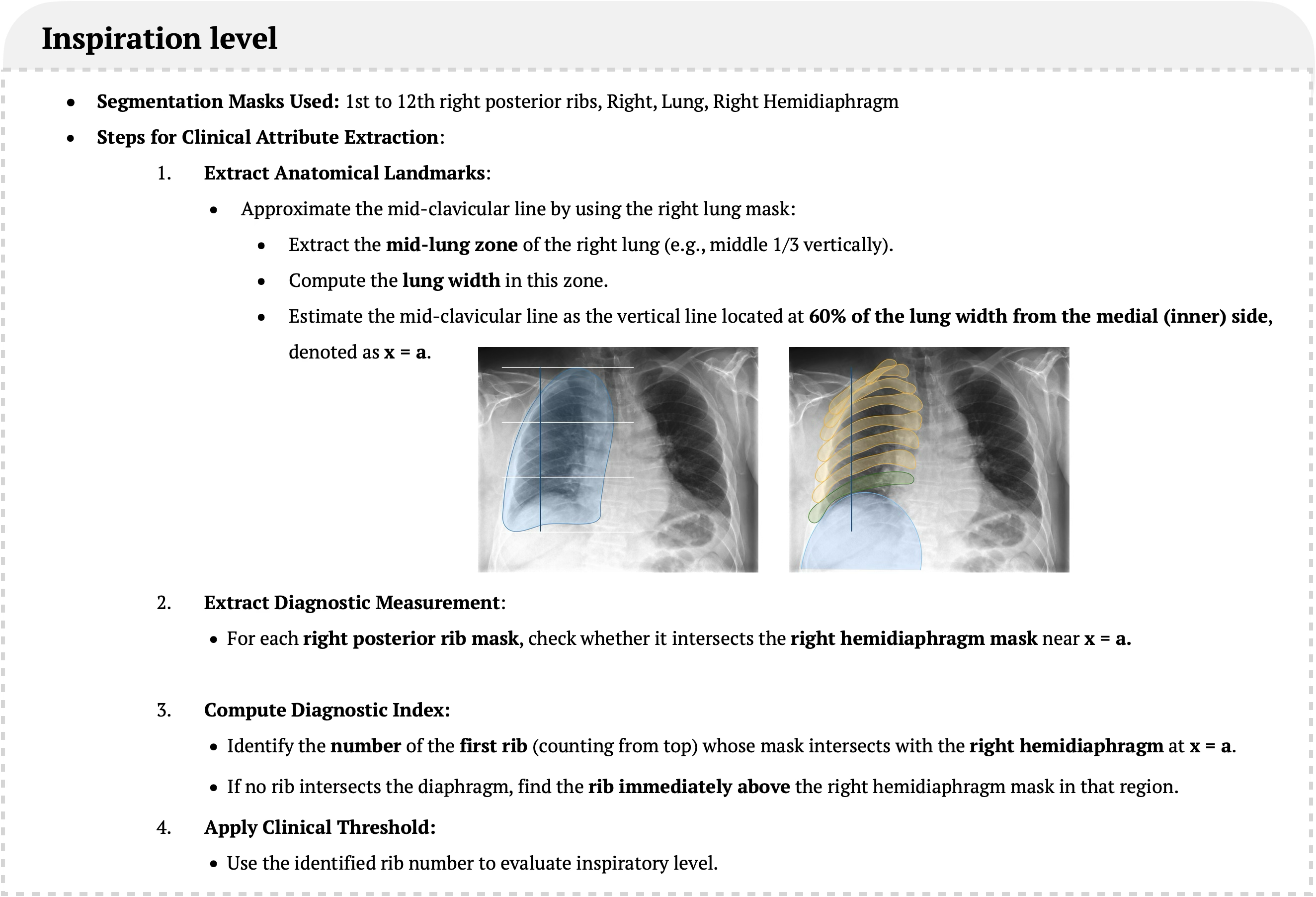}
    \caption{
    Quantification of inspiration level by counting posterior ribs intersecting the diaphragm along the mid-clavicular line.
    }
\label{figure:chexstruct_inspiration}
\end{figure}

Each of the 12 diagnostic tasks is grounded in clinical measurement guidelines, which fall into one of two categories:
\begin{itemize}[leftmargin=*]
    \item \textbf{Standardized, Quantifiable Criteria}
    For tasks with well-established criteria that do not depend on imaging metadata (\textit{e.g.}, pixel spacing that represents the real-world distance of each pixel), we adopt standard clinical measurement rules.
    
    \begin{itemize}[leftmargin=*]
        \item \textbf{Cardiomegaly} refers to an enlarged heart silhouette on chest X-rays. It is diagnosed by calculating the cardiothoracic ratio (CTR), defined as the maximal horizontal cardiac diameter divided by the maximal horizontal thoracic diameter.
        In the PA view, a CTR exceeding 0.50 indicates cardiomegaly.
        See Figure~\ref{figure:chexstruct_cardiomegaly}.
    
        \item \textbf{Carina Angle} represents the angle formed at the bifurcation of the trachea into the right and left main bronchi.
        It is measured directly at the point of bifurcation, typically ranging between 40-80 degrees, although normal values may vary slightly across literature.
        See Figure~\ref{figure:chexstruct_carina}.
        
        \item \textbf{Trachea Deviation} refers to the lateral displacement of the trachea from the midline on a chest X-ray.
        It is assessed by determining whether the trachea lies along the midline defined by the spinous processes or is deviated from it.
        See Figure~\ref{figure:chexstruct_tracheadev}.
        
        \item \textbf{Inclusion} describes whether all relevant thoracic anatomy is fully captured in the image.
        Proper inclusion is confirmed when the image shows the lung apices, inner margins of the lateral ribs, and costophrenic angles.
        See Figure~\ref{figure:chexstruct_inclusion}.
        
        \item \textbf{Rotation} refers to the misalignment of the patient during image acquisition.
        It is evaluated by checking the symmetry of the clavicles relative to the spinous processes.
        The spinous processes should be equidistant from the medial ends of both clavicles; any asymmetry indicates patient rotation.
        See Figure~\ref{figure:chexstruct_rotation}.
    \end{itemize}

    \item \textbf{Expert-Defined Criteria}
    To ensure reliable evaluation in tasks lacking standardized criteria or involving measurement ambiguity (\textit{e.g.}, when diagnosis depends on absolute measurements derived from imaging metadata), we collaborated with clinical experts to define quantifiable, clinically meaningful criteria that guide structured diagnostic reasoning.
    \begin{itemize}[leftmargin=*]
        \item \textbf{Mediastinal Widening} refers to abnormal broadening of the mediastinum. Traditionally defined as a mediastinal width >8~cm at the aortic arch level on PA chest X-rays, this absolute measurement is not feasible in image-only settings. Instead, we assess the ratio of mediastinal width to thoracic width at the same level, providing a consistent and image-based standard. See Figure~\ref{figure:chexstruct_mw}.
    
        \item \textbf{Aortic Knob Enlargement} refers to the prominence of the aortic arch along the left mediastinal border.
        Traditionally, it has been assessed through visual estimation, which is susceptible to inter-observer variability.
        We instead quantify it as the ratio between the maximum width of the aortic knob and the median width of the trachea.
        The trachea is chosen as a reference due to its stable anatomy, ensuring reproducible assessment.
        See Figure~\ref{figure:chexstruct_aorticknob}.
    
        \item \textbf{Ascending Aorta Enlargement} indicates abnormal dilation of the ascending aorta, visible along the right mediastinal border.
        Conventional evaluation relied on subjective visual inspection, which lacked standardized criteria.
        We define enlargement based on whether the aorta extends beyond an imaginary line connecting the right heart border and the inner margin of the right lung, providing a consistent and interpretable rule. See Figure~\ref{figure:chexstruct_ascaorta}.

        \item \textbf{Descending Aorta Enlargement} denotes widening of the descending thoracic aorta, often seen as a prominent left paraspinal contour.
        Visual estimation was the conventional approach, but it often led to inconsistency.
        We introduce a ratio-based measurement between the aorta’s maximum width and the median tracheal width, offering a consistent evaluation.
        See Figure~\ref{figure:chexstruct_dscaorta}.
    
        \item \textbf{Descending Aorta Tortuous} describes excessive curvature of the descending thoracic aorta.
        It was previously evaluated by subjective visual impression, leading to variability.
        To overcome this, we quantify tortuosity by dividing the aorta into five equal-length sections and annotating six coordinates.
        Curvature at each point is computed using finite difference methods (central, forward, and backward difference), enabling objective and reproducible evaluation.
        See Figure~\ref{figure:chexstruct_tortuous}.
    
        \item \textbf{Inspiration Level} assesses the degree of lung expansion during image acquisition, as observed on a chest X-ray.
        Adequate inspiration is typically indicated by the visualization of 9–10 posterior ribs above the diaphragm.
        It is measured by counting the number of right posterior ribs intersecting the right hemidiaphragm along the mid-clavicular line.
        Originally, this criterion lacked a clearly defined reference line, which led to inconsistency in rib counts and variability across evaluators.
        By introducing the mid-clavicular line as a standardized landmark, we ensure more consistent and reproducible assessment.
        See Figure~\ref{figure:chexstruct_inspiration}.
    
        \item \textbf{Projection} refers to the orientation of the X-ray beam relative to the patient, typically posteroanterior (PA) or anteroposterior (AP) on chest X-rays.
        Originally, projection was assessed by visually inspecting scapular positioning, whether the scapulae were retracted (suggesting PA) or overlapping the lung fields (suggesting AP).
        However, this method lacked clear, objective criteria, leading to variability across evaluators.
        To address this, we compute the ratio of the overlapping area between each scapula and the lung field to the scapular area.
        A higher overlap ratio indicates an AP view. This quantitative method provides a standardized and reproducible approach to classification. See Figure~\ref{figure:chexstruct_projection}.
        
    \end{itemize}
\end{itemize}

\subsubsection{Segmenting Anatomical Structures}
We define our diagnostic tasks based on the anatomical structures segmented by CXAS~\cite{seibold2023accurate}, a chest X-ray anatomy segmentation model trained on high-quality, expert-curated data from the PAX-Ray++ dataset.
PAX-Ray++\cite{seibold2023accurate} was constructed through a multi-stage pipeline: expert-annotated CT datasets were aggregated, used to train nnUNet models, and refined via inference on additional datasets such as AutoPET.
These models were then applied to thoracic CT volumes to generate accurate 2D anatomical projections, which were post-processed to ensure consistency and label quality.
The final dataset contains 14,754 projected X-ray images with 157 anatomical structures.
Trained on this dataset, CXAS enables accurate multi-class segmentation of anatomical structures in chest X-rays.
Table\ref{table:cxas_performance} presents the performance of CXAS on the PAX-Ray++ test set.

\begin{figure}[htb!]
    \centering
\includegraphics[width=0.9\linewidth]{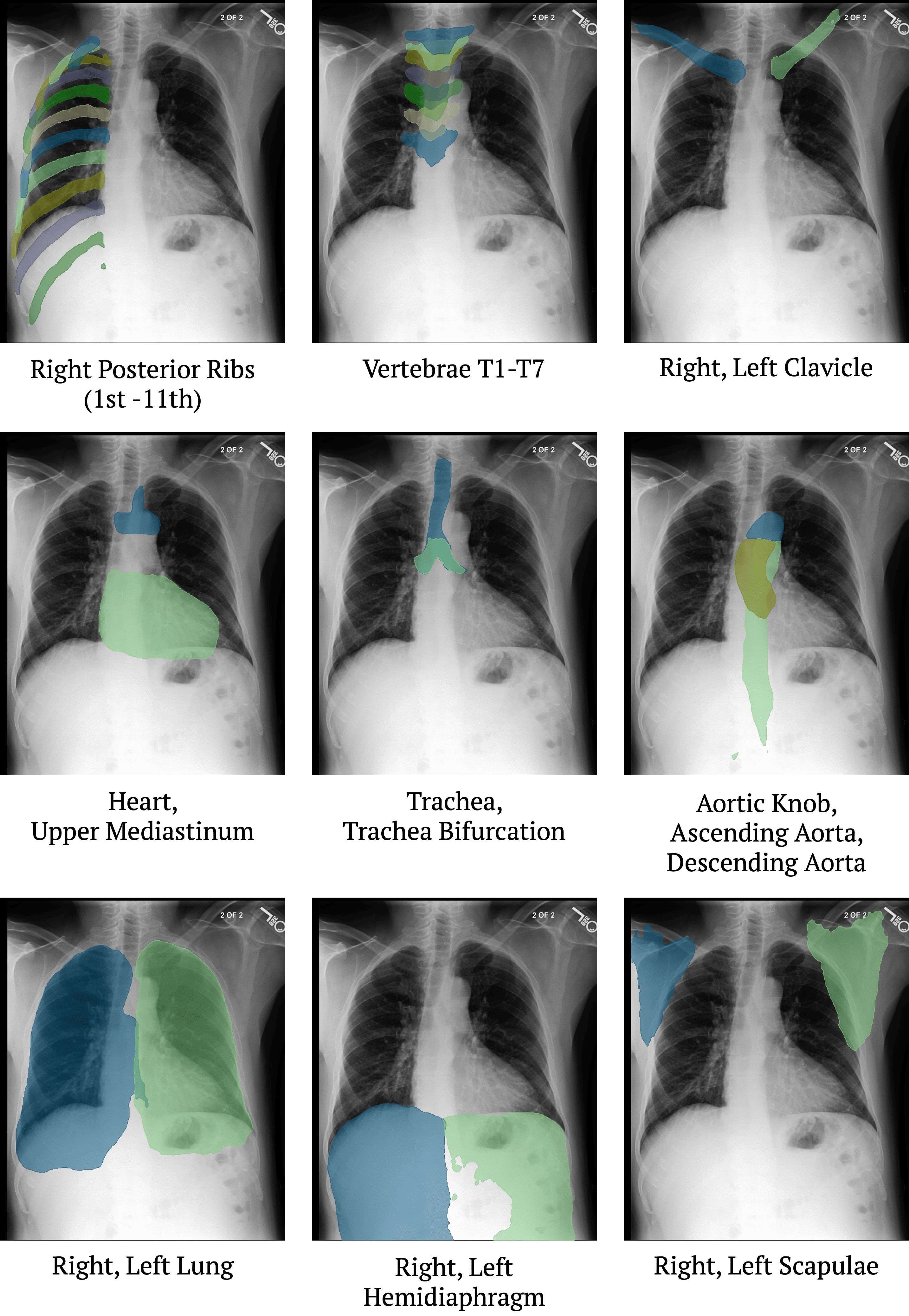}
    \caption{
    Visualization of the anatomical structures of CXAS used in \extractor.
    }
\label{figure:cxas_masks}
\end{figure}

Figure~\ref{figure:cxas_masks} and Table~\ref{table:chexstruct_anatomy} present the subset of anatomical structures segmented by CXAS that were used in \extractor for each diagnostic task.

\input{Tables/Supple_CheXStruct_Bodyparts}
\input{Tables/Supple_valid_num_sample_per_dx}  
\input{Tables/Supple_CXAS_performance}

\begin{figure}[htb!]
    \centering
    \includegraphics[width=\linewidth]{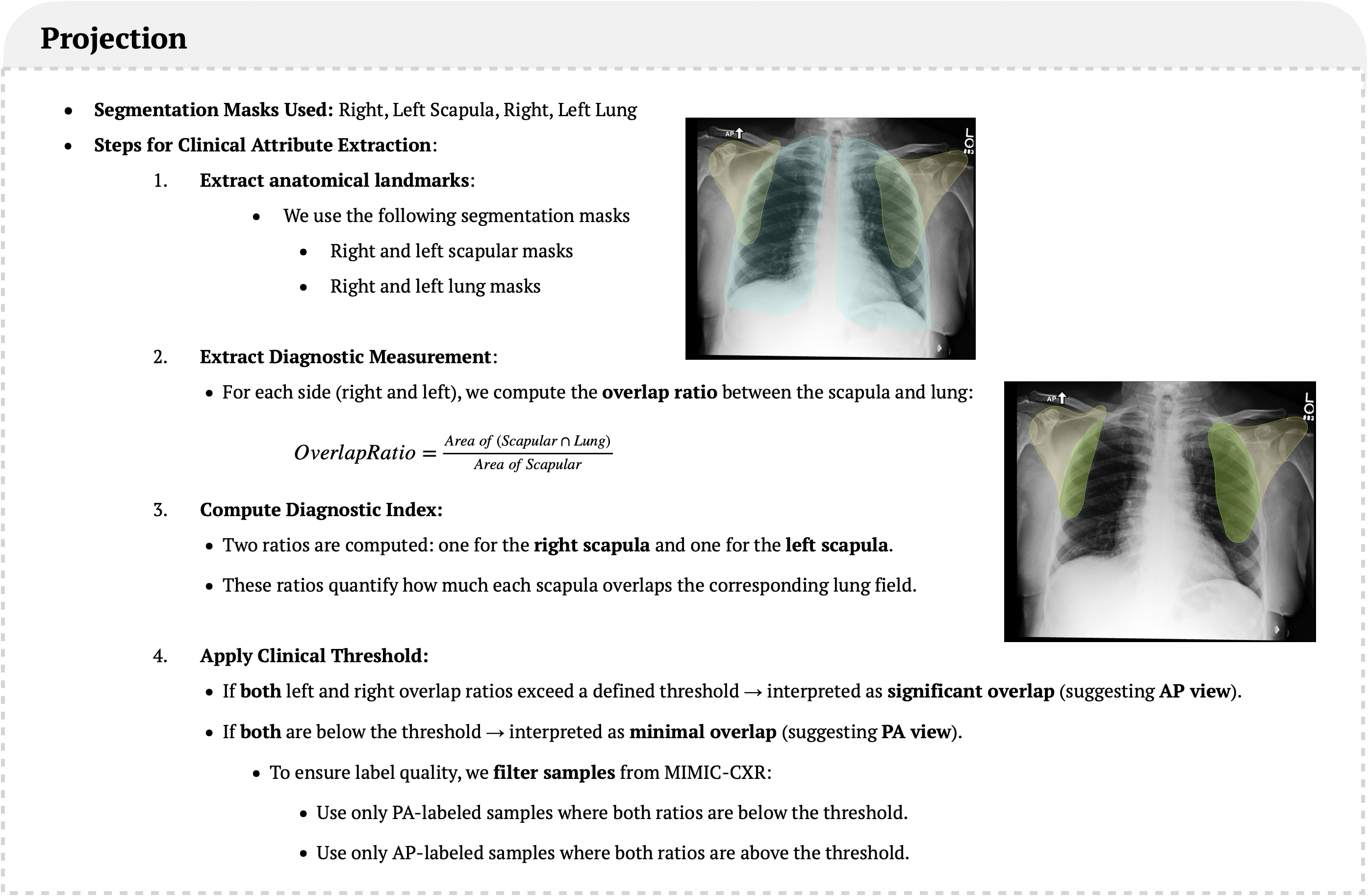}
    \caption{
    Assessment of projection type (PA vs. AP) via scapula-lung overlap ratio.
    }
\label{figure:chexstruct_projection}
\end{figure}

%% file: Tables/Supple_CheXStruct_Bodyparts.tex
\begin{table}[htb!]
  \caption{Anatomical structures of CXAS used in \extractor per diagnostic task}
  \label{table:chexstruct_anatomy}
  \centering
  \begin{tabular}{llll}
    \toprule
    Task     & Anatomical Structures  \\
    \midrule
    Cardiomegaly & Heart, Right Lung, Left Lung \\
    Mediastinal Widening & Upper Mediastinum, Right Lung, Left Lung \\
    Trachea Deviation & Trachea, Trachea Bifurcation, Vertebrae T1--T7 \\
    Carina Angle & Trachea Bifurcation \\
    Aortic Knob Enlargement & Aortic Arch, Descending Aorta, Trachea, Trachea Bifurcation \\
    Ascending Aorta Enlargement & Ascending Aorta, Heart, Trachea, Trachea Bifurcation \\
    Descending Aorta Enlargement & Descending Aorta, Trachea, Trachea Bifurcation \\
    Descending Aorta Tortuous & Descending Aorta \\
    Inclusion & Right Lung, Left Lung, Right Hemidiaphragm, Left Hemidiaphragm \\
    Inspiration Level & Right Posterior 1st--11th Ribs, Right Lung, Right Hemidiaphragm \\
    Rotation & Clavicle Right, Clavicle Left, Vertebrae T1--T7 \\
    Projection & Scapula Right, Scapula Left, Right Lung, Left Lung \\
    
    \bottomrule
  \end{tabular}
\end{table}

%% file: Tables/Supple_valid_num_sample_per_dx.tex
\begin{table}[htb!]
  \caption{
  Number of cases and diagnostic label distribution per diagnostic task after quality control.
  }
  \label{table:chexstruct_label_distribution}
  \centering
  \begin{tabular}{ll}
    \toprule
    Diagnostic Task     & Total Cases (Label Distribution) \\
    \midrule
    Cardiomegaly & 184,169 (Normal: 112,329, Abnormal: 71,840) \\
    Mediastinal Widening & 174,905 (Normal: 118,092, Abnormal: 56,813)\\
    Trachea Deviation & 66,588 (Normal: 57,878, Deviated: 8,710) \\
    Carina Angle & 122,085 (Normal: 29,010, Abnormal: 93,075)\\
    Aortic Knob Enlarged & 174,949 (Normal: 156,183, Abnormal: 18,766)\\
    Ascending Aorta Enlarged & 75,009 (Normal: 71,126, Abnormal: 3,883)\\
    Descending Aorta Enlarged & 106,634 (Normal: 106,274, Abnormal: 360)\\
    Descending Aorta Tortuous & 120,330 (Normal: 116,420, Abnormal: 3,910) \\
    Inclusion & 144,221 (Included: 132,594, Excluded: 11,627) \\
    Inspiration Level & 148,222 (Good: 127,848, Poor: 20,374) \\
    Rotation & 51,095 (Not Rotated: 41,155, Rotated: 9,940)\\
    Projection & 92,938 (Small Overlap: 53,524, Large Overlap: 39,414) \\
    
    \bottomrule
  \end{tabular}
\end{table}

%% file: Tables/Supple_CXAS_performance.tex
\begin{table}[htb!]
  \caption{Performance of CXAS on PAX-Ray++}
  \label{table:cxas_performance}
  \noindent 
  \begin{minipage}[t]{0.4\textwidth}
  \raggedright
  \vspace{0pt}
  \begin{tabular}{llll}
  \toprule
  Anatomical structure     & F1     & IoU  & Dice \\
  \midrule
    Right Lung & 0.98 & 0.96 & 0.98 \\
    Left Lung & 0.98 & 0.96 & 0.98 \\
    Right Hemidiaphragm & 0.97 & 0.95 & 0.97 \\
    Left Hemidiaphragm & 0.96 & 0.93 & 0.96 \\
    Heart & 0.97 & 0.94 & 0.97 \\
    Trachea & 0.91 & 0.84 & 0.91 \\
    Tracheal Bifurcation & 0.87 & 0.78 & 0.87 \\
    Scapula Right & 0.96 & 0.92 & 0.96 \\
    Scapula Left & 0.95 & 0.92 & 0.95 \\
    Clavicle Right & 0.90 & 0.83 & 0.90 \\
    Clavicle Left & 0.90 & 0.82 & 0.90 \\
    Aortic Arch & 0.89 & 0.81 & 0.89 \\
    Ascending Aorta & 0.90 & 0.83 & 0.90 \\
    Descending Aorta & 0.93 & 0.87 & 0.93 \\
    Upper Mediastinum & 0.82 & 0.72 & 0.82 \\
  \bottomrule
  \end{tabular}
  \end{minipage}
  \hfill
  \begin{minipage}[t]{0.4\textwidth}
  \raggedright
  \vspace{0pt}
  \begin{tabular}{llll}
  \toprule
  Anatomical structure     & F1     & IoU  & Dice \\
  \midrule
    Vertebrae T1 & 0.89 & 0.82 & 0.89 \\
    Vertebrae T2 & 0.89 & 0.81 & 0.89 \\
    Vertebrae T3 & 0.88 & 0.79 & 0.88 \\
    Vertebrae T4 & 0.87 & 0.79 & 0.87 \\
    Vertebrae T5 & 0.86 & 0.78 & 0.86 \\
    Vertebrae T6 & 0.85 & 0.76 & 0.85 \\
    Vertebrae T7 & 0.85 & 0.75 & 0.85 \\

    Posterior 1st Rib Right & 0.72 & 0.64 & 0.72 \\
    Posterior 2nd Rib Right & 0.74 & 0.65 & 0.74 \\
    Posterior 3rd Rib Right & 0.72 & 0.65 & 0.72 \\
    Posterior 4th Rib Right & 0.73 & 0.66 & 0.73 \\
    Posterior 5th Rib Right & 0.74 & 0.68 & 0.74 \\
    Posterior 6th Rib Right & 0.74 & 0.68 & 0.74 \\
    Posterior 7th Rib Right & 0.75 & 0.69 & 0.75 \\
    Posterior 8th Rib Right & 0.76 & 0.70 & 0.76 \\
    Posterior 9th Rib Right & 0.77 & 0.71 & 0.77 \\
    Posterior 10th Rib Right & 0.78 & 0.72 & 0.78 \\
    Posterior 11th Rib Right & 0.77 & 0.70 & 0.77 \\
  \bottomrule
  \end{tabular}
  \end{minipage}
\end{table}

%% file: Supple/1.CheXStruct_QC.tex
\clearpage

\subsection{Quality Control}\label{apd:chexstruct_QC}

To ensure the reliability and anatomical validity of the extracted clinical information, we apply a multi-step quality control (QC) process.
This includes (1) global image-level filtering, applied uniformly across all tasks, and (2) task-specific QC criteria tailored to the anatomical and clinical requirements of each diagnostic task.

All quality control criteria were developed in close collaboration with clinical experts, who also determined appropriate thresholds based on their clinical knowledge and expectations regarding anatomical characteristics.

\subsubsection{Global Filtering of Raw Images}
We use the MIMIC-CXR-JPG dataset~\cite{johnson2019mimic} as our base image source.
Initially, we select all images labeled with a ViewPosition of PA or AP in the metadata.
However, we observed that these labels alone are not sufficient to ensure the validity of the chest X-ray content.

To construct a clean and trustworthy benchmark, we apply the following multi-step filtering process to exclude non-relevant or corrupted images:

\textbf{Step1. Segmentation-Based Filtering Using CXAS}
We utilize CXAS~\cite{seibold2023accurate}, a chest X-ray anatomy segmentation model that can segment up to 157 anatomical structures relevant to chest X-ray.
For each image, we apply CXAS to obtain the corresponding segmentation masks.

We discard images in which the number of segmented anatomical structures falls below a expert-defined threshold.
A low number of detected structures typically indicates one of the following:
\begin{itemize}[leftmargin=*]
    \item The image is not a chest X-ray (\textit{e.g.}, abdominal or spinal radiograph, or an irrelevant modality).
    
    \item The image is corrupted or blank (\textit{e.g.}, all black, all white, or contains artifacts).
    
    \item Anatomical structures are not clearly visible due to poor patient positioning or motion artifacts.
\end{itemize}

\begin{figure}[htb!]
    \centering
    \includegraphics[width=0.75\linewidth]{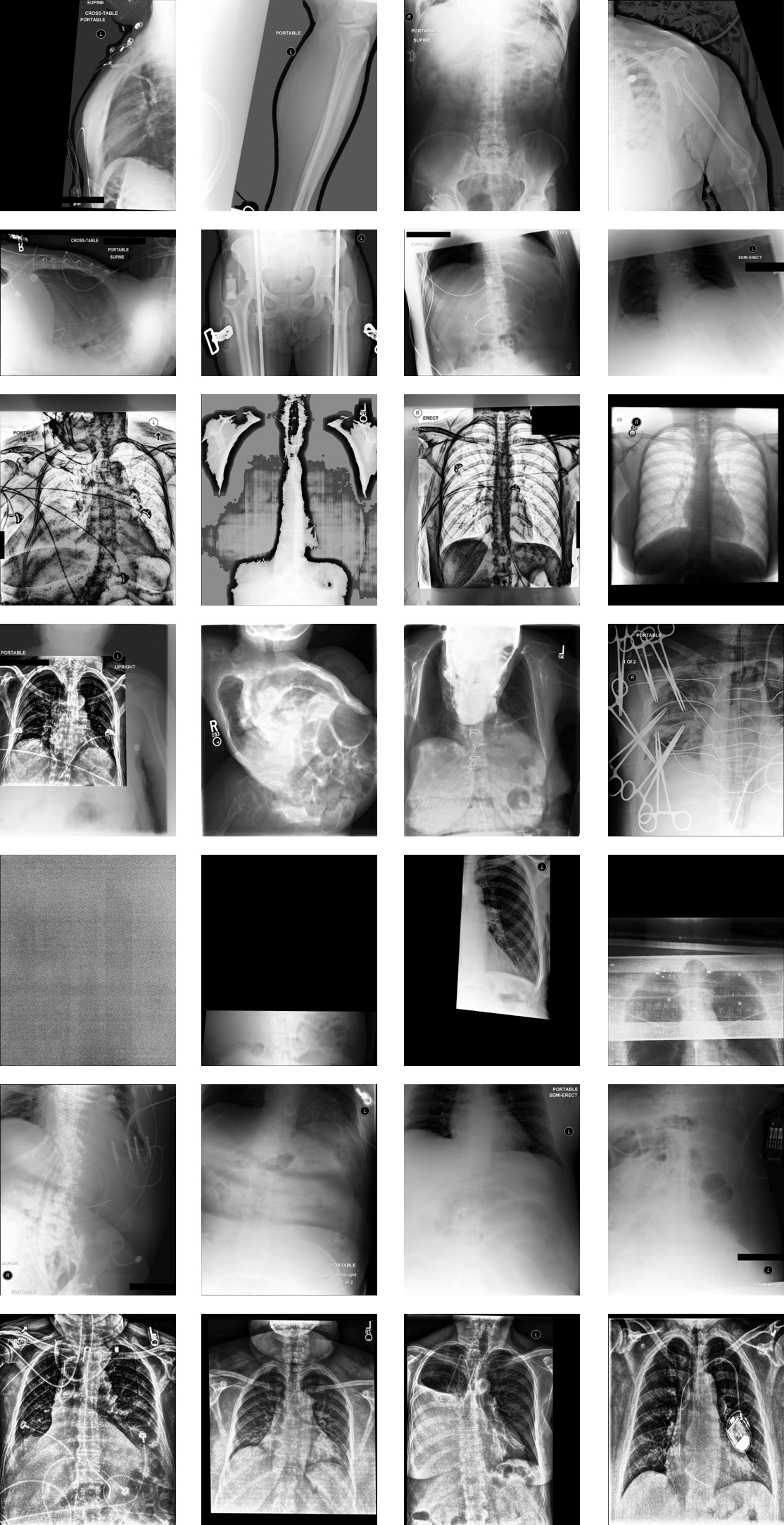}
    \caption{
    \textbf{Representative Examples of Globally Filtered Images.}
    Rows 1–5: Images removed due to a small number of segmented anatomical structures using CXAS (Step 1).
    Row 6: Image removed due to abnormal anatomical positioning of heart mask (Step 2).
    Row 7: Image removed due to post-processing artifacts and low gradient strength (Step 3).
    }
\label{figure:chexstruct_noncxr}
\end{figure}

This step serves as a robust first-pass filter that eliminates obviously invalid images based on anatomical structure segmentation.
See Figure~\ref{figure:chexstruct_noncxr}, Rows 1–5 for representative examples.

\textbf{Step 2: Anatomical Position-Based Filtering (Heart Mask Heuristic)}

For the remaining images, we apply a heuristic based on the vertical position of the heart mask, one of the reliably segmented structures from CXAS.

We compute the vertical ratio of the image above and below the heart mask.
If the heart is located disproportionately high in the image (\textit{i.e.}, a large portion of the image appears below the heart), this often indicates that the image contains substantial abdominal or pelvic anatomy.

These images are excluded based on this anatomical positioning heuristic.
See Figure~\ref{figure:chexstruct_noncxr}, Row 6 for representative examples.

\textbf{Step 3: Gradient-Based Filtering to Remove Post-Processing Artifacts}

Even after anatomical filtering, some images suffer from post-processing artifacts, such as low contrast or improper windowing adjustments that obscure anatomical structures.

To detect such cases, we compute the average edge strength across each image using gradient magnitude:
\begin{itemize}[leftmargin=*]
    \item Horizontal and vertical edge maps are computed using the Sobel operator.
    
    \item Gradient magnitude is calculated to quantify edge strength.
    
    \item The mean gradient magnitude across the image is used as a filtering metric.
\end{itemize}

Images with abnormally low edge strength typically lack sufficient anatomical contrast and are deemed visually uninformative.
These are removed during this step.
See Figure~\ref{figure:chexstruct_noncxr}, Row 7 for representative examples.

After applying the global filtering steps, we excluded 7,587 images, resulting in 235,747 high-quality PA and AP chest X-rays.
This corresponds to 3.12\% of the images initially labeled as PA or AP in the metadata being removed due to quality concerns.

\clearpage

\subsubsection{Task-Specific Quality Control}
To ensure reliable measurements for each diagnostic task, we apply task-specific quality control (QC) criteria following global filtering steps. These QC criteria are designed to verify that the anatomical structures relevant to each task are accurately segmented and that the extracted clinical information derived from the segmentation masks is also reliable. The QC criteria are applied based on the task-specific masks and extracted landmark points used in each diagnostic task, as described in Section~\ref{apd:chexstruct_criteria}.

\begin{itemize}[leftmargin=*]

    \item \textbf{Inclusion}
    \begin{itemize}[leftmargin=*]
        \item \textbf{Lung Apex Points}

        1) Left–Right Y-Coordinate Symmetry: The absolute vertical difference between the left and right lung apex points is calculated.
        A large discrepancy indicates a potential segmentation error, as the lung apices are typically located at similar vertical levels.
        
        2) Anatomical Plausibility of Vertical Location: The average y-coordinate of the apex points is used to assess their relative vertical positioning within the image. If the apex points are located too far down in the image, outside the typical thoracic boundary, they are considered anatomically implausible and filtered out.

        \item \textbf{Lung Side Points}
        
        1) Relative Width Consistency: For each lung, the width between the outermost (side) point and the nearest inner lung border is measured. Based on the inclusion or exclusion label of each side point, we compute the left-to-right width ratio. Images showing abnormally imbalanced widths are excluded, as this suggests potential segmentation errors or inaccuracies in the extracted anatomical points.
        
        \item \textbf{Lung Bottom Points}
        
        1) Left–Right Y-Coordinate Symmetry:
        Similar to the apex points, a large vertical discrepancy between the left and right bottom lung points suggests possible misidentification or segmentation errors and is used as a criterion for filtering.
        
        2) Anatomical Plausibility of Vertical Location: The bottom points are expected to lie in the lower region of the thoracic cavity. If they are found too high in the image, they are likely erroneous and thus filtered.
    \end{itemize}
    
    \item \textbf{Inspiration Level}
    \begin{itemize}[leftmargin=*]
        \item \textbf{Right Posterior Ribs}
        
        1) Mask Existence: If any of the right posterior rib masks are missing, the image is excluded. The absence of even a single rib mask indicates that other ribs may not be correctly segmented or positioned, compromising anatomical reliability.
        
        2) Overlap Area: Excessive overlap between adjacent rib masks suggests that two ribs were merged or wrongly localized.
        
        3) Inter-Rib Distance: Unusual spacing (either too close or too far) indicates segmentation failure or anatomical misalignment.

        \item \textbf{Right Hemidiaphragm}
        If the diaphragm mask inappropriately invades the lung region, the image is excluded, as this indicates a segmentation failure.
        Diaphragm masks located significantly below their expected anatomical position (\textit{e.g.}, well beneath the lung bases) are also excluded, as such positions reflect inaccurate segmentation rather than valid anatomical placement.

        \item \textbf{Right Lung}
        We compare lung masks from CXAS and CheXmask~\cite{gaggion2023chexmaskPhysioNet}.
        If both masks are well-segmented, approximated mid-clavicular lines from each should produce similar results.
        If the mid-clavicular line calculated using the CXAS mask differs significantly from that calculated using the combined CXAS–CheXmask mask, we exclude the image.
        This discrepancy suggests that the lung boundary is poorly segmented in at least one of the masks.
    \end{itemize}

    \item \textbf{Rotation}
    \begin{itemize}[leftmargin=*]
        \item \textbf{Clavicles}

        1) Distance to Lung Inner Edge: The distance between each clavicle medial endpoint and the inner border of the corresponding lung mask is computed. Anatomically implausible distances, either too small (indicating over-penetration into the lung) or too large (indicating external or misplaced points), are filtered out.

        2) Left–Right Y-Coordinate Symmetry:
        The absolute difference in y-coordinates between the left and right medial endpoints is measured. A large vertical discrepancy suggests that one of the points may have been incorrectly detected, possibly outside the clavicle region.

        3) Anatomical Plausibility of Vertical Location:
        The average y-coordinate of the left and right medial endpoints is used to estimate the vertical positioning of the clavicles within the image. Points located too close to the image top or bottom (outside the typical thoracic region) are excluded to ensure anatomical plausibility.

        \item \textbf{Spinous Process}
        
        1) Standard Deviation of Point Positions:
        High variation among vertebral point coordinates indicate imprecise detection. Images with such inconsistencies are excluded.

        2) Residual from Spinal Line Fitting:
        We fit a smooth spinal line to the vertebral points. If the residuals (deviation from the fitted line) are large, it suggests poor alignment or noise in the points, leading to the exclusion of the image.
    \end{itemize}

    \item \textbf{Projection}
    \begin{itemize}[leftmargin=*]
        \item \textbf{Scapular}
        The width and height of the scapular mask must exceed a expert-defined threshold relative to the overall image dimensions.
        Images with scapular masks that are too small or barely visible are excluded, as they indicate incomplete scapular segmentation.

        \item \textbf{Lungs}
        For each image, we locate the horizontal band corresponding to the height of the scapular mask.
        Within this band, we assess the segmented lung regions.
        We require that the lung regions in this zone exhibit sufficient width and height relative to the image dimensions.
        If the lung region is too narrow or compressed within this scapular band, the image is excluded.
    \end{itemize}

    \item \textbf{Cardiomegaly}
    \begin{itemize}[leftmargin=*]
        \item \textbf{Heart}

        1) IoU-Based Filtering:
        We compute the Intersection-over-Union (IoU) between the heart mask and the Chest ImaGenome heart bounding box~\cite{wu2021chest}. Images with low IoU are discarded, as this typically indicates a mismatch in localization, either one of the annotations includes non-heart regions, or fails to capture the heart fully.

        2) Center Alignment Filtering:
        The centroid coordinates of both the heart mask and the bounding box are extracted. Samples with large differences between these centroids are excluded, as such deviations imply at least one of the annotations may be misaligned with the actual heart location.

        \item \textbf{Lungs}
        We include only images labeled as having both left and right lung sides included, obtained during the Inclusion task. The same lung quality filtering criteria used in the Inclusion task are applied here to ensure anatomical plausibility.
    \end{itemize}

    \item \textbf{Mediastinal Widening}
    \begin{itemize}[leftmargin=*]
        \item \textbf{Mediastinum}
        Filtering is conducted using the same criteria as for the heart mask in the Cardiomegaly task.
        The IoU between the upper mediastinum mask and the Chest ImaGenome mediastinum bounding box~\cite{wu2021chest} must exceed a expert-defined threshold to ensure both are localized to the correct anatomical region.
        The distance between their centroids must be within an acceptable range, as a large discrepancy may indicate failed segmentation or misaligned structure prediction.
        
        \item \textbf{Lungs}
        We include only images labeled as having both left and right lung sides included, obtained during the Inclusion task. The same lung quality filtering criteria used in the Inclusion task are applied here to ensure anatomical plausibility.
    \end{itemize}

    \item \textbf{Trachea Deviation}
    \begin{itemize}[leftmargin=*]
        \item \textbf{Trachea}

        1) Height: The total height of the trachea mask must exceed a minimum threshold to ensure the full trachea structure is captured within the thorax.

        2) Normalized Width: The median width of the mask (computed across rows and normalized by image width) must fall within an anatomically reasonable range.

        3) Width Variability: The standard deviation of the row-wise width should be low, as large variations indicate irregular or failed segmentation.
        
        \item \textbf{Midline}
        The same midline estimation and quality filtering strategy used in the Rotation task is applied here.

    \end{itemize}

    \item \textbf{Carina Angle}
    \begin{itemize}[leftmargin=*]
        \item \textbf{Carina}

        1) Carina Mask Shape: We filter out images where the carina mask has an abnormal aspect ratio, either too narrow, too wide, too short, or too tall, as these are likely mis-segmentations.

        2) Carina Point Location: For a well-predicted mask, the carina point should lie near the center of the mask. If the point is too far from the center, the mask or point may be misaligned.

        3) Angle Stability: We compute carina angles using the carina point and three leftmost and three rightmost edge points of the mask. Samples with high standard deviation across these angles are excluded, as they indicate unreliable estimation.

        4) Vertical Alignment: The coordinate difference between the carina point and the topmost point of the mask should be small, a large difference suggests an inaccurately predicted carina point.
    \end{itemize}

    \item \textbf{Aortic Knob Enlargement}
    \begin{itemize}[leftmargin=*]
        \item \textbf{Aortic Knob}

        1) Mask Alignment Check: We compare the x-coordinates of the outermost points predicted from the aortic arch mask and the descending aorta mask. If these coordinates are similar, it suggests that both masks are correctly segmented.

        2) Descending Aorta Upper Region Height: We measure the height of the upper 30\% portion of the descending aorta mask. A mask that is too short in this region is likely not properly segmented, leading to an inaccurate estimation of the aortic border.

        \item \textbf{Trachea}
        The trachea mask is filtered using the same criteria described in the Trachea Deviation task.
    \end{itemize}
    
    \item \textbf{Ascending Aorta Enlargement}
    \begin{itemize}[leftmargin=*]
        \item \textbf{Ascending Aorta}

        1) Height: The total height of the ascending aorta mask must exceed a expert-defined threshold. A short mask may indicate improper segmentation.

        2) Relative Vertical Position: We compute the vertical distance between the bottom-right point of the trachea mask and the top-right point of the ascending aorta mask.
        A large difference suggests that the ascending aorta mask is not located in its expected anatomical position and the image is discarded.
        
        \item \textbf{Border Line}
        This line is defined using the right heart border from the heart mask and the right edge point of the trachea mask. Only images that passed the quality checks in the Cardiomegaly task (for heart mask quality) and the Trachea Deviation task (for trachea mask quality) are included.
    \end{itemize}

    \item \textbf{Descending Aorta Tortuous}
    \begin{itemize}[leftmargin=*]
        \item \textbf{Descending Aorta}

        1) Height: The total height of the mask must exceed a expert-defined threshold, as a short mask suggests incomplete segmentation of the aorta.

        2) Width Consistency: During tortuosity computation, we extract six points from the right edge of the descending aorta mask at fixed vertical intervals. For each of these six right-edge points, we find the corresponding point on the left edge at the same y-coordinate. We calculate the width at each level and compute the standard deviation across the six measurements. A large standard deviation suggests irregular or faulty segmentation and such images are discarded.
    \end{itemize}
    
    \item \textbf{Descending Aorta Enlargement}
    \begin{itemize}[leftmargin=*]
        \item \textbf{Descending Aorta}
        The descending aorta mask is filtered using the same criteria described in the Descending Aorta Tortuous task.

        \item \textbf{Trachea}
        The trachea mask is filtered using the same criteria described in the Trachea Deviation task.
    \end{itemize}
\end{itemize}

Table~\ref{table:chexstruct_label_distribution} shows the number of cases extracted by \extractor for each diagnostic task remaining after quality control, along with their corresponding diagnostic label distribution.

\subsection{Clinician Review} \label{apd:chexstruct_clinician_review}

\subsubsection{Clinician-Guided Thresholding and Label Mapping}

While task-specific extraction and quality control (QC) criteria were defined in collaboration with clinicians (as described in Sections~\ref{apd:chexstruct_criteria} and~\ref{apd:chexstruct_QC}), each criterion required a corresponding threshold to interpret extracted measurements in a clinically meaningful way.

To establish these thresholds, we conducted a structured review process with representative samples spanning a wide range of measured values (e.g., anatomical lengths, ratios, distances).
Three board-certified clinical experts, two radiologists and one radiation oncologist, participated in this review.
One radiologist primarily led the initial establishment of QC criteria for each anatomical structure.
The review involved the following two steps:

\textbf{Step 1: Threshold Definition for Quality Control (QC).}  
For each criterion (e.g., “the descending aorta must be sufficiently tall”), we computed the relevant geometric indicator (e.g., aorta mask height), grouped images by value ranges, and presented representative examples to clinicians. Based on these examples, they determined value thresholds that distinguish anatomically valid segmentations from invalid ones. These thresholds were then used to filter out unreliable cases.

\textbf{Step 2: Diagnostic Label Mapping.}  
For images that passed QC, we used the same range-based interface to collect diagnostic labels (e.g., \textit{enlarged}, \textit{normal}). Clinicians assigned labels to each measurement range, resulting in value-to-label mappings that enabled consistent and scalable annotation of the full dataset.

This process ensured that both QC filtering and diagnostic labeling were grounded in clinical expertise and aligned with the semantics of the task-specific criteria.

\subsubsection{Manual Segmentation Mask Review for Benchmark Validation}

As described in Section~\ref{sec:benchmark_structure}, we randomly sampled 100 cases per diagnostic task (1,200 in total) from the dataset constructed by applying \extractor to MIMIC-CXR-JPG~\cite{johnson2019mimic}.
All sampled cases had previously passed anatomical quality control and were deemed suitable for evaluation.

To further ensure the reliability of the segmentation masks used to extract diagnostic measurements, a clinical expert manually reviewed all task-relevant masks for each of the 12 diagnostic tasks.
Among the 1,200 reviewed cases, 1,198 (99.83\%) were confirmed to have high-quality segmentation masks.
Two cases exhibited minor issues (e.g., slight truncation of the clavicle), but were retained in the benchmark because the diagnostic measurements were unaffected.

This manual review confirms that \benchmark consists of high-quality samples with reliable anatomical segmentations, establishing it as a robust and clinically meaningful benchmark for evaluating diagnostic reasoning across 12 tasks.

\begin{figure}[htb!]
    \centering
    \includegraphics[width=\linewidth]{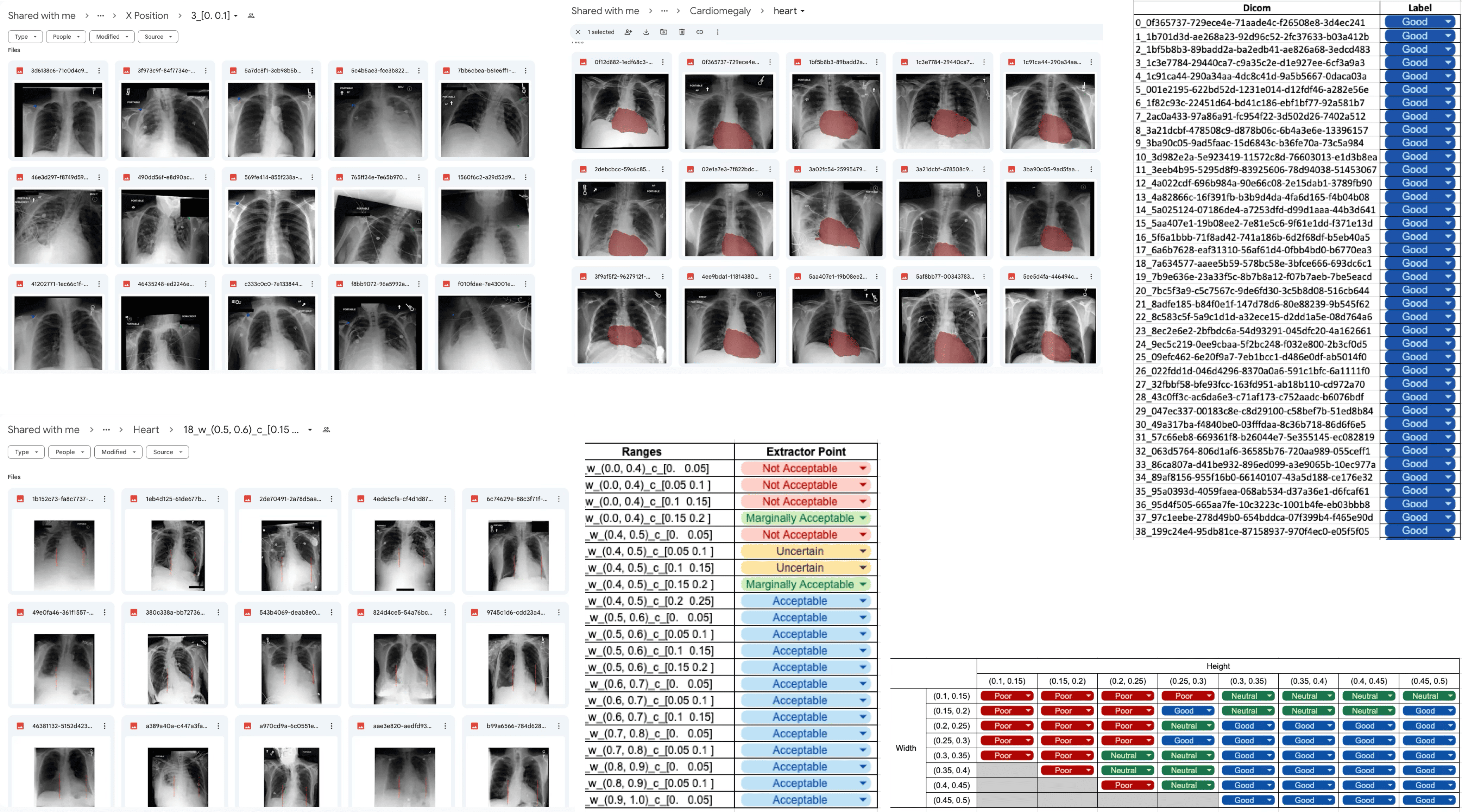}
    \caption{
    Screenshots from clinician review.
    Representative screenshots of the image review interface and label entries provided by clinicians in Google Sheets.
    }
\label{figure:clinician_review}
\end{figure}

A representative screenshot of the clinician review interface used in both the thresholding and segmentation review processes is shown in Figure~\ref{figure:clinician_review}.

%% file: Supple/2.CXReaonBench.tex
\section{Details of \benchmark} \label{apd:cxreasonbench_details}

\subsection{Overview}

The following sections present diagnostic questions and corresponding answers used in \benchmark.
Section~\ref{apd:cxreasonbench_details_init} introduces the initial diagnostic question for each diagnostic task. 
Section~\ref{apd:cxreasonbench_eval_pipeline} provides the full set of questions and answers for each diagnostic task across all stages of each path.
Finally, Section~\ref{apd:cxreasonbench_structure} outlines the overall benchmark structure.
Figures~\ref{figure:qualitative_pro_cardio_final_unmatch},~\ref{figure:qualitative_pro_carina_final_unmatch},~\ref{figure:qualitative_pro_cardio_final_match},~\ref{figure:qualitative_pro_carina_final_match},~\ref{figure:qualitative_pro_path2_cardio_final_match}, and~\ref{figure:qualitative_pro_path2_knob_final_match} illustrate examples of the model evaluation pipeline based on the diagnostic questions and answers.

\subsection{Initial Diagnostic Question} \label{apd:cxreasonbench_details_init}
The corresponding question for each diagnostic task is presented below.
A chest X-ray image is provided to the model along with the question.

Each question is presented to the model along with the instruction: \textit{Please base your decision on the most established and clearly defined diagnostic criterion used in standard radiologic references. Avoid relying on indirect factors, which, while potentially relevant, are not the direct and primary criteria. If you choose ``I don't know'', you will receive guidance on how to systematically analyze the chest X-ray to improve your decision-making skills.}

\begin{itemize}[leftmargin=*]
        \item \textbf{Cardiomegaly} Does this patient have cardiomegaly? The chest X-ray was taken in the \texttt{\{view position\}} view, where \texttt{\{view position\}} is either PA or AP.

        \item \textbf{Mediastinal Widening} Does this patient have mediastinal widening?

        \item \textbf{Carina Angle} Does this chest X-ray show a normal carina angle?

        \item \textbf{Trachea Deviation} Is the trachea deviated in this chest X-ray?

        \item \textbf{Aortic Knob Enlargement} Does the aortic knob appear enlarged in this chest X-ray?
 
        \item \textbf{Ascending Aorta Enlargement} Does the ascending aorta appear enlarged in this chest X-ray?

        \item \textbf{Descending Aorta Enlargement} Does the descending aorta appear enlarged in this chest X-ray?

        \item \textbf{Descending Aorta Tortuous} Is the descending aorta tortuous in this chest X-ray?

        \item \textbf{Inclusion} Is the entire thoracic cage including the lung apices, inner margins of the lateral ribs, and costophrenic angles (CPAs) fully visible in this chest X-ray without being cropped?

        \item \textbf{Inspiration} Assess the level of inspiration in this chest X-ray. Was it taken with good or poor inspiration?

        \item \textbf{Rotation} Was the patient rotated during the chest X-ray?

        \item \textbf{Projection} Identify the view of this chest X-ray.
    \end{itemize}

\subsection{Diagnostic Questions per Path} \label{apd:cxreasonbench_eval_pipeline}

Based on the model’s response to the initial diagnostic question, it proceeds to Path 1 if a binary answer is selected (Section~\ref{apd:cxreasonbench_path1}), or to Path 2 if \textit{I don’t know} is chosen (Section~\ref{apd:cxreasonbench_path2}).

\subsubsection{Path 1: Direct Reasoning Process Evaluation} \label{apd:cxreasonbench_path1}

The following outlines the questions and answers for each task, organized by stage.
\begin{itemize}[leftmargin=*]
    \item \textbf{Stage 1. Diagnostic Criterion Selection}
    
    Question: \textit{What criterion was used to make the decision for the first question?}

    Below are the answers for each diagnostic task.
    \begin{itemize}[leftmargin=*]
        \item \textbf{Cardiomegaly} By calculating the cardiothoracic ratio, which is the ratio of the maximal horizontal cardiac diameter to the maximal horizontal thoracic diameter.

        \item \textbf{Mediastinal Widening} By evaluating the width of the mediastinum.

        \item \textbf{Carina Angle} By evaluating the angle of the carina.

        \item \textbf{Trachea Deviation} By checking if the trachea is displaced to one side from the midline.

        \item \textbf{Aortic Knob Enlargement} By checking for any prominent bulge or abnormal widening of the aortic knob.
 
        \item \textbf{Ascending Aorta Enlargement} By checking for any bulge or abnormal widening of the ascending aorta.

        \item \textbf{Descending Aorta Enlargement} By checking for any abnormal dilation or widening of the descending aorta.

        \item \textbf{Descending Aorta Tortuous} By evaluating the shape of the descending aorta and checking for any signs of tortuosity, such as irregular bends or twists.

        \item \textbf{Inclusion} By checking if the chest X-ray includes the lung apices, inner margins of the lateral ribs, and costophrenic angles.

        \item \textbf{Inspiration} By counting the number of right posterior ribs or anterior rib visible above the right hemidiaphragm.

        \item \textbf{Rotation} By checking if the spinous processes are equidistant from the medial ends of the clavicles.

        \item \textbf{Projection} By checking if the scapulae were laterally retracted or if they overlapped the lung fields.
    \end{itemize}

    \item \textbf{Stage 1.5. Refined Criterion Adoption (Expert-Defined Criteria Only)}

    For diagnostic tasks where the original criteria are difficult to apply to images (e.g., due to reliance on imaging metadata) or are inherently ambiguous, the model is provided with expert-defined criteria from Section~\ref{apd:chexstruct_criteria}.
    The corresponding questions are listed below.

    Each question is presented to the model along with the instruction: \textit{``If you choose `No', you will receive guidance on how to systematically analyze the chest X-ray to improve your decision-making skills.''}
    Answer options are \texttt{Yes} or \texttt{No}.
    \begin{itemize}[leftmargin=*]
        \item \textbf{Mediastinal Widening} The original criterion for assessing mediastinal widening involves measuring the mediastinal width in centimeters, typically using metadata or physical markers. However, in this evaluation setting where only images are provided without accompanying metadata, absolute measurements are not feasible. To maintain the fundamental diagnostic approach while adapting to this constraint, the criterion has been refined to use a ratio-based measurement. Specifically, evaluators are now instructed to measure both the mediastinal width and the thoracic width at the same level, then calculate the ratio by dividing the mediastinal widthby the thoracic width. This updated approach enables consistent and objective assessments even in the absence of physical measurement units.

        \item \textbf{Aortic Knob Enlargement} The original criterion for assessing an enlarged aortic knob involved visually inspecting for any prominent bulge or abnormal widening of the aortic knob. However, this approach can be subjective and may vary among evaluators. To enhance consistency and objectivity in assessment, the criterion has been refined. The refined criterion involves measuring both the maximum width of the aortic knob and the median width of the trachea. Then, calculate the ratio by dividing the aortic knob width by the trachea width. This updated approach provides a more standardized andreliable method for identifying an enlarged aortic knob.

        \item \textbf{Ascending Aorta Enlargement} The original criterion for assessing an enlarged ascending aorta involved visually inspecting for any bulge or abnormal widening of the ascending aorta. However, this approach can be subjective and may vary among evaluators. To enhance consistency and objectivity in assessment, the criterion has been refined. The refined criterion involves drawing an imaginary straight line connecting the inner boundary of the right lung and the right heart side. Then, determine whether the ascending aorta extends beyond this line. This approach provides a more reproducible and objective way to assess aortic enlargement.

        \item \textbf{Descending Aorta Enlargement} The original criterion for assessing an enlarged descending aorta involved visually inspecting for any bulge or abnormal widening of the descending aorta. However, this approach can be subjective and may vary among evaluators. To enhance consistency and objectivity in assessment, the criterion has been refined. The refined criterion involves measuring both the maximum width of the descending aorta and the median width of the trachea. Then, calculate the ratio by dividing the descending aorta width by the trachea width. This updated approach provides a more standardized and reliable method for identifying an enlarged descending aorta.

        \item \textbf{Descending Aorta Tortuous} The original criterion for assessing descending aorta tortuosity involved visually inspecting the shape of the descending aorta and checking for any signs of tortuosity. However, this approach can be subjective and may vary among evaluators. To enhance consistency and objectivity in assessment, the criterion has been refined to include a more quantitative approach.This refined method involves focusing on the thoracic portion of the descending aorta, particularly the region at the upper part of the heart. The region is divided into five equal sections, and six coordinates are determined at the top-left lung side of each division. Curvature is then calculated at each of these six coordinates using finite difference methods: forward and backward differences for the first and last points, and central differences for the middle points to ensure greater accuracy. The average curvature is computed across all six points to quantify the tortuosity of the descending aorta. This refined approach minimizes evaluator variability and provides a more objective and reproducible assessment of aortic tortuosity.

        \item \textbf{Inspiration} The criterion has been refined to reduce ambiguity and ensure consistency in measurements between individuals, as the previous criterion lacked a clear reference point, which could result in variations in interpretation. Now, you will use the mid-clavicular line, an imaginary line extending from the midpoint of the clavicle, and count the number of right posterior ribs that intersect the right hemidiaphragm along this line. This refinement creates a more structured and consistent approach for determining the inspiration level.

        \item \textbf{Projection} The previous method for determining chest X-ray projection (PA or AP) relied on visually assessing scapular retraction or overlap with the lung fields. While this approach reflects the correct fundamental concept — that scapular positioning is indicative of projection — it lacked clear thresholds, leading to variability in interpretation among individuals. To reduce this ambiguity and enhance measurement consistency between evaluators, the criterion has been refined. The refined method maintains the original diagnostic logic but introduces a more structured process: measuring the ratio of the overlapping area between the scapula and the lung to the total scapular area for both the right and left scapula. This update ensures a more objective and standardized application of the same underlying principle.
    \end{itemize}
    
    \item \textbf{Stage 2. Anatomical Structure Identification}

    Depending on whether the criteria are standardized and the number of anatomical structures involved, the question is phrased accordingly.
    For the anatomical structures involved in each task, refer to Figure~\ref{figure:cxreasonbench_p1s2}.
    
    Prefix for the questions: \textit{In the following images, either a segmentation mask of a specific body part or a reference line is shown.}
    \begin{itemize}[leftmargin=*]
        \item \textbf{Standardized, Quantifiable Criteria, Single Anatomical Structure} Based on \textit{the selected criterion}, select \textit{the image} that include the relevant body part or reference line required for applying that criterion.
        
        \item  \textbf{Standardized, Quantifiable Criteria, Anatomical Multiple Structures} Based on \textit{the selected criterion}, select \textit{all images} that include the relevant body part or reference line required for applying that criterion.
        
        \item \textbf{Expert-Defined Criteria, Single Anatomical Structure} Based on \textit{the refined criterion}, select \textit{the image} that include the relevant body part or reference line required for applying that criterion.

        \item \textbf{Expert-Defined Criteria, Multiple Anatomical Structures} Based on \textit{the refined criterion}, select \textit{all images} that include the relevant body part or reference line required for applying that criterion.
    \end{itemize}

    \item \textbf{Stage 3. Measurement or Recognition}
    Depending on the task type, the model either computes diagnostic indices from anatomical measurements (measurement-type), or interprets anatomical patterns and positional changes to assign a label (recognition-type).
    
    Each task type has a distinct format for both question and answer options, as follows:
    \begin{itemize}[leftmargin=*]
        \item \textbf{Measurement-type tasks:}
        The model calculates a diagnostic index from anatomical measurements and selects the appropriate \textit{value range} (\textit{e.g.}, [0.50–0.52], [30–50]).
        \begin{itemize}[leftmargin=*]
            \item \textbf{Rotation} Measure the horizontal distance (x-coordinate difference) from the medial end of each clavicle to the nearest point on the vertical line through the spinous processes. Then, compute the ratio of the shorter distance to the longer distance. Round the result to two decimal places, and determine which range the measured value falls into and select the correct option.
    
            \item \textbf{Projection} Use the refined criterion outlined earlier to measure the ratio. Round the ratio to two decimal places, then determine which range it falls into for both the right and left sides, and select the correct option.
    
            \item \textbf{Cardiomegaly} Use the criterion you selected to make the decision for the first question to measure the ratio. Round the ratio to two decimal places, then determine which range it falls into and select the correct option.
    
            \item \textbf{Mediastinal Widening} Use the refined criterion outlined earlier to measure the ratio. Round the ratio to two decimal places, then determine which range it falls into and select the correct option.
    
            \item \textbf{Carina Angle} Then, use the criterion you selected to make the decision for the first question to measure the angle. Round the value to the nearest whole number, then determine which range it falls into and select the correct option.
            
            \item \textbf{Aortic Knob Enlargement} Use the refined criterion outlined earlier to measure the ratio. Round the ratio to two decimal places, then determine which range it falls into and select the correct option.

            \item \textbf{Descending Aorta Enlargement} Use the refined criterion outlined earlier to measure the ratio. Round the ratio to two decimal places, then determine which range it falls into and select the correct option.
    
            \item \textbf{Descending Aorta Tortuous} Use the refined criterion outlined earlier to measure the ratio. Round the ratio to four decimal places, then determine which range it falls into and select the correct option.
        \end{itemize}

        \item \textbf{Recognition-type tasks:} The model interprets anatomical patterns or positional changes and selects the appropriate \textit{label}.
        \begin{itemize}[leftmargin=*]
            \item \textbf{Inclusion} Use the criterion you selected to assess which parts of the thoracic cage are included or excluded, and select the correct option (\textit{e.g.}, ``Right apex: Included, Left apex: Included, Right rib edge: Included, Left rib edge: Included, Right costophrenic angle: Included, Left costophrenic angle: Included,'', ``Right apex: Excluded, Left apex: Included, Right rib edge: Included, Left rib edge: Excluded, Right costophrenic angle: Excluded, Left costophrenic angle: Excluded,'').
    
            \item \textbf{Inspiration} Use the refined criterion outlined earlier to count the rib number and select the correct option (\textit{e.g.}, ``7th rib'', ``8th rib'')
    
            \item \textbf{Trachea Deviation} Use the criterion you selected to assess the deviation and select the correct option (\textit{e.g.}, ``Deviated to the right'', ``Deviated to the left'', ``Not deviated'').

            \item \textbf{Ascending Aorta Enlargement} Use the refined criterion outlined earlier to assess whether the ascending aorta extends beyond the line. Select the appropriate label based on your measurement (\textit{e.g.}, ``Extend beyond the line'', ``Does not extend beyond the line'').
    
        \end{itemize}
    \end{itemize}

    \item \textbf{Stage 4. Final Decision}

    Depending on whether a standardized threshold is available, the question is divided as follows:
    \begin{itemize}[leftmargin=*]
        \item \textbf{For tasks with standardized thresholds}:  
        The model must rely on commonly known diagnostic thresholds.

        \begin{itemize}[leftmargin=*]
            \item \textbf{Cardiomegaly (PA view)} Does this patient have cardiomegaly?

            \item \textbf{Trachea Deviation} Is the trachea deviated in this chest X-ray?

            \item \textbf{Inclusion} Is the entire thoracic cage - including the lung apices, inner margins of the lateral ribs, and costophrenic angles (CPAs) fully visible in this chest X-ray without being cropped?

        \end{itemize}
        
        \item \textbf{For other tasks with expert-defined thresholds}:
        The model is provided with thresholds defined by clinical experts to guide diagnostic decisions, but specific values are omitted here for brevity.

        \begin{itemize}[leftmargin=*]
            \item \textbf{Cardiomegaly (AP view)} Does this patient have cardiomegaly?

            \item \textbf{Mediastinal Widening} Does this patient have mediastinal widening?

            \item \textbf{Carina Angle} Does this chest X-ray show a normal carina angle?

            \item \textbf{Aortic Knob Enlargement} Does the aortic knob appear enlarged in the chest X-ray?

            \item \textbf{Descending Aorta Enlargement} Does the descending aorta appear enlarged in the chest X-ray?

            \item \textbf{Descending Aorta Tortuous} Is the descending aorta tortuous in the chest X-ray?
        
            \item \textbf{Inspiration} What is the inspiration level of the chest X-ray?

            \item \textbf{Ascending Aorta Enlargement} Does the ascending aorta appear enlarged in the chest X-ray?

            \item \textbf{Rotation} Was the patient rotated during the chest X-ray?

            \item  \textbf{Projection} Identify the view of the chest X-ray.

        \end{itemize}

    \end{itemize}

\end{itemize}

\begin{figure}[htb!]
    \centering
    \includegraphics[width=0.87\linewidth]{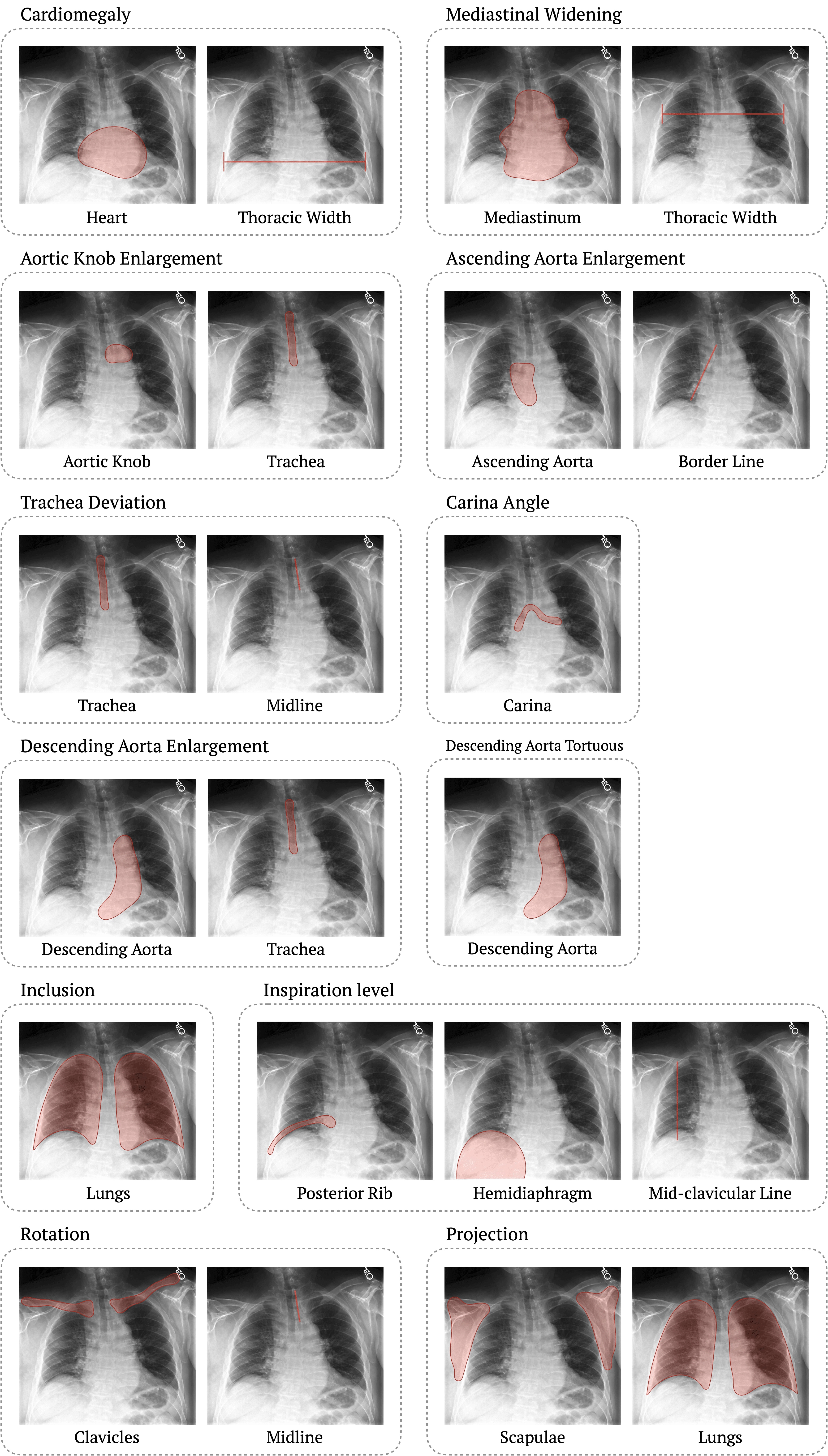}
    \caption{
    Visualization of anatomical structures required for each diagnostic task in Stage 2 of Path 1 and Stage 1 of Path 2.
    }
\label{figure:cxreasonbench_p1s2}
\end{figure}

\clearpage

\subsubsection{Path 2: Guided Reasoning and Re-evaluation}\label{apd:cxreasonbench_path2}
\begin{figure}[htb!]
    \centering
    \includegraphics[width=\linewidth]{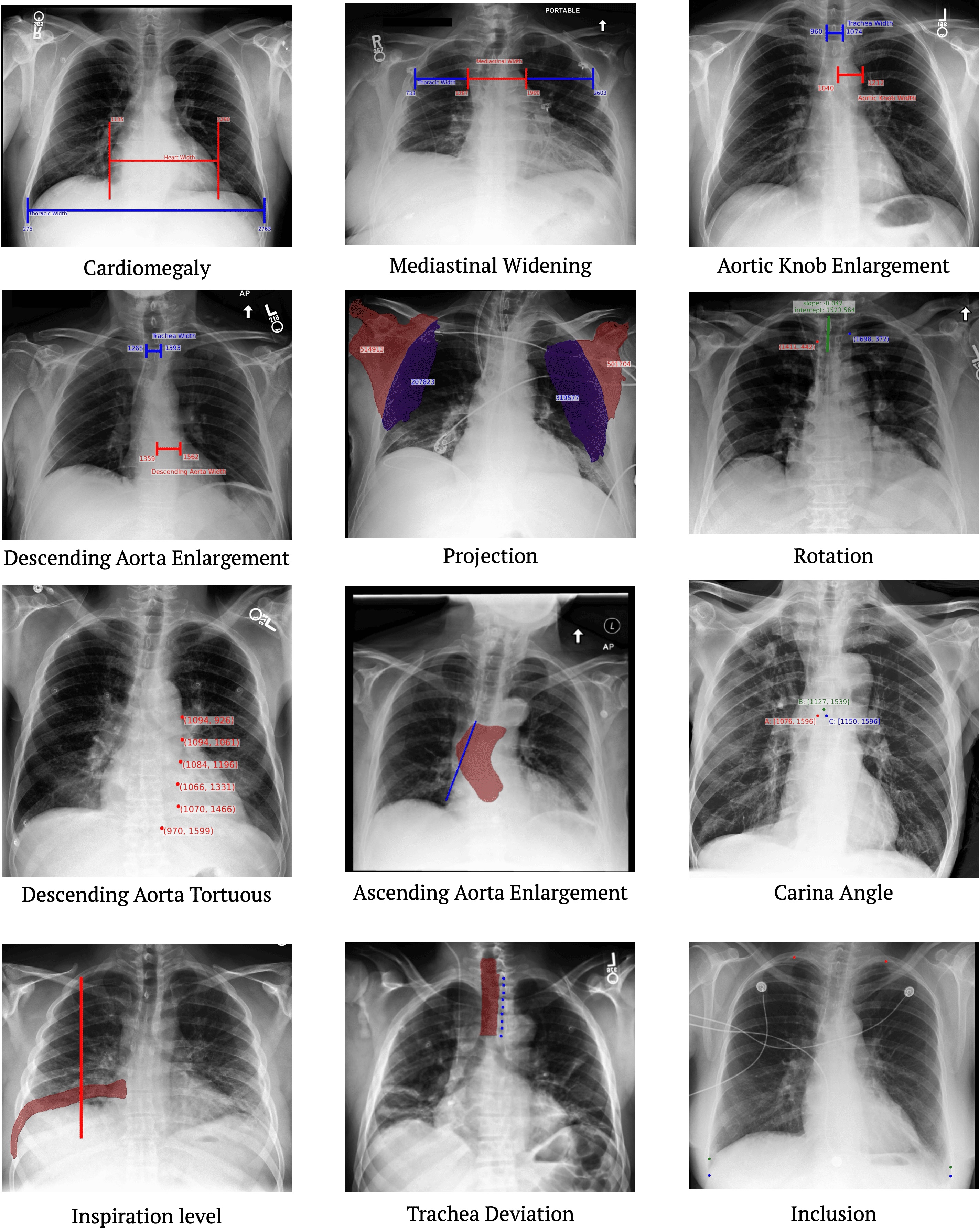}
    \caption{
    Visualization of the detailed visual annotations provided per diagnostic task in Stage 2 of Path 2.
    }
\label{figure:cxreasonbench_p2s2}
\end{figure}

The following outlines the questions and answers for each task, organized by stage.
\begin{itemize}[leftmargin=*]
   
    \item \textbf{Stage 1. Anatomical Structure Identification}

    The following is the question format used in this stage. The placeholder \texttt{\{anatomical structure\}} is replaced with the anatomical structure name involved in each diagnostic task.
    
    Refer to Figure~\ref{figure:cxreasonbench_p1s2} for the list of anatomical structures associated with each task.

    \begin{itemize}[leftmargin=*]
        \item Among the following images, each image either contains a segmentation mask highlighting a specific body part or a reference line necessary for a decision. Which image represents \texttt{\{anatomical structure\}}?
    
        \item Continuing from the previous question, which image corresponds to \texttt{\{anatomical structure\}}?

    \end{itemize}

    \item \textbf{Stage 2. Guided Measurement or Recognition}
    
    For each diagnostic task, the model receives a chest X-ray image with detailed visual annotations (e.g., segmentation masks, reference lines, landmarks) and a task-specific instruction.
    
    Figure~\ref{figure:cxreasonbench_p2s2} shows examples of the visual annotations used for each task.
    
    Instructions for each diagnostic task are shown below.
    \begin{itemize}[leftmargin=*]

        \item \textbf{Inclusion} To check whether the entire thoracic cage is shown in the chest X-ray, look to see if three important parts are visible: the tops of the lungs (lung apices), the inner edges of the side ribs, and the costophrenic angles (CPAs), which are the corners at the bottom of the lungs. In the image, a colored point has been placed at each of these areas: red for the lung apices, green for the inner rib edges, and blue for the CPAs. Keep in mind: the point doesn't always mark the exact part of the body. If the part is visible in the image, the point shows its actual location. If it's missing from the image (for example, if the top of the lung is cut off), the point just shows the general level where it should have appeared. If a part is excluded, that means the body part isn't visible at the point's location - it was cut off in the X-ray. Examine each point and decide whether the corresponding body part is visible in the image.

        \item \textbf{Inspiration} To assess the level of inspiration, draw an imaginary vertical line from the midpoint of the clavicle (mid-clavicular line). Then, count how many right posterior ribs intersect the right hemidiaphragm along this line. In the provided image, the mid-clavicular line and the right posterior rib intersecting the right hemidiaphragm along the line are marked. Look at the image and count which rib is intersecting.

        \item \textbf{Rotation} To assess the rotation of the patient during an X-ray, check whether the spinous processes are equidistant from the medial ends of the clavicles. In the provided image, the points corresponding to the medial ends of the clavicle and their coordinate values are marked. Additionally, a straight line representing the spinous processes is given, with its slope and intercept provided. This line is defined in a way that for any given y-coordinate, the corresponding x-coordinate can be determined using the slope and intercept. To measure the rotation: 1. For each medial clavicle point, use its y-coordinate to determine the corresponding x-coordinate on the spinous process line. 2.	Compute the difference between this x-coordinate and the original x-coordinate of the medial clavicle point to obtain the distance. 3.	Compare the two distances and determine the ratio of the shorter distance to the longer distance. 4. Round the result to two decimal places, check which range it falls into, and select the correct option.

        \item \textbf{Projection} To assess the projection, check whether the scapulae are laterally retracted or overlapping with the lung fields. In the provided image, segmentation masks for the left and right scapulae are drawn. The overlapping regions between the scapulae and the lung fields are highlighted in purple, while the remaining scapular regions are marked in red. Each mask also displays numerical values indicating the overlapping area and the total scapular area. If there is no overlapping region, no purple markings are displayed. To determine whether the scapulae are retracted or overlapping, calculate the ratio of the overlapping area between the scapula and the lung to the total scapular area for both the right and left scapula. Round the ratio to two decimal places, then determine which range it falls into for both the right and left sides, and select the correct option, choosing one option for each.

        \item  \textbf{Cardiomegaly} To assess cardiomegaly, calculate the cardiothoracic ratio, which is the ratio of the maximal horizontal cardiac diameter to the maximal horizontal thoracic diameter. In the provided image, the x-coordinates for measuring cardiac width and thoracic width are marked, along with lines representing both measurements. The coordinates and lines associated with the heart are highlighted in red, while those related to the lungs are highlighted in blue. Round the ratio to two decimal places, then determine which range it falls into and select the correct option.

        \item \textbf{Mediastinal Widening} To assess mediastinal widening, measure the mediastinal width and the thoracic width at the same level, then calculate their ratio by dividing the mediastinal width by the thoracic width. In the provided image, the x-coordinates for measuring mediastinal width and thoracic width are marked, along with lines representing both measurements. The coordinates and lines associated with the mediastinum are highlighted in red, while those related to the lungs are highlighted in blue. Round the ratio to two decimal places, then determine which range it falls into and select the correct option.

        \item \textbf{Carina Angle} To assess whether the carina angle is normal, measure the angle between the left and right main bronchi. In the provided image, the central point at the carina is marked as B, the right main bronchus as A, and the left main bronchus as C, with their respective coordinate values also indicated. Use these points to determine the angle formed between the two bronchi. Round the value to the nearest whole number, then determine which range it falls into and select the correct option.

        \item \textbf{Trachea Deviation} To assess tracheal deviation, use the spinous processes as a reference to draw an imaginary straight line down the center of the vertebral bodies and evaluate whether the trachea aligns with this line. In the provided image, the trachea segmentation mask and nine points along the line are marked. For each point, determine whether the trachea is on, deviated toward the left (left lung side), or deviated toward the right (right lung side) of the point. The final label is determined by majority vote. If multiple labels share the highest count, assign the label based on the order in which the majority count was first reached. Based on your assessment, select the appropriate option.

        \item \textbf{Aortic knob Enlargement} To assess aortic knob enlargement, measure the maximum width of the aortic knob and the median width of the trachea, then calculate their ratio by dividing the aortic knob width by the trachea width. In the provided image, the x-coordinates for measuring the trachea and aortic knob widths are marked, along with lines representing both measurements. The coordinates and lines associated with the aortic knob are highlighted in red, while those related to the trachea are highlighted in blue. Round the ratio to two decimal places, then determine the corresponding range and select the appropriate option.

        \item \textbf{Ascending Aorta Enlargement} To assess ascending aorta enlargement, determine whether the ascending aorta extends beyond an imaginary straight line connecting the inner boundary of the right lung and the right heart border. In the provided image, the ascending aorta segmentation mask and this reference line are marked. Select the appropriate label based on your measurement.

        \item \textbf{Descending Aorta Enlargement} To assess descending aorta enlargement, measure the maximum width of the descending aorta and the median width of the trachea, then calculate their ratio by dividing the descending aorta width by the trachea width. In the provided image, the x-coordinates for measuring the trachea and descending aorta widths are marked, along with lines representing both measurements. The coordinates and lines associated with the descending aorta are highlighted in red, while those related to the trachea are highlighted in blue. Round the ratio to two decimal places, then determine the corresponding range and select the appropriate option.

        \item \textbf{Descending Aorta Tortuous} To assess descending aorta tortuosity, focus on the thoracic portion of the descending aorta, specifically the region at the upper part of the heart. Divide this region into five equal sections and determine the coordinates at the top-left lung side of each division, resulting in six total coordinates. Calculate the curvature at each of these six coordinates using finite difference methods: - The first and last points use forward and backward differences, respectively. - The middle points use central differences for higher accuracy. Compute the average curvature across all six points to quantify tortuosity. In the provided images, these six coordinates are marked. Round the ratio to four decimal places, then determine which range it falls into and select the correct option.

    \end{itemize}

    \item \textbf{Stage 3. Final Decision}

    The same diagnostic question format used in Stage 4 of Path 1 is applied here. However, unlike Path 1, a threshold is provided for all tasks, regardless of whether a standardized threshold is available.
    
\end{itemize}

\clearpage

\subsection{Benchmark Structure} \label{apd:cxreasonbench_structure}

\benchmark employs a multiple-choice format with both single-choice (\textit{e.g.}, one diagnostic criterion) and multi-choice questions (\textit{e.g.}, all relevant anatomical structures) depending on the stage.

\subsubsection{Evaluation Formats: Basic and Two-Round with \textit{“Need new option”}}

Except for Path 2, certain stages adopt either a basic or a two-round evaluation format, depending on whether the correct option is included in the initial round. The two-round format is enabled by including a “Need new option” choice.

This mechanism assesses whether the model can recognize that its current reasoning paths are insufficient and adapt to more clinically valid alternatives.

\textbf{Stage 1. Diagnostic Criterion Selection}
    \begin{itemize}[leftmargin=*] 
        \item \textit{Basic Format}: The correct diagnostic criterion is included in the initial options. The model selects one and proceeds.
    
        \item \textit{Two-Round Format}: The correct option is intentionally excluded. A “Need new option” choice is included instead. If selected, a second round presents the correct option. A valid second-round response is required to proceed.
    \end{itemize}

\textbf{Stage 2. Anatomical Structures Identification} 
The format varies depending on how many anatomical structures are clinically required for the diagnostic criterion:
    \begin{itemize}[leftmargin=*] 
        \item \textbf{Task with Single Anatomical Structure}: Follows the same format as Stage 1 (\textit{e.g.}, Carina angle - carina). 
        
        \item \textbf{Task with Multiple Anatomical Structures}: one of three formats is used (\textit{e.g.}, Cardiomegaly – heart, lung):
        \begin{itemize}[leftmargin=*]
            \item \textit{Basic Format}: All correct anatomical structures are included in the initial options.
            \item \textit{Two-Round (Partial Inclusion)}: Only one correct anatomical structure is included in the initial round. The model must select both the included correct anatomical structure and "Need new option" to access the second round, where the remaining anatomical structure becomes available.
            \item \textit{Two-Round (None Included)}: Neither anatomical structure is included initially. The model must select "Need new option" to proceed to a second round where all correct anatomical structures are presented.
        \end{itemize}
    \end{itemize}

\textbf{\textit{``None of the above''} option}
When the \textit{“Need new option”} is not included, a \textit{“None of the above”} option is provided instead.
In this case, the correct answer is always included in the initial options.
Selecting \textit{“None of the above”} results in failure, and the model is expected to explain its reasoning.

%% file: Supple/3.Experiments.tex
\section{Experimental Setup Details}\label{apd:experiment_setup_details}


\subsection{Dataset Resources} \label{apd:chexstruct_resources}

\begin{itemize}[leftmargin=*] 
    \item \textbf{MIMIC-CXR-JPG ~\cite{johnson2019mimic}}
    \begin{itemize}[leftmargin=*] 
        \item URL: https://physionet.org/content/mimic-cxr-jpg/ 
        \item License: PhysioNet Credentialed Health Data License 1.5.0
    \end{itemize}

    \item \textbf{CXAS ~\cite{seibold2023accurate}}
    \begin{itemize}[leftmargin=*] 
        \item URL: https://github.com/ConstantinSeibold/ChestXRayAnatomySegmentation
        \item License: CC BY NC 4.0
    \end{itemize}

    \item \textbf{CheXmask \cite{gaggion2023chexmaskPhysioNet}}
    \begin{itemize}[leftmargin=*] 
        \item URL: https://physionet.org/content/chexmask-cxr-segmentation-data/
        \item License: CC BY 4.0
    \end{itemize}

    \item \textbf{Chest ImaGenome \cite{wu2021chest}}
    \begin{itemize}[leftmargin=*] 
        \item URL: https://physionet.org/content/chest-imagenome/1.0.0/
        \item License: PhysioNet Credentialed Health Data License 1.5.0
    \end{itemize}
    
    \item \textbf{CheXStruct pipeline and CXReasonBench evaluation}
    \begin{itemize}[leftmargin=*] 
        \item URL: https://github.com/ttumyche/CXReasonBench
        \item License: CC BY NC 4.0
        \item Description: Detailed instructions on how to run the code are provided at the provided URL.
    \end{itemize}
    
    \item \textbf{CheXStruct and CXReasonBench}
    \begin{itemize}[leftmargin=*] 
        \item URL: https://physionet.org/content/chexstruct-cxreasonbench/1.0.1/
        \item License: PhysioNet Credentialed Health Data License 1.5.0
        \item Description: Detailed information about the dataset, including file contents and structure, is available at the provided URL. Access requires credentials, which can be obtained by following the instructions in the GitHub (https://github.com/ttumyche/CXReasonBench).
    \end{itemize}

\end{itemize}

\subsection{Model Resources} \label{apd:experiment_model_resource}

The models used in this paper are categorized into closed-source and open-source models.

The closed-source models include Gemini-2.5-Pro, Gemini-2.5-Flash, and GPT-4.1, all of which are accessed through proprietary APIs and subject to the respective provider's terms of service.
\revision{Table~\ref{table:api_cost} summarizes the estimated total cost (in USD) for performing inference using each closed-source model.}
The specific model versions are as follows:
\begin{itemize}[leftmargin=*]
    \item \textbf{Gemini-2.5-Pro~\cite{team2023gemini}}: gemini-2.5-pro-preview-03-25

    \item \textbf{Gemini-2.5-Flash~\cite{team2023gemini}}: gemini-2.5-flash-preview-04-17

    \item \textbf{GPT-4.1~\cite{achiam2023gpt}}: GPT-4.1 (2025-04-14)
\end{itemize}

\input{Tables/Supple_API_Cost}

The open-source models were obtained from Hugging Face~\cite{wolf2019huggingface} and GitHub.
For each model, the corresponding Hugging Face repository or GitHub link, along with its license, is listed below:

\begin{itemize}[leftmargin=*]
    \item \textbf{Pixtral-Large-Instruct-2411~\cite{agrawal2024pixtral}}
    \begin{itemize}[leftmargin=*]
        \item Hugging Face Path: mistralai/Pixtral-Large-Instruct-2411
        \item License: Apache-2.0 license
    \end{itemize}
    
    \item \textbf{Qwen2.5-VL-72B-Instruct~\cite{qwen2.5-VL}}
    \begin{itemize}[leftmargin=*]
        \item Hugging Face Path: Qwen/Qwen2.5-VL-72B-Instruct
        \item License: Qwen LICENSE AGREEMENT
    \end{itemize}
    
    \item \textbf{Llama-3.2-90b-Vision-Instruct~\cite{grattafiori2024llama}}
    \begin{itemize}[leftmargin=*]
        \item Hugging Face Path: meta-llama/Llama-3.2-90B-Vision-Instruct
        \item License: Meta Llama 3 Community License Agreement
    \end{itemize}

    \item \textbf{Pixtral 12B~\cite{agrawal2024pixtral}}
    \begin{itemize}[leftmargin=*]
        \item Hugging Face Path: mistralai/Pixtral-12B-2409
        \item License: Apache-2.0 license
    \end{itemize}
    
    \item \textbf{Qwen2.5-VL-7B-Instruct~\cite{qwen2.5-VL}}
    \begin{itemize}[leftmargin=*]
        \item Hugging Face Path: Qwen/Qwen2.5-VL-7B-Instruct
        \item License: Qwen LICENSE AGREEMENT
    \end{itemize}

    \item \textbf{MedGemma 4B / 27B ~\cite{sellergren2025medgemma}}
    \begin{itemize}[leftmargin=*]
        \item Hugging Face Path: google/medgemma-4b-it, google/medgemma-27b-it
        \item License: Health AI Developer Foundations License
    \end{itemize}
    
    \item \textbf{HealthGPT-L14~\cite{lin2025healthgpt}}
    \begin{itemize}[leftmargin=*]
        \item Hugging Face Path: lintw/HealthGPT-L14
        \item License: MIT License
    \end{itemize}
    
    \item \textbf{RadVLM~\cite{deperrois2025radvlm}}
    \begin{itemize}[leftmargin=*]
        \item GitHub: https://github.com/uzh-dqbm-cmi/RadVLM
        \item License: CC BY-NC 4.0
    \end{itemize}
\end{itemize}

\textbf{Compute Requirements}
Our evaluation pipeline comprises three paths: Path 1, Path 2, and Re-evaluated Path 1. Each diagnostic task involves a multi-stage reasoning process, where progression to later stages depends on correct responses from earlier ones.
Due to this dependency, the total evaluation time per model varies.

On average, evaluations with closed-source models required approximately 2 GPU-hours per diagnostic task.
For open-source models, evaluation times varied:
Pixtral-Large-Instruct-2411, Qwen2.5-VL-72B-Instruct, and LLaVA-3.2-90B-Vision-Instruct required around 4 GPU-hours per task, while other models completed evaluations within 2 GPU-hours.

Under stochastic sampling, we generated three responses per query, effectively doubling the evaluation time compared to greedy sampling. Thus, stochastic sampling generally requires twice as much computation time as greedy sampling.

Experiments were conducted on the following hardware:
\begin{itemize}[leftmargin=*]
    \item Pixtral-Large-Instruct-2411, Qwen2.5-VL-72B-Instruct, and LLaVA-3.2-90B-Vision-Instruct were run on 4 A100 GPUs.
    \item Pixtral-12B, Qwen2.5-VL-7B-Instruct, MedGemma 4B, MedGemma 27B, HealthGPT-L14, and RadVLM were run on 1-4 RTX A6000 GPUs, depending on the model size.
\end{itemize}

To comply with the MIMIC Data Use Agreement, Gemini models are run on Vertex AI, the GPT model is deployed on Azure’s HIPAA-compliant platform, and open-source LVLMs are evaluated locally.

\subsection{Sampling Methods} \label{apd:experiment_sampling_method}
We adopt two sampling strategies as follows:

\textbf{Greedy Sampling} is used to generate a single, near-deterministic response by setting the temperature to zero, minimizing output randomness.

\textbf{Stochastic Sampling} generates three distinct responses per question to evaluate model stability and robustness. This is achieved by configuring the temperature to 0.7 and the top-$p$ parameter to 0.95, which approximates the model’s default behavior. The reliability of responses is assessed using majority voting, where a response is considered correct if at least two out of the three generated responses match the ground truth.

\subsection{Evaluation Metric Details}\label{apd:experiment_evaluation_metric}

As mentioned in Section~\ref{sec:experimental_setup}, all evaluation metrics, except for \textit{Average Reasoning Depth}, are computed using the Wilson score~\cite{wilson1927probable}, which accounts for the number of attempts to enable fairer comparisons across models with varying attempt counts at each stage.
The following section introduces an additional evaluation metric, the \textit{Stage-wise Score}, to assess model performance at individual stages, followed by a detailed explanation of the Wilson score and our adjusted Wilson score used across all proportion-based metrics.

\subsubsection{Additional Evaluation Metric}
The \textbf{Stage-wise Score} metric measures the proportion of correct responses at each stage.
For each stage, this metric is calculated as the number of correct responses divided by the number of attempts, using the adjusted Wilson score (described
below) to account for varying attempt counts and penalize uncertainty from small sample sizes.
This metric evaluates the model's ability to meet stage-specific requirements, such as selecting diagnostic criteria or identifying anatomical structures.

\subsubsection{Wilson Score}
The Wilson score interval is a statistical method used to estimate the confidence interval for a proportion, such as the success rate of a model in a binary outcome (e.g., success or failure in completing a task). Unlike a simple proportion (e.g., number of successes divided by total attempts), the Wilson score accounts for sample size variability and provides a more robust estimate.

For a given success proportion \( p = \frac{x}{n} \), where \( x \) is the number of successes and \( n \) is the number of attempts, the Wilson score interval is calculated as follows:

\begin{equation}
\text{Wilson score} = \frac{\hat{p} + \frac{z^2}{2n}}{1 + \frac{z^2}{n}}
\end{equation}

\begin{equation}
\text{Lower bound} = \frac{\hat{p} + \frac{z^2}{2n} - z \sqrt{\frac{\hat{p}(1-\hat{p})}{n} + \frac{z^2}{4n^2}}}{1 + \frac{z^2}{n}}
\end{equation}

\begin{equation}
\text{Upper bound} = \frac{\hat{p} + \frac{z^2}{2n} + z \sqrt{\frac{\hat{p}(1-\hat{p})}{n} + \frac{z^2}{4n^2}}}{1 + \frac{z^2}{n}}
\end{equation}

Here, \( \hat{p} = \frac{x}{n} \) is the observed proportion, \( z \) is the z-score corresponding to the desired confidence level (e.g., \( z = 1.96 \) for a 95\% confidence interval), and \( n \) is the total number of attempts. The Wilson score (the midpoint of the interval) provides a point estimate that adjusts for sample size, while the lower and upper bounds define the confidence interval.

This method is particularly valuable in our evaluation because it allows fairer comparisons across models with different numbers of attempts at each stage of the reasoning process. For instance, a model with fewer attempts might have a higher simple proportion but greater uncertainty, which the Wilson score accounts for by penalizing small sample sizes.

\subsubsection{Adjusted Wilson Score}
To simplify the interpretation of our metrics, we modify the original Wilson score by incorporating the width of the confidence interval as a penalty term, resulting in an adjusted Wilson score. This adjustment addresses the challenge of analyzing results with both a point estimate (the Wilson score) and a confidence interval, providing a single, interpretable metric that balances performance with estimate reliability.

The adjusted Wilson score is calculated as:

\begin{equation}
\text{Adjusted Wilson score} = \text{Wilson score} \times \left(1 - \frac{\text{Upper bound} - \text{Lower bound}}{2}\right)
\end{equation}

The term \( \frac{\text{Upper bound} - \text{Lower bound}}{2} \) represents half the width of the confidence interval, reflecting the uncertainty in the estimate. A wider interval (indicating greater uncertainty, often due to a smaller sample size) results in a larger penalty, reducing the adjusted score. Conversely, a narrower interval (indicating lower uncertainty) applies a smaller penalty, preserving more of the original Wilson score.

The adjusted Wilson score is applied to our metrics (\textit{Stage-wise Score}, \textit{Final Stage Completion}, \textit{Final Stage Completion}, \textit{Decision Alignment}, and \textit{Measurement Consistency}), except for \textit{Average Reasoning Depth}, which is a simple average and does not involve proportions. For each metric, we first compute the Wilson score and its confidence interval based on the number of successes and attempts at the relevant stage. We then apply the adjustment to produce a single score that reflects both the model’s performance and the reliability of the estimate.

%% file: Tables/Supple_API_Cost.tex
\begin{table}[htb!]
  \caption{Estimated inference costs for closed-source models}
  \label{table:api_cost}
  \centering
  \begin{tabular}{cccc}
    \toprule
    Model     & Task & Greedy Sampling & Stochastic Sampling  \\
    \midrule
    Gemini-2.5-Pro & Path1 & \$40 & \$91 \\
     & Path2 \& Re-evaluated Path1 & \$14 & \$69 \\
     \midrule
     Gemini-2.5-Flash & Path1 & \$12 & \$25 \\
     & Path2 \& Re-evaluated Path1 & \$5 & \$17 \\
     \midrule
     GPT-4.1 & Path1 & \$18 & \$40 \\
     & Path2 \& Re-evaluated Path1 & \$7 & \$30 \\
    \bottomrule
  \end{tabular}
\end{table}

%% file: Supple/4.Results.tex
\section{Additional Results} \label{apd:additional_results}

\textbf{Qualitative Analysis of Measurement Consistency Results}
In Section~\ref{sec:results}, we concluded that the primary limitation of current LVLMs lies not in their ability to maintain coherence across reasoning stages, but in their tendency to rely on visual cues or shortcuts when making diagnostic decisions, rather than applying diagnostic criteria through structured, measurement-based reasoning.

This section provides qualitative evidence supporting that conclusion, focusing on the top-performing model, \textit{Gemini-2.5-Pro}.

\begin{figure}[htb!]
    \centering
\includegraphics[width=\linewidth]{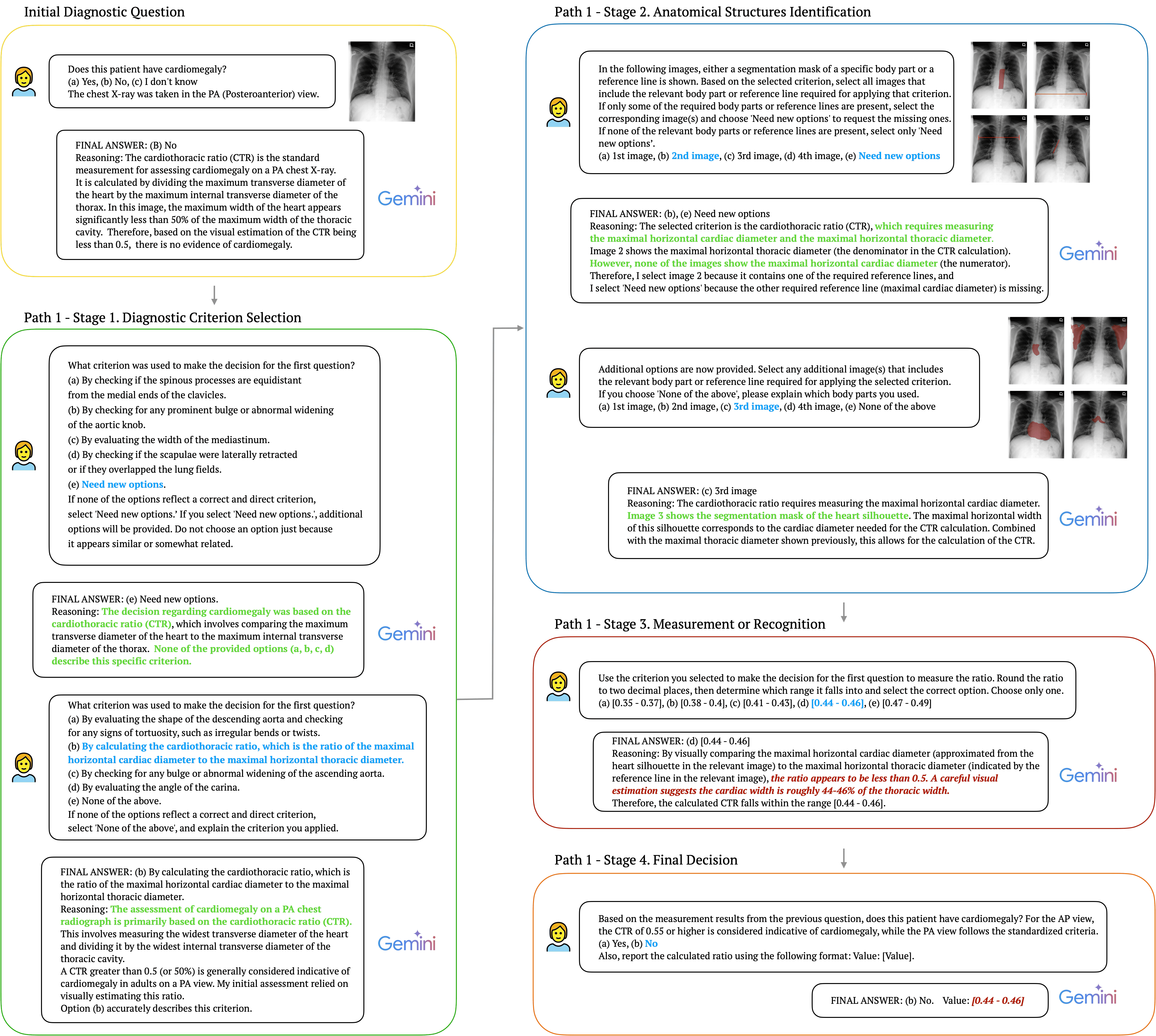}
    \caption{
    Qualitative example from the cardiomegaly task under Path 1, where the model returns the earlier selected value range in the final stage, rather than the specific value that had to be computed.
    Although the model selects a correct range in Stage 3, it fails to perform an actual calculation and instead relies on visual estimation, as reflected in the heuristic reasoning shown.
    }
\label{figure:qualitative_pro_cardio_final_unmatch}
\end{figure}

\begin{figure}[htb!]
    \centering
\includegraphics[width=\linewidth]{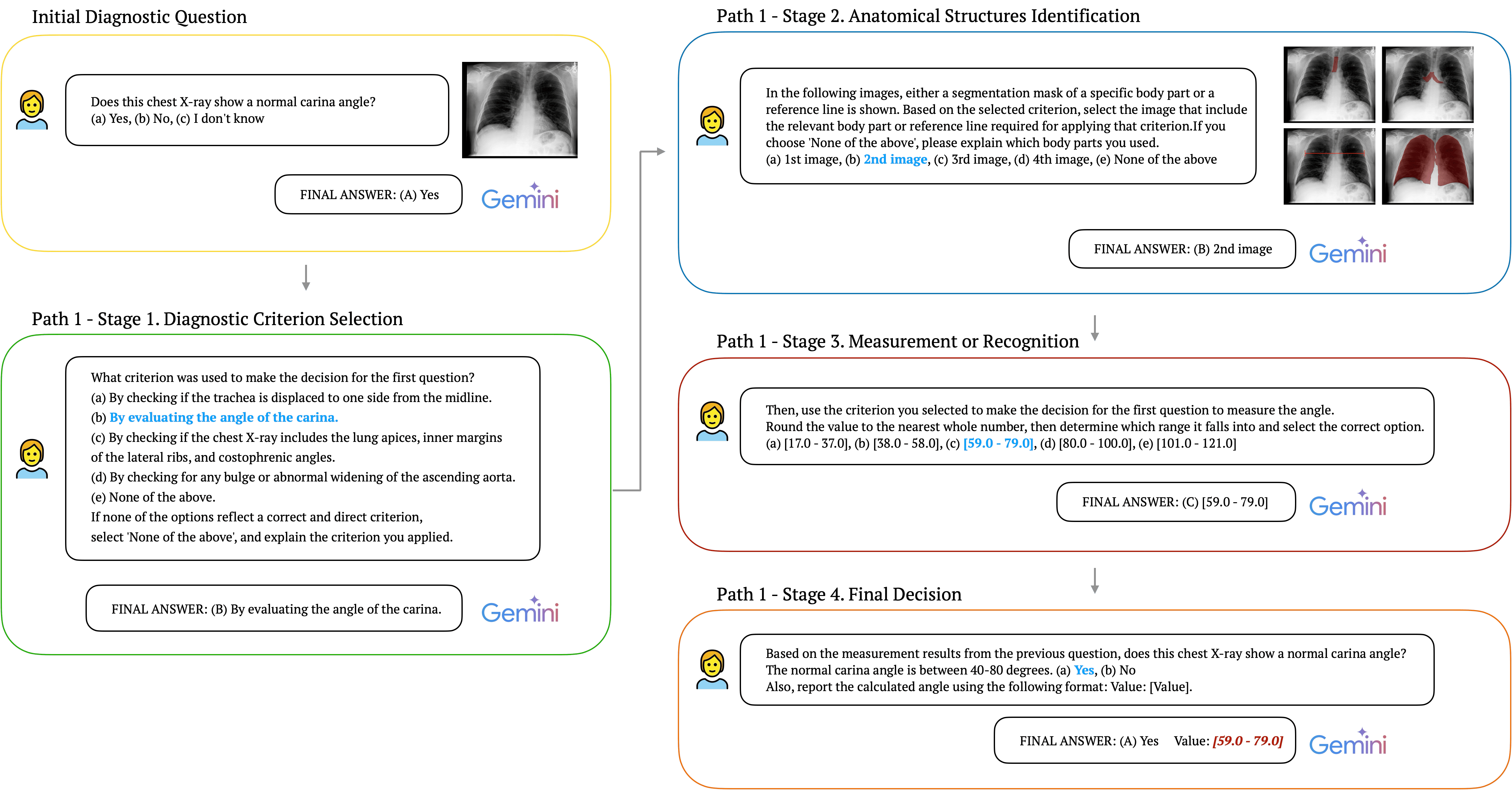}
    \caption{
    Qualitative example from the carina angle task under Path 1, where the model returns the earlier selected value range in the final stage, rather than the specific value that had to be computed.
    The model chooses a range based on visual heuristics and does not return a concrete numeric value, indicating the absence of true measurement-based reasoning.
    }
\label{figure:qualitative_pro_carina_final_unmatch}
\end{figure}

As illustrated in Figures~\ref{figure:qualitative_pro_cardio_final_unmatch} and~\ref{figure:qualitative_pro_carina_final_unmatch}, the model often omits explicit calculations under Path 1.
In many cases, it returns only a value range in Stage 4, rather than a concrete numeric value, indicating that no actual computation was conducted in Stage 3.
A closer inspection of the Stage 3 responses confirms this: the model frequently justifies its choices with heuristic statements such as \textit{``By visually comparing...''} or \textit{``The ratio appears to be less than 0.5. A careful visual estimation suggests the cardiac width is roughly 44–46\% of the thoracic width.''}
These examples demonstrate that even the best-performing model tends to rely on visual estimation, rather than systematically applying diagnostic criteria.

\begin{figure}[htb!]
    \centering
\includegraphics[width=\linewidth]{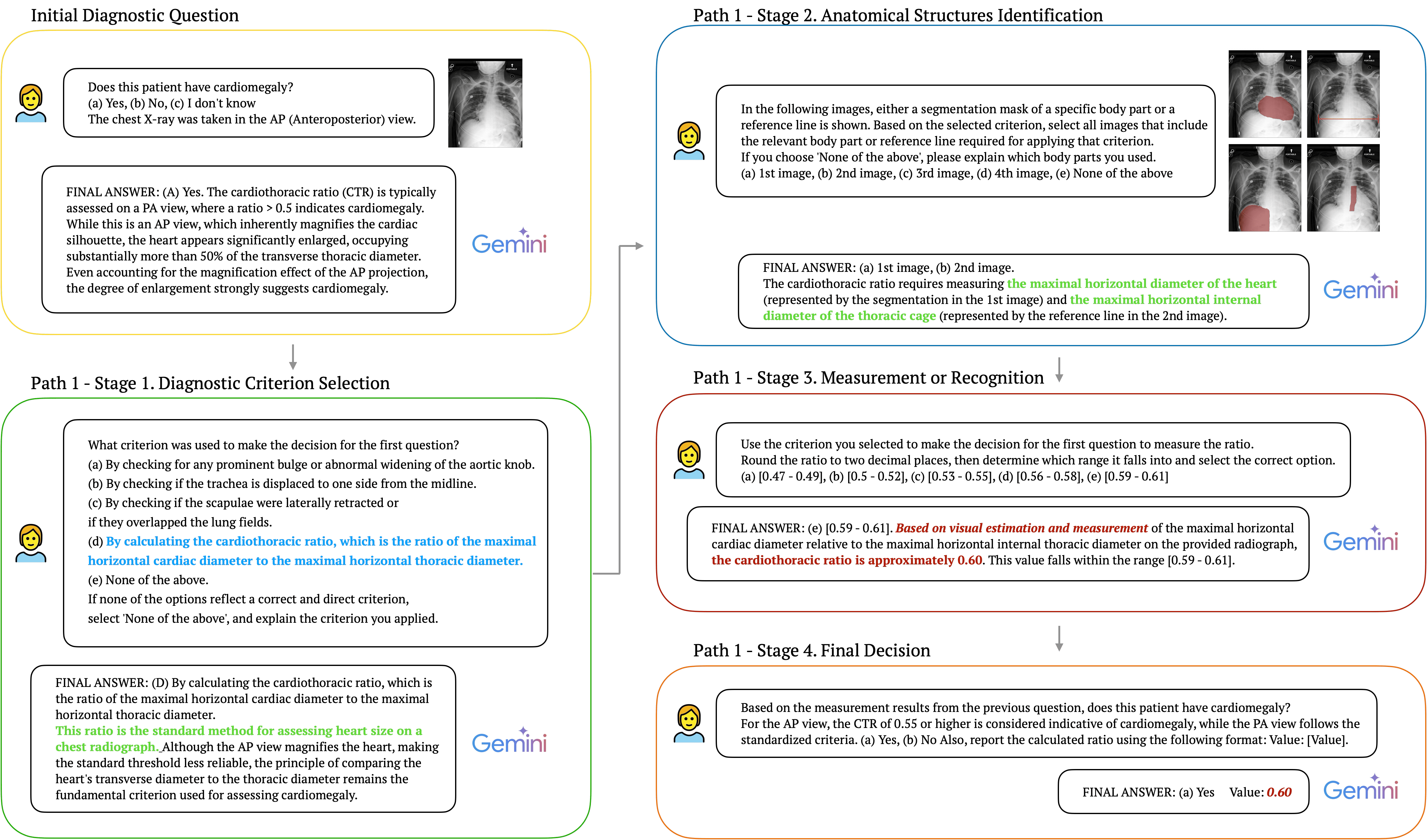}
    \caption{
    Qualitative example from the cardiomegaly task under Path 1, where the returned value matches the selected range but is still based on visual approximation.
    Even though consistency is observed in the outputs, the reasoning lacks structured application of diagnostic criteria.
    }
\label{figure:qualitative_pro_cardio_final_match}
\end{figure}

\begin{figure}[htb!]
    \centering
\includegraphics[width=\linewidth]{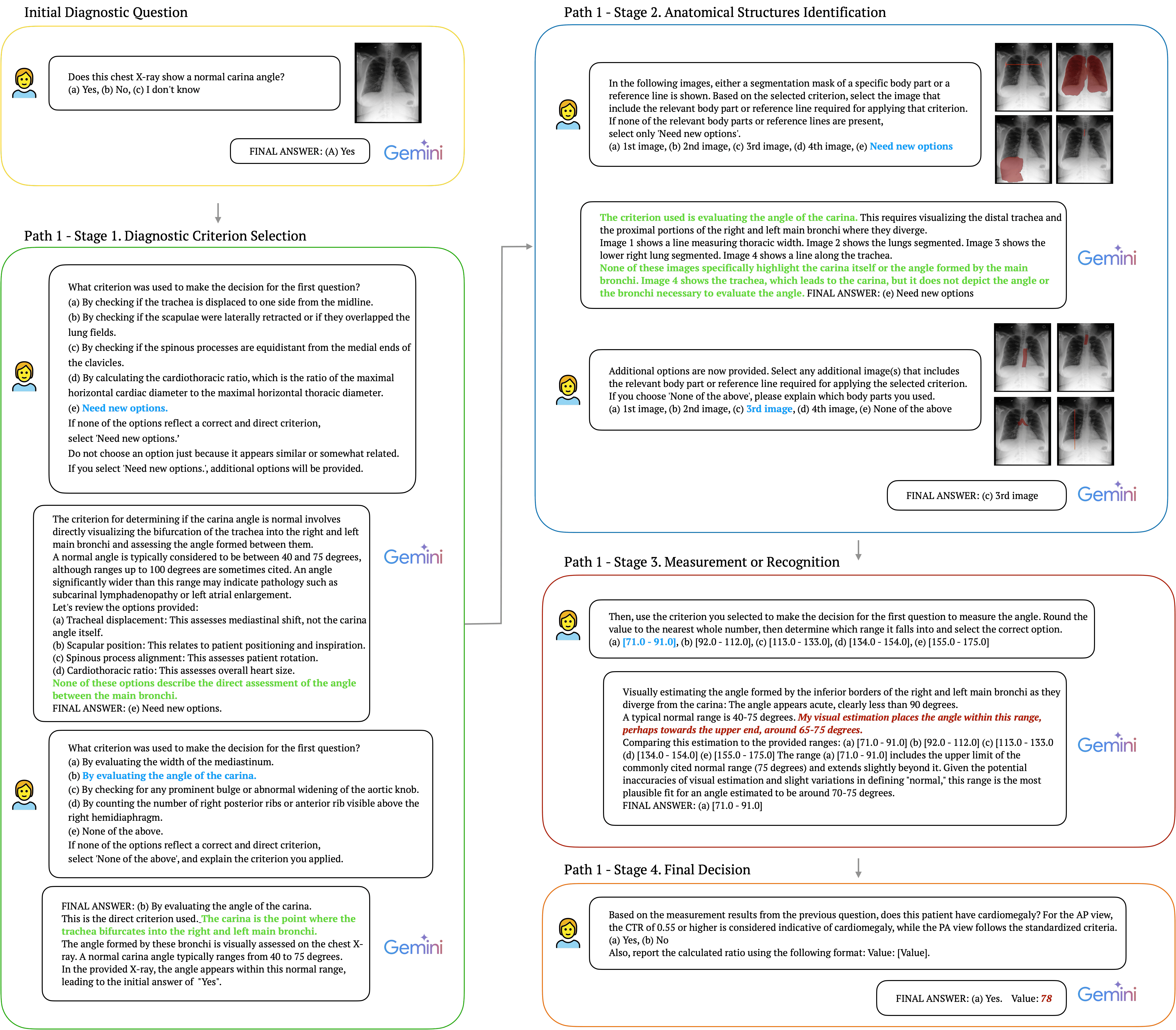}
    \caption{
    Qualitative example from the carina angle task under Path 1 showing apparent consistency.
    The model’s output falls within the selected range, but the reasoning is again heuristic rather than computational.
    }
\label{figure:qualitative_pro_carina_final_match}
\end{figure}

Moreover, as shown in Figures~\ref{figure:qualitative_pro_cardio_final_match} and~\ref{figure:qualitative_pro_carina_final_match}, even when the returned value in Stage 4 falls within the value range selected in Stage 3, the underlying reasoning remains heuristic.
For example, the model states: \textit{``My visual estimation places the angle within this range, perhaps towards the upper end, around 65–75 degrees.''}

\begin{figure}[htb!]
    \centering
\includegraphics[width=\linewidth]{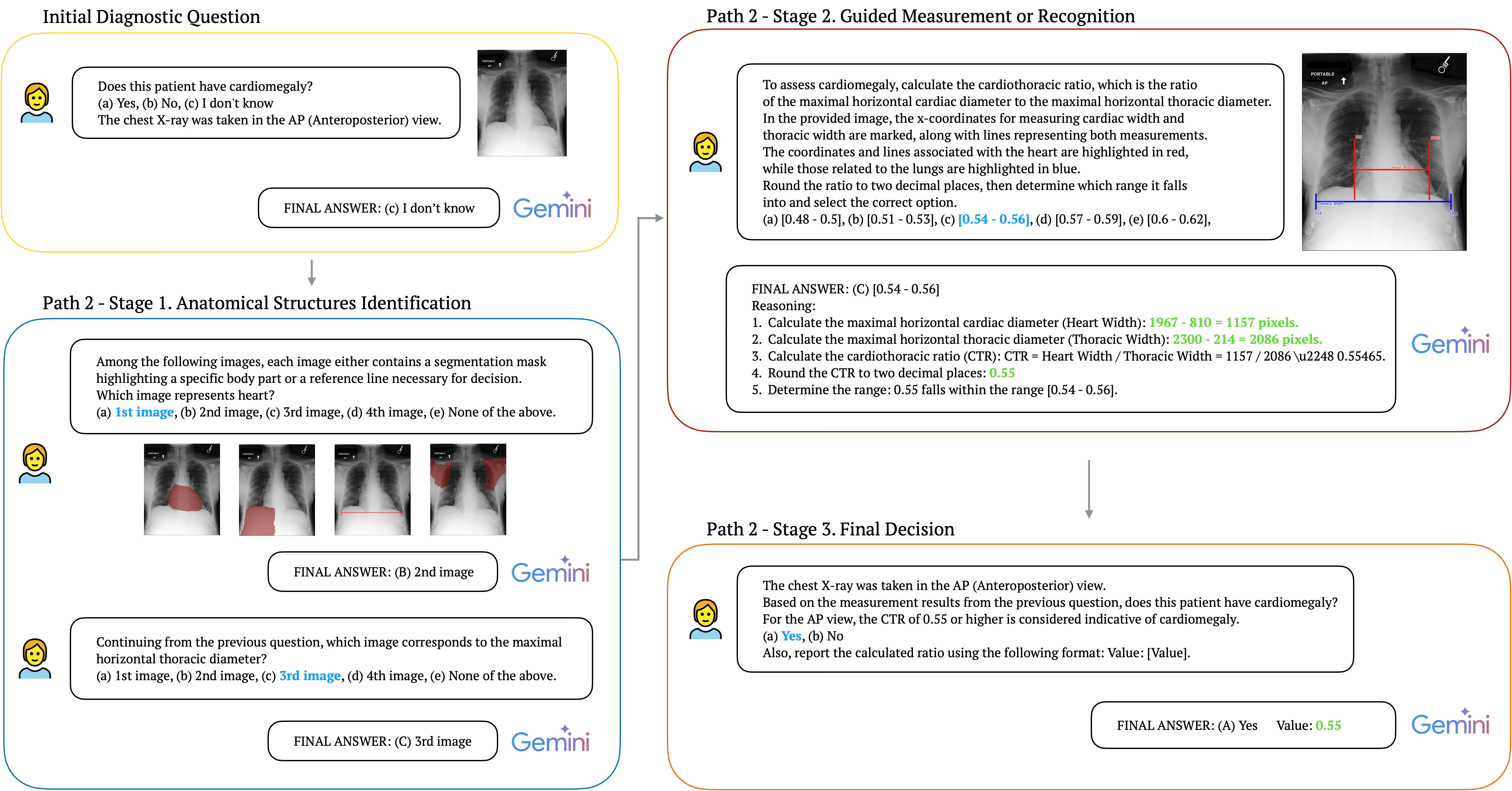}
    \caption{
    Qualitative example from the cardiomegaly task under Path 2, where the model follows the given instructions and performs an explicit computation.
    The final-stage output reflects the calculated value, which aligns with the previously selected range.
    }
\label{figure:qualitative_pro_path2_cardio_final_match}
\end{figure}

\begin{figure}[htb!]
    \centering
\includegraphics[width=\linewidth]{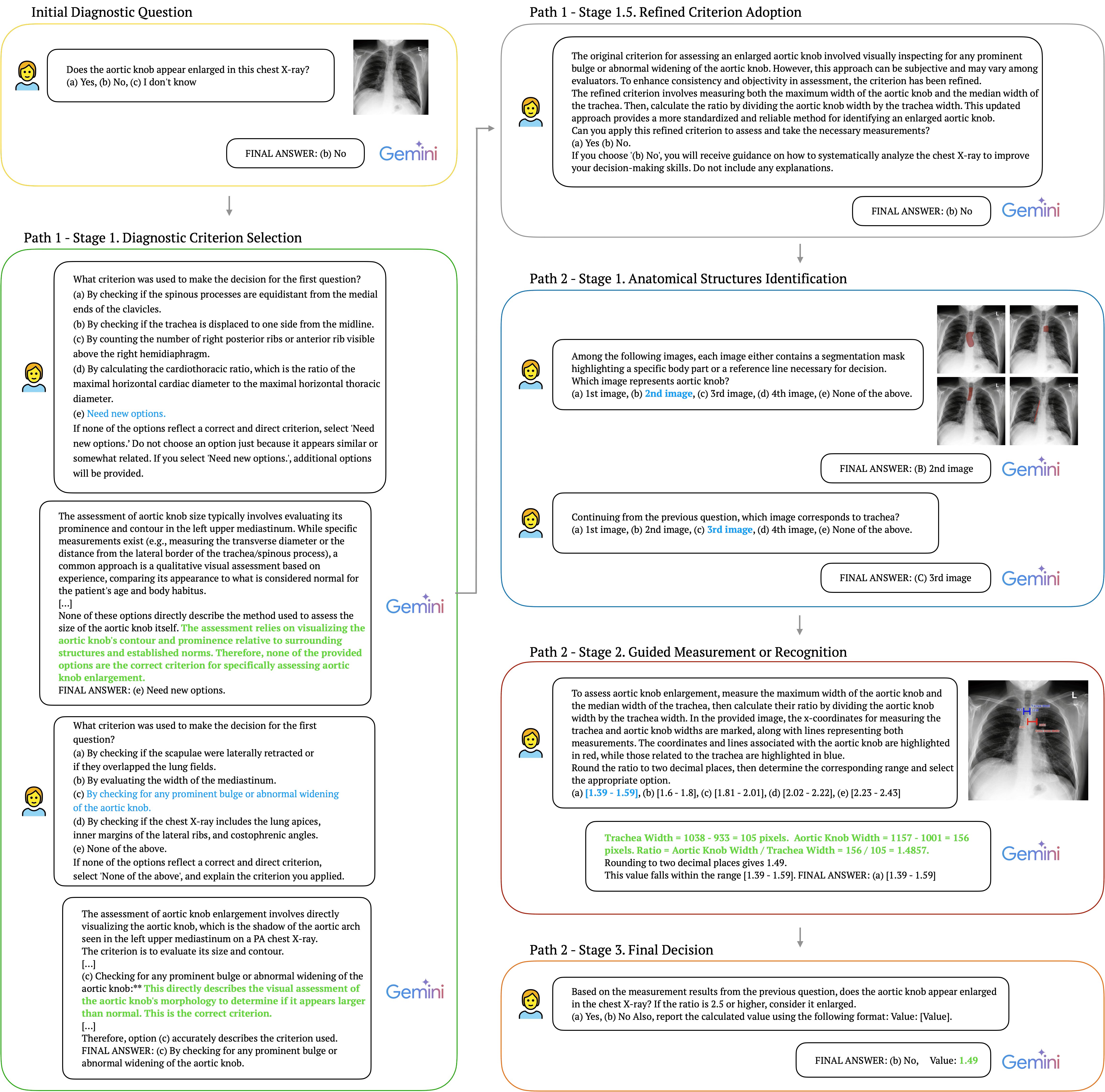}
    \caption{
    Qualitative example from the aortic knob enlargement task under Path 2, demonstrating structured measurement-based reasoning.
    The model applies the instructed computation using visual landmarks and selects the value range accordingly.
    }
\label{figure:qualitative_pro_path2_knob_final_match}
\end{figure}

In contrast, as shown in Figures~\ref{figure:qualitative_pro_path2_cardio_final_match} and~\ref{figure:qualitative_pro_path2_knob_final_match}, under Path 2, where explicit measurement instructions and visual landmarks are provided, the model demonstrates consistent and structured behavior.
The returned value in the final stage reliably falls within the range selected in the previous stage.
More importantly, the Stage 2 responses reveal that the model explicitly follows instructions, performs concrete computations based on visual landmarks, and selects a value range grounded in the calculated result.

These findings highlight a critical weakness in current LVLMs: although their diagnostic decisions may appear reasonable, they often fail to engage with the diagnostic criteria necessary to support those conclusions, frequently resorting to visual estimation and heuristic reasoning instead of performing the required measurements.
As a result, their outputs, while superficially plausible, lack the structured, criteria-driven reasoning expected in clinical decision-making.
This underscores the need for future research to enable LVLMs to internalize and autonomously apply diagnostic criteria through structured, measurement-based reasoning.


\textbf{Decision Alignment Analysis}
As shown in the Alignment columns of Tables~\ref{table:reasoning_greedy_all},
\ref{table:reasoning_greedy_all_recognition}, and\ref{table:reasoning_greedy_all_arithmetic} for Path 1 results, and Tables~\ref{table:review_greedy_all},
\ref{table:review_greedy_all_recognition}, and\ref{table:review_greedy_all_arithmetic} for re-evaluated Path 1 results, all models consistently show higher decision alignment in recognition-type tasks than in measurement-type tasks.

This trend stems from the qualitative nature of recognition tasks, which depend on visual pattern recognition rather than precise numerical calculations.
Recognition tasks typically involve simple, discrete labels (\textit{e.g.}, deviated, included, or excluded), leading to clearer decision boundaries and reducing ambiguity during reasoning.
In contrast, measurement-type tasks require accurate anatomical measurements and comparisons against specific thresholds (\textit{e.g.}, calculating CTR), increasing the likelihood of divergence between initial and final decisions.

The discrete and visually intuitive nature of recognition tasks makes them less susceptible to misinterpretation, contributing to more consistent alignment across stages.

\textbf{Sampling Method Analysis}
Tables~\ref{table:reasoning_shot_overall}, \ref{table:guidance_shot_overall}, and \ref{table:review_shot_overall} present stochastic sampling results, while Tables~\ref{table:reasoning_greedy_overall} and~\ref{table:guidance_greedy_overall} present greedy sampling results.
In the Path 1 evaluation, closed-source models demonstrate robust performance across all stages under both greedy and stochastic sampling. In contrast, open-source models show minimal performance variation in Stage 1 but substantial variability in other stages (See Tables~\ref{table:reasoning_greedy_all} and~\ref{table:reasoning_shot_all}).

Stage 1 involves selecting diagnostic criteria based solely on textual input, requiring no visual grounding. This likely contributes to the stable performance of open-source models at this stage, regardless of sampling strategy.
However, Stages 2 (anatomical structure identification), 3 (measurement or recognition), and 4 (final decision) require visual grounding, application of diagnostic criteria, and retention of prior reasoning steps. Performance in these stages fluctuates considerably for open-source models under stochastic sampling, with some responses improving while others decline.

This instability suggests that open-source models possess less certain or robust multimodal diagnostic reasoning capabilities, especially when visual processing and multi-step reasoning are involved. In contrast, closed-source models maintain consistent performance across all stages of Path 1 and show similarly stable results across corresponding stages in Path 2.

\textbf{Multiple Run Stability Analysis}
\revision{We report averaged results over three seeds (11, 33, 77) with temperature set to 0.7 and top-$p$ to 0.95.}
Results are presented in Tables~\ref{table:reasoning_3run_overall}, ~\ref{table:guidance_3runs_overall}, and ~\ref{table:review_3runs_overall}.

\revision{Overall, closed-source models continue to outperform open-source models; however, even the strongest model, Gemini-2.5-Pro, typically reaches only Stage 2 in Path 1 and rarely completes the full reasoning trajectory.
The results from Path 2 and Re-evaluated Path 1 further demonstrate the limitations of these models in internalizing and applying guidance across multi-stage reasoning tasks.
Notably, the standard deviations for Consistency and Alignment metrics are relatively large, reflecting instability in maintaining coherent reasoning across runs.
These findings underscore the challenges that LVLMs face in achieving structured and consistent reasoning, particularly in clinical settings that demand interpretability and reproducibility.
}

\textbf{Qualitative Analysis of JPG Compression Effects}
\revision{
We evaluated the impact of JPG compression on the diagnostic quality of chest radiographs used in our benchmark. Although JPG is a lossy compression format, the coarse anatomical structures critical for our tasks, such as the heart and lungs, remain clearly distinguishable. Our quality control pipeline was specifically designed to filter out images where compression or suboptimal windowing could compromise anatomical visibility. As shown in Appendix~\ref{apd:chexstruct_clinician_review}, manual review of 1,200 cases (100 per diagnostic task) confirmed that 99.83\% of images retained high-quality delineation of anatomical boundaries. Furthermore, prior work such as CheXZero~\cite{tiu2022expert} has demonstrated radiologist-level performance using JPG-format images, supporting the suitability of JPG compression for research purposes. We note that our benchmark is intended for controlled research evaluation rather than direct clinical deployment, providing a practical tradeoff between accessibility and sufficient image fidelity. The choice of JPG also enabled manageable storage and processing, avoiding the complexities and large size of DICOM datasets.
}

\input{Tables/1.Reasoning/Result_Main_Shot_Overall}
\input{Tables/2.Guidance/Result_Main_Shot_Overall}
\input{Tables/3.Review/Result_Main_Shot_Overall}

\input{Tables/1.Reasoning/Result_Supple_Greedy_ALL}

\input{Tables/1.Reasoning/Result_Supple_Greedy_ALL_Recognition}

\input{Tables/1.Reasoning/Result_Supple_Greedy_ALL_Arithmetic}

\input{Tables/1.Reasoning/Result_Supple_Shot_ALL}

\input{Tables/1.Reasoning/Result_Supple_Shot_ALL_Recognition}

\input{Tables/1.Reasoning/Result_Supple_Shot_ALL_Arithmetic}

\input{Tables/2.Guidance/Result_Supple_Greedy_ALL}

\input{Tables/2.Guidance/Result_Supple_Greedy_ALL_Recognition}

\input{Tables/2.Guidance/Result_Supple_Greedy_ALL_Arithmetic}

\input{Tables/2.Guidance/Result_Supple_Shot_ALL}

\input{Tables/2.Guidance/Result_Supple_Shot_ALL_Recognition}

\input{Tables/2.Guidance/Result_Supple_Shot_ALL_Arithmetic}

\input{Tables/3.Review/Result_Supple_Greedy_ALL}

\input{Tables/3.Review/Result_Supple_Greedy_ALL_Recognition}

\input{Tables/3.Review/Result_Supple_Greedy_ALL_Arithmetic}

\input{Tables/3.Review/Result_Supple_Shot_ALL}

\input{Tables/3.Review/Result_Supple_Shot_ALL_Recognition}

\input{Tables/3.Review/Result_Supple_Shot_ALL_Arithmetic}

\input{Tables/1.Reasoning/Result_Supple_3runs_ALL}
\input{Tables/2.Guidance/Result_Supple_3runs_ALL}
\input{Tables/3.Review/Result_Supple_3runs_ALL}

%% file: Tables/1.Reasoning/Result_Main_Shot_Overall.tex
\begin{table}[htb!]
  \caption{Path 1 evaluation results with stochastic sampling for overall tasks.
  Completion: Percentage of cases completing all reasoning stages; 
  Depth: Average number of reasoning stages reached;
  Consistency: Percentage of cases where the value returned at Stage 4 matches the Stage 3 response;
  Alignment: Percentage of agreement between initial and final decisions;
  Refined scores incorporating measurement consistency are shown in parentheses.
  `N/A' indicates that the model did not reach the required stage to compute the corresponding metric.
  }
  \label{table:reasoning_shot_overall}
  \centering
  {\begin{adjustbox}{width=0.9\linewidth}
  \begin{tabular}{ccccc}
    \toprule
    \textbf{Models} & \textbf{Completion (0-100) $\uparrow$} & \textbf{Depth (0-4) $\uparrow$} & \textbf{Consistency (0-100) $\uparrow$} & \textbf{Alignment (0-100)$\uparrow$} \\
    \midrule

 Gemini-2.5-Pro  & 17.26 (16.72) & 1.99 (1.97) & 70.15  & 61.66 (59.62) \\
Gemini-2.5-Flash  & 11.66 (8.51) & 1.37 (1.3) & 47.56  & 51.43 (39.54) \\
GPT-4.1  & 8.69 (8.69) & 1.15 (1.15) & 55.86  & 40.89 (40.89) \\
\midrule
Pixtral-Large-Instruct-2411  & 5.08 (4.04) & 1.09 (1.07) & 44.96  & 45.63 (37.84) \\
Llama-3.2-90B-Vision-Instruct  & 0.3 (0.3) & 0.57 (0.57) & 45.05  & 29.73 (29.73) \\
Qwen2.5-VL-72B-Instruct  & 2.7 (2.7) & 0.4 (0.4) & N/A  & 20.8 (20.8) \\
Pixtral 12B  & 0.0 (0.0) & 0.38 (0.38) & 45.05  & 0.0 (0.0) \\
Qwen2.5-VL-7B-Instruct  & 0.0 (0.0) & 0.36 (0.36) & 0.0  & 0.0 (0.0) \\
HealthGPT-L14  & 0.48 (0.23) & 0.28 (0.28) & 0.0  & 36.39 (18.2) \\
RadVLM  & 0.29 (0.29) & 0.22 (0.22) & 36.39  & 18.2 (18.2) \\
 
    \bottomrule
  \end{tabular}
  \end{adjustbox}
  }
\end{table}

%% file: Tables/2.Guidance/Result_Main_Shot_Overall.tex
\begin{table}[htb!]
  \caption{Path 2 evaluation results with stochastic sampling for overall tasks.
  Completion: Percentage of cases completing all reasoning stages; 
  Depth: Average number of reasoning stages reached;
  Consistency: Percentage of cases where the value returned at Stage 3 matches the Stage 2 response;
  Refined scores incorporating measurement consistency are shown in parentheses.
  `N/A' indicates that the model did not reach the required stage to compute the corresponding metric.
  }
  \label{table:guidance_shot_overall}
  \centering
  {\begin{adjustbox}{width=0.9\linewidth}
  \begin{tabular}{cccc}
    \toprule
    \textbf{Models} & \textbf{Completion (0-100) $\uparrow$} & \textbf{Depth (0-3) $\uparrow$} & \textbf{Consistency (0-100) $\uparrow$} \\
    \midrule
Gemini-2.5-Pro  & 49.12 (46.56) & 2.12 (2.07) & 73.44  \\
Gemini-2.5-Flash  & 32.44 (28.08) & 1.61 (1.52) & 59.73  \\
GPT-4.1  & 22.14 (21.7) & 1.19 (1.18) & 71.93  \\
\midrule
Pixtral-Large-Instruct-2411  & 26.2 (23.53) & 1.44 (1.38) & 60.53  \\
Llama-3.2-90B-Vision-Instruct  & 2.58 (0.7) & 0.12 (0.09) & 11.89  \\
Qwen2.5-VL-72B-Instruct  & 10.7 (6.41) & 0.64 (0.55) & 37.52  \\
Pixtral 12B  & 7.92 (2.5) & 0.46 (0.33) & 24.26  \\
Qwen2.5-VL-7B-Instruct  & 4.56 (3.32) & 0.37 (0.35) & 34.06  \\
HealthGPT-L14  & 0.0 (0.0) & 0.04 (0.04) & N/A  \\
RadVLM  & 3.43 (1.02) & 0.08 (0.05) & 12.13  \\
 
    \bottomrule
  \end{tabular}
  \end{adjustbox}
  }
\end{table}

%% file: Tables/3.Review/Result_Main_Shot_Overall.tex
\begin{table}[htb!]
  \caption{Re-evaluated Path 1 result with stochastic sampling for overall tasks.
  Completion: Percentage of cases completing all reasoning stages; 
  Depth: Average number of reasoning stages reached;
  Consistency: Percentage of cases where the value returned at Stage 4 matches the Stage 3 response;
  Refined scores incorporating measurement consistency are shown in parentheses.
  `N/A' indicates that the model did not reach the required stage to compute the corresponding metric.
  }
  \label{table:review_shot_overall}
  \centering
  {\begin{adjustbox}{width=0.9\linewidth}
  \begin{tabular}{cccc}
    \toprule
    \textbf{Models} & \textbf{Completion (0-100) $\uparrow$} & \textbf{Depth (0-4) $\uparrow$} & \textbf{Consistency (0-100) $\uparrow$} \\
    \midrule
Gemini-2.5-Pro  & 0.0 (0.0) & 1.93 (1.86) & 49.82  \\
Gemini-2.5-Flash  & 0.0 (0.0) & 1.24 (1.21) & 43.97  \\
GPT-4.1  & 2.72 (2.72) & 1.48 (1.48) & 48.39  \\
\midrule
Pixtral-Large-Instruct-2411  & 0.0 (0.0) & 1.81 (1.66) & 0.0  \\
Llama-3.2-90B-Vision-Instruct  & 0.0 (0.0) & 0.4 (0.4) & N/A  \\
Qwen2.5-VL-72B-Instruct  & 0.0 (0.0) & 1.1 (1.1) & N/A  \\
Pixtral 12B  & 0.0 (0.0) & 1.0 (1.0) & N/A  \\
Qwen2.5-VL-7B-Instruct  & 0.0 (0.0) & 0.7 (0.7) & N/A  \\
HealthGPT-L14  & N/A (N/A) & N/A (N/A) & N/A  \\
RadVLM  & 0.0 (0.0) & 0.0 (0.0) & N/A  \\

    \bottomrule
  \end{tabular}
  \end{adjustbox}
  }
\end{table}

%% file: Tables/1.Reasoning/Result_Supple_Greedy_ALL.tex
\begin{table}[htb!]
  \caption{Path 1 evaluation results with greedy sampling for overall, measurement-type, and recognition-type tasks, respectively.
  Stage-wise Score: Percentage of cases that completed each stage;
  Completion: Percentage of cases completing all reasoning stages; 
  Depth: Average number of reasoning stages reached;
  Consistency: Percentage of cases where the value returned at Stage 4 matches the Stage 3 response;
  Alignment: Percentage of agreement between initial and final decisions;
  Refined scores incorporating measurement consistency are shown in parentheses.
  `N/A' indicates that the model did not reach the required stage to compute the corresponding metric.
  }
  \label{table:reasoning_greedy_all}
  \centering
  {\begin{adjustbox}{width=\linewidth}
  \begin{tabular}{cccccccccc}
    \toprule
    & & \multicolumn{4}{c}{Stage-wise Score} \\
    \cmidrule(r){3-6}
    Task & Models & Stage 1 & Stage 2 & Stage 3 & Stage 4 & Completion & Depth & Consistency & Alignment \\
    
    \midrule
  & Gemini-2.5-Pro & 92.91  & 56.03  & 30.18 (29.08) & 71.58 (68.51) & 17.03 (16.24) & 1.96 (1.95) & 68.4  & 60.88 (58.61) \\
 & Gemini-2.5-Flash & 90.03  & 41.61  & 35.6 (22.81) & 64.4 (43.98) & 12.83 (8.56) & 1.4 (1.31) & 43.76  & 50.29 (37.67) \\
 & GPT-4.1 & 85.34  & 28.46  & 25.98 (25.98) & 52.3 (52.3) & 8.32 (8.32) & 1.15 (1.15) & 61.22  & 39.8 (39.8) \\
 & Pixtral-Large-Instruct-2411 & 85.9  & 19.61  & 28.56 (17.66) & 37.02 (25.37) & 3.73 (2.31) & 1.0 (0.96) & 28.5  & 36.74 (25.3) \\
 & Llama-3.2-90B-Vision-Instruct & 72.24  & 4.07  & 10.49 (10.49) & 37.52 (37.52) & 0.38 (0.38) & 0.53 (0.53) & 61.27  & 23.32 (23.32) \\
Overall & Qwen2.5-VL-72B-Instruct & 76.2  & 14.14  & 22.73 (17.79) & 43.2 (39.12) & 2.34 (2.12) & 0.67 (0.67) & 34.67  & 38.45 (34.36) \\
 & Pixtral 12B & 38.9  & 2.04  & 31.56 (31.56) & 0.0 (0.0) & 0.0 (0.0) & 0.3 (0.3) & 36.39  & 0.0 (0.0) \\
 & Qwen2.5-VL-7B-Instruct & 47.54  & 4.98  & 17.69 (10.15) & 45.44 (28.35) & 1.21 (0.87) & 0.49 (0.48) & 14.86  & 45.44 (28.35) \\
 & MedGemma 27B & 76.52  & 7.15  & 39.08 (19.87) & 47.19 (31.2) & 3.31 (2.34) & 0.69 (0.68) & 17.06  & 47.19 (31.2) \\
 & HealthGPT-L14 & 40.82  & 5.82  & 34.8 (34.8) & 56.8 (56.8) & 1.27 (1.27) & 0.32 (0.32) & 36.39  & 33.87 (33.87) \\
 & RadVLM & 25.12  & 5.94  & 24.56 (14.06) & 32.5 (20.32) & 0.64 (0.39) & 0.28 (0.28) & 25.07  & 32.5 (20.32) \\
 & MedGemma 4B & 35.97  & 5.07  & 34.64 (19.63) & 39.12 (31.61) & 1.13 (0.98) & 0.32 (0.32) & 14.86  & 39.12 (31.61) \\
\midrule
 & Gemini-2.5-Pro & 92.0  & 59.8  & 21.28 (19.55) & 67.26 (62.14) & 13.44 (12.21) & 1.92 (1.89) & 68.4  & 49.71 (45.93) \\
 & Gemini-2.5-Flash & 91.05  & 52.26  & 36.79 (17.6) & 66.74 (36.12) & 14.66 (8.25) & 1.45 (1.32) & 43.76  & 48.72 (29.8) \\
 & GPT-4.1 & 85.44  & 30.07  & 30.06 (30.06) & 48.0 (48.0) & 8.68 (8.68) & 1.07 (1.07) & 61.22  & 36.98 (36.98) \\
 & Pixtral-Large-Instruct-2411 & 88.34  & 24.4  & 30.91 (12.74) & 32.27 (14.8) & 3.89 (1.76) & 0.98 (0.93) & 28.5  & 31.86 (14.68) \\
 & Llama-3.2-90B-Vision-Instruct & 69.67  & 3.93  & 15.74 (15.74) & 37.52 (37.52) & 0.57 (0.57) & 0.4 (0.4) & 61.27  & 23.32 (23.32) \\
Measurement & Qwen2.5-VL-72B-Instruct & 73.5  & 14.8  & 26.59 (18.69) & 40.61 (34.48) & 2.48 (2.16) & 0.59 (0.58) & 34.67  & 37.31 (31.18) \\
 & Pixtral 12B & 38.56  & 0.73  & 36.39 (36.39) & 0.0 (0.0) & 0.0 (0.0) & 0.24 (0.24) & 36.39  & 0.0 (0.0) \\
 & Qwen2.5-VL-7B-Instruct & 47.73  & 4.09  & 15.21 (2.65) & 33.06 (7.43) & 0.68 (0.17) & 0.43 (0.43) & 14.86  & 33.06 (7.43) \\
 & MedGemma 27B & 76.05  & 8.39  & 45.82 (13.8) & 37.66 (16.34) & 3.54 (2.02) & 0.61 (0.59) & 17.06  & 37.66 (16.34) \\
 & HealthGPT-L14 & 42.4  & 4.33  & 16.13 (16.13) & 36.39 (36.39) & 0.34 (0.34) & 0.26 (0.26) & 36.39  & 0.0 (0.0) \\
 & RadVLM & 23.02  & 5.83  & 21.01 (5.25) & 32.5 (8.13) & 0.51 (0.13) & 0.24 (0.24) & 25.07  & 32.5 (8.13) \\
  & MedGemma 4B & 38.55  & 4.47  & 26.83 (4.32) & 22.52 (11.26) & 0.46 (0.23) & 0.3 (0.29) & 14.86  & 22.52 (11.26) \\

\midrule
 & Gemini-2.5-Pro & 94.72  & 49.42  & 45.74  & 78.05  & 23.3  & 2.04  & N/A  & 77.63  \\
 & Gemini-2.5-Flash & 87.99  & 25.64  & 33.23  & 59.71  & 9.16  & 1.28  & N/A  & 53.42  \\
 & GPT-4.1 & 85.16  & 25.63  & 19.85  & 60.9  & 7.61  & 1.3  & N/A  & 45.46  \\
 & Pixtral-Large-Instruct-2411 & 81.01  & 12.42  & 25.04  & 46.52  & 3.41  & 1.03  & N/A  & 46.52  \\
 & Llama-3.2-90B-Vision-Instruct & 77.38  & 4.23  & 0.0  & N/A  & 0.0  & 0.78  & N/A  & N/A  \\
Recognition & Qwen2.5-VL-72B-Instruct & 81.6  & 13.05  & 16.29  & 48.39  & 2.05  & 0.83  & N/A  & 40.73  \\
 & Pixtral 12B & 39.58  & 4.65  & 26.74  & 0.0  & 0.0  & 0.42  & N/A  & 0.0  \\
 & Qwen2.5-VL-7B-Instruct & 47.16  & 6.54  & 21.4  & 70.19  & 2.26  & 0.59  & N/A  & 70.19  \\
 & MedGemma 27B & 77.47  & 5.61  & 28.96  & 75.79  & 2.92  & 0.83  & N/A  & 75.79  \\
 & HealthGPT-L14 & 37.67  & 7.67  & 53.48  & 77.22  & 3.14  & 0.45  & N/A  & 67.74  \\
 & RadVLM & 29.31  & 6.16  & 31.67  & 32.5  & 0.91  & 0.35  & N/A  & 32.5  \\
  & MedGemma 4B & 30.82  & 6.11  & 50.27  & 72.32  & 2.48  & 0.38  & N/A  & 72.32  \\

    \bottomrule
  \end{tabular}
  \end{adjustbox}
  }
\end{table}

%% file: Tables/1.Reasoning/Result_Supple_Greedy_ALL_Recognition.tex
\begin{table}
  \caption{Path 1 evaluation results with greedy sampling on recognition-type tasks.
  Stage-wise Score: Percentage of cases that completed each stage;
  Completion: Percentage of cases completing all reasoning stages; 
  Depth: Average number of reasoning stages reached;
  Alignment: Percentage of agreement between initial and final decisions;
  Refined scores incorporating measurement consistency are shown in parentheses.
  `N/A' indicates that the model did not reach the required stage to compute the corresponding metric.
  }
  \label{table:reasoning_greedy_all_recognition}
  \centering
  {\begin{adjustbox}{width=\linewidth}
  \begin{tabular}{cccccccccc}
    \toprule
    & & \multicolumn{4}{c}{Stage-wise Score} \\
    \cmidrule(r){3-6}

    Task & Models & Stage 1 & Stage 2 & Stage 3 & Stage 4 & Completion & Depth & Consistency & Alignment \\
    
    \midrule
   & Gemini-2.5-Pro & 93.12  & 47.81  & 46.57  & 88.3  & 26.31  & 2.06  & N/A  & 88.3  \\
 & Gemini-2.5-Flash & 96.33  & 32.29  & 30.46  & 77.22  & 12.55  & 1.59  & N/A  & 77.22  \\
 & GPT-4.1 & 78.72  & 31.36  & 23.1  & 67.71  & 8.22  & 1.29  & N/A  & 67.71  \\
 & Pixtral-Large-Instruct-2411 & 85.52  & 24.65  & 12.43  & 45.05  & 3.65  & 1.2  & N/A  & 45.05  \\
 & Llama-3.2-90B-Vision-Instruct & 76.82  & 0.0  & N/A  & N/A  & 0.0  & 0.84  & N/A  & N/A  \\
Trachea Deviation & Qwen2.5-VL-72B-Instruct & 90.4  & 6.61  & 37.52  & 51.73  & 4.56  & 1.07  & N/A  & 51.73  \\
 & Pixtral 12B & 49.57  & 0.0  & N/A  & N/A  & 0.0  & 0.55  & N/A  & N/A  \\
 & Qwen2.5-VL-7B-Instruct & 77.86  & 3.19  & 0.0  & N/A  & 0.0  & 0.86  & N/A  & N/A  \\
 & MedGemma 27B & 89.13  & 7.61  & 0.0  & N/A  & 0.0  & 1.01  & N/A  & N/A  \\
 & HealthGPT-L14 & 41.72  & 0.0  & N/A  & N/A  & 0.0  & 0.46  & N/A  & N/A  \\
 & RadVLM & 42.59  & 0.0  & N/A  & N/A  & 0.0  & 0.47  & N/A  & N/A  \\
  & MedGemma 4B & 17.75  & 0.0  & N/A  & N/A  & 0.0  & 0.18  & N/A  & N/A  \\

\midrule
 & Gemini-2.5-Pro & 96.33  & 89.13  & 53.85  & 69.09  & 40.86  & 2.97  & N/A  & 67.41  \\
 & Gemini-2.5-Flash & 94.63  & 65.66  & 32.83  & 65.53  & 20.33  & 2.18  & N/A  & 46.64  \\
 & GPT-4.1 & 96.33  & 59.44  & 29.22  & 78.61  & 19.47  & 2.07  & N/A  & 32.27  \\
 & Pixtral-Large-Instruct-2411 & 81.05  & 6.15  & 32.5  & 29.73  & 2.74  & 0.95  & N/A  & 29.73  \\
 & Llama-3.2-90B-Vision-Instruct & 87.89  & 16.93  & 0.0  & N/A  & 0.0  & 1.1  & N/A  & N/A  \\
Inclusion & Qwen2.5-VL-72B-Instruct & 96.33  & 25.46  & 11.36  & 45.05  & 3.65  & 1.31  & N/A  & 29.73  \\
 & Pixtral 12B & 67.87  & 13.95  & 26.74  & 0.0  & 0.0  & 0.88  & N/A  & 0.0  \\
 & Qwen2.5-VL-7B-Instruct & 52.22  & 22.97  & 42.81  & 70.19  & 9.03  & 0.88  & N/A  & 70.19  \\
 & MedGemma 27B & 89.13  & 7.61  & 0.0  & N/A  & 0.0  & 1.01  & N/A  & N/A  \\
 & MedGemma 27B & 90.4  & 14.82  & 57.93  & 75.79  & 11.67  & 1.32  & N/A  & 75.79  \\
 & HealthGPT-L14 & 45.19  & 30.68  & 53.48  & 77.22  & 12.55  & 0.91  & N/A  & 67.74  \\
 & RadVLM & 34.85  & 24.64  & 31.67  & 32.5  & 3.65  & 0.54  & N/A  & 32.5  \\
  & MedGemma 4B & 45.19  & 24.43  & 50.27  & 72.32  & 9.92  & 0.81  & N/A  & 72.32  \\

\midrule
 & Gemini-2.5-Pro & 96.33  & 46.94  & 26.67  & 80.65  & 15.16  & 1.82  & N/A  & 80.65  \\
 & Gemini-2.5-Flash & 89.13  & 0.0  & N/A  & N/A  & 0.0  & 0.5  & N/A  & N/A  \\
 & GPT-4.1 & 96.3  & 4.61  & 27.09  & 36.39  & 2.76  & 1.05  & N/A  & 36.39  \\
 & Pixtral-Large-Instruct-2411 & 86.69  & 8.7  & 0.0  & N/A  & 0.0  & 1.0  & N/A  & N/A  \\
 & Llama-3.2-90B-Vision-Instruct & 72.76  & 0.0  & N/A  & N/A  & 0.0  & 0.7  & N/A  & N/A  \\
Inspiration & Qwen2.5-VL-72B-Instruct & 82.14  & 7.08  & 0.0  & N/A  & 0.0  & 0.94  & N/A  & N/A  \\
 & Pixtral 12B & 0.0  & N/A  & N/A  & N/A  & 0.0  & 0.0  & N/A  & N/A  \\
 & Qwen2.5-VL-7B-Instruct & 6.36  & 0.0  & N/A  & N/A  & 0.0  & 0.05  & N/A  & N/A  \\
 & MedGemma 27B & 57.61  & 0.0  & N/A  & N/A  & 0.0  & 0.57  & N/A  & N/A  \\
 & HealthGPT-L14 & 21.18  & 0.0  & N/A  & N/A  & 0.0  & 0.22  & N/A  & N/A  \\
 & RadVLM & 8.38  & 0.0  & N/A  & N/A  & 0.0  & 0.07  & N/A  & N/A  \\
  & MedGemma 4B & 16.02  & 0.0  & N/A  & N/A  & 0.0  & 0.16  & N/A  & N/A  \\

\midrule
 & Gemini-2.5-Pro & 93.12  & 13.79  & 55.89  & 74.17  & 10.89  & 1.31  & N/A  & 74.17  \\
 & Gemini-2.5-Flash & 71.86  & 4.59  & 36.39  & 36.39  & 3.76  & 0.85  & N/A  & 36.39  \\
 & GPT-4.1 & 69.27  & 7.1  & 0.0  & N/A  & 0.0  & 0.81  & N/A  & N/A  \\
 & Pixtral-Large-Instruct-2411 & 70.78  & 10.19  & 55.25  & 64.77  & 7.26  & 0.97  & N/A  & 64.77  \\
 & Llama-3.2-90B-Vision-Instruct & 72.03  & 0.0  & N/A  & N/A  & 0.0  & 0.49  & N/A  & N/A  \\
Ascending Aorta Enlargement & Qwen2.5-VL-72B-Instruct & 57.51  & N/A  & N/A  & N/A  & 0.0  & 0.0  & N/A  & N/A  \\
 & Pixtral 12B & 40.86  & 0.0  & N/A  & N/A  & 0.0  & 0.26  & N/A  & N/A  \\
 & Qwen2.5-VL-7B-Instruct & 52.22  & 0.0  & N/A  & N/A  & 0.0  & 0.58  & N/A  & N/A  \\
 & MedGemma 27B & 72.76  & 0.0  & N/A  & N/A  & 0.0  & 0.43  & N/A  & N/A  \\
 & HealthGPT-L14 & 42.59  & 0.0  & N/A  & N/A  & 0.0  & 0.21  & N/A  & N/A  \\
 & RadVLM & 31.43  & 0.0  & N/A  & N/A  & 0.0  & 0.33  & N/A  & N/A  \\
  & MedGemma 4B & 44.32  & 0.0  & N/A  & N/A  & 0.0  & 0.38  & N/A  & N/A  \\

    \bottomrule
  \end{tabular}
  \end{adjustbox}
  }
\end{table}

%% file: Tables/1.Reasoning/Result_Supple_Greedy_ALL_Arithmetic.tex
\begin{table}[htb!]
  \caption{Path 1 evaluation results with greedy sampling on measurement-type tasks.
  Stage-wise Score: Percentage of cases that completed each stage;
  Completion: Percentage of cases completing all reasoning stages; 
  Depth: Average number of reasoning stages reached;
  Consistency: Percentage of cases where the value returned at Stage 4 matches the Stage 3 response;
  Alignment: Percentage of agreement between initial and final decisions;
  Refined scores incorporating measurement consistency are shown in parentheses.
  `N/A' indicates that the model did not reach the required stage to compute the corresponding metric.
  }
  \label{table:reasoning_greedy_all_arithmetic}
  \centering
  {\begin{adjustbox}{width=\linewidth, height=0.4\textheight, keepaspectratio=true}
  \begin{tabular}{cccccccccc}
    \toprule
    & & \multicolumn{4}{c}{Stage-wise Score} \\
    \cmidrule(r){3-6}

    Task & Models & Stage 1 & Stage 2 & Stage 3 & Stage 4 & Completion & Depth & Consistency & Alignment \\
    
    \midrule
 & Gemini-2.5-Pro & 94.41  & 74.66  & 17.81 (16.54) & 64.01 (59.43) & 13.02 (12.09) & 2.07 (2.05) & 71.0  & 64.01 (59.43) \\
 & Gemini-2.5-Flash & 91.57  & 57.63  & 22.01 (15.72) & 42.81 (30.58) & 9.2 (6.57) & 1.82 (1.76) & 52.47  & 42.81 (30.58) \\
 & GPT-4.1 & 93.12  & 25.57  & 18.03 (18.03) & 0.0 (0.0) & 0.0 (0.0) & 1.31 (1.31) & 45.05  & 0.0 (0.0) \\
 & Pixtral-Large-Instruct-2411 & 96.33  & 22.9  & 32.14 (25.0) & 52.65 (40.95) & 8.15 (6.34) & 1.4 (1.36) & 52.65  & 52.65 (40.95) \\
 & Llama-3.2-90B-Vision-Instruct & 96.23  & 0.0  & N/A (N/A) & N/A (N/A) & 0.0 (0.0) & 1.0 (1.0) & N/A  & N/A (N/A) \\
Cardiomegaly & Qwen2.5-VL-72B-Instruct & 87.92  & 14.12  & 17.03 (17.03) & 36.39 (36.39) & 3.45 (3.45) & 1.1 (1.1) & 36.39  & 36.39 (36.39) \\
 & Pixtral 12B & 54.01  & 4.38  & 36.39 (36.39) & 0.0 (0.0) & 0.0 (0.0) & 0.62 (0.62) & 36.39  & 0.0 (0.0) \\
 & Qwen2.5-VL-7B-Instruct & 66.91  & 4.82  & 29.73 (0.0) & 36.39 (0.0) & 2.74 (0.0) & 0.78 (0.76) & 0.0  & 36.39 (0.0) \\
 & MedGemma 27B & 96.33  & 2.74  & 36.39 (0.0) & 36.39 (0.0) & 2.74 (0.0) & 1.03 (1.01) & 0.0  & 36.39 (0.0) \\
 & HealthGPT-L14 & 45.19  & 0.0  & N/A (N/A) & N/A (N/A) & 0.0 (0.0) & 0.5 (0.5) & N/A  & N/A (N/A) \\
 & RadVLM & 26.31  & 0.0  & N/A (N/A) & N/A (N/A) & 0.0 (0.0) & 0.28 (0.28) & N/A  & N/A (N/A) \\
 & MedGemma 4B & 46.06  & 5.06  & 36.39 (0.0) & 0.0 (0.0) & 0.0 (0.0) & 0.53 (0.52) & 0.0  & 0.0 (0.0) \\
\midrule
 & Gemini-2.5-Pro & 96.33  & 78.91  & 36.92 (31.64) & 67.08 (57.49) & 26.31 (22.56) & 2.49 (2.43) & 72.68  & 64.42 (55.22) \\
 & Gemini-2.5-Flash & 96.3  & 91.65  & 34.37 (32.46) & 86.33 (81.54) & 32.58 (30.77) & 2.69 (2.65) & 82.67  & 82.67 (78.08) \\
 & GPT-4.1 & 96.15  & 67.42  & 50.93 (50.93) & 84.51 (84.51) & 37.38 (37.38) & 2.59 (2.59) & 91.62  & 71.0 (71.0) \\
 & Pixtral-Large-Instruct-2411 & 94.63  & 63.76  & 35.99 (10.28) & 28.68 (8.2) & 9.92 (2.83) & 2.06 (1.84) & 26.2  & 26.2 (7.49) \\
 & Llama-3.2-90B-Vision-Instruct & 87.89  & 14.18  & 31.48 (31.48) & 37.52 (37.52) & 4.56 (4.56) & 1.15 (1.15) & 61.27  & 23.32 (23.32) \\
Carina Angle & Qwen2.5-VL-72B-Instruct & 90.53  & 13.8  & 45.48 (45.48) & 64.77 (64.77) & 9.88 (9.88) & 1.27 (1.27) & 64.77  & 51.57 (51.57) \\
 & Pixtral 12B & 48.69  & 0.0  & N/A (N/A) & N/A (N/A) & 0.0 (0.0) & 0.54 (0.54) & N/A  & N/A (N/A) \\
 & Qwen2.5-VL-7B-Instruct & 75.79  & 19.92  & 15.89 (7.95) & 29.73 (14.86) & 2.74 (1.37) & 1.03 (1.02) & 29.73  & 29.73 (14.86) \\
 & MedGemma 27B & 96.33  & 29.73  & 64.7 (41.41) & 76.58 (49.01) & 22.04 (14.11) & 1.8 (1.64) & 51.19  & 76.58 (49.01) \\
 & HealthGPT-L14 & 44.32  & 21.66  & 16.13 (16.13) & 36.39 (36.39) & 2.74 (2.74) & 0.62 (0.62) & 36.39  & 0.0 (0.0) \\
 & RadVLM & 13.5  & 25.39  & 0.0 (0.0) & N/A (N/A) & 0.0 (0.0) & 0.16 (0.16) & N/A  & N/A (N/A) \\
  & MedGemma 4B & 47.81  & 26.26  & 17.26 (8.63) & 45.05 (22.52) & 3.65 (1.83) & 0.72 (0.7) & 29.73  & 45.05 (22.52) \\

\midrule
 & Gemini-2.5-Pro & 96.33  & 65.01  & 25.97 (22.08) & 77.78 (66.11) & 18.61 (15.82) & 2.11 (2.06) & 67.14  & 39.96 (33.97) \\
 & Gemini-2.5-Flash & 96.33  & 48.69  & 21.45 (12.51) & 67.74 (39.51) & 11.67 (6.81) & 1.77 (1.68) & 42.59  & 67.74 (39.51) \\
 & GPT-4.1 & 96.26  & 31.13  & 25.32 (25.32) & 72.32 (72.32) & 10.1 (10.1) & 1.52 (1.52) & 72.32  & 72.32 (72.32) \\
 & Pixtral-Large-Instruct-2411 & 93.12  & 20.7  & 19.98 (5.0) & 42.24 (10.56) & 4.56 (1.14) & 1.26 (1.21) & 25.07  & 42.24 (10.56) \\
 & Llama-3.2-90B-Vision-Instruct & 74.77  & 5.48  & 0.0 (0.0) & N/A (N/A) & 0.0 (0.0) & 0.85 (0.85) & N/A  & N/A (N/A) \\
Rotation & Qwen2.5-VL-72B-Instruct & 94.63  & 13.54  & 18.87 (0.0) & 0.0 (0.0) & 0.0 (0.0) & 1.14 (1.12) & 0.0  & 0.0 (0.0) \\
 & Pixtral 12B & 22.9  & 0.0  & N/A (N/A) & N/A (N/A) & 0.0 (0.0) & 0.24 (0.24) & N/A  & N/A (N/A) \\
 & Qwen2.5-VL-7B-Instruct & 15.16  & 0.0  & N/A (N/A) & N/A (N/A) & 0.0 (0.0) & 0.15 (0.15) & N/A  & N/A (N/A) \\
 & MedGemma 27B & 40.0  & 9.47  & 36.37 (0.0) & 0.0 (0.0) & 0.0 (0.0) & 0.49 (0.47) & 0.0  & 0.0 (0.0) \\
 & HealthGPT-L14 & 45.19  & 0.0  & N/A (N/A) & N/A (N/A) & 0.0 (0.0) & 0.5 (0.5) & N/A  & N/A (N/A) \\
 & RadVLM & 9.03  & 0.0  & N/A (N/A) & N/A (N/A) & 0.0 (0.0) & 0.08 (0.08) & N/A  & N/A (N/A) \\
  & MedGemma 4B & 31.43  & 0.0  & N/A (N/A) & N/A (N/A) & 0.0 (0.0) & 0.34 (0.34) & N/A  & N/A (N/A) \\

\midrule
 & Gemini-2.5-Pro & 81.05  & 30.7  & 23.72 (23.72) & 55.25 (55.25) & 7.53 (7.53) & 1.3 (1.3) & 67.71  & 38.47 (38.47) \\
 & Gemini-2.5-Flash & 94.63  & 19.63  & 42.59 (6.08) & 55.25 (7.89) & 11.28 (1.61) & 1.39 (1.21) & 20.37  & 31.97 (4.57) \\
 & GPT-4.1 & 88.06  & 21.76  & 37.4 (37.4) & 58.44 (58.44) & 10.08 (10.08) & 1.35 (1.35) & 70.19  & 42.16 (42.16) \\
 & Pixtral-Large-Instruct-2411 & 77.86  & 12.49  & 31.67 (7.92) & 42.24 (10.56) & 4.56 (1.14) & 1.02 (0.97) & 25.07  & 42.24 (10.56) \\
 & Llama-3.2-90B-Vision-Instruct & 81.29  & 0.0  & N/A (N/A) & N/A (N/A) & 0.0 (0.0) & 0.21 (0.21) & N/A  & N/A (N/A) \\
Mediastinal Widening & Qwen2.5-VL-72B-Instruct & 96.33  & 7.46  & 51.57 (30.94) & 61.27 (36.76) & 6.54 (3.93) & 1.16 (1.12) & 37.52  & 61.27 (36.76) \\
 & Pixtral 12B & 40.0  & 0.0  & N/A (N/A) & N/A (N/A) & 0.0 (0.0) & 0.44 (0.44) & N/A  & N/A (N/A) \\
 & Qwen2.5-VL-7B-Instruct & 64.07  & 3.91  & 0.0 (0.0) & N/A (N/A) & 0.0 (0.0) & 0.71 (0.71) & N/A  & N/A (N/A) \\
 & MedGemma 27B & 94.63  & 0.0  & N/A (N/A) & N/A (N/A) & 0.0 (0.0) & 0.97 (0.97) & N/A  & N/A (N/A) \\
 & HealthGPT-L14 & 45.19  & 0.0  & N/A (N/A) & N/A (N/A) & 0.0 (0.0) & 0.45 (0.45) & N/A  & N/A (N/A) \\
 & RadVLM & 24.61  & 0.0  & N/A (N/A) & N/A (N/A) & 0.0 (0.0) & 0.2 (0.2) & N/A  & N/A (N/A) \\
  & MedGemma 4B & 49.57  & 0.0  & N/A (N/A) & N/A (N/A) & 0.0 (0.0) & 0.55 (0.55) & N/A  & N/A (N/A) \\

\midrule
 & Gemini-2.5-Pro & 96.33  & 57.01  & 0.0 (0.0) & N/A (N/A) & 0.0 (0.0) & 2.0 (2.0) & N/A  & N/A (N/A) \\
 & Gemini-2.5-Flash & 87.89  & N/A  & N/A (N/A) & N/A (N/A) & 0.0 (0.0) & 0.0 (0.0) & N/A  & N/A (N/A) \\
 & GPT-4.1 & 82.94  & 42.88  & 12.29 (12.29) & 36.37 (36.37) & 4.71 (4.71) & 1.36 (1.36) & 51.73  & 36.37 (36.37) \\
 & Pixtral-Large-Instruct-2411 & 79.97  & N/A  & N/A (N/A) & N/A (N/A) & 0.0 (0.0) & 0.0 (0.0) & N/A  & N/A (N/A) \\
 & Llama-3.2-90B-Vision-Instruct & 33.14  & 0.0  & N/A (N/A) & N/A (N/A) & 0.0 (0.0) & 0.02 (0.02) & N/A  & N/A (N/A) \\
Projection & Qwen2.5-VL-72B-Instruct & 15.16  & 25.07  & 0.0 (0.0) & N/A (N/A) & 0.0 (0.0) & 0.06 (0.06) & N/A  & N/A (N/A) \\
 & Pixtral 12B & 8.15  & 0.0  & N/A (N/A) & N/A (N/A) & 0.0 (0.0) & 0.05 (0.05) & N/A  & N/A (N/A) \\
 & Qwen2.5-VL-7B-Instruct & 14.75  & 0.0  & N/A (N/A) & N/A (N/A) & 0.0 (0.0) & 0.04 (0.04) & N/A  & N/A (N/A) \\
 & MedGemma 27B & 11.67  & 0.0  & N/A (N/A) & N/A (N/A) & 0.0 (0.0) & 0.01 (0.01) & N/A  & N/A (N/A) \\
 & HealthGPT-L14 & 25.46  & 0.0  & N/A (N/A) & N/A (N/A) & 0.0 (0.0) & 0.01 (0.01) & N/A  & N/A (N/A) \\
 & RadVLM & 2.74  & 0.0  & N/A (N/A) & N/A (N/A) & 0.0 (0.0) & 0.01 (0.01) & N/A  & N/A (N/A) \\
  & MedGemma 4B & 16.02  & N/A  & N/A (N/A) & N/A (N/A) & 0.0 (0.0) & 0.0 (0.0) & N/A  & N/A (N/A) \\

\midrule
 & Gemini-2.5-Pro & 96.33  & 58.34  & 28.34 (26.67) & 82.42 (77.57) & 19.71 (18.55) & 2.06 (2.04) & 74.83  & 49.17 (46.27) \\
 & Gemini-2.5-Flash & 94.63  & 48.39  & 50.22 (17.38) & 77.3 (26.76) & 27.28 (9.44) & 2.13 (1.74) & 30.31  & 32.95 (11.41) \\
 & GPT-4.1 & 77.66  & 0.0  & N/A (N/A) & N/A (N/A) & 0.0 (0.0) & 0.17 (0.17) & N/A  & N/A (N/A) \\
 & Pixtral-Large-Instruct-2411 & 89.13  & 13.34  & 23.32 (0.0) & 0.0 (0.0) & 0.0 (0.0) & 1.02 (1.0) & 0.0  & 0.0 (0.0) \\
 & Llama-3.2-90B-Vision-Instruct & 50.61  & N/A  & N/A (N/A) & N/A (N/A) & 0.0 (0.0) & 0.0 (0.0) & N/A  & N/A (N/A) \\
Aortic Knob Enlargement & Qwen2.5-VL-72B-Instruct & 71.76  & N/A  & N/A (N/A) & N/A (N/A) & 0.0 (0.0) & 0.0 (0.0) & N/A  & N/A (N/A) \\
 & Pixtral 12B & 46.06  & N/A  & N/A (N/A) & N/A (N/A) & 0.0 (0.0) & 0.0 (0.0) & N/A  & N/A (N/A) \\
 & Qwen2.5-VL-7B-Instruct & 74.77  & 0.0  & N/A (N/A) & N/A (N/A) & 0.0 (0.0) & 0.4 (0.4) & N/A  & N/A (N/A) \\
 & MedGemma 27B & 79.97  & N/A  & N/A (N/A) & N/A (N/A) & 0.0 (0.0) & 0.0 (0.0) & N/A  & N/A (N/A) \\
 & HealthGPT-L14 & 45.19  & N/A  & N/A (N/A) & N/A (N/A) & 0.0 (0.0) & 0.0 (0.0) & N/A  & N/A (N/A) \\
 & RadVLM & 36.56  & 0.0  & N/A (N/A) & N/A (N/A) & 0.0 (0.0) & 0.4 (0.4) & N/A  & N/A (N/A) \\
  & MedGemma 4B & 40.86  & 0.0  & N/A (N/A) & N/A (N/A) & 0.0 (0.0) & 0.07 (0.07) & N/A  & N/A (N/A) \\

\midrule
 & Gemini-2.5-Pro & 78.91  & 53.99  & 16.2 (16.2) & 57.01 (57.01) & 8.92 (8.92) & 1.38 (1.38) & 57.01  & 42.24 (42.24) \\
 & Gemini-2.5-Flash & 81.51  & 47.54  & 50.11 (21.48) & 71.0 (30.43) & 25.27 (10.83) & 1.83 (1.52) & 34.18  & 34.18 (14.65) \\
 & GPT-4.1 & 64.08  & 21.76  & 36.39 (36.39) & 36.39 (36.39) & 7.16 (7.16) & 0.26 (0.26) & 36.39  & 0.0 (0.0) \\
 & Pixtral-Large-Instruct-2411 & 81.05  & 13.18  & 42.37 (28.25) & 27.82 (18.55) & 3.95 (2.63) & 1.07 (1.03) & 42.02  & 27.82 (18.55) \\
 & Llama-3.2-90B-Vision-Instruct & 50.31  & N/A  & N/A (N/A) & N/A (N/A) & 0.0 (0.0) & 0.0 (0.0) & N/A  & N/A (N/A) \\
Descending Aorta Enlargement & Qwen2.5-VL-72B-Instruct & 57.41  & N/A  & N/A (N/A) & N/A (N/A) & 0.0 (0.0) & 0.0 (0.0) & N/A  & N/A (N/A) \\
 & Pixtral 12B & 46.06  & 0.0  & N/A (N/A) & N/A (N/A) & 0.0 (0.0) & 0.06 (0.06) & N/A  & N/A (N/A) \\
 & Qwen2.5-VL-7B-Instruct & 61.28  & 0.0  & N/A (N/A) & N/A (N/A) & 0.0 (0.0) & 0.35 (0.35) & N/A  & N/A (N/A) \\
 & MedGemma 27B & 93.12  & N/A  & N/A (N/A) & N/A (N/A) & 0.0 (0.0) & 0.0 (0.0) & N/A  & N/A (N/A) \\
 & HealthGPT-L14 & 43.45  & N/A  & N/A (N/A) & N/A (N/A) & 0.0 (0.0) & 0.0 (0.0) & N/A  & N/A (N/A) \\
 & RadVLM & 35.71  & 0.0  & N/A (N/A) & N/A (N/A) & 0.0 (0.0) & 0.36 (0.36) & N/A  & N/A (N/A) \\
  & MedGemma 4B & 31.43  & 0.0  & N/A (N/A) & N/A (N/A) & 0.0 (0.0) & 0.01 (0.01) & N/A  & N/A (N/A) \\

\midrule
 & Gemini-2.5-Pro & 96.33  & N/A  & N/A (N/A) & N/A (N/A) & N/A (N/A) & N/A (N/A) & N/A  & N/A (N/A) \\
 & Gemini-2.5-Flash & 85.52  & N/A  & N/A (N/A) & N/A (N/A) & 0.0 (0.0) & 0.0 (0.0) & N/A  & N/A (N/A) \\
 & GPT-4.1 & 85.25  & N/A  & N/A (N/A) & N/A (N/A) & 0.0 (0.0) & 0.0 (0.0) & N/A  & N/A (N/A) \\
 & Pixtral-Large-Instruct-2411 & 94.63  & N/A  & N/A (N/A) & N/A (N/A) & 0.0 (0.0) & 0.0 (0.0) & N/A  & N/A (N/A) \\
 & Llama-3.2-90B-Vision-Instruct & 83.09  & N/A  & N/A (N/A) & N/A (N/A) & 0.0 (0.0) & 0.0 (0.0) & N/A  & N/A (N/A) \\
Descending Aorta Tortuous & Qwen2.5-VL-72B-Instruct & 74.22  & N/A  & N/A (N/A) & N/A (N/A) & 0.0 (0.0) & 0.0 (0.0) & N/A  & N/A (N/A) \\
 & Pixtral 12B & 42.59  & N/A  & N/A (N/A) & N/A (N/A) & 0.0 (0.0) & 0.0 (0.0) & N/A  & N/A (N/A) \\
 & Qwen2.5-VL-7B-Instruct & 9.12  & N/A  & N/A (N/A) & N/A (N/A) & 0.0 (0.0) & 0.0 (0.0) & N/A  & N/A (N/A) \\
 & MedGemma 27B & 96.33  & N/A  & N/A (N/A) & N/A (N/A) & N/A (N/A) & N/A (N/A) & N/A  & N/A (N/A) \\
 & HealthGPT-L14 & 45.19  & N/A  & N/A (N/A) & N/A (N/A) & 0.0 (0.0) & 0.0 (0.0) & N/A  & N/A (N/A) \\
 & RadVLM & 35.71  & 21.23  & 42.02 (10.5) & 32.5 (8.13) & 4.07 (1.02) & 0.45 (0.4) & 25.07  & 32.5 (8.13) \\
  & MedGemma 4B & 45.19  & 0.0  & N/A (N/A) & N/A (N/A) & 0.0 (0.0) & 0.14 (0.14) & N/A  & N/A (N/A) \\

    \bottomrule
  \end{tabular}
  \end{adjustbox}
  }
\end{table}

%% file: Tables/1.Reasoning/Result_Supple_Shot_ALL.tex
\begin{table}[htb!]
  \caption{
  Path 1 evaluation results with stochstic sampling for overall, measurement-type, and recognition-type tasks, respectively.
  Stage-wise Score: Percentage of cases that completed each stage;
  Completion: Percentage of cases completing all reasoning stages; 
  Depth: Average number of reasoning stages reached;
  Consistency: Percentage of cases where the value returned at Stage 4 matches the Stage 3 response;
  Alignment: Percentage of agreement between initial and final decisions;
  Refined scores incorporating measurement consistency are shown in parentheses.
  `N/A' indicates that the model did not reach the required stage to compute the corresponding metric.
  }
  \label{table:reasoning_shot_all}
  \centering
  {\begin{adjustbox}{width=\linewidth, height=0.5\textheight, keepaspectratio=true}
  \begin{tabular}{cccccccccc}
    \toprule
    & & \multicolumn{4}{c}{Stage-wise Score} \\
    \cmidrule(r){3-6}

    Task & Models & Stage 1 & Stage 2 & Stage 3 & Stage 4 & Completion & Depth & Consistency & Alignment \\
    
    \midrule
  & Gemini-2.5-Pro & 93.39  & 56.37  & 30.06 (29.32) & 71.83 (69.56) & 17.26 (16.72) & 1.99 (1.97) & 70.15  & 61.66 (59.62) \\
 & Gemini-2.5-Flash & 87.58  & 41.12  & 32.29 (23.76) & 62.29 (45.73) & 11.66 (8.51) & 1.37 (1.3) & 47.56  & 51.43 (39.54) \\
 & GPT-4.1 & 85.36  & 30.64  & 25.17 (25.17) & 53.41 (53.41) & 8.69 (8.69) & 1.15 (1.15) & 55.86  & 40.89 (40.89) \\
 & Pixtral-Large-Instruct-2411 & 86.26  & 17.03  & 25.9 (21.32) & 49.47 (40.62) & 5.08 (4.04) & 1.09 (1.07) & 44.96  & 45.63 (37.84) \\
 & Llama-3.2-90B-Vision-Instruct & 72.26  & 1.48  & 14.95 (14.95) & 29.73 (29.73) & 0.3 (0.3) & 0.57 (0.57) & 45.05  & 29.73 (29.73) \\
Overall & Qwen2.5-VL-72B-Instruct & 46.63  & 13.38  & 49.17 (49.17) & 51.2 (51.2) & 2.7 (2.7) & 0.4 (0.4) & N/A  & 20.8 (20.8) \\
 & Pixtral 12B & 40.21  & 5.63  & 14.61 (14.61) & 0.0 (0.0) & 0.0 (0.0) & 0.38 (0.38) & 45.05  & 0.0 (0.0) \\
 & Qwen2.5-VL-7B-Instruct & 33.42  & 5.49  & 9.91 (0.0) & 0.0 (0.0) & 0.0 (0.0) & 0.36 (0.36) & 0.0  & 0.0 (0.0) \\
 & HealthGPT-L14 & 38.17  & 2.48  & 22.54 (10.88) & 36.39 (18.2) & 0.48 (0.23) & 0.28 (0.28) & 0.0  & 36.39 (18.2) \\
 & RadVLM & 21.56  & 5.27  & 17.39 (17.39) & 18.2 (18.2) & 0.29 (0.29) & 0.22 (0.22) & 36.39  & 18.2 (18.2) \\
\midrule
 & Gemini-2.5-Pro & 92.14  & 60.23  & 20.42 (19.26) & 65.74 (61.95) & 13.41 (12.56) & 1.95 (1.93) & 70.15  & 49.82 (46.42) \\
 & Gemini-2.5-Flash & 86.64  & 51.32  & 32.76 (19.97) & 65.81 (40.97) & 13.69 (8.97) & 1.38 (1.28) & 47.56  & 50.69 (32.85) \\
 & GPT-4.1 & 85.87  & 31.5  & 24.39 (24.39) & 52.07 (52.07) & 8.51 (8.51) & 1.05 (1.05) & 55.86  & 37.48 (37.48) \\
 & Pixtral-Large-Instruct-2411 & 86.35  & 20.49  & 28.15 (20.52) & 52.68 (39.4) & 6.28 (4.66) & 1.11 (1.08) & 44.96  & 46.92 (35.24) \\
 & Llama-3.2-90B-Vision-Instruct & 71.63  & 1.48  & 29.9 (29.9) & 29.73 (29.73) & 0.45 (0.45) & 0.48 (0.48) & 45.05  & 29.73 (29.73) \\
Measurement & Qwen2.5-VL-72B-Instruct & 38.39  & N/A  & N/A (N/A) & N/A (N/A) & 0.0 (0.0) & 0.0 (0.0) & N/A  & N/A (N/A) \\
 & Pixtral 12B & 39.31  & 5.56  & 12.12 (12.12) & 0.0 (0.0) & 0.0 (0.0) & 0.33 (0.33) & 45.05  & 0.0 (0.0) \\
 & Qwen2.5-VL-7B-Instruct & 34.99  & 5.53  & 14.86 (0.0) & 0.0 (0.0) & 0.0 (0.0) & 0.37 (0.36) & 0.0  & 0.0 (0.0) \\
 & HealthGPT-L14 & 38.6  & 2.02  & 23.32 (0.0) & 36.39 (0.0) & 0.37 (0.0) & 0.24 (0.23) & 0.0  & 36.39 (0.0) \\
 & RadVLM & 19.21  & 6.28  & 13.54 (13.54) & 36.39 (36.39) & 0.44 (0.44) & 0.18 (0.18) & 36.39  & 36.39 (36.39) \\
\midrule
 & Gemini-2.5-Pro & 95.9  & 49.62  & 46.93  & 80.98  & 24.0  & 2.05  & N/A  & 79.43  \\
 & Gemini-2.5-Flash & 89.46  & 25.82  & 31.34  & 55.23  & 7.59  & 1.35  & N/A  & 52.91  \\
 & GPT-4.1 & 84.34  & 29.14  & 26.34  & 55.43  & 9.06  & 1.35  & N/A  & 45.99  \\
 & Pixtral-Large-Instruct-2411 & 86.08  & 10.98  & 22.53  & 43.05  & 2.96  & 1.06  & N/A  & 43.05  \\
 & Llama-3.2-90B-Vision-Instruct & 73.53  & 1.48  & 0.0  & N/A  & 0.0  & 0.74  & N/A  & N/A  \\
Recognition & Qwen2.5-VL-72B-Instruct & 59.0  & 13.38  & 49.17  & 51.2  & 5.4  & 0.8  & N/A  & 20.8  \\
 & Pixtral 12B & 41.79  & 5.81  & 22.06  & 0.0  & 0.0  & 0.47  & N/A  & 0.0  \\
 & Qwen2.5-VL-7B-Instruct & 30.28  & 5.44  & 0.0  & N/A  & 0.0  & 0.34  & N/A  & N/A  \\
 & HealthGPT-L14 & 37.32  & 3.29  & 21.76  & 36.39  & 0.68  & 0.38  & N/A  & 36.39  \\
 & RadVLM & 26.26  & 3.26  & 25.07  & 0.0  & 0.0  & 0.29  & N/A  & 0.0  \\
    \bottomrule
  \end{tabular}
  \end{adjustbox}
  }
\end{table}

%% file: Tables/1.Reasoning/Result_Supple_Shot_ALL_Recognition.tex
\begin{table}
  \caption{Path 1 evaluation results with stochastic sampling on recognition-type tasks.
  Stage-wise Score: Percentage of cases that completed each stage;
  Completion: Percentage of cases completing all reasoning stages; 
  Depth: Average number of reasoning stages reached;
  Alignment: Percentage of agreement between initial and final decisions;
  Refined scores incorporating measurement consistency are shown in parentheses.
  `N/A' indicates that the model did not reach the required stage to compute the corresponding metric.
  }
  \label{table:reasoning_shot_all_recognition}
  \centering
  {\begin{adjustbox}{width=\linewidth, height=0.5\textheight, keepaspectratio=true}
  \begin{tabular}{cccccccccc}
    \toprule
    & & \multicolumn{4}{c}{Stage-wise Score} \\
    \cmidrule(r){3-6}

    Task & Models & Stage 1 & Stage 2 & Stage 3 & Stage 4 & Completion & Depth & Consistency & Alignment \\
    
    \midrule
  & Gemini-2.5-Pro & 96.33  & 46.94  & 49.75  & 88.97  & 28.02  & 2.12  & N/A  & 88.97  \\
 & Gemini-2.5-Flash & 96.33  & 31.43  & 31.16  & 77.22  & 12.55  & 1.58  & N/A  & 77.22  \\
 & GPT-4.1 & 77.45  & 38.1  & 28.37  & 75.79  & 11.89  & 1.43  & N/A  & 75.79  \\
 & Pixtral-Large-Instruct-2411 & 90.4  & 21.08  & 13.71  & 45.05  & 3.65  & 1.21  & N/A  & 45.05  \\
 & Llama-3.2-90B-Vision-Instruct & 74.9  & 0.0  & N/A  & N/A  & 0.0  & 0.82  & N/A  & N/A  \\
Trachea Deviation & Qwen2.5-VL-72B-Instruct & 48.48  & 0.0  & N/A  & N/A  & 0.0  & 0.7  & N/A  & N/A  \\
 & Pixtral 12B & 48.71  & 0.0  & N/A  & N/A  & 0.0  & 0.54  & N/A  & N/A  \\
 & Qwen2.5-VL-7B-Instruct & 40.79  & 0.0  & N/A  & N/A  & 0.0  & 0.53  & N/A  & N/A  \\
 & HealthGPT-L14 & 38.28  & 0.0  & N/A  & N/A  & 0.0  & 0.42  & N/A  & N/A  \\
 & RadVLM & 29.33  & 0.0  & N/A  & N/A  & 0.0  & 0.32  & N/A  & N/A  \\
\midrule
 & Gemini-2.5-Pro & 96.33  & 93.12  & 44.26  & 74.55  & 37.42  & 2.87  & N/A  & 68.33  \\
 & Gemini-2.5-Flash & 94.63  & 67.59  & 26.48  & 52.09  & 14.29  & 2.08  & N/A  & 45.12  \\
 & GPT-4.1 & 96.3  & 62.81  & 28.13  & 73.16  & 18.78  & 2.1  & N/A  & 35.39  \\
 & Pixtral-Large-Instruct-2411 & 90.4  & 7.53  & 34.38  & 27.09  & 2.74  & 1.06  & N/A  & 27.09  \\
 & Llama-3.2-90B-Vision-Instruct & 90.03  & 5.9  & 0.0  & N/A  & 0.0  & 1.0  & N/A  & N/A  \\
Inclusion & Qwen2.5-VL-72B-Instruct & 90.59  & 40.13  & 49.17  & 51.2  & 21.62  & 2.0  & N/A  & 20.8  \\
 & Pixtral 12B & 74.31  & 17.42  & 22.06  & 0.0  & 0.0  & 0.99  & N/A  & 0.0  \\
 & Qwen2.5-VL-7B-Instruct & 31.36  & 21.76  & 0.0  & N/A  & 0.0  & 0.44  & N/A  & N/A  \\
 & HealthGPT-L14 & 46.06  & 13.15  & 21.76  & 36.39  & 2.74  & 0.59  & N/A  & 36.39  \\
 & RadVLM & 35.92  & 13.02  & 25.07  & 0.0  & 0.0  & 0.45  & N/A  & 0.0  \\
\midrule
 & Gemini-2.5-Pro & 96.33  & 42.98  & 33.93  & 83.18  & 17.91  & 1.84  & N/A  & 83.18  \\
 & Gemini-2.5-Flash & 93.12  & 0.0  & N/A  & N/A  & 0.0  & 0.88  & N/A  & N/A  \\
 & GPT-4.1 & 96.3  & 3.69  & 29.73  & 36.39  & 2.76  & 1.04  & N/A  & 36.39  \\
 & Pixtral-Large-Instruct-2411 & 86.69  & 6.8  & 0.0  & N/A  & 0.0  & 0.98  & N/A  & N/A  \\
 & Llama-3.2-90B-Vision-Instruct & 68.56  & 0.0  & N/A  & N/A  & 0.0  & 0.69  & N/A  & N/A  \\
Inspiration & Qwen2.5-VL-72B-Instruct & 29.73  & 0.0  & N/A  & N/A  & 0.0  & 0.5  & N/A  & N/A  \\
 & Pixtral 12B & 0.0  & N/A  & N/A  & N/A  & 0.0  & 0.0  & N/A  & N/A  \\
 & Qwen2.5-VL-7B-Instruct & 13.01  & 0.0  & N/A  & N/A  & 0.0  & 0.1  & N/A  & N/A  \\
 & HealthGPT-L14 & 17.6  & 0.0  & N/A  & N/A  & 0.0  & 0.17  & N/A  & N/A  \\
 & RadVLM & 10.69  & 0.0  & N/A  & N/A  & 0.0  & 0.1  & N/A  & N/A  \\
\midrule
 & Gemini-2.5-Pro & 94.63  & 15.43  & 59.79  & 77.22  & 12.66  & 1.38  & N/A  & 77.22  \\
 & Gemini-2.5-Flash & 73.74  & 4.25  & 36.39  & 36.39  & 3.53  & 0.86  & N/A  & 36.39  \\
 & GPT-4.1 & 67.3  & 11.97  & 19.13  & 36.39  & 2.79  & 0.85  & N/A  & 36.39  \\
 & Pixtral-Large-Instruct-2411 & 76.82  & 8.5  & 42.02  & 57.01  & 5.47  & 0.98  & N/A  & 57.01  \\
 & Llama-3.2-90B-Vision-Instruct & 60.62  & 0.0  & N/A  & N/A  & 0.0  & 0.47  & N/A  & N/A  \\
Ascending Aorta Enlargement & Qwen2.5-VL-72B-Instruct & 67.19  & N/A  & N/A  & N/A  & 0.0  & 0.0  & N/A  & N/A  \\
 & Pixtral 12B & 44.13  & 0.0  & N/A  & N/A  & 0.0  & 0.33  & N/A  & N/A  \\
 & Qwen2.5-VL-7B-Instruct & 35.96  & 0.0  & N/A  & N/A  & 0.0  & 0.3  & N/A  & N/A  \\
 & HealthGPT-L14 & 47.35  & 0.0  & N/A  & N/A  & 0.0  & 0.35  & N/A  & N/A  \\
 & RadVLM & 29.09  & 0.0  & N/A  & N/A  & 0.0  & 0.28  & N/A  & N/A  \\
    \bottomrule
  \end{tabular}
  \end{adjustbox}
  }
\end{table}

%% file: Tables/1.Reasoning/Result_Supple_Shot_ALL_Arithmetic.tex
\begin{table}[htb!]
  \caption{Path 1 evaluation results with stochastic sampling on measurement-type tasks.
  Stage-wise Score: Percentage of cases that completed each stage;
  Completion: Percentage of cases completing all reasoning stages; 
  Depth: Average number of reasoning stages reached;
  Consistency: Percentage of cases where the value returned at Stage 4 matches the Stage 3 response;
  Alignment: Percentage of agreement between initial and final decisions;
  Refined scores incorporating measurement consistency are shown in parentheses.
  `N/A' indicates that the model did not reach the required stage to compute the corresponding metric.
  }
  \label{table:reasoning_shot_all_arithmetic}
  \centering
  {\begin{adjustbox}{width=\linewidth, height=0.5\textheight, keepaspectratio=true}
  \begin{tabular}{cccccccccc}
    \toprule
    & & \multicolumn{4}{c}{Stage-wise Score} \\
    \cmidrule(r){3-6}

    Task & Models & Stage 1 & Stage 2 & Stage 3 & Stage 4 & Completion & Depth & Consistency & Alignment \\
    
    \midrule
  & Gemini-2.5-Pro & 96.33  & 78.91  & 21.27 (16.79) & 61.18 (48.3) & 15.16 (11.97) & 2.2 (2.14) & 61.18  & 61.18 (48.3) \\
 & Gemini-2.5-Flash & 79.6  & 55.58  & 23.28 (17.91) & 45.13 (34.72) & 9.2 (7.08) & 1.62 (1.58) & 55.89  & 40.35 (31.04) \\
 & GPT-4.1 & 92.86  & 29.6  & 20.28 (20.28) & 36.37 (36.37) & 6.95 (6.95) & 1.43 (1.43) & 51.73  & 27.09 (27.09) \\
 & Pixtral-Large-Instruct-2411 & 96.33  & 19.47  & 33.4 (25.05) & 58.44 (43.83) & 8.15 (6.11) & 1.35 (1.31) & 49.52  & 58.44 (43.83) \\
 & Llama-3.2-90B-Vision-Instruct & 84.93  & 0.0  & N/A (N/A) & N/A (N/A) & 0.0 (0.0) & 0.93 (0.93) & N/A  & N/A (N/A) \\
Cardiomegaly & Qwen2.5-VL-72B-Instruct & N/A  & N/A  & N/A (N/A) & N/A (N/A) & N/A (N/A) & N/A (N/A) & N/A  & N/A (N/A) \\
 & Pixtral 12B & 51.12  & 0.0  & N/A (N/A) & N/A (N/A) & 0.0 (0.0) & 0.57 (0.57) & N/A  & N/A (N/A) \\
 & Qwen2.5-VL-7B-Instruct & 31.36  & 27.82  & 29.73 (0.0) & 0.0 (0.0) & 0.0 (0.0) & 0.56 (0.5) & 0.0  & 0.0 (0.0) \\
 & HealthGPT-L14 & 44.62  & 0.0  & N/A (N/A) & N/A (N/A) & 0.0 (0.0) & 0.49 (0.49) & N/A  & N/A (N/A) \\
 & RadVLM & 23.31  & 0.0  & N/A (N/A) & N/A (N/A) & 0.0 (0.0) & 0.24 (0.24) & N/A  & N/A (N/A) \\
\midrule
 & Gemini-2.5-Pro & 96.33  & 83.25  & 36.4 (34.38) & 67.8 (64.03) & 27.17 (25.66) & 2.55 (2.51) & 82.67  & 65.2 (61.58) \\
 & Gemini-2.5-Flash & 96.26  & 87.67  & 38.47 (35.51) & 87.25 (80.54) & 35.49 (32.76) & 2.72 (2.66) & 80.69  & 75.0 (69.23) \\
 & GPT-4.1 & 96.11  & 71.24  & 46.74 (46.74) & 83.82 (83.82) & 35.92 (35.92) & 2.6 (2.6) & 91.23  & 69.8 (69.8) \\
 & Pixtral-Large-Instruct-2411 & 96.33  & 60.35  & 27.63 (19.34) & 46.87 (32.81) & 12.55 (8.78) & 1.99 (1.92) & 54.31  & 27.02 (18.91) \\
 & Llama-3.2-90B-Vision-Instruct & 84.26  & 8.9  & 29.9 (29.9) & 29.73 (29.73) & 3.58 (3.58) & 1.03 (1.03) & 45.05  & 29.73 (29.73) \\
Carina Angle & Qwen2.5-VL-72B-Instruct & 0.0  & N/A  & N/A (N/A) & N/A (N/A) & 0.0 (0.0) & 0.0 (0.0) & N/A  & N/A (N/A) \\
 & Pixtral 12B & 48.77  & 8.49  & 36.37 (36.37) & 0.0 (0.0) & 0.0 (0.0) & 0.6 (0.6) & 45.05  & 0.0 (0.0) \\
 & Qwen2.5-VL-7B-Instruct & 46.0  & 0.0  & N/A (N/A) & N/A (N/A) & 0.0 (0.0) & 0.71 (0.71) & N/A  & N/A (N/A) \\
 & HealthGPT-L14 & 39.78  & 14.16  & 23.32 (0.0) & 36.39 (0.0) & 2.99 (0.0) & 0.52 (0.49) & 0.0  & 36.39 (0.0) \\
 & RadVLM & 10.92  & 29.97  & 0.0 (0.0) & N/A (N/A) & 0.0 (0.0) & 0.14 (0.14) & N/A  & N/A (N/A) \\
\midrule
 & Gemini-2.5-Pro & 96.33  & 70.78  & 26.36 (26.36) & 85.69 (85.69) & 21.18 (21.18) & 2.22 (2.22) & 85.69  & 50.11 (50.11) \\
 & Gemini-2.5-Flash & 96.33  & 49.57  & 29.78 (18.2) & 64.54 (39.44) & 15.16 (9.26) & 1.88 (1.76) & 47.01  & 64.54 (39.44) \\
 & GPT-4.1 & 96.26  & 33.74  & 19.56 (19.56) & 67.71 (67.71) & 8.3 (8.3) & 1.51 (1.51) & 67.71  & 55.25 (55.25) \\
 & Pixtral-Large-Instruct-2411 & 85.52  & 20.97  & 17.46 (17.46) & 51.73 (51.73) & 4.56 (4.56) & 1.18 (1.18) & 51.73  & 51.73 (51.73) \\
 & Llama-3.2-90B-Vision-Instruct & 87.06  & 0.0  & N/A (N/A) & N/A (N/A) & 0.0 (0.0) & 0.94 (0.94) & N/A  & N/A (N/A) \\
Rotation & Qwen2.5-VL-72B-Instruct & N/A  & N/A  & N/A (N/A) & N/A (N/A) & N/A (N/A) & N/A (N/A) & N/A  & N/A (N/A) \\
 & Pixtral 12B & 23.29  & 0.0  & N/A (N/A) & N/A (N/A) & 0.0 (0.0) & 0.24 (0.24) & N/A  & N/A (N/A) \\
 & Qwen2.5-VL-7B-Instruct & 25.29  & 0.0  & N/A (N/A) & N/A (N/A) & 0.0 (0.0) & 0.27 (0.27) & N/A  & N/A (N/A) \\
 & HealthGPT-L14 & 37.72  & 0.0  & N/A (N/A) & N/A (N/A) & 0.0 (0.0) & 0.42 (0.42) & N/A  & N/A (N/A) \\
 & RadVLM & 5.03  & 0.0  & N/A (N/A) & N/A (N/A) & 0.0 (0.0) & 0.03 (0.03) & N/A  & N/A (N/A) \\
\midrule
 & Gemini-2.5-Pro & 84.37  & 31.39  & 24.27 (24.27) & 49.52 (49.52) & 7.26 (7.26) & 1.36 (1.36) & 70.19  & 35.76 (35.76) \\
 & Gemini-2.5-Flash & 93.12  & 20.98  & 27.25 (13.63) & 57.01 (28.51) & 8.41 (4.21) & 1.31 (1.24) & 32.5  & 42.24 (21.12) \\
 & GPT-4.1 & 90.04  & 18.68  & 21.14 (21.14) & 51.73 (51.73) & 5.48 (5.48) & 1.22 (1.22) & 51.73  & 36.37 (36.37) \\
 & Pixtral-Large-Instruct-2411 & 79.97  & 9.24  & 26.06 (13.03) & 45.05 (22.52) & 3.65 (1.83) & 0.98 (0.96) & 29.73  & 45.05 (22.52) \\
 & Llama-3.2-90B-Vision-Instruct & 91.6  & 0.0  & N/A (N/A) & N/A (N/A) & 0.0 (0.0) & 0.92 (0.92) & N/A  & N/A (N/A) \\
Mediastinal Widening & Qwen2.5-VL-72B-Instruct & 36.39  & N/A  & N/A (N/A) & N/A (N/A) & N/A (N/A) & N/A (N/A) & N/A  & N/A (N/A) \\
 & Pixtral 12B & 45.97  & 5.34  & 0.0 (0.0) & N/A (N/A) & 0.0 (0.0) & 0.52 (0.52) & N/A  & N/A (N/A) \\
 & Qwen2.5-VL-7B-Instruct & 59.17  & 10.86  & 0.0 (0.0) & N/A (N/A) & 0.0 (0.0) & 0.72 (0.72) & N/A  & N/A (N/A) \\
 & HealthGPT-L14 & 38.47  & 0.0  & N/A (N/A) & N/A (N/A) & 0.0 (0.0) & 0.29 (0.29) & N/A  & N/A (N/A) \\
 & RadVLM & 16.89  & 0.0  & N/A (N/A) & N/A (N/A) & 0.0 (0.0) & 0.13 (0.13) & N/A  & N/A (N/A) \\
\midrule
 & Gemini-2.5-Pro & 96.33  & 36.39  & 0.0 (0.0) & N/A (N/A) & 0.0 (0.0) & 2.0 (2.0) & N/A  & N/A (N/A) \\
 & Gemini-2.5-Flash & 71.77  & N/A  & N/A (N/A) & N/A (N/A) & 0.0 (0.0) & 0.0 (0.0) & N/A  & N/A (N/A) \\
 & GPT-4.1 & 81.98  & 37.35  & 8.9 (8.9) & 36.39 (36.39) & 3.67 (3.67) & 1.23 (1.23) & 36.39  & 36.39 (36.39) \\
 & Pixtral-Large-Instruct-2411 & 71.77  & 0.0  & N/A (N/A) & N/A (N/A) & 0.0 (0.0) & 0.05 (0.05) & N/A  & N/A (N/A) \\
 & Llama-3.2-90B-Vision-Instruct & 28.87  & 0.0  & N/A (N/A) & N/A (N/A) & 0.0 (0.0) & 0.01 (0.01) & N/A  & N/A (N/A) \\
Projection & Qwen2.5-VL-72B-Instruct & 10.12  & N/A  & N/A (N/A) & N/A (N/A) & 0.0 (0.0) & 0.0 (0.0) & N/A  & N/A (N/A) \\
 & Pixtral 12B & 14.05  & 25.07  & 0.0 (0.0) & N/A (N/A) & 0.0 (0.0) & 0.06 (0.06) & N/A  & N/A (N/A) \\
 & Qwen2.5-VL-7B-Instruct & 9.46  & 0.0  & N/A (N/A) & N/A (N/A) & 0.0 (0.0) & 0.03 (0.03) & N/A  & N/A (N/A) \\
 & HealthGPT-L14 & 19.82  & 0.0  & N/A (N/A) & N/A (N/A) & 0.0 (0.0) & 0.05 (0.05) & N/A  & N/A (N/A) \\
 & RadVLM & 2.74  & 0.0  & N/A (N/A) & N/A (N/A) & 0.0 (0.0) & 0.01 (0.01) & N/A  & N/A (N/A) \\
\midrule
 & Gemini-2.5-Pro & 96.33  & 64.55  & 20.98 (19.36) & 78.49 (72.45) & 15.84 (14.62) & 2.04 (2.01) & 69.46  & 50.27 (46.4) \\
 & Gemini-2.5-Flash & 93.05  & 49.72  & 46.52 (21.15) & 85.69 (38.95) & 27.92 (12.69) & 2.12 (1.79) & 37.27  & 43.52 (19.78) \\
 & GPT-4.1 & 75.57  & 0.0  & N/A (N/A) & N/A (N/A) & 0.0 (0.0) & 0.11 (0.11) & N/A  & N/A (N/A) \\
 & Pixtral-Large-Instruct-2411 & 86.69  & 15.53  & 35.76 (26.82) & 57.01 (42.76) & 8.53 (6.4) & 1.15 (1.11) & 42.24  & 57.01 (42.76) \\
 & Llama-3.2-90B-Vision-Instruct & 57.07  & 0.0  & N/A (N/A) & N/A (N/A) & 0.0 (0.0) & 0.04 (0.04) & N/A  & N/A (N/A) \\
Aortic Knob Enlargement & Qwen2.5-VL-72B-Instruct & 80.65  & N/A  & N/A (N/A) & N/A (N/A) & N/A (N/A) & N/A (N/A) & N/A  & N/A (N/A) \\
 & Pixtral 12B & 45.97  & 0.0  & N/A (N/A) & N/A (N/A) & 0.0 (0.0) & 0.16 (0.16) & N/A  & N/A (N/A) \\
 & Qwen2.5-VL-7B-Instruct & 51.74  & 0.0  & N/A (N/A) & N/A (N/A) & 0.0 (0.0) & 0.48 (0.48) & N/A  & N/A (N/A) \\
 & HealthGPT-L14 & 42.74  & 0.0  & N/A (N/A) & N/A (N/A) & 0.0 (0.0) & 0.02 (0.02) & N/A  & N/A (N/A) \\
 & RadVLM & 34.77  & 0.0  & N/A (N/A) & N/A (N/A) & 0.0 (0.0) & 0.34 (0.34) & N/A  & N/A (N/A) \\
\midrule
 & Gemini-2.5-Pro & 74.77  & 56.32  & 13.64 (13.64) & 51.73 (51.73) & 7.25 (7.25) & 1.27 (1.27) & 51.73  & 36.37 (36.37) \\
 & Gemini-2.5-Flash & 75.12  & 44.37  & 31.28 (13.41) & 55.25 (23.68) & 13.37 (5.73) & 1.36 (1.22) & 31.97  & 38.47 (16.49) \\
 & GPT-4.1 & 66.33  & 29.9  & 29.73 (29.73) & 36.39 (36.39) & 7.73 (7.73) & 0.29 (0.29) & 36.39  & 0.0 (0.0) \\
 & Pixtral-Large-Instruct-2411 & 77.86  & 17.86  & 28.57 (21.43) & 57.01 (42.76) & 6.55 (4.92) & 1.06 (1.04) & 42.24  & 42.24 (31.68) \\
 & Llama-3.2-90B-Vision-Instruct & 58.69  & N/A  & N/A (N/A) & N/A (N/A) & 0.0 (0.0) & 0.0 (0.0) & N/A  & N/A (N/A) \\
Descending Aorta Enlargement & Qwen2.5-VL-72B-Instruct & 39.52  & N/A  & N/A (N/A) & N/A (N/A) & 0.0 (0.0) & 0.0 (0.0) & N/A  & N/A (N/A) \\
 & Pixtral 12B & 45.97  & 0.0  & N/A (N/A) & N/A (N/A) & 0.0 (0.0) & 0.13 (0.13) & N/A  & N/A (N/A) \\
 & Qwen2.5-VL-7B-Instruct & 33.15  & 0.0  & N/A (N/A) & N/A (N/A) & 0.0 (0.0) & 0.18 (0.18) & N/A  & N/A (N/A) \\
 & HealthGPT-L14 & 42.7  & 0.0  & N/A (N/A) & N/A (N/A) & 0.0 (0.0) & 0.09 (0.09) & N/A  & N/A (N/A) \\
 & RadVLM & 29.88  & 0.0  & N/A (N/A) & N/A (N/A) & 0.0 (0.0) & 0.28 (0.28) & N/A  & N/A (N/A) \\
\midrule
 & Gemini-2.5-Pro & 96.3  & N/A  & N/A (N/A) & N/A (N/A) & N/A (N/A) & N/A (N/A) & N/A  & N/A (N/A) \\
 & Gemini-2.5-Flash & 87.89  & N/A  & N/A (N/A) & N/A (N/A) & 0.0 (0.0) & 0.0 (0.0) & N/A  & N/A (N/A) \\
 & GPT-4.1 & 87.78  & N/A  & N/A (N/A) & N/A (N/A) & 0.0 (0.0) & 0.0 (0.0) & N/A  & N/A (N/A) \\
 & Pixtral-Large-Instruct-2411 & 96.33  & N/A  & N/A (N/A) & N/A (N/A) & N/A (N/A) & N/A (N/A) & N/A  & N/A (N/A) \\
 & Llama-3.2-90B-Vision-Instruct & 80.56  & N/A  & N/A (N/A) & N/A (N/A) & 0.0 (0.0) & 0.0 (0.0) & N/A  & N/A (N/A) \\
Descending Aorta Tortuous & Qwen2.5-VL-72B-Instruct & 63.66  & N/A  & N/A (N/A) & N/A (N/A) & 0.0 (0.0) & 0.0 (0.0) & N/A  & N/A (N/A) \\
 & Pixtral 12B & N/A  & N/A  & N/A (N/A) & N/A (N/A) & N/A (N/A) & N/A (N/A) & N/A  & N/A (N/A) \\
 & Qwen2.5-VL-7B-Instruct & 23.72  & N/A  & N/A (N/A) & N/A (N/A) & 0.0 (0.0) & 0.0 (0.0) & N/A  & N/A (N/A) \\
 & HealthGPT-L14 & 42.94  & N/A  & N/A (N/A) & N/A (N/A) & 0.0 (0.0) & 0.0 (0.0) & N/A  & N/A (N/A) \\
 & RadVLM & 30.12  & 20.28  & 27.09 (27.09) & 36.39 (36.39) & 3.53 (3.53) & 0.28 (0.28) & 36.39  & 36.39 (36.39) \\
    \bottomrule
  \end{tabular}
  \end{adjustbox}
  }
\end{table}

%% file: Tables/2.Guidance/Result_Supple_Greedy_ALL.tex
\begin{table}[htb!]
  \caption{Path 2 evaluation results with greedy sampling for overall, measurement-type, and recognition-type tasks.
  Stage-wise Score: Percentage of cases that completed each stage;
  Completion: Percentage of cases completing all reasoning stages; 
  Depth: Average number of reasoning stages reached;
  Consistency: Percentage of cases where the value returned at Stage 3 matches the Stage 2 response;
  Refined scores incorporating measurement consistency are shown in parentheses.
  `N/A' indicates that the model did not reach the required stage to compute the corresponding metric.
  }
  \label{table:guidance_greedy_all}
  \centering
  {\begin{adjustbox}{width=\linewidth, height=0.5\textheight, keepaspectratio=true}
  \begin{tabular}{cccccccc}
    \toprule
    & & \multicolumn{3}{c}{Stage-wise Score} \\
    \cmidrule(r){3-5}

    Task & Models & Stage 1 & Stage 2 & Stage 3 & Completion & Depth & Consistency \\
    
    \midrule

& Gemini-2.5-Pro & 69.61  & 57.85 (56.2) & 71.54 (66.1) & 55.81 (54.29) & 2.62 (2.58) & 69.93  \\
 & Gemini-2.5-Flash & 60.54  & 41.42 (35.77) & 66.66 (58.75) & 34.65 (30.11) & 1.75 (1.64) & 57.73  \\
 & GPT-4.1 & 44.89  & 32.37 (31.07) & 73.92 (69.81) & 19.94 (19.15) & 0.99 (0.97) & 68.7  \\
 & Pixtral-Large-Instruct-2411 & 70.88  & 22.17 (18.87) & 64.38 (50.98) & 19.04 (16.16) & 1.21 (1.16) & 51.89  \\
 & Llama-3.2-90B-Vision-Instruct & 15.36  & 37.41 (28.61) & 50.27 (38.64) & 9.55 (8.03) & 0.37 (0.34) & 38.18  \\
Overall & Qwen2.5-VL-72B-Instruct & 40.45  & 47.52 (41.0) & 67.26 (55.0) & 28.45 (25.48) & 1.2 (1.14) & 52.81  \\
 & Pixtral 12B & 28.16  & 10.66 (10.66) & 43.26 (43.26) & 4.22 (4.22) & 0.45 (0.45) & 43.26  \\
 & Qwen2.5-VL-7B-Instruct & 29.09  & 19.15 (6.96) & 23.96 (7.43) & 5.38 (3.6) & 0.46 (0.44) & 11.26  \\
 & MedGemma 27B & 35.15  & 38.81 (10.81) & 62.95 (19.85) & 17.81 (4.1) & 0.84 (0.5) & 18.83  \\
 & HealthGPT-L14 & 18.25  & 11.14 (8.64) & 35.5 (31.33) & 2.89 (2.46) & 0.21 (0.2) & 37.98  \\
 & RadVLM & 8.3  & 0.0 (0.0) & N/A (N/A) & 0.0 (0.0) & 0.05 (0.05) & N/A  \\
  & MedGemma 4B & 13.87  & 16.93 (11.51) & 15.02 (15.02) & 1.59 (1.59) & 0.15 (0.14) & 27.15  \\

\midrule
 & Gemini-2.5-Pro & 75.15  & 61.42 (59.5) & 77.4 (71.05) & 59.05 (57.28) & 2.55 (2.52) & 69.93  \\
 & Gemini-2.5-Flash & 61.28  & 47.86 (40.6) & 67.53 (57.37) & 40.88 (35.04) & 1.99 (1.86) & 57.73  \\
 & GPT-4.1 & 58.01  & 37.0 (35.51) & 73.92 (69.81) & 27.41 (26.33) & 1.3 (1.27) & 68.7  \\
 & Pixtral-Large-Instruct-2411 & 70.88  & 22.17 (18.87) & 64.38 (50.98) & 19.04 (16.16) & 1.21 (1.16) & 51.89  \\
 & Llama-3.2-90B-Vision-Instruct & 12.42  & 33.27 (19.19) & 51.7 (31.35) & 5.63 (3.35) & 0.22 (0.18) & 38.18  \\
Measurment & Qwen2.5-VL-72B-Instruct & 41.83  & 46.15 (38.69) & 64.7 (50.7) & 29.32 (25.93) & 1.26 (1.19) & 52.81  \\
 & Pixtral 12B & 32.28  & 13.33 (13.33) & 43.26 (43.26) & 5.27 (5.27) & 0.54 (0.54) & 43.26  \\
 & Qwen2.5-VL-7B-Instruct & 29.09  & 19.15 (6.96) & 23.96 (7.43) & 5.38 (3.6) & 0.46 (0.44) & 11.26  \\
 & MedGemma 27B & 43.54  & 43.08 (9.48) & 68.26 (16.55) & 23.85 (4.66) & 1.1 (0.62) & 18.83  \\
 & HealthGPT-L14 & 20.43  & 13.37 (10.37) & 35.5 (31.33) & 3.46 (2.95) & 0.25 (0.24) & 37.98  \\
 & RadVLM & 12.44  & 0.0 (0.0) & N/A (N/A) & 0.0 (0.0) & 0.08 (0.08) & N/A  \\
  & MedGemma 4B & 13.51  & 21.16 (14.39) & 15.02 (15.02) & 1.98 (1.98) & 0.15 (0.15) & 27.15  \\

\midrule
 & Gemini-2.5-Pro & 36.39  & 36.39  & 36.39  & 36.39  & 3.0  & N/A  \\
 & Gemini-2.5-Flash & 57.94  & 18.88  & 63.6  & 12.86  & 0.88  & N/A  \\
 & GPT-4.1 & 9.91  & 0.0  & N/A  & 0.0  & 0.17  & N/A  \\
 & Pixtral-Large-Instruct-2411 & N/A  & N/A  & N/A  & N/A  & N/A  & N/A  \\
 & Llama-3.2-90B-Vision-Instruct & 21.23  & 44.31  & 48.37  & 17.39  & 0.66  & N/A  \\
Recognition & Qwen2.5-VL-72B-Instruct & 30.76  & 57.16  & 85.13  & 22.32  & 0.8  & N/A  \\
 & Pixtral 12B & 11.69  & 0.0  & N/A  & 0.0  & 0.08  & N/A  \\
 & Qwen2.5-VL-7B-Instruct & N/A  & N/A  & N/A  & N/A  & N/A  & N/A  \\
 & MedGemma 27B & 14.18  & 17.46  & 36.37  & 2.71  & 0.19  & N/A  \\
 & HealthGPT-L14 & 7.34  & 0.0  & N/A  & 0.0  & 0.03  & N/A  \\
 & RadVLM & 0.0  & N/A  & N/A  & 0.0  & 0.0  & N/A  \\
  & MedGemma 4B & 15.29  & 0.0  & N/A  & 0.0  & 0.11  & N/A  \\

    \bottomrule
  \end{tabular}
  \end{adjustbox}
  }
\end{table}

%% file: Tables/2.Guidance/Result_Supple_Greedy_ALL_Recognition.tex
\begin{table}[htb!]
  \caption{Path 2 evaluation results with greedy sampling on recognition-type tasks.
  Stage-wise Score: Percentage of cases that completed each stage;
  Completion: Percentage of cases completing all reasoning stages; 
  Depth: Average number of reasoning stages reached;
  Refined scores incorporating measurement consistency are shown in parentheses.
  `N/A' indicates that the model did not reach the required stage to compute the corresponding metric.
  }
  \label{table:guidance_greedy_all_recognition}
  \centering
  {\begin{adjustbox}{width=\linewidth, height=0.5\textheight, keepaspectratio=true}
  \begin{tabular}{cccccccc}
    \toprule
    & & \multicolumn{3}{c}{Stage-wise Score} \\
    \cmidrule(r){3-5}

    Task & Models & Stage 1 & Stage 2 & Stage 3 & Completion & Depth & Consistency \\
    
    \midrule

  & Gemini-2.5-Pro & N/A  & N/A  & N/A  & N/A  & N/A  & N/A  \\
 & Gemini-2.5-Flash & N/A  & N/A  & N/A  & N/A  & N/A  & N/A  \\
 & GPT-4.1 & 0.0  & N/A  & N/A  & 0.0  & 0.0  & N/A  \\
 & Pixtral-Large-Instruct-2411 & N/A  & N/A  & N/A  & N/A  & N/A  & N/A  \\
 & Llama-3.2-90B-Vision-Instruct & 37.52  & 36.37  & 45.05  & 29.9  & 1.4  & N/A  \\
Trachea Deviation & Qwen2.5-VL-72B-Instruct & N/A  & N/A  & N/A  & N/A  & N/A  & N/A  \\
 & Pixtral 12B & N/A  & N/A  & N/A  & N/A  & N/A  & N/A  \\
 & Qwen2.5-VL-7B-Instruct & N/A  & N/A  & N/A  & N/A  & N/A  & N/A  \\
 & MedGemma 27B & N/A  & N/A  & N/A  & N/A  & N/A  & N/A  \\
 & HealthGPT-L14 & N/A  & N/A  & N/A  & N/A  & N/A  & N/A  \\
 & RadVLM & N/A  & N/A  & N/A  & N/A  & N/A  & N/A  \\
 & MedGemma 4B & N/A  & N/A  & N/A  & N/A  & N/A  & N/A  \\
\midrule
 & Gemini-2.5-Pro & N/A  & N/A  & N/A  & N/A  & N/A  & N/A  \\
 & Gemini-2.5-Flash & N/A  & N/A  & N/A  & N/A  & N/A  & N/A  \\
 & GPT-4.1 & N/A  & N/A  & N/A  & N/A  & N/A  & N/A  \\
 & Pixtral-Large-Instruct-2411 & N/A  & N/A  & N/A  & N/A  & N/A  & N/A  \\
 & Llama-3.2-90B-Vision-Instruct & N/A  & N/A  & N/A  & N/A  & N/A  & N/A  \\
Inclusion & Qwen2.5-VL-72B-Instruct & N/A  & N/A  & N/A  & N/A  & N/A  & N/A  \\
 & Pixtral 12B & N/A  & N/A  & N/A  & N/A  & N/A  & N/A  \\
 & Qwen2.5-VL-7B-Instruct & N/A  & N/A  & N/A  & N/A  & N/A  & N/A  \\
 & MedGemma 27B & N/A  & N/A  & N/A  & N/A  & N/A  & N/A  \\
 & HealthGPT-L14 & N/A  & N/A  & N/A  & N/A  & N/A  & N/A  \\
 & RadVLM & N/A  & N/A  & N/A  & N/A  & N/A  & N/A  \\
 & MedGemma 4B & N/A  & N/A  & N/A  & N/A  & N/A  & N/A  \\
\midrule
 & Gemini-2.5-Pro & N/A  & N/A  & N/A  & N/A  & N/A  & N/A  \\
 & Gemini-2.5-Flash & 75.69  & 11.69  & 70.19  & 9.94  & 1.01  & N/A  \\
 & GPT-4.1 & 0.0  & N/A  & N/A  & 0.0  & 0.0  & N/A  \\
 & Pixtral-Large-Instruct-2411 & N/A  & N/A  & N/A  & N/A  & N/A  & N/A  \\
 & Llama-3.2-90B-Vision-Instruct & 7.34  & 36.39  & 36.39  & 7.34  & 0.09  & N/A  \\
Inspiration & Qwen2.5-VL-72B-Instruct & N/A  & N/A  & N/A  & N/A  & N/A  & N/A  \\
 & Pixtral 12B & N/A  & N/A  & N/A  & N/A  & N/A  & N/A  \\
 & Qwen2.5-VL-7B-Instruct & N/A  & N/A  & N/A  & N/A  & N/A  & N/A  \\
 & MedGemma 27B & 0.0  & N/A  & N/A  & 0.0  & 0.0  & N/A  \\
 & HealthGPT-L14 & N/A  & N/A  & N/A  & N/A  & N/A  & N/A  \\
 & RadVLM & 0.0  & N/A  & N/A  & 0.0  & 0.0  & N/A  \\
 & MedGemma 4B & N/A  & N/A  & N/A  & N/A  & N/A  & N/A  \\
\midrule
 & Gemini-2.5-Pro & 36.39  & 36.39  & 36.39  & 36.39  & 3.0  & N/A  \\
 & Gemini-2.5-Flash & 40.19  & 26.06  & 57.01  & 15.77  & 0.76  & N/A  \\
 & GPT-4.1 & 29.73  & 0.0  & N/A  & 0.0  & 0.5  & N/A  \\
 & Pixtral-Large-Instruct-2411 & N/A  & N/A  & N/A  & N/A  & N/A  & N/A  \\
 & Llama-3.2-90B-Vision-Instruct & 18.83  & 60.16  & 63.66  & 14.93  & 0.49  & N/A  \\
Ascending Aorta Enlargement & Qwen2.5-VL-72B-Instruct & 30.76  & 57.16  & 85.13  & 22.32  & 0.8  & N/A  \\
 & Pixtral 12B & 11.69  & 0.0  & N/A  & 0.0  & 0.08  & N/A  \\
 & Qwen2.5-VL-7B-Instruct & N/A  & N/A  & N/A  & N/A  & N/A  & N/A  \\
 & MedGemma 27B & 28.35  & 17.46  & 36.37  & 5.42  & 0.38  & N/A  \\
 & HealthGPT-L14 & 7.34  & 0.0  & N/A  & 0.0  & 0.03  & N/A  \\
 & RadVLM & 0.0  & N/A  & N/A  & 0.0  & 0.0  & N/A  \\
  & MedGemma 4B & 15.29  & 0.0  & N/A  & 0.0  & 0.11  & N/A  \\

    \bottomrule
  \end{tabular}
  \end{adjustbox}
  }
\end{table}

%% file: Tables/2.Guidance/Result_Supple_Greedy_ALL_Arithmetic.tex
\begin{table}[htb!]
  \caption{Path 2 evaluation results with greedy sampling on measurement-type tasks.
  Stage-wise Score: Percentage of cases that completed each stage;
  Completion: Percentage of cases completing all reasoning stages; 
  Depth: Average number of reasoning stages reached;
  Consistency: Percentage of cases where the value returned at Stage 3 matches the Stage 2 response;
  Refined scores incorporating measurement consistency are shown in parentheses.
  `N/A' indicates that the model did not reach the required stage to compute the corresponding metric.
  }
  \label{table:guidance_greedy_all_arithmetic}
  \centering
  {\begin{adjustbox}{width=\linewidth, height=0.45\textheight, keepaspectratio=true}
  \begin{tabular}{ccccccccc}
    \toprule
    & & \multicolumn{3}{c}{Stage-wise Score} \\
    \cmidrule(r){3-5}

    Task & Models & Stage 1 & Stage 2 & Stage 3 & Completion & Depth & Consistency \\
    
    \midrule
   & Gemini-2.5-Pro & 57.01  & 57.01 (57.01) & 57.01 (57.01) & 57.01 (57.01) & 3.0 (3.0) & 57.01  \\
 & Gemini-2.5-Flash & 45.05  & 29.73 (29.73) & 36.39 (36.39) & 29.73 (29.73) & 2.0 (2.0) & 36.39  \\
 & GPT-4.1 & 86.75  & 30.64 (30.64) & 81.58 (81.58) & 29.6 (29.6) & 1.61 (1.61) & 81.58  \\
 & Pixtral-Large-Instruct-2411 & N/A  & N/A (N/A) & N/A (N/A) & N/A (N/A) & N/A (N/A) & N/A  \\
 & Llama-3.2-90B-Vision-Instruct & 0.0  & N/A (N/A) & N/A (N/A) & 0.0 (0.0) & 0.0 (0.0) & N/A  \\
Cardiomegaly & Qwen2.5-VL-72B-Instruct & 61.18  & 65.67 (65.67) & 78.49 (78.49) & 52.71 (52.71) & 2.16 (2.16) & 78.49  \\
 & Pixtral 12B & N/A  & N/A (N/A) & N/A (N/A) & N/A (N/A) & N/A (N/A) & N/A  \\
 & Qwen2.5-VL-7B-Instruct & N/A  & N/A (N/A) & N/A (N/A) & N/A (N/A) & N/A (N/A) & N/A  \\
 & MedGemma 27B & N/A  & N/A (N/A) & N/A (N/A) & N/A (N/A) & N/A (N/A) & N/A  \\
 & HealthGPT-L14 & N/A  & N/A (N/A) & N/A (N/A) & N/A (N/A) & N/A (N/A) & N/A  \\
 & RadVLM & N/A  & N/A (N/A) & N/A (N/A) & N/A (N/A) & N/A (N/A) & N/A  \\
  & MedGemma 4B & N/A  & N/A (N/A) & N/A (N/A) & N/A (N/A) & N/A (N/A) & N/A  \\

\midrule
 & Gemini-2.5-Pro & N/A  & N/A (N/A) & N/A (N/A) & N/A (N/A) & N/A (N/A) & N/A  \\
 & Gemini-2.5-Flash & 36.39  & 36.39 (36.39) & 36.39 (36.39) & 36.39 (36.39) & 3.0 (3.0) & 36.39  \\
 & GPT-4.1 & 47.27  & 25.07 (25.07) & 36.39 (36.39) & 23.32 (23.32) & 1.2 (1.2) & 36.39  \\
 & Pixtral-Large-Instruct-2411 & N/A  & N/A (N/A) & N/A (N/A) & N/A (N/A) & N/A (N/A) & N/A  \\
 & Llama-3.2-90B-Vision-Instruct & N/A  & N/A (N/A) & N/A (N/A) & N/A (N/A) & N/A (N/A) & N/A  \\
Carina Angle & Qwen2.5-VL-72B-Instruct & 40.19  & 57.93 (36.86) & 65.82 (41.88) & 30.36 (19.32) & 1.21 (0.97) & 45.24  \\
 & Pixtral 12B & N/A  & N/A (N/A) & N/A (N/A) & N/A (N/A) & N/A (N/A) & N/A  \\
 & Qwen2.5-VL-7B-Instruct & N/A  & N/A (N/A) & N/A (N/A) & N/A (N/A) & N/A (N/A) & N/A  \\
 & MedGemma 27B & N/A  & N/A (N/A) & N/A (N/A) & N/A (N/A) & N/A (N/A) & N/A  \\
 & HealthGPT-L14 & N/A  & N/A (N/A) & N/A (N/A) & N/A (N/A) & N/A (N/A) & N/A  \\
 & RadVLM & 13.28  & 0.0 (0.0) & N/A (N/A) & 0.0 (0.0) & 0.07 (0.07) & N/A  \\
  & MedGemma 4B & N/A  & N/A (N/A) & N/A (N/A) & N/A (N/A) & N/A (N/A) & N/A  \\

\midrule
 & Gemini-2.5-Pro & N/A  & N/A (N/A) & N/A (N/A) & N/A (N/A) & N/A (N/A) & N/A  \\
 & Gemini-2.5-Flash & N/A  & N/A (N/A) & N/A (N/A) & N/A (N/A) & N/A (N/A) & N/A  \\
 & GPT-4.1 & 0.0  & N/A (N/A) & N/A (N/A) & 0.0 (0.0) & 0.0 (0.0) & N/A  \\
 & Pixtral-Large-Instruct-2411 & N/A  & N/A (N/A) & N/A (N/A) & N/A (N/A) & N/A (N/A) & N/A  \\
 & Llama-3.2-90B-Vision-Instruct & N/A  & N/A (N/A) & N/A (N/A) & N/A (N/A) & N/A (N/A) & N/A  \\
Rotation & Qwen2.5-VL-72B-Instruct & N/A  & N/A (N/A) & N/A (N/A) & N/A (N/A) & N/A (N/A) & N/A  \\
 & Pixtral 12B & N/A  & N/A (N/A) & N/A (N/A) & N/A (N/A) & N/A (N/A) & N/A  \\
 & Qwen2.5-VL-7B-Instruct & N/A  & N/A (N/A) & N/A (N/A) & N/A (N/A) & N/A (N/A) & N/A  \\
 & MedGemma 27B & N/A  & N/A (N/A) & N/A (N/A) & N/A (N/A) & N/A (N/A) & N/A  \\
 & HealthGPT-L14 & N/A  & N/A (N/A) & N/A (N/A) & N/A (N/A) & N/A (N/A) & N/A  \\
 & RadVLM & N/A  & N/A (N/A) & N/A (N/A) & N/A (N/A) & N/A (N/A) & N/A  \\
  & MedGemma 4B & N/A  & N/A (N/A) & N/A (N/A) & N/A (N/A) & N/A (N/A) & N/A  \\

\midrule
 & Gemini-2.5-Pro & 57.01  & 57.01 (57.01) & 57.01 (57.01) & 57.01 (57.01) & 3.0 (3.0) & 57.01  \\
 & Gemini-2.5-Flash & 55.92  & 38.3 (38.3) & 77.22 (77.22) & 27.97 (27.97) & 1.28 (1.28) & 77.22  \\
 & GPT-4.1 & 48.54  & 27.25 (27.25) & 57.01 (57.01) & 19.98 (19.98) & 1.0 (1.0) & 57.01  \\
 & Pixtral-Large-Instruct-2411 & N/A  & N/A (N/A) & N/A (N/A) & N/A (N/A) & N/A (N/A) & N/A  \\
 & Llama-3.2-90B-Vision-Instruct & 17.33  & 40.79 (30.59) & 58.44 (43.83) & 9.34 (7.01) & 0.35 (0.3) & 49.52  \\
Mediastinal Widening & Qwen2.5-VL-72B-Instruct & 27.09  & 36.39 (36.39) & 36.39 (36.39) & 27.09 (27.09) & 1.0 (1.0) & 36.39  \\
 & Pixtral 12B & N/A  & N/A (N/A) & N/A (N/A) & N/A (N/A) & N/A (N/A) & N/A  \\
 & Qwen2.5-VL-7B-Instruct & 0.0  & N/A (N/A) & N/A (N/A) & 0.0 (0.0) & 0.0 (0.0) & N/A  \\
 & MedGemma 27B & 35.55  & 28.48 (0.0) & 49.52 (0.0) & 10.97 (0.0) & 0.62 (0.4) & 0.0  \\
 & HealthGPT-L14 & 23.14  & 0.0 (0.0) & N/A (N/A) & 0.0 (0.0) & 0.22 (0.22) & N/A  \\
 & RadVLM & 20.37  & 0.0 (0.0) & N/A (N/A) & 0.0 (0.0) & 0.14 (0.14) & N/A  \\
  & MedGemma 4B & N/A  & N/A (N/A) & N/A (N/A) & N/A (N/A) & N/A (N/A) & N/A  \\

\midrule
 & Gemini-2.5-Pro & 96.19  & 61.82 (58.08) & 94.58 (88.84) & 61.82 (58.08) & 2.38 (2.29) & 85.94  \\
 & Gemini-2.5-Flash & 96.11  & 88.48 (74.57) & 94.0 (79.21) & 87.18 (73.47) & 2.88 (2.59) & 76.6  \\
 & GPT-4.1 & 75.81  & 50.11 (46.53) & 79.63 (73.94) & 46.69 (43.36) & 2.08 (2.0) & 71.0  \\
 & Pixtral-Large-Instruct-2411 & 89.07  & 29.0 (24.54) & 87.54 (74.07) & 27.83 (23.55) & 1.55 (1.46) & 69.18  \\
 & Llama-3.2-90B-Vision-Instruct & 10.74  & 0.0 (0.0) & N/A (N/A) & 0.0 (0.0) & 0.07 (0.07) & N/A  \\
Projection & Qwen2.5-VL-72B-Instruct & 65.82  & 74.17 (74.17) & 63.66 (63.66) & 57.92 (57.92) & 2.64 (2.64) & 74.17  \\
 & Pixtral 12B & 45.05  & 0.0 (0.0) & N/A (N/A) & 0.0 (0.0) & 1.0 (1.0) & N/A  \\
 & Qwen2.5-VL-7B-Instruct & 62.56  & 12.72 (0.0) & 0.0 (0.0) & 0.0 (0.0) & 0.85 (0.8) & 0.0  \\
 & MedGemma 27B & 48.48  & 67.71 (0.0) & 67.71 (0.0) & 48.48 (0.0) & 2.1 (0.7) & 0.0  \\
 & HealthGPT-L14 & 14.43  & 0.0 (0.0) & N/A (N/A) & 0.0 (0.0) & 0.12 (0.12) & N/A  \\
 & RadVLM & N/A  & N/A (N/A) & N/A (N/A) & N/A (N/A) & N/A (N/A) & N/A  \\
  & MedGemma 4B & 16.55  & 29.73 (29.73) & 0.0 (0.0) & 0.0 (0.0) & 0.19 (0.19) & 36.39  \\

\midrule
 & Gemini-2.5-Pro & 81.58  & 81.58 (81.58) & 81.58 (81.58) & 81.58 (81.58) & 3.0 (3.0) & 81.58  \\
 & Gemini-2.5-Flash & 75.9  & 41.17 (41.17) & 72.32 (72.32) & 39.52 (39.52) & 1.94 (1.94) & 72.32  \\
 & GPT-4.1 & 60.81  & 51.07 (51.07) & 89.84 (89.84) & 36.46 (36.46) & 1.49 (1.49) & 89.84  \\
 & Pixtral-Large-Instruct-2411 & 54.62  & 18.73 (18.73) & 64.77 (64.77) & 12.94 (12.94) & 0.87 (0.87) & 64.77  \\
 & Llama-3.2-90B-Vision-Instruct & 13.01  & 20.37 (20.37) & 36.39 (36.39) & 4.45 (4.45) & 0.15 (0.15) & 36.39  \\
Aortic Knob Enlargement & Qwen2.5-VL-72B-Instruct & 36.04  & 31.84 (19.6) & 78.49 (48.3) & 14.6 (8.99) & 0.68 (0.57) & 45.13  \\
 & Pixtral 12B & 14.73  & 20.37 (20.37) & 36.39 (36.39) & 5.06 (5.06) & 0.18 (0.18) & 36.39  \\
 & Qwen2.5-VL-7B-Instruct & 19.57  & 18.03 (0.0) & 29.73 (0.0) & 3.81 (0.0) & 0.24 (0.2) & 0.0  \\
 & MedGemma 27B & 43.38  & 46.53 (24.27) & 80.11 (41.8) & 23.96 (12.5) & 1.0 (0.76) & 42.06  \\
 & HealthGPT-L14 & 22.87  & 19.79 (19.79) & 45.05 (45.05) & 6.83 (6.83) & 0.32 (0.32) & 45.05  \\
 & RadVLM & N/A  & N/A (N/A) & N/A (N/A) & N/A (N/A) & N/A (N/A) & N/A  \\
  & MedGemma 4B & 10.05  & 27.09 (0.0) & 0.0 (0.0) & 0.0 (0.0) & 0.1 (0.07) & 0.0  \\

\midrule
 & Gemini-2.5-Pro & 79.12  & 91.03 (91.03) & 91.03 (91.03) & 79.12 (79.12) & 2.71 (2.71) & 91.03  \\
 & Gemini-2.5-Flash & 43.44  & 52.51 (52.51) & 82.42 (82.42) & 29.69 (29.69) & 1.15 (1.15) & 82.42  \\
 & GPT-4.1 & 77.37  & 50.27 (50.27) & 89.84 (89.84) & 44.14 (44.14) & 1.86 (1.86) & 89.84  \\
 & Pixtral-Large-Instruct-2411 & 49.52  & 27.82 (27.82) & 45.05 (45.05) & 24.51 (24.51) & 1.25 (1.25) & 45.05  \\
 & Llama-3.2-90B-Vision-Instruct & 11.13  & 45.48 (15.16) & 64.77 (21.59) & 8.15 (2.72) & 0.24 (0.15) & 27.82  \\
Descending Aorta Enlargement & Qwen2.5-VL-72B-Instruct & 33.34  & 35.8 (27.54) & 78.49 (60.38) & 15.51 (11.93) & 0.67 (0.6) & 55.89  \\
 & Pixtral 12B & 23.65  & 15.31 (15.31) & 36.39 (36.39) & 5.34 (5.34) & 0.29 (0.29) & 36.39  \\
 & Qwen2.5-VL-7B-Instruct & 17.85  & 18.02 (0.0) & 36.39 (0.0) & 5.06 (0.0) & 0.22 (0.18) & 0.0  \\
 & MedGemma 27B & 40.73  & 42.92 (21.46) & 74.1 (37.05) & 19.82 (9.91) & 0.88 (0.67) & 40.36  \\
 & HealthGPT-L14 & 20.4  & 17.03 (17.03) & 36.39 (36.39) & 5.34 (5.34) & 0.25 (0.25) & 36.39  \\
 & RadVLM & 0.0  & N/A (N/A) & N/A (N/A) & 0.0 (0.0) & 0.0 (0.0) & N/A  \\
  & MedGemma 4B & 11.99  & 0.0 (0.0) & N/A (N/A) & 0.0 (0.0) & 0.09 (0.09) & N/A  \\

\midrule
 & Gemini-2.5-Pro & 79.97  & 20.08 (12.27) & 83.18 (50.84) & 17.75 (10.85) & 1.23 (1.09) & 47.01  \\
 & Gemini-2.5-Flash & 76.15  & 48.43 (11.53) & 73.98 (17.62) & 35.7 (8.5) & 1.68 (1.04) & 22.75  \\
 & GPT-4.1 & 67.54  & 24.57 (17.74) & 83.18 (60.08) & 19.11 (13.81) & 1.14 (1.03) & 55.21  \\
 & Pixtral-Large-Instruct-2411 & 90.31  & 13.14 (4.38) & 60.16 (20.05) & 10.89 (3.63) & 1.18 (1.04) & 28.57  \\
 & Llama-3.2-90B-Vision-Instruct & 22.32  & 59.71 (29.85) & 47.2 (23.6) & 11.87 (5.94) & 0.52 (0.39) & 39.0  \\
Descending Aorta Tortuous & Qwen2.5-VL-72B-Instruct & 29.17  & 21.23 (10.61) & 51.57 (25.78) & 7.08 (3.54) & 0.44 (0.38) & 34.38  \\
 & Pixtral 12B & 45.69  & 17.63 (17.63) & 57.01 (57.01) & 10.68 (10.68) & 0.7 (0.7) & 57.01  \\
 & Qwen2.5-VL-7B-Instruct & 45.48  & 27.82 (27.82) & 29.73 (29.73) & 18.02 (18.02) & 1.0 (1.0) & 45.05  \\
 & MedGemma 27B & 49.57  & 29.78 (1.65) & 69.87 (3.88) & 16.02 (0.89) & 0.89 (0.57) & 11.72  \\
 & HealthGPT-L14 & 21.3  & 30.03 (15.01) & 25.07 (12.54) & 5.15 (2.57) & 0.32 (0.28) & 32.5  \\
 & RadVLM & 16.13  & 0.0 (0.0) & N/A (N/A) & 0.0 (0.0) & 0.09 (0.09) & N/A  \\
  & MedGemma 4B & 15.45  & 27.82 (27.82) & 45.05 (45.05) & 7.93 (7.93) & 0.24 (0.24) & 45.05  \\

    \bottomrule
  \end{tabular}
  \end{adjustbox}
  }
\end{table}

%% file: Tables/2.Guidance/Result_Supple_Shot_ALL.tex
\begin{table}[htb!]
  \caption{Path 2 evaluation results with stochastic sampling for overall, measurement-type, and recognition-type tasks.
  Stage-wise Score: Percentage of cases that completed each stage;
  Completion: Percentage of cases completing all reasoning stages; 
  Depth: Average number of reasoning stages reached;
  Consistency: Percentage of cases where the value returned at Stage 3 matches the Stage 2 response;
  Refined scores incorporating measurement consistency are shown in parentheses.
  `N/A' indicates that the model did not reach the required stage to compute the corresponding metric.
  }
  \label{table:guidance_shot_all}
  \centering
  {\begin{adjustbox}{width=\linewidth, height=0.5\textheight, keepaspectratio=true}
  \begin{tabular}{cccccccc}
    \toprule
    & & \multicolumn{3}{c}{Stage-wise Score} \\
    \cmidrule(r){3-5}

    Task & Models & Stage 1 & Stage 2 & Stage 3 & Completion & Depth & Consistency \\
    
    \midrule
  & Gemini-2.5-Pro & 64.78  & 59.31 (56.05) & 77.56 (67.03) & 49.12 (46.56) & 2.12 (2.07) & 73.44  \\
 & Gemini-2.5-Flash & 59.98  & 38.9 (33.77) & 74.08 (63.03) & 32.44 (28.08) & 1.61 (1.52) & 59.73  \\
 & GPT-4.1 & 47.04  & 31.69 (31.03) & 71.23 (68.04) & 22.14 (21.7) & 1.19 (1.18) & 71.93  \\
 & Pixtral-Large-Instruct-2411 & 74.79  & 30.37 (27.22) & 71.86 (58.78) & 26.2 (23.53) & 1.44 (1.38) & 60.53  \\
 & Llama-3.2-90B-Vision-Instruct & 9.91  & 15.15 (3.47) & 36.66 (9.01) & 2.58 (0.7) & 0.12 (0.09) & 11.89  \\
Overall & Qwen2.5-VL-72B-Instruct & 38.91  & 26.6 (18.37) & 52.93 (37.63) & 10.7 (6.41) & 0.64 (0.55) & 37.52  \\
 & Pixtral 12B & 25.78  & 15.46 (8.18) & 36.39 (24.26) & 7.92 (2.5) & 0.46 (0.33) & 24.26  \\
 & Qwen2.5-VL-7B-Instruct & 27.87  & 15.07 (12.48) & 42.63 (32.07) & 4.56 (3.32) & 0.37 (0.35) & 34.06  \\
 & HealthGPT-L14 & 5.47  & 0.0 (0.0) & N/A (N/A) & 0.0 (0.0) & 0.04 (0.04) & N/A  \\
 & RadVLM & 5.87  & 25.63 (7.43) & 36.39 (12.13) & 3.43 (1.02) & 0.08 (0.05) & 12.13  \\
\midrule
 & Gemini-2.5-Pro & 88.07  & 65.04 (60.96) & 87.85 (74.69) & 64.58 (60.75) & 2.43 (2.35) & 73.44  \\
 & Gemini-2.5-Flash & 60.21  & 42.54 (35.93) & 75.29 (60.56) & 36.3 (30.7) & 1.77 (1.65) & 59.73  \\
 & GPT-4.1 & 58.58  & 40.07 (39.13) & 76.21 (72.56) & 29.5 (28.83) & 1.41 (1.39) & 71.93  \\
 & Pixtral-Large-Instruct-2411 & 74.79  & 30.37 (27.22) & 71.86 (58.78) & 26.2 (23.53) & 1.44 (1.38) & 60.53  \\
 & Llama-3.2-90B-Vision-Instruct & 14.08  & 17.04 (3.9) & 36.66 (9.01) & 3.87 (1.05) & 0.17 (0.13) & 11.89  \\
Measurement & Qwen2.5-VL-72B-Instruct & 42.4  & 25.35 (13.0) & 53.17 (30.21) & 11.84 (5.41) & 0.71 (0.57) & 37.52  \\
 & Pixtral 12B & 28.91  & 19.32 (10.22) & 36.39 (24.26) & 9.9 (3.12) & 0.56 (0.39) & 24.26  \\
 & Qwen2.5-VL-7B-Instruct & 27.88  & 16.02 (12.13) & 41.59 (28.01) & 4.73 (2.87) & 0.37 (0.34) & 34.06  \\
 & HealthGPT-L14 & 6.84  & 0.0 (0.0) & N/A (N/A) & 0.0 (0.0) & 0.05 (0.05) & N/A  \\
 & RadVLM & 5.89  & 34.17 (9.91) & 36.39 (12.13) & 5.4 (1.61) & 0.09 (0.05) & 12.13  \\
\midrule
 & Gemini-2.5-Pro & 18.2  & 36.39  & 36.39  & 18.2  & 1.5  & N/A  \\
 & Gemini-2.5-Flash & 59.17  & 26.18  & 70.45  & 18.92  & 1.05  & N/A  \\
 & GPT-4.1 & 23.96  & 12.13  & 36.39  & 7.43  & 0.75  & N/A  \\
 & Pixtral-Large-Instruct-2411 & N/A  & N/A  & N/A  & N/A  & N/A  & N/A  \\
 & Llama-3.2-90B-Vision-Instruct & 1.57  & 0.0  & N/A  & 0.0  & 0.01  & N/A  \\
Recognition & Qwen2.5-VL-72B-Instruct & 31.91  & 29.1  & 52.45  & 8.41  & 0.49  & N/A  \\
 & Pixtral 12B & 13.25  & 0.0  & N/A  & 0.0  & 0.09  & N/A  \\
 & Qwen2.5-VL-7B-Instruct & 27.84  & 13.16  & 46.26  & 4.22  & 0.37  & N/A  \\
 & HealthGPT-L14 & 0.0  & N/A  & N/A  & 0.0  & 0.0  & N/A  \\
 & RadVLM & 5.83  & 0.0  & N/A  & 0.0  & 0.05  & N/A  \\

    \bottomrule
  \end{tabular}
  \end{adjustbox}
  }
\end{table}

%% file: Tables/2.Guidance/Result_Supple_Shot_ALL_Recognition.tex
\begin{table}[htb!]
  \caption{Path 2 evaluation results with stochastic sampling on recognition-type tasks.
  Stage-wise Score: Percentage of cases that completed each stage;
  Completion: Percentage of cases completing all reasoning stages; 
  Depth: Average number of reasoning stages reached;
  Refined scores incorporating measurement consistency are shown in parentheses.
  `N/A' indicates that the model did not reach the required stage to compute the corresponding metric.
  }
  \label{table:guidance_shot_all_recognition}
  \centering
  {\begin{adjustbox}{width=\linewidth, height=0.5\textheight, keepaspectratio=true}
  \begin{tabular}{cccccccc}
    \toprule
    & & \multicolumn{3}{c}{Stage-wise Score} \\
    \cmidrule(r){3-5}
    Task & Models & Stage 1 & Stage 2 & Stage 3 & Completion & Depth & Consistency \\
    
    \midrule
 & Gemini-2.5-Pro & N/A  & N/A  & N/A  & N/A  & N/A  & N/A  \\
 & Gemini-2.5-Flash & N/A  & N/A  & N/A  & N/A  & N/A  & N/A  \\
 & GPT-4.1 & 29.73  & 36.39  & 36.39  & 29.73  & 1.5  & N/A  \\
 & Pixtral-Large-Instruct-2411 & N/A  & N/A  & N/A  & N/A  & N/A  & N/A  \\
 & Llama-3.2-90B-Vision-Instruct & 0.0  & N/A  & N/A  & 0.0  & 0.0  & N/A  \\
Trachea Deviation & Qwen2.5-VL-72B-Instruct & 14.75  & 35.83  & 64.77  & 7.99  & 0.28  & N/A  \\
 & Pixtral 12B & N/A  & N/A  & N/A  & N/A  & N/A  & N/A  \\
 & Qwen2.5-VL-7B-Instruct & 3.19  & 0.0  & N/A  & 0.0  & 0.01  & N/A  \\
 & HealthGPT-L14 & N/A  & N/A  & N/A  & N/A  & N/A  & N/A  \\
 & RadVLM & 23.32  & 0.0  & N/A  & 0.0  & 0.2  & N/A  \\
\midrule
 & Gemini-2.5-Pro & N/A  & N/A  & N/A  & N/A  & N/A  & N/A  \\
 & Gemini-2.5-Flash & N/A  & N/A  & N/A  & N/A  & N/A  & N/A  \\
 & GPT-4.1 & 36.39  & 0.0  & N/A  & 0.0  & 1.0  & N/A  \\
 & Pixtral-Large-Instruct-2411 & N/A  & N/A  & N/A  & N/A  & N/A  & N/A  \\
 & Llama-3.2-90B-Vision-Instruct & 0.0  & N/A  & N/A  & 0.0  & 0.0  & N/A  \\
Inclusion & Qwen2.5-VL-72B-Instruct & 91.84  & 23.99  & 51.63  & 17.3  & 1.41  & N/A  \\
 & Pixtral 12B & N/A  & N/A  & N/A  & N/A  & N/A  & N/A  \\
 & Qwen2.5-VL-7B-Instruct & 90.26  & 18.27  & 40.79  & 10.57  & 1.24  & N/A  \\
 & HealthGPT-L14 & N/A  & N/A  & N/A  & N/A  & N/A  & N/A  \\
 & RadVLM & 0.0  & N/A  & N/A  & 0.0  & 0.0  & N/A  \\
\midrule
 & Gemini-2.5-Pro & 36.39  & 36.39  & 36.39  & 36.39  & 3.0  & N/A  \\
 & Gemini-2.5-Flash & 77.62  & 19.33  & 79.63  & 16.86  & 1.19  & N/A  \\
 & GPT-4.1 & 0.0  & N/A  & N/A  & 0.0  & 0.0  & N/A  \\
 & Pixtral-Large-Instruct-2411 & N/A  & N/A  & N/A  & N/A  & N/A  & N/A  \\
 & Llama-3.2-90B-Vision-Instruct & 0.0  & N/A  & N/A  & 0.0  & 0.0  & N/A  \\
Inspiration & Qwen2.5-VL-72B-Instruct & 13.66  & 14.57  & 36.39  & 2.79  & 0.15  & N/A  \\
 & Pixtral 12B & N/A  & N/A  & N/A  & N/A  & N/A  & N/A  \\
 & Qwen2.5-VL-7B-Instruct & 7.93  & 0.0  & N/A  & 0.0  & 0.05  & N/A  \\
 & HealthGPT-L14 & 0.0  & N/A  & N/A  & 0.0  & 0.0  & N/A  \\
 & RadVLM & 0.0  & N/A  & N/A  & 0.0  & 0.0  & N/A  \\
\midrule
 & Gemini-2.5-Pro & 0.0  & N/A  & N/A  & 0.0  & 0.0  & N/A  \\
 & Gemini-2.5-Flash & 40.71  & 33.04  & 61.27  & 20.98  & 0.92  & N/A  \\
 & GPT-4.1 & 29.73  & 0.0  & N/A  & 0.0  & 0.5  & N/A  \\
 & Pixtral-Large-Instruct-2411 & N/A  & N/A  & N/A  & N/A  & N/A  & N/A  \\
 & Llama-3.2-90B-Vision-Instruct & 6.29  & 0.0  & N/A  & 0.0  & 0.04  & N/A  \\
Ascending Aorta Enlargement & Qwen2.5-VL-72B-Instruct & 7.39  & 42.02  & 57.01  & 5.57  & 0.14  & N/A  \\
 & Pixtral 12B & 13.25  & 0.0  & N/A  & 0.0  & 0.09  & N/A  \\
 & Qwen2.5-VL-7B-Instruct & 10.0  & 34.38  & 51.73  & 6.32  & 0.17  & N/A  \\
 & HealthGPT-L14 & 0.0  & N/A  & N/A  & 0.0  & 0.0  & N/A  \\
 & RadVLM & 0.0  & N/A  & N/A  & 0.0  & 0.0  & N/A  \\
    \bottomrule
  \end{tabular}
  \end{adjustbox}
  }
\end{table}

%% file: Tables/2.Guidance/Result_Supple_Shot_ALL_Arithmetic.tex
\begin{table}[htb!]
  \caption{Path 2 evaluation results with stochastic sampling on measurement-type tasks.
  Stage-wise Score: Percentage of cases that completed each stage;
  Completion: Percentage of cases completing all reasoning stages; 
  Depth: Average number of reasoning stages reached;
  Consistency: Percentage of cases where the value returned at Stage 3 matches the Stage 2 response;
  Refined scores incorporating measurement consistency are shown in parentheses.
  `N/A' indicates that the model did not reach the required stage to compute the corresponding metric.
  }
  \label{table:guidance_shot_all_arithmetic}
  \centering
  {\begin{adjustbox}{width=\linewidth, height=0.5\textheight, keepaspectratio=true}
  \begin{tabular}{cccccccc}
    \toprule
    & & \multicolumn{3}{c}{Stage-wise Score} \\
    \cmidrule(r){3-5}

    Task & Models & Stage 1 & Stage 2 & Stage 3 & Completion & Depth & Consistency \\
    
    \midrule
  & Gemini-2.5-Pro & N/A  & N/A (N/A) & N/A (N/A) & N/A (N/A) & N/A (N/A) & N/A  \\
 & Gemini-2.5-Flash & 45.05  & 0.0 (0.0) & N/A (N/A) & 0.0 (0.0) & 1.0 (1.0) & N/A  \\
 & GPT-4.1 & 87.22  & 31.19 (31.19) & 82.42 (82.42) & 30.18 (30.18) & 1.63 (1.63) & 82.42  \\
 & Pixtral-Large-Instruct-2411 & N/A  & N/A (N/A) & N/A (N/A) & N/A (N/A) & N/A (N/A) & N/A  \\
 & Llama-3.2-90B-Vision-Instruct & 15.36  & 32.5 (0.0) & 29.73 (0.0) & 7.94 (0.0) & 0.23 (0.13) & 0.0  \\
Cardiomegaly & Qwen2.5-VL-72B-Instruct & 73.76  & 24.48 (13.99) & 59.71 (34.12) & 16.02 (9.16) & 1.18 (1.04) & 45.12  \\
 & Pixtral 12B & N/A  & N/A (N/A) & N/A (N/A) & N/A (N/A) & N/A (N/A) & N/A  \\
 & Qwen2.5-VL-7B-Instruct & 56.2  & 17.29 (9.6) & 52.65 (29.25) & 9.54 (5.3) & 0.82 (0.75) & 39.22  \\
 & HealthGPT-L14 & 0.0  & N/A (N/A) & N/A (N/A) & 0.0 (0.0) & 0.0 (0.0) & N/A  \\
 & RadVLM & 0.0  & N/A (N/A) & N/A (N/A) & 0.0 (0.0) & 0.0 (0.0) & N/A  \\
\midrule
 & Gemini-2.5-Pro & N/A  & N/A (N/A) & N/A (N/A) & N/A (N/A) & N/A (N/A) & N/A  \\
 & Gemini-2.5-Flash & 45.05  & 45.05 (45.05) & 45.05 (45.05) & 45.05 (45.05) & 3.0 (3.0) & 45.05  \\
 & GPT-4.1 & 51.57  & 47.27 (47.27) & 57.01 (57.01) & 42.02 (42.02) & 2.17 (2.17) & 57.01  \\
 & Pixtral-Large-Instruct-2411 & N/A  & N/A (N/A) & N/A (N/A) & N/A (N/A) & N/A (N/A) & N/A  \\
 & Llama-3.2-90B-Vision-Instruct & 20.36  & 29.9 (14.95) & 45.05 (22.52) & 12.05 (6.02) & 0.36 (0.28) & 29.73  \\
Carina Angle & Qwen2.5-VL-72B-Instruct & 10.1  & 28.26 (18.84) & 36.37 (24.24) & 3.72 (2.48) & 0.14 (0.13) & 36.37  \\
 & Pixtral 12B & 27.09  & 36.39 (0.0) & 36.39 (0.0) & 27.09 (0.0) & 1.0 (0.33) & 0.0  \\
 & Qwen2.5-VL-7B-Instruct & 10.59  & 18.02 (18.02) & 36.39 (36.39) & 2.93 (2.93) & 0.12 (0.12) & 36.39  \\
 & HealthGPT-L14 & 0.0  & N/A (N/A) & N/A (N/A) & 0.0 (0.0) & 0.0 (0.0) & N/A  \\
 & RadVLM & 14.72  & 29.73 (29.73) & 36.39 (36.39) & 11.27 (11.27) & 0.21 (0.21) & 36.39  \\
\midrule
 & Gemini-2.5-Pro & N/A  & N/A (N/A) & N/A (N/A) & N/A (N/A) & N/A (N/A) & N/A  \\
 & Gemini-2.5-Flash & N/A  & N/A (N/A) & N/A (N/A) & N/A (N/A) & N/A (N/A) & N/A  \\
 & GPT-4.1 & 0.0  & N/A (N/A) & N/A (N/A) & 0.0 (0.0) & 0.0 (0.0) & N/A  \\
 & Pixtral-Large-Instruct-2411 & N/A  & N/A (N/A) & N/A (N/A) & N/A (N/A) & N/A (N/A) & N/A  \\
 & Llama-3.2-90B-Vision-Instruct & 20.37  & 0.0 (0.0) & N/A (N/A) & 0.0 (0.0) & 0.14 (0.14) & N/A  \\
Rotation & Qwen2.5-VL-72B-Instruct & 55.8  & 16.46 (11.52) & 74.17 (51.92) & 10.8 (7.56) & 0.82 (0.76) & 48.48  \\
 & Pixtral 12B & N/A  & N/A (N/A) & N/A (N/A) & N/A (N/A) & N/A (N/A) & N/A  \\
 & Qwen2.5-VL-7B-Instruct & 37.03  & 9.93 (4.97) & 45.05 (22.52) & 4.6 (2.3) & 0.46 (0.44) & 29.73  \\
 & HealthGPT-L14 & 0.0  & N/A (N/A) & N/A (N/A) & 0.0 (0.0) & 0.0 (0.0) & N/A  \\
 & RadVLM & 17.03  & 36.39 (0.0) & 36.39 (0.0) & 17.03 (0.0) & 0.3 (0.1) & 0.0  \\
\midrule
 & Gemini-2.5-Pro & N/A  & N/A (N/A) & N/A (N/A) & N/A (N/A) & N/A (N/A) & N/A  \\
 & Gemini-2.5-Flash & 52.86  & 46.69 (43.36) & 79.63 (73.94) & 32.47 (30.15) & 1.37 (1.32) & 71.0  \\
 & GPT-4.1 & 47.01  & 25.39 (25.39) & 51.73 (51.73) & 18.77 (18.77) & 0.94 (0.94) & 51.73  \\
 & Pixtral-Large-Instruct-2411 & N/A  & N/A (N/A) & N/A (N/A) & N/A (N/A) & N/A (N/A) & N/A  \\
 & Llama-3.2-90B-Vision-Instruct & 7.23  & 23.32 (0.0) & 36.39 (0.0) & 3.12 (0.0) & 0.08 (0.06) & 0.0  \\
Mediastinal Widening & Qwen2.5-VL-72B-Instruct & 46.06  & 24.03 (12.94) & 50.27 (27.07) & 9.92 (5.34) & 0.73 (0.64) & 40.35  \\
 & Pixtral 12B & N/A  & N/A (N/A) & N/A (N/A) & N/A (N/A) & N/A (N/A) & N/A  \\
 & Qwen2.5-VL-7B-Instruct & 28.6  & 19.34 (9.67) & 42.24 (21.12) & 6.24 (3.12) & 0.41 (0.35) & 32.5  \\
 & HealthGPT-L14 & 12.05  & 0.0 (0.0) & N/A (N/A) & 0.0 (0.0) & 0.08 (0.08) & N/A  \\
 & RadVLM & 0.0  & N/A (N/A) & N/A (N/A) & 0.0 (0.0) & 0.0 (0.0) & N/A  \\
\midrule
 & Gemini-2.5-Pro & 96.3  & 67.59 (61.19) & 95.13 (86.13) & 67.59 (61.19) & 2.49 (2.35) & 82.52  \\
 & Gemini-2.5-Flash & 95.42  & 89.7 (71.99) & 93.04 (74.68) & 88.06 (70.68) & 2.91 (2.53) & 71.83  \\
 & GPT-4.1 & 77.98  & 45.19 (45.19) & 79.63 (79.63) & 42.52 (42.52) & 1.96 (1.96) & 79.63  \\
 & Pixtral-Large-Instruct-2411 & 89.57  & 40.55 (36.97) & 90.11 (82.16) & 39.17 (35.72) & 1.83 (1.76) & 78.33  \\
 & Llama-3.2-90B-Vision-Instruct & 12.93  & 0.0 (0.0) & N/A (N/A) & 0.0 (0.0) & 0.1 (0.1) & N/A  \\
Projection & Qwen2.5-VL-72B-Instruct & 80.68  & 54.07 (13.83) & 74.49 (19.05) & 41.93 (10.73) & 1.9 (1.15) & 24.09  \\
 & Pixtral 12B & 52.65  & 0.0 (0.0) & N/A (N/A) & 0.0 (0.0) & 0.78 (0.78) & N/A  \\
 & Qwen2.5-VL-7B-Instruct & 67.21  & 13.15 (4.38) & 42.02 (14.01) & 7.86 (2.62) & 0.91 (0.81) & 27.82  \\
 & HealthGPT-L14 & 11.72  & 0.0 (0.0) & N/A (N/A) & 0.0 (0.0) & 0.06 (0.06) & N/A  \\
 & RadVLM & N/A  & N/A (N/A) & N/A (N/A) & N/A (N/A) & N/A (N/A) & N/A  \\
\midrule
 & Gemini-2.5-Pro & 82.42  & 82.42 (82.42) & 82.42 (82.42) & 82.42 (82.42) & 3.0 (3.0) & 82.42  \\
 & Gemini-2.5-Flash & 77.98  & 51.19 (47.99) & 81.58 (76.48) & 48.01 (45.01) & 2.11 (2.04) & 73.67  \\
 & GPT-4.1 & 63.81  & 58.4 (58.4) & 91.23 (91.23) & 43.1 (43.1) & 1.68 (1.68) & 91.23  \\
 & Pixtral-Large-Instruct-2411 & 55.92  & 32.95 (32.95) & 74.17 (74.17) & 24.14 (24.14) & 1.18 (1.18) & 74.17  \\
 & Llama-3.2-90B-Vision-Instruct & 8.57  & 0.0 (0.0) & N/A (N/A) & 0.0 (0.0) & 0.07 (0.07) & N/A  \\
Aortic Knob Enlargement & Qwen2.5-VL-72B-Instruct & 12.55  & 24.18 (8.06) & 51.73 (17.24) & 4.56 (1.52) & 0.18 (0.14) & 27.09  \\
 & Pixtral 12B & 16.41  & 21.76 (21.76) & 36.39 (36.39) & 6.38 (6.38) & 0.21 (0.21) & 36.39  \\
 & Qwen2.5-VL-7B-Instruct & 7.88  & 23.32 (23.32) & 36.39 (36.39) & 3.41 (3.41) & 0.09 (0.09) & 36.39  \\
 & HealthGPT-L14 & 7.82  & 0.0 (0.0) & N/A (N/A) & 0.0 (0.0) & 0.05 (0.05) & N/A  \\
 & RadVLM & 0.0  & N/A (N/A) & N/A (N/A) & 0.0 (0.0) & 0.0 (0.0) & N/A  \\
\midrule
 & Gemini-2.5-Pro & 91.43  & 91.43 (91.43) & 91.43 (91.43) & 91.43 (91.43) & 3.0 (3.0) & 91.43  \\
 & Gemini-2.5-Flash & 36.95  & 35.39 (35.39) & 72.32 (72.32) & 18.15 (18.15) & 0.78 (0.78) & 72.32  \\
 & GPT-4.1 & 70.82  & 51.91 (51.91) & 89.84 (89.84) & 42.44 (42.44) & 1.75 (1.75) & 89.84  \\
 & Pixtral-Large-Instruct-2411 & 64.54  & 36.68 (36.68) & 67.71 (67.71) & 32.45 (32.45) & 1.61 (1.61) & 67.71  \\
 & Llama-3.2-90B-Vision-Instruct & 7.19  & 32.5 (16.25) & 45.05 (22.52) & 4.82 (2.41) & 0.11 (0.08) & 29.73  \\
Descending Aorta Enlargement & Qwen2.5-VL-72B-Instruct & 15.55  & 26.06 (19.55) & 42.24 (31.68) & 4.98 (3.73) & 0.23 (0.22) & 42.24  \\
 & Pixtral 12B & 19.49  & 19.13 (19.13) & 36.39 (36.39) & 6.12 (6.12) & 0.24 (0.24) & 36.39  \\
 & Qwen2.5-VL-7B-Instruct & 5.42  & 27.09 (27.09) & 36.39 (36.39) & 3.26 (3.26) & 0.06 (0.06) & 36.39  \\
 & HealthGPT-L14 & 9.12  & 0.0 (0.0) & N/A (N/A) & 0.0 (0.0) & 0.07 (0.07) & N/A  \\
 & RadVLM & 0.0  & N/A (N/A) & N/A (N/A) & 0.0 (0.0) & 0.0 (0.0) & N/A  \\
\midrule
 & Gemini-2.5-Pro & 82.14  & 18.73 (8.81) & 82.42 (38.78) & 16.89 (7.95) & 1.23 (1.05) & 37.4  \\
 & Gemini-2.5-Flash & 68.14  & 29.75 (7.76) & 80.11 (20.9) & 22.38 (5.84) & 1.23 (0.88) & 24.5  \\
 & GPT-4.1 & 70.2  & 21.14 (14.53) & 81.58 (56.08) & 16.93 (11.64) & 1.12 (1.01) & 51.63  \\
 & Pixtral-Large-Instruct-2411 & 89.13  & 11.31 (2.26) & 55.43 (11.09) & 9.03 (1.81) & 1.13 (0.99) & 21.91  \\
 & Llama-3.2-90B-Vision-Instruct & 20.64  & 18.09 (0.0) & 27.09 (0.0) & 3.05 (0.0) & 0.26 (0.21) & 0.0  \\
Descending Aorta Tortuous & Qwen2.5-VL-72B-Instruct & 44.73  & 5.24 (5.24) & 36.39 (36.39) & 2.76 (2.76) & 0.52 (0.52) & 36.39  \\
 & Pixtral 12B & N/A  & N/A (N/A) & N/A (N/A) & N/A (N/A) & N/A (N/A) & N/A  \\
 & Qwen2.5-VL-7B-Instruct & 10.08  & 0.0 (0.0) & N/A (N/A) & 0.0 (0.0) & 0.09 (0.09) & N/A  \\
 & HealthGPT-L14 & 14.03  & 0.0 (0.0) & N/A (N/A) & 0.0 (0.0) & 0.13 (0.13) & N/A  \\
 & RadVLM & 9.47  & 36.39 (0.0) & 36.39 (0.0) & 9.47 (0.0) & 0.12 (0.04) & 0.0  \\
    \bottomrule
  \end{tabular}
  \end{adjustbox}
  }
\end{table}

%% file: Tables/3.Review/Result_Supple_Greedy_ALL.tex
\begin{table}[htb!]
  \caption{Re-evaluated Path 1 results with greedy sampling for overall, measurement-type, and recognition-type tasks.
  Stage-wise Score: Percentage of cases that completed each stage;
  Completion: Percentage of cases completing all reasoning stages; 
  Depth: Average number of reasoning stages reached;
  Consistency: Percentage of cases where the value returned at Stage 4 matches the Stage 3 response;
  Refined scores incorporating measurement consistency are shown in parentheses.
  `N/A' indicates that the model did not reach the required stage to compute the corresponding metric.
  }
  \label{table:review_greedy_all}
  \centering
  {\begin{adjustbox}{width=\linewidth}
  \begin{tabular}{cccccccccc}
    \toprule
    & & \multicolumn{4}{c}{Stage-wise Score} \\
    \cmidrule(r){3-6}

    Task & Models & Stage 1 & Stage 2 & Stage 3 & Stage 4 & Completion & Depth & Consistency & Alignment \\
    \midrule

 & Gemini-2.5-Pro & 54.2  & 56.24  & 23.15 (22.88) & 0.0 (0.0) & 0.0 (0.0) & 1.78 (1.77) & 49.71  & 0.0 (0.0) \\
 & Gemini-2.5-Flash & 55.21  & 43.97  & 16.78 (14.84) & 19.58 (15.68) & 6.03 (5.16) & 1.41 (1.4) & 46.98  & 19.58 (15.68) \\
 & GPT-4.1 & 46.11  & 26.98  & 17.96 (17.96) & 0.0 (0.0) & 0.0 (0.0) & 1.11 (1.11) & 36.39  & 0.0 (0.0) \\
 & Pixtral-Large-Instruct-2411 & 76.22  & 14.12  & 0.0 (0.0) & N/A (N/A) & 0.0 (0.0) & 1.13 (1.13) & N/A  & N/A (N/A) \\
 & Llama-3.2-90B-Vision-Instruct & 52.41  & 16.98  & 0.0 (0.0) & N/A (N/A) & 0.0 (0.0) & 1.17 (1.17) & N/A  & N/A (N/A) \\
Overall & Qwen2.5-VL-72B-Instruct & 52.58  & 12.39  & 18.7 (18.7) & 22.52 (22.52) & 2.74 (2.74) & 1.08 (1.08) & 36.39  & 22.52 (22.52) \\
 & Pixtral 12B & 51.73  & 0.0  & N/A (N/A) & N/A (N/A) & 0.0 (0.0) & 1.0 (1.0) & N/A  & N/A (N/A) \\
 & Qwen2.5-VL-7B-Instruct & 36.39  & 0.0  & N/A (N/A) & N/A (N/A) & 0.0 (0.0) & 1.0 (1.0) & N/A  & N/A (N/A) \\
 & MedGemma 27B & 61.5  & 6.76  & 14.86 (14.86) & 0.0 (0.0) & 0.0 (0.0) & 1.0 (1.0) & 36.39  & 0.0 (0.0) \\
 & HealthGPT-L14 & 0.0  & N/A  & N/A (N/A) & N/A (N/A) & 0.0 (0.0) & 0.0 (0.0) & N/A  & N/A (N/A) \\
 & RadVLM & N/A  & N/A  & N/A (N/A) & N/A (N/A) & N/A (N/A) & N/A (N/A) & N/A  & N/A (N/A) \\
  & MedGemma 4B & 0.0  & N/A  & N/A (N/A) & N/A (N/A) & 0.0 (0.0) & 0.0 (0.0) & N/A  & N/A (N/A) \\
\midrule
 & Gemini-2.5-Pro & 54.2  & 56.24  & 23.15 (22.88) & 0.0 (0.0) & 0.0 (0.0) & 1.78 (1.77) & 49.71  & 0.0 (0.0) \\
 & Gemini-2.5-Flash & 60.19  & 46.59  & 16.08 (13.16) & 13.98 (8.78) & 3.02 (1.81) & 1.44 (1.41) & 46.98  & 13.98 (8.78) \\
 & GPT-4.1 & 46.11  & 26.98  & 17.96 (17.96) & 0.0 (0.0) & 0.0 (0.0) & 1.11 (1.11) & 36.39  & 0.0 (0.0) \\
 & Pixtral-Large-Instruct-2411 & 76.22  & 14.12  & 0.0 (0.0) & N/A (N/A) & 0.0 (0.0) & 1.13 (1.13) & N/A  & N/A (N/A) \\
 & Llama-3.2-90B-Vision-Instruct & 51.16  & 18.04  & 0.0 (0.0) & N/A (N/A) & 0.0 (0.0) & 1.13 (1.13) & N/A  & N/A (N/A) \\
Measurement & Qwen2.5-VL-72B-Instruct & 50.99  & 10.86  & 9.91 (9.91) & 0.0 (0.0) & 0.0 (0.0) & 1.02 (1.02) & 36.39  & 0.0 (0.0) \\
 & Pixtral 12B & 51.73  & 0.0  & N/A (N/A) & N/A (N/A) & 0.0 (0.0) & 1.0 (1.0) & N/A  & N/A (N/A) \\
 & Qwen2.5-VL-7B-Instruct & 36.39  & 0.0  & N/A (N/A) & N/A (N/A) & 0.0 (0.0) & 1.0 (1.0) & N/A  & N/A (N/A) \\
 & MedGemma 27B & 66.52  & 8.12  & 14.86 (14.86) & 0.0 (0.0) & 0.0 (0.0) & 1.0 (1.0) & 36.39  & 0.0 (0.0) \\
 & HealthGPT-L14 & 0.0  & N/A  & N/A (N/A) & N/A (N/A) & 0.0 (0.0) & 0.0 (0.0) & N/A  & N/A (N/A) \\
 & RadVLM & N/A  & N/A  & N/A (N/A) & N/A (N/A) & N/A (N/A) & N/A (N/A) & N/A  & N/A (N/A) \\
  & MedGemma 4B & 0.0  & N/A  & N/A (N/A) & N/A (N/A) & 0.0 (0.0) & 0.0 (0.0) & N/A  & N/A (N/A) \\

\midrule
 & Gemini-2.5-Pro & N/A  & N/A  & N/A  & N/A  & N/A  & N/A  & N/A  & N/A  \\
 & Gemini-2.5-Flash & 42.76  & 37.43  & 18.2  & 36.39  & 13.54  & 1.35  & N/A  & 36.39  \\
 & GPT-4.1 & N/A  & N/A  & N/A  & N/A  & N/A  & N/A  & N/A  & N/A  \\
 & Pixtral-Large-Instruct-2411 & N/A  & N/A  & N/A  & N/A  & N/A  & N/A  & N/A  & N/A  \\
 & Llama-3.2-90B-Vision-Instruct & 54.91  & 14.86  & 0.0  & N/A  & 0.0  & 1.25  & N/A  & N/A  \\
Recognition & Qwen2.5-VL-72B-Instruct & 63.66  & 23.14  & 45.05  & 45.05  & 21.91  & 1.5  & N/A  & 45.05  \\
 & Pixtral 12B & N/A  & N/A  & N/A  & N/A  & N/A  & N/A  & N/A  & N/A  \\
 & Qwen2.5-VL-7B-Instruct & N/A  & N/A  & N/A  & N/A  & N/A  & N/A  & N/A  & N/A  \\
 & MedGemma 27B & 36.39  & 0.0  & N/A  & N/A  & 0.0  & 1.0  & N/A  & N/A  \\
 & HealthGPT-L14 & N/A  & N/A  & N/A  & N/A  & N/A  & N/A  & N/A  & N/A  \\
 & RadVLM & N/A  & N/A  & N/A  & N/A  & N/A  & N/A  & N/A  & N/A  \\
  & MedGemma 4B & N/A  & N/A  & N/A  & N/A  & N/A  & N/A  & N/A  & N/A  \\

    \bottomrule
  \end{tabular}
  \end{adjustbox}
  }
\end{table}

%% file: Tables/3.Review/Result_Supple_Greedy_ALL_Recognition.tex
\begin{table}
  \caption{Re-evaluated Path 1 results with greedy sampling for recognition-type tasks.
  Stage-wise Score: Percentage of cases that completed each stage;
  Completion: Percentage of cases completing all reasoning stages; 
  Depth: Average number of reasoning stages reached;
  Refined scores incorporating measurement consistency are shown in parentheses.
  `N/A' indicates that the model did not reach the required stage to compute the corresponding metric.
  }
  \label{table:review_greedy_all_recognition}
  \centering
  {\begin{adjustbox}{width=\linewidth}
  \begin{tabular}{cccccccccc}
    \toprule
    & & \multicolumn{4}{c}{Stage-wise Score} \\
    \cmidrule(r){3-6}

    Task & Models & Stage 1 & Stage 2 & Stage 3 & Stage 4 & Completion & Depth & Consistency & Alignment \\
    
    \midrule

 & Gemini-2.5-Pro & N/A  & N/A  & N/A  & N/A  & N/A  & N/A  & N/A  & N/A  \\
 & Gemini-2.5-Flash & N/A  & N/A  & N/A  & N/A  & N/A  & N/A  & N/A  & N/A  \\
 & GPT-4.1 & N/A  & N/A  & N/A  & N/A  & N/A  & N/A  & N/A  & N/A  \\
 & Pixtral-Large-Instruct-2411 & N/A  & N/A  & N/A  & N/A  & N/A  & N/A  & N/A  & N/A  \\
 & Llama-3.2-90B-Vision-Instruct & 45.05  & 29.73  & 0.0  & N/A  & 0.0  & 1.5  & N/A  & N/A  \\
Trachea Deviation & Qwen2.5-VL-72B-Instruct & N/A  & N/A  & N/A  & N/A  & N/A  & N/A  & N/A  & N/A  \\
 & Pixtral 12B & N/A  & N/A  & N/A  & N/A  & N/A  & N/A  & N/A  & N/A  \\
 & Qwen2.5-VL-7B-Instruct & N/A  & N/A  & N/A  & N/A  & N/A  & N/A  & N/A  & N/A  \\
 & MedGemma 27B & N/A  & N/A  & N/A  & N/A  & N/A  & N/A  & N/A  & N/A  \\
 & HealthGPT-L14 & N/A  & N/A  & N/A  & N/A  & N/A  & N/A  & N/A  & N/A  \\
 & RadVLM & N/A  & N/A  & N/A  & N/A  & N/A  & N/A  & N/A  & N/A  \\
  & MedGemma 4B & N/A  & N/A  & N/A  & N/A  & N/A  & N/A  & N/A  & N/A  \\
\midrule
 & Gemini-2.5-Pro & N/A  & N/A  & N/A  & N/A  & N/A  & N/A  & N/A  & N/A  \\
 & Gemini-2.5-Flash & N/A  & N/A  & N/A  & N/A  & N/A  & N/A  & N/A  & N/A  \\
 & GPT-4.1 & N/A  & N/A  & N/A  & N/A  & N/A  & N/A  & N/A  & N/A  \\
 & Pixtral-Large-Instruct-2411 & N/A  & N/A  & N/A  & N/A  & N/A  & N/A  & N/A  & N/A  \\
 & Llama-3.2-90B-Vision-Instruct & N/A  & N/A  & N/A  & N/A  & N/A  & N/A  & N/A  & N/A  \\
Inclusion & Qwen2.5-VL-72B-Instruct & N/A  & N/A  & N/A  & N/A  & N/A  & N/A  & N/A  & N/A  \\
 & Pixtral 12B & N/A  & N/A  & N/A  & N/A  & N/A  & N/A  & N/A  & N/A  \\
 & Qwen2.5-VL-7B-Instruct & N/A  & N/A  & N/A  & N/A  & N/A  & N/A  & N/A  & N/A  \\
 & MedGemma 27B & N/A  & N/A  & N/A  & N/A  & N/A  & N/A  & N/A  & N/A  \\
 & HealthGPT-L14 & N/A  & N/A  & N/A  & N/A  & N/A  & N/A  & N/A  & N/A  \\
 & RadVLM & N/A  & N/A  & N/A  & N/A  & N/A  & N/A  & N/A  & N/A  \\
  & MedGemma 4B & N/A  & N/A  & N/A  & N/A  & N/A  & N/A  & N/A  & N/A  \\
\midrule
 & Gemini-2.5-Pro & N/A  & N/A  & N/A  & N/A  & N/A  & N/A  & N/A  & N/A  \\
 & Gemini-2.5-Flash & 58.44  & 38.47  & 0.0  & N/A  & 0.0  & 1.38  & N/A  & N/A  \\
 & GPT-4.1 & N/A  & N/A  & N/A  & N/A  & N/A  & N/A  & N/A  & N/A  \\
 & Pixtral-Large-Instruct-2411 & N/A  & N/A  & N/A  & N/A  & N/A  & N/A  & N/A  & N/A  \\
 & Llama-3.2-90B-Vision-Instruct & N/A  & N/A  & N/A  & N/A  & N/A  & N/A  & N/A  & N/A  \\
Inspiration & Qwen2.5-VL-72B-Instruct & N/A  & N/A  & N/A  & N/A  & N/A  & N/A  & N/A  & N/A  \\
 & Pixtral 12B & N/A  & N/A  & N/A  & N/A  & N/A  & N/A  & N/A  & N/A  \\
 & Qwen2.5-VL-7B-Instruct & N/A  & N/A  & N/A  & N/A  & N/A  & N/A  & N/A  & N/A  \\
 & MedGemma 27B & N/A  & N/A  & N/A  & N/A  & N/A  & N/A  & N/A  & N/A  \\
 & HealthGPT-L14 & N/A  & N/A  & N/A  & N/A  & N/A  & N/A  & N/A  & N/A  \\
 & RadVLM & N/A  & N/A  & N/A  & N/A  & N/A  & N/A  & N/A  & N/A  \\
  & MedGemma 4B & N/A  & N/A  & N/A  & N/A  & N/A  & N/A  & N/A  & N/A  \\
\midrule
 & Gemini-2.5-Pro & N/A  & N/A  & N/A  & N/A  & N/A  & N/A  & N/A  & N/A  \\
 & Gemini-2.5-Flash & 27.09  & 36.39  & 36.39  & 36.39  & 27.09  & 1.33  & N/A  & 36.39  \\
 & GPT-4.1 & N/A  & N/A  & N/A  & N/A  & N/A  & N/A  & N/A  & N/A  \\
 & Pixtral-Large-Instruct-2411 & N/A  & N/A  & N/A  & N/A  & N/A  & N/A  & N/A  & N/A  \\
 & Llama-3.2-90B-Vision-Instruct & 64.77  & 0.0  & N/A  & N/A  & 0.0  & 1.0  & N/A  & N/A  \\
Ascending Aorta Enlargement & Qwen2.5-VL-72B-Instruct & 63.66  & 23.14  & 45.05  & 45.05  & 21.91  & 1.5  & N/A  & 45.05  \\
 & Pixtral 12B & N/A  & N/A  & N/A  & N/A  & N/A  & N/A  & N/A  & N/A  \\
 & Qwen2.5-VL-7B-Instruct & N/A  & N/A  & N/A  & N/A  & N/A  & N/A  & N/A  & N/A  \\
 & MedGemma 27B & 36.39  & 0.0  & N/A  & N/A  & 0.0  & 1.0  & N/A  & N/A  \\
 & HealthGPT-L14 & N/A  & N/A  & N/A  & N/A  & N/A  & N/A  & N/A  & N/A  \\
 & RadVLM & N/A  & N/A  & N/A  & N/A  & N/A  & N/A  & N/A  & N/A  \\
  & MedGemma 4B & N/A  & N/A  & N/A  & N/A  & N/A  & N/A  & N/A  & N/A  \\
    \bottomrule
  \end{tabular}
  \end{adjustbox}
  }
\end{table}

%% file: Tables/3.Review/Result_Supple_Greedy_ALL_Arithmetic.tex
\begin{table}[htb!]
  \caption{Re-evaluated Path 1 results with greedy sampling for measurement-type tasks.
  Stage-wise Score: Percentage of cases that completed each stage;
  Completion: Percentage of cases completing all reasoning stages; 
  Depth: Average number of reasoning stages reached;
  Consistency: Percentage of cases where the value returned at Stage 4 matches the Stage 3 response;
  Refined scores incorporating measurement consistency are shown in parentheses.
  `N/A' indicates that the model did not reach the required stage to compute the corresponding metric.
  }
  \label{table:review_greedy_all_arithmetic}
  \centering
  {\begin{adjustbox}{width=\linewidth, height=0.4\textheight, keepaspectratio=true}
  \begin{tabular}{cccccccccc}
    \toprule
    & & \multicolumn{4}{c}{Stage-wise Score} \\
    \cmidrule(r){3-6}

    Task & Models & Stage 1 & Stage 2 & Stage 3 & Stage 4 & Completion & Depth & Consistency  & Alignment \\
    
    \midrule

  & Gemini-2.5-Pro & 42.24  & 51.73  & 27.09 (27.09) & 0.0 (0.0) & 0.0 (0.0) & 1.75 (1.75) & 36.39  & 0.0 (0.0) \\
 & Gemini-2.5-Flash & N/A  & N/A  & N/A (N/A) & N/A (N/A) & N/A (N/A) & N/A (N/A) & N/A  & N/A (N/A) \\
 & GPT-4.1 & 52.65  & 20.37  & 36.39 (36.39) & 0.0 (0.0) & 0.0 (0.0) & 1.0 (1.0) & 36.39  & 0.0 (0.0) \\
 & Pixtral-Large-Instruct-2411 & N/A  & N/A  & N/A (N/A) & N/A (N/A) & N/A (N/A) & N/A (N/A) & N/A  & N/A (N/A) \\
 & Llama-3.2-90B-Vision-Instruct & N/A  & N/A  & N/A (N/A) & N/A (N/A) & N/A (N/A) & N/A (N/A) & N/A  & N/A (N/A) \\
Cardiomegaly & Qwen2.5-VL-72B-Instruct & 55.25  & 27.82  & 0.0 (0.0) & N/A (N/A) & 0.0 (0.0) & 1.14 (1.14) & N/A  & N/A (N/A) \\
 & Pixtral 12B & N/A  & N/A  & N/A (N/A) & N/A (N/A) & N/A (N/A) & N/A (N/A) & N/A  & N/A (N/A) \\
 & Qwen2.5-VL-7B-Instruct & N/A  & N/A  & N/A (N/A) & N/A (N/A) & N/A (N/A) & N/A (N/A) & N/A  & N/A (N/A) \\
 & MedGemma 27B & N/A  & N/A  & N/A (N/A) & N/A (N/A) & N/A (N/A) & N/A (N/A) & N/A  & N/A (N/A) \\
 & HealthGPT-L14 & N/A  & N/A  & N/A (N/A) & N/A (N/A) & N/A (N/A) & N/A (N/A) & N/A  & N/A (N/A) \\
 & RadVLM & N/A  & N/A  & N/A (N/A) & N/A (N/A) & N/A (N/A) & N/A (N/A) & N/A  & N/A (N/A) \\
 & MedGemma 4B & N/A  & N/A  & N/A (N/A) & N/A (N/A) & N/A (N/A) & N/A (N/A) & N/A  & N/A (N/A) \\
\midrule
 & Gemini-2.5-Pro & N/A  & N/A  & N/A (N/A) & N/A (N/A) & N/A (N/A) & N/A (N/A) & N/A  & N/A (N/A) \\
 & Gemini-2.5-Flash & N/A  & N/A  & N/A (N/A) & N/A (N/A) & N/A (N/A) & N/A (N/A) & N/A  & N/A (N/A) \\
 & GPT-4.1 & 0.0  & N/A  & N/A (N/A) & N/A (N/A) & 0.0 (0.0) & 0.0 (0.0) & N/A  & N/A (N/A) \\
 & Pixtral-Large-Instruct-2411 & N/A  & N/A  & N/A (N/A) & N/A (N/A) & N/A (N/A) & N/A (N/A) & N/A  & N/A (N/A) \\
 & Llama-3.2-90B-Vision-Instruct & N/A  & N/A  & N/A (N/A) & N/A (N/A) & N/A (N/A) & N/A (N/A) & N/A  & N/A (N/A) \\
Carina Angle & Qwen2.5-VL-72B-Instruct & 64.77  & 27.82  & 29.73 (29.73) & 0.0 (0.0) & 0.0 (0.0) & 1.5 (1.5) & 36.39  & 0.0 (0.0) \\
 & Pixtral 12B & N/A  & N/A  & N/A (N/A) & N/A (N/A) & N/A (N/A) & N/A (N/A) & N/A  & N/A (N/A) \\
 & Qwen2.5-VL-7B-Instruct & N/A  & N/A  & N/A (N/A) & N/A (N/A) & N/A (N/A) & N/A (N/A) & N/A  & N/A (N/A) \\
 & MedGemma 27B & N/A  & N/A  & N/A (N/A) & N/A (N/A) & N/A (N/A) & N/A (N/A) & N/A  & N/A (N/A) \\
 & HealthGPT-L14 & N/A  & N/A  & N/A (N/A) & N/A (N/A) & N/A (N/A) & N/A (N/A) & N/A  & N/A (N/A) \\
 & RadVLM & N/A  & N/A  & N/A (N/A) & N/A (N/A) & N/A (N/A) & N/A (N/A) & N/A  & N/A (N/A) \\
 & MedGemma 4B & N/A  & N/A  & N/A (N/A) & N/A (N/A) & N/A (N/A) & N/A (N/A) & N/A  & N/A (N/A) \\
\midrule
 & Gemini-2.5-Pro & N/A  & N/A  & N/A (N/A) & N/A (N/A) & N/A (N/A) & N/A (N/A) & N/A  & N/A (N/A) \\
 & Gemini-2.5-Flash & N/A  & N/A  & N/A (N/A) & N/A (N/A) & N/A (N/A) & N/A (N/A) & N/A  & N/A (N/A) \\
 & GPT-4.1 & N/A  & N/A  & N/A (N/A) & N/A (N/A) & N/A (N/A) & N/A (N/A) & N/A  & N/A (N/A) \\
 & Pixtral-Large-Instruct-2411 & N/A  & N/A  & N/A (N/A) & N/A (N/A) & N/A (N/A) & N/A (N/A) & N/A  & N/A (N/A) \\
 & Llama-3.2-90B-Vision-Instruct & N/A  & N/A  & N/A (N/A) & N/A (N/A) & N/A (N/A) & N/A (N/A) & N/A  & N/A (N/A) \\
Rotation & Qwen2.5-VL-72B-Instruct & N/A  & N/A  & N/A (N/A) & N/A (N/A) & N/A (N/A) & N/A (N/A) & N/A  & N/A (N/A) \\
 & Pixtral 12B & N/A  & N/A  & N/A (N/A) & N/A (N/A) & N/A (N/A) & N/A (N/A) & N/A  & N/A (N/A) \\
 & Qwen2.5-VL-7B-Instruct & N/A  & N/A  & N/A (N/A) & N/A (N/A) & N/A (N/A) & N/A (N/A) & N/A  & N/A (N/A) \\
 & MedGemma 27B & N/A  & N/A  & N/A (N/A) & N/A (N/A) & N/A (N/A) & N/A (N/A) & N/A  & N/A (N/A) \\
 & HealthGPT-L14 & N/A  & N/A  & N/A (N/A) & N/A (N/A) & N/A (N/A) & N/A (N/A) & N/A  & N/A (N/A) \\
 & RadVLM & N/A  & N/A  & N/A (N/A) & N/A (N/A) & N/A (N/A) & N/A (N/A) & N/A  & N/A (N/A) \\
 & MedGemma 4B & N/A  & N/A  & N/A (N/A) & N/A (N/A) & N/A (N/A) & N/A (N/A) & N/A  & N/A (N/A) \\
\midrule
 & Gemini-2.5-Pro & 36.39  & 36.39  & 0.0 (0.0) & N/A (N/A) & 0.0 (0.0) & 2.0 (2.0) & N/A  & N/A (N/A) \\
 & Gemini-2.5-Flash & 46.0  & 29.9  & 0.0 (0.0) & N/A (N/A) & 0.0 (0.0) & 1.0 (1.0) & N/A  & N/A (N/A) \\
 & GPT-4.1 & 45.05  & 0.0  & N/A (N/A) & N/A (N/A) & 0.0 (0.0) & 1.0 (1.0) & N/A  & N/A (N/A) \\
 & Pixtral-Large-Instruct-2411 & N/A  & N/A  & N/A (N/A) & N/A (N/A) & N/A (N/A) & N/A (N/A) & N/A  & N/A (N/A) \\
 & Llama-3.2-90B-Vision-Instruct & 51.73  & 27.09  & 0.0 (0.0) & N/A (N/A) & 0.0 (0.0) & 1.33 (1.33) & N/A  & N/A (N/A) \\
Mediastinal Widening & Qwen2.5-VL-72B-Instruct & 36.39  & 0.0  & N/A (N/A) & N/A (N/A) & 0.0 (0.0) & 1.0 (1.0) & N/A  & N/A (N/A) \\
 & Pixtral 12B & N/A  & N/A  & N/A (N/A) & N/A (N/A) & N/A (N/A) & N/A (N/A) & N/A  & N/A (N/A) \\
 & Qwen2.5-VL-7B-Instruct & N/A  & N/A  & N/A (N/A) & N/A (N/A) & N/A (N/A) & N/A (N/A) & N/A  & N/A (N/A) \\
 & MedGemma 27B & 64.77  & 0.0  & N/A (N/A) & N/A (N/A) & 0.0 (0.0) & 1.0 (1.0) & N/A  & N/A (N/A) \\
 & HealthGPT-L14 & N/A  & N/A  & N/A (N/A) & N/A (N/A) & N/A (N/A) & N/A (N/A) & N/A  & N/A (N/A) \\
 & RadVLM & N/A  & N/A  & N/A (N/A) & N/A (N/A) & N/A (N/A) & N/A (N/A) & N/A  & N/A (N/A) \\
 & MedGemma 4B & N/A  & N/A  & N/A (N/A) & N/A (N/A) & N/A (N/A) & N/A (N/A) & N/A  & N/A (N/A) \\
\midrule
 & Gemini-2.5-Pro & 91.6  & 94.16  & 16.69 (15.02) & 0.0 (0.0) & 0.0 (0.0) & 2.13 (2.11) & 63.66  & 0.0 (0.0) \\
 & Gemini-2.5-Flash & 95.47  & 86.65  & 21.76 (20.48) & 12.2 (11.48) & 3.37 (3.17) & 2.16 (2.15) & 74.83  & 12.2 (11.48) \\
 & GPT-4.1 & 69.46  & 60.16  & 17.03 (17.03) & 0.0 (0.0) & 0.0 (0.0) & 1.77 (1.77) & 36.39  & 0.0 (0.0) \\
 & Pixtral-Large-Instruct-2411 & 80.11  & 28.25  & 0.0 (0.0) & N/A (N/A) & 0.0 (0.0) & 1.26 (1.26) & N/A  & N/A (N/A) \\
 & Llama-3.2-90B-Vision-Instruct & N/A  & N/A  & N/A (N/A) & N/A (N/A) & N/A (N/A) & N/A (N/A) & N/A  & N/A (N/A) \\
Projection & Qwen2.5-VL-72B-Instruct & 58.44  & 20.37  & 0.0 (0.0) & N/A (N/A) & 0.0 (0.0) & 1.0 (1.0) & N/A  & N/A (N/A) \\
 & Pixtral 12B & N/A  & N/A  & N/A (N/A) & N/A (N/A) & N/A (N/A) & N/A (N/A) & N/A  & N/A (N/A) \\
 & Qwen2.5-VL-7B-Instruct & N/A  & N/A  & N/A (N/A) & N/A (N/A) & N/A (N/A) & N/A (N/A) & N/A  & N/A (N/A) \\
 & MedGemma 27B & 51.57  & 23.32  & 0.0 (0.0) & N/A (N/A) & 0.0 (0.0) & 1.0 (1.0) & N/A  & N/A (N/A) \\
 & HealthGPT-L14 & N/A  & N/A  & N/A (N/A) & N/A (N/A) & N/A (N/A) & N/A (N/A) & N/A  & N/A (N/A) \\
 & RadVLM & N/A  & N/A  & N/A (N/A) & N/A (N/A) & N/A (N/A) & N/A (N/A) & N/A  & N/A (N/A) \\
 & MedGemma 4B & N/A  & N/A  & N/A (N/A) & N/A (N/A) & N/A (N/A) & N/A (N/A) & N/A  & N/A (N/A) \\
\midrule
 & Gemini-2.5-Pro & 49.52  & 42.02  & 32.5 (32.5) & 0.0 (0.0) & 0.0 (0.0) & 1.5 (1.5) & 45.05  & 0.0 (0.0) \\
 & Gemini-2.5-Flash & 34.38  & 0.0  & N/A (N/A) & N/A (N/A) & 0.0 (0.0) & 0.5 (0.5) & N/A  & N/A (N/A) \\
 & GPT-4.1 & 74.17  & 21.91  & 0.0 (0.0) & N/A (N/A) & 0.0 (0.0) & 1.2 (1.2) & N/A  & N/A (N/A) \\
 & Pixtral-Large-Instruct-2411 & N/A  & N/A  & N/A (N/A) & N/A (N/A) & N/A (N/A) & N/A (N/A) & N/A  & N/A (N/A) \\
 & Llama-3.2-90B-Vision-Instruct & 36.39  & 0.0  & N/A (N/A) & N/A (N/A) & 0.0 (0.0) & 1.0 (1.0) & N/A  & N/A (N/A) \\
Aortic Knob Enlargement & Qwen2.5-VL-72B-Instruct & 45.48  & 0.0  & N/A (N/A) & N/A (N/A) & 0.0 (0.0) & 0.67 (0.67) & N/A  & N/A (N/A) \\
 & Pixtral 12B & N/A  & N/A  & N/A (N/A) & N/A (N/A) & N/A (N/A) & N/A (N/A) & N/A  & N/A (N/A) \\
 & Qwen2.5-VL-7B-Instruct & 36.39  & 0.0  & N/A (N/A) & N/A (N/A) & 0.0 (0.0) & 1.0 (1.0) & N/A  & N/A (N/A) \\
 & MedGemma 27B & 74.83  & 0.0  & N/A (N/A) & N/A (N/A) & 0.0 (0.0) & 0.94 (0.94) & N/A  & N/A (N/A) \\
 & HealthGPT-L14 & 0.0  & N/A  & N/A (N/A) & N/A (N/A) & 0.0 (0.0) & 0.0 (0.0) & N/A  & N/A (N/A) \\
 & RadVLM & N/A  & N/A  & N/A (N/A) & N/A (N/A) & N/A (N/A) & N/A (N/A) & N/A  & N/A (N/A) \\
 & MedGemma 4B & N/A  & N/A  & N/A (N/A) & N/A (N/A) & N/A (N/A) & N/A (N/A) & N/A  & N/A (N/A) \\
\midrule
 & Gemini-2.5-Pro & 47.01  & 57.92  & 28.26 (28.26) & 0.0 (0.0) & 0.0 (0.0) & 1.28 (1.28) & 51.73  & 0.0 (0.0) \\
 & Gemini-2.5-Flash & 55.25  & 64.77  & 21.76 (21.76) & 0.0 (0.0) & 0.0 (0.0) & 1.86 (1.86) & 36.39  & 0.0 (0.0) \\
 & GPT-4.1 & 36.37  & 29.73  & 36.39 (36.39) & 0.0 (0.0) & 0.0 (0.0) & 1.33 (1.33) & 36.39  & 0.0 (0.0) \\
 & Pixtral-Large-Instruct-2411 & N/A  & N/A  & N/A (N/A) & N/A (N/A) & N/A (N/A) & N/A (N/A) & N/A  & N/A (N/A) \\
 & Llama-3.2-90B-Vision-Instruct & 61.27  & 23.32  & 0.0 (0.0) & N/A (N/A) & 0.0 (0.0) & 1.2 (1.2) & N/A  & N/A (N/A) \\
Descending Aorta Enlargement & Qwen2.5-VL-72B-Instruct & 51.57  & 0.0  & N/A (N/A) & N/A (N/A) & 0.0 (0.0) & 0.83 (0.83) & N/A  & N/A (N/A) \\
 & Pixtral 12B & N/A  & N/A  & N/A (N/A) & N/A (N/A) & N/A (N/A) & N/A (N/A) & N/A  & N/A (N/A) \\
 & Qwen2.5-VL-7B-Instruct & N/A  & N/A  & N/A (N/A) & N/A (N/A) & N/A (N/A) & N/A (N/A) & N/A  & N/A (N/A) \\
 & MedGemma 27B & 73.67  & 17.26  & 29.73 (29.73) & 0.0 (0.0) & 0.0 (0.0) & 1.12 (1.12) & 36.39  & 0.0 (0.0) \\
 & HealthGPT-L14 & N/A  & N/A  & N/A (N/A) & N/A (N/A) & N/A (N/A) & N/A (N/A) & N/A  & N/A (N/A) \\
 & RadVLM & N/A  & N/A  & N/A (N/A) & N/A (N/A) & N/A (N/A) & N/A (N/A) & N/A  & N/A (N/A) \\
 & MedGemma 4B & N/A  & N/A  & N/A (N/A) & N/A (N/A) & N/A (N/A) & N/A (N/A) & N/A  & N/A (N/A) \\
\midrule
 & Gemini-2.5-Pro & 58.44  & 55.25  & 34.38 (34.38) & 0.0 (0.0) & 0.0 (0.0) & 2.0 (2.0) & 51.73  & 0.0 (0.0) \\
 & Gemini-2.5-Flash & 69.87  & 51.63  & 20.8 (10.4) & 29.73 (14.86) & 11.72 (5.86) & 1.67 (1.56) & 29.73  & 29.73 (14.86) \\
 & GPT-4.1 & 45.05  & 29.73  & 0.0 (0.0) & N/A (N/A) & 0.0 (0.0) & 1.5 (1.5) & N/A  & N/A (N/A) \\
 & Pixtral-Large-Instruct-2411 & 72.32  & 0.0  & N/A (N/A) & N/A (N/A) & 0.0 (0.0) & 1.0 (1.0) & N/A  & N/A (N/A) \\
 & Llama-3.2-90B-Vision-Instruct & 55.25  & 21.76  & 0.0 (0.0) & N/A (N/A) & 0.0 (0.0) & 1.0 (1.0) & N/A  & N/A (N/A) \\
Descending Aorta Tortuous & Qwen2.5-VL-72B-Instruct & 45.05  & 0.0  & N/A (N/A) & N/A (N/A) & 0.0 (0.0) & 1.0 (1.0) & N/A  & N/A (N/A) \\
 & Pixtral 12B & 51.73  & 0.0  & N/A (N/A) & N/A (N/A) & 0.0 (0.0) & 1.0 (1.0) & N/A  & N/A (N/A) \\
 & Qwen2.5-VL-7B-Instruct & N/A  & N/A  & N/A (N/A) & N/A (N/A) & N/A (N/A) & N/A (N/A) & N/A  & N/A (N/A) \\
 & MedGemma 27B & 67.74  & 0.0  & N/A (N/A) & N/A (N/A) & 0.0 (0.0) & 0.92 (0.92) & N/A  & N/A (N/A) \\
 & HealthGPT-L14 & N/A  & N/A  & N/A (N/A) & N/A (N/A) & N/A (N/A) & N/A (N/A) & N/A  & N/A (N/A) \\
 & RadVLM & N/A  & N/A  & N/A (N/A) & N/A (N/A) & N/A (N/A) & N/A (N/A) & N/A  & N/A (N/A) \\
 & MedGemma 4B & 0.0  & N/A  & N/A (N/A) & N/A (N/A) & 0.0 (0.0) & 0.0 (0.0) & N/A  & N/A (N/A) \\
      
    \bottomrule
  \end{tabular}
  \end{adjustbox}
  }
\end{table}

%% file: Tables/3.Review/Result_Supple_Shot_ALL.tex
\begin{table}[htb!]
  \caption{Re-evaluated Path 1 results with stochastic sampling for overall, measurement-type, and recognition-type tasks.
  Stage-wise Score: Percentage of cases that completed each stage;
  Completion: Percentage of cases completing all reasoning stages; 
  Depth: Average number of reasoning stages reached;
  Consistency: Percentage of cases where the value returned at Stage 4 matches the Stage 3 response;
  Refined scores incorporating measurement consistency are shown in parentheses.
  `N/A' indicates that the model did not reach the required stage to compute the corresponding metric.
  }
  \label{table:review_shot_all}
  \centering
  {\begin{adjustbox}{width=\linewidth, height=0.5\textheight, keepaspectratio=true}
    \begin{tabular}{ccccccccc}
    \toprule
    & & \multicolumn{4}{c}{Stage-wise Score} \\
    \cmidrule(r){3-6}
    Task & Models & Stage 1 & Stage 2 & Stage 3 & Stage 4 & Completion & Depth & Consistency \\
    
    \midrule
  & Gemini-2.5-Pro & 70.86  & 67.96  & 27.55 (19.48) & 0.0 (0.0) & 0.0 (0.0) & 1.93 (1.86) & 49.82  \\
 & Gemini-2.5-Flash & 50.38  & 44.15  & 17.49 (13.12) & 0.0 (0.0) & 0.0 (0.0) & 1.24 (1.21) & 43.97  \\
 & GPT-4.1 & 51.92  & 30.32  & 14.04 (14.04) & 14.86 (14.86) & 2.72 (2.72) & 1.48 (1.48) & 48.39  \\
 & Pixtral-Large-Instruct-2411 & 55.06  & 36.5  & 12.72 (0.0) & 0.0 (0.0) & 0.0 (0.0) & 1.81 (1.66) & 0.0  \\
 & Llama-3.2-90B-Vision-Instruct & 14.56  & 0.0  & N/A (N/A) & N/A (N/A) & 0.0 (0.0) & 0.4 (0.4) & N/A  \\
Overall & Qwen2.5-VL-72B-Instruct & 38.97  & 8.94  & 0.0 (0.0) & N/A (N/A) & 0.0 (0.0) & 1.1 (1.1) & N/A  \\
 & Pixtral 12B & 36.39  & 0.0  & N/A (N/A) & N/A (N/A) & 0.0 (0.0) & 1.0 (1.0) & N/A  \\
 & Qwen2.5-VL-7B-Instruct & 30.01  & 0.0  & N/A (N/A) & N/A (N/A) & 0.0 (0.0) & 0.7 (0.7) & N/A  \\
 & HealthGPT-L14 & N/A  & N/A  & N/A (N/A) & N/A (N/A) & N/A (N/A) & N/A (N/A) & N/A  \\
 & RadVLM & 0.0  & N/A  & N/A (N/A) & N/A (N/A) & 0.0 (0.0) & 0.0 (0.0) & N/A  \\
\midrule
 & Gemini-2.5-Pro & 70.86  & 67.96  & 27.55 (19.48) & 0.0 (0.0) & 0.0 (0.0) & 1.93 (1.86) & 49.82  \\
 & Gemini-2.5-Flash & 58.92  & 45.77  & 15.57 (10.33) & 0.0 (0.0) & 0.0 (0.0) & 1.44 (1.41) & 43.97  \\
 & GPT-4.1 & 54.14  & 34.66  & 14.04 (14.04) & 14.86 (14.86) & 3.11 (3.11) & 1.55 (1.55) & 48.39  \\
 & Pixtral-Large-Instruct-2411 & 55.06  & 36.5  & 12.72 (0.0) & 0.0 (0.0) & 0.0 (0.0) & 1.81 (1.66) & 0.0  \\
 & Llama-3.2-90B-Vision-Instruct & 14.56  & 0.0  & N/A (N/A) & N/A (N/A) & 0.0 (0.0) & 0.4 (0.4) & N/A  \\
Measurement & Qwen2.5-VL-72B-Instruct & 40.36  & 14.91  & 0.0 (0.0) & N/A (N/A) & 0.0 (0.0) & 1.25 (1.25) & N/A  \\
 & Pixtral 12B & 36.39  & 0.0  & N/A (N/A) & N/A (N/A) & 0.0 (0.0) & 1.0 (1.0) & N/A  \\
 & Qwen2.5-VL-7B-Instruct & 41.83  & 0.0  & N/A (N/A) & N/A (N/A) & 0.0 (0.0) & 0.9 (0.9) & N/A  \\
 & HealthGPT-L14 & N/A  & N/A  & N/A (N/A) & N/A (N/A) & N/A (N/A) & N/A (N/A) & N/A  \\
 & RadVLM & 0.0  & N/A  & N/A (N/A) & N/A (N/A) & 0.0 (0.0) & 0.0 (0.0) & N/A  \\
\midrule
 & Gemini-2.5-Pro & N/A  & N/A  & N/A  & N/A  & N/A  & N/A  & N/A  \\
 & Gemini-2.5-Flash & 24.76  & 34.38  & 27.09  & 0.0  & 0.0  & 0.62  & N/A  \\
 & GPT-4.1 & 36.39  & 0.0  & N/A  & N/A  & 0.0  & 1.0  & N/A  \\
 & Pixtral-Large-Instruct-2411 & N/A  & N/A  & N/A  & N/A  & N/A  & N/A  & N/A  \\
 & Llama-3.2-90B-Vision-Instruct & N/A  & N/A  & N/A  & N/A  & N/A  & N/A  & N/A  \\
Recognition & Qwen2.5-VL-72B-Instruct & 36.89  & 0.0  & N/A  & N/A  & 0.0  & 0.88  & N/A  \\
 & Pixtral 12B & N/A  & N/A  & N/A  & N/A  & N/A  & N/A  & N/A  \\
 & Qwen2.5-VL-7B-Instruct & 18.2  & 0.0  & N/A  & N/A  & 0.0  & 0.5  & N/A  \\
 & HealthGPT-L14 & N/A  & N/A  & N/A  & N/A  & N/A  & N/A  & N/A  \\
 & RadVLM & N/A  & N/A  & N/A  & N/A  & N/A  & N/A  & N/A  \\
    \bottomrule
  \end{tabular}
  \end{adjustbox}
  }
\end{table}

%% file: Tables/3.Review/Result_Supple_Shot_ALL_Recognition.tex
\begin{table}[htb!]
  \caption{Re-evaluated Path 1 results with stochastic sampling for recognition-type tasks.
  Stage-wise Score: Percentage of cases that completed each stage;
  Completion: Percentage of cases completing all reasoning stages; 
  Depth: Average number of reasoning stages reached;
  Refined scores incorporating measurement consistency are shown in parentheses.
  `N/A' indicates that the model did not reach the required stage to compute the corresponding metric.
  }
  \label{table:review_shot_all_recognition}
  \centering
  {\begin{adjustbox}{width=\linewidth, height=0.5\textheight, keepaspectratio=true}
  \begin{tabular}{ccccccccc}
    \toprule
    & & \multicolumn{4}{c}{Stage-wise Score} \\
    \cmidrule(r){3-6}
    Task & Models & Stage 1 & Stage 2 & Stage 3 & Stage 4 & Completion & Depth & Consistency \\

    \midrule
  & Gemini-2.5-Pro & N/A  & N/A  & N/A  & N/A  & N/A  & N/A  & N/A  \\
 & Gemini-2.5-Flash & N/A  & N/A  & N/A  & N/A  & N/A  & N/A  & N/A  \\
 & GPT-4.1 & 36.39  & 0.0  & N/A  & N/A  & 0.0  & 1.0  & N/A  \\
 & Pixtral-Large-Instruct-2411 & N/A  & N/A  & N/A  & N/A  & N/A  & N/A  & N/A  \\
 & Llama-3.2-90B-Vision-Instruct & N/A  & N/A  & N/A  & N/A  & N/A  & N/A  & N/A  \\
Trachea Deviation & Qwen2.5-VL-72B-Instruct & 29.73  & 0.0  & N/A  & N/A  & 0.0  & 0.5  & N/A  \\
 & Pixtral 12B & N/A  & N/A  & N/A  & N/A  & N/A  & N/A  & N/A  \\
 & Qwen2.5-VL-7B-Instruct & N/A  & N/A  & N/A  & N/A  & N/A  & N/A  & N/A  \\
 & HealthGPT-L14 & N/A  & N/A  & N/A  & N/A  & N/A  & N/A  & N/A  \\
 & RadVLM & N/A  & N/A  & N/A  & N/A  & N/A  & N/A  & N/A  \\
\midrule
 & Gemini-2.5-Pro & N/A  & N/A  & N/A  & N/A  & N/A  & N/A  & N/A  \\
 & Gemini-2.5-Flash & N/A  & N/A  & N/A  & N/A  & N/A  & N/A  & N/A  \\
 & GPT-4.1 & N/A  & N/A  & N/A  & N/A  & N/A  & N/A  & N/A  \\
 & Pixtral-Large-Instruct-2411 & N/A  & N/A  & N/A  & N/A  & N/A  & N/A  & N/A  \\
 & Llama-3.2-90B-Vision-Instruct & N/A  & N/A  & N/A  & N/A  & N/A  & N/A  & N/A  \\
Inclusion & Qwen2.5-VL-72B-Instruct & 36.39  & 0.0  & N/A  & N/A  & 0.0  & 1.0  & N/A  \\
 & Pixtral 12B & N/A  & N/A  & N/A  & N/A  & N/A  & N/A  & N/A  \\
 & Qwen2.5-VL-7B-Instruct & 36.39  & 0.0  & N/A  & N/A  & 0.0  & 1.0  & N/A  \\
 & HealthGPT-L14 & N/A  & N/A  & N/A  & N/A  & N/A  & N/A  & N/A  \\
 & RadVLM & N/A  & N/A  & N/A  & N/A  & N/A  & N/A  & N/A  \\
\midrule
 & Gemini-2.5-Pro & N/A  & N/A  & N/A  & N/A  & N/A  & N/A  & N/A  \\
 & Gemini-2.5-Flash & 49.52  & 34.38  & 27.09  & 0.0  & 0.0  & 1.25  & N/A  \\
 & GPT-4.1 & N/A  & N/A  & N/A  & N/A  & N/A  & N/A  & N/A  \\
 & Pixtral-Large-Instruct-2411 & N/A  & N/A  & N/A  & N/A  & N/A  & N/A  & N/A  \\
 & Llama-3.2-90B-Vision-Instruct & N/A  & N/A  & N/A  & N/A  & N/A  & N/A  & N/A  \\
Inspiration & Qwen2.5-VL-72B-Instruct & 36.39  & 0.0  & N/A  & N/A  & 0.0  & 1.0  & N/A  \\
 & Pixtral 12B & N/A  & N/A  & N/A  & N/A  & N/A  & N/A  & N/A  \\
 & Qwen2.5-VL-7B-Instruct & N/A  & N/A  & N/A  & N/A  & N/A  & N/A  & N/A  \\
 & HealthGPT-L14 & N/A  & N/A  & N/A  & N/A  & N/A  & N/A  & N/A  \\
 & RadVLM & N/A  & N/A  & N/A  & N/A  & N/A  & N/A  & N/A  \\
\midrule
 & Gemini-2.5-Pro & N/A  & N/A  & N/A  & N/A  & N/A  & N/A  & N/A  \\
 & Gemini-2.5-Flash & 0.0  & N/A  & N/A  & N/A  & 0.0  & 0.0  & N/A  \\
 & GPT-4.1 & N/A  & N/A  & N/A  & N/A  & N/A  & N/A  & N/A  \\
 & Pixtral-Large-Instruct-2411 & N/A  & N/A  & N/A  & N/A  & N/A  & N/A  & N/A  \\
 & Llama-3.2-90B-Vision-Instruct & N/A  & N/A  & N/A  & N/A  & N/A  & N/A  & N/A  \\
Ascending Aorta Enlargement & Qwen2.5-VL-72B-Instruct & 45.05  & 0.0  & N/A  & N/A  & 0.0  & 1.0  & N/A  \\
 & Pixtral 12B & N/A  & N/A  & N/A  & N/A  & N/A  & N/A  & N/A  \\
 & Qwen2.5-VL-7B-Instruct & 0.0  & N/A  & N/A  & N/A  & 0.0  & 0.0  & N/A  \\
 & HealthGPT-L14 & N/A  & N/A  & N/A  & N/A  & N/A  & N/A  & N/A  \\
 & RadVLM & N/A  & N/A  & N/A  & N/A  & N/A  & N/A  & N/A  \\
    \bottomrule
  \end{tabular}
  \end{adjustbox}
  }
\end{table}

%% file: Tables/3.Review/Result_Supple_Shot_ALL_Arithmetic.tex
\begin{table}[htb!]
  \caption{Re-evaluated Path 1 results with stochastic sampling for measurement-type tasks.
  Stage-wise Score: Percentage of cases that completed each stage;
  Completion: Percentage of cases completing all reasoning stages; 
  Depth: Average number of reasoning stages reached;
  Consistency: Percentage of cases where the value returned at Stage 4 matches the Stage 3 response;
  Refined scores incorporating measurement consistency are shown in parentheses.
  `N/A' indicates that the model did not reach the required stage to compute the corresponding metric.
  }
  \label{table:review_shot_all_arithmetic}
  \centering
  {\begin{adjustbox}{width=\linewidth, height=0.5\textheight, keepaspectratio=true}
    \begin{tabular}{ccccccccc}
    \toprule
    & & \multicolumn{4}{c}{Stage-wise Score} \\
    \cmidrule(r){3-6}

    Task & Models & Stage 1 & Stage 2 & Stage 3 & Stage 4 & Completion & Depth & Consistency \\

    \midrule
  & Gemini-2.5-Pro & N/A  & N/A  & N/A (N/A) & N/A (N/A) & N/A (N/A) & N/A (N/A) & N/A  \\
 & Gemini-2.5-Flash & N/A  & N/A  & N/A (N/A) & N/A (N/A) & N/A (N/A) & N/A (N/A) & N/A  \\
 & GPT-4.1 & 46.0  & 0.0  & N/A (N/A) & N/A (N/A) & 0.0 (0.0) & 0.71 (0.71) & N/A  \\
 & Pixtral-Large-Instruct-2411 & N/A  & N/A  & N/A (N/A) & N/A (N/A) & N/A (N/A) & N/A (N/A) & N/A  \\
 & Llama-3.2-90B-Vision-Instruct & 0.0  & N/A  & N/A (N/A) & N/A (N/A) & 0.0 (0.0) & 0.0 (0.0) & N/A  \\
Cardiomegaly & Qwen2.5-VL-72B-Instruct & 51.57  & 23.32  & 0.0 (0.0) & N/A (N/A) & 0.0 (0.0) & 1.0 (1.0) & N/A  \\
 & Pixtral 12B & N/A  & N/A  & N/A (N/A) & N/A (N/A) & N/A (N/A) & N/A (N/A) & N/A  \\
 & Qwen2.5-VL-7B-Instruct & 47.27  & 0.0  & N/A (N/A) & N/A (N/A) & 0.0 (0.0) & 0.8 (0.8) & N/A  \\
 & HealthGPT-L14 & N/A  & N/A  & N/A (N/A) & N/A (N/A) & N/A (N/A) & N/A (N/A) & N/A  \\
 & RadVLM & N/A  & N/A  & N/A (N/A) & N/A (N/A) & N/A (N/A) & N/A (N/A) & N/A  \\
\midrule
 & Gemini-2.5-Pro & N/A  & N/A  & N/A (N/A) & N/A (N/A) & N/A (N/A) & N/A (N/A) & N/A  \\
 & Gemini-2.5-Flash & 36.39  & 0.0  & N/A (N/A) & N/A (N/A) & 0.0 (0.0) & 1.0 (1.0) & N/A  \\
 & GPT-4.1 & 36.37  & 45.05  & 0.0 (0.0) & N/A (N/A) & 0.0 (0.0) & 1.33 (1.33) & N/A  \\
 & Pixtral-Large-Instruct-2411 & N/A  & N/A  & N/A (N/A) & N/A (N/A) & N/A (N/A) & N/A (N/A) & N/A  \\
 & Llama-3.2-90B-Vision-Instruct & 0.0  & N/A  & N/A (N/A) & N/A (N/A) & 0.0 (0.0) & 0.0 (0.0) & N/A  \\
Carina Angle & Qwen2.5-VL-72B-Instruct & 36.39  & 0.0  & N/A (N/A) & N/A (N/A) & 0.0 (0.0) & 1.0 (1.0) & N/A  \\
 & Pixtral 12B & N/A  & N/A  & N/A (N/A) & N/A (N/A) & N/A (N/A) & N/A (N/A) & N/A  \\
 & Qwen2.5-VL-7B-Instruct & N/A  & N/A  & N/A (N/A) & N/A (N/A) & N/A (N/A) & N/A (N/A) & N/A  \\
 & HealthGPT-L14 & N/A  & N/A  & N/A (N/A) & N/A (N/A) & N/A (N/A) & N/A (N/A) & N/A  \\
 & RadVLM & N/A  & N/A  & N/A (N/A) & N/A (N/A) & N/A (N/A) & N/A (N/A) & N/A  \\
\midrule
 & Gemini-2.5-Pro & N/A  & N/A  & N/A (N/A) & N/A (N/A) & N/A (N/A) & N/A (N/A) & N/A  \\
 & Gemini-2.5-Flash & N/A  & N/A  & N/A (N/A) & N/A (N/A) & N/A (N/A) & N/A (N/A) & N/A  \\
 & GPT-4.1 & N/A  & N/A  & N/A (N/A) & N/A (N/A) & N/A (N/A) & N/A (N/A) & N/A  \\
 & Pixtral-Large-Instruct-2411 & N/A  & N/A  & N/A (N/A) & N/A (N/A) & N/A (N/A) & N/A (N/A) & N/A  \\
 & Llama-3.2-90B-Vision-Instruct & N/A  & N/A  & N/A (N/A) & N/A (N/A) & N/A (N/A) & N/A (N/A) & N/A  \\
Rotation & Qwen2.5-VL-72B-Instruct & 45.05  & 29.73  & 0.0 (0.0) & N/A (N/A) & 0.0 (0.0) & 1.5 (1.5) & N/A  \\
 & Pixtral 12B & N/A  & N/A  & N/A (N/A) & N/A (N/A) & N/A (N/A) & N/A (N/A) & N/A  \\
 & Qwen2.5-VL-7B-Instruct & N/A  & N/A  & N/A (N/A) & N/A (N/A) & N/A (N/A) & N/A (N/A) & N/A  \\
 & HealthGPT-L14 & N/A  & N/A  & N/A (N/A) & N/A (N/A) & N/A (N/A) & N/A (N/A) & N/A  \\
 & RadVLM & 0.0  & N/A  & N/A (N/A) & N/A (N/A) & 0.0 (0.0) & 0.0 (0.0) & N/A  \\
\midrule
 & Gemini-2.5-Pro & N/A  & N/A  & N/A (N/A) & N/A (N/A) & N/A (N/A) & N/A (N/A) & N/A  \\
 & Gemini-2.5-Flash & 63.66  & 39.22  & 37.52 (25.01) & 0.0 (0.0) & 0.0 (0.0) & 1.7 (1.6) & 36.37  \\
 & GPT-4.1 & 36.37  & 29.73  & 0.0 (0.0) & N/A (N/A) & 0.0 (0.0) & 1.0 (1.0) & N/A  \\
 & Pixtral-Large-Instruct-2411 & N/A  & N/A  & N/A (N/A) & N/A (N/A) & N/A (N/A) & N/A (N/A) & N/A  \\
 & Llama-3.2-90B-Vision-Instruct & 0.0  & N/A  & N/A (N/A) & N/A (N/A) & 0.0 (0.0) & 0.0 (0.0) & N/A  \\
Mediastinal Widening & Qwen2.5-VL-72B-Instruct & 36.39  & 0.0  & N/A (N/A) & N/A (N/A) & 0.0 (0.0) & 1.0 (1.0) & N/A  \\
 & Pixtral 12B & N/A  & N/A  & N/A (N/A) & N/A (N/A) & N/A (N/A) & N/A (N/A) & N/A  \\
 & Qwen2.5-VL-7B-Instruct & 36.39  & 0.0  & N/A (N/A) & N/A (N/A) & 0.0 (0.0) & 1.0 (1.0) & N/A  \\
 & HealthGPT-L14 & N/A  & N/A  & N/A (N/A) & N/A (N/A) & N/A (N/A) & N/A (N/A) & N/A  \\
 & RadVLM & N/A  & N/A  & N/A (N/A) & N/A (N/A) & N/A (N/A) & N/A (N/A) & N/A  \\
\midrule
 & Gemini-2.5-Pro & 94.65  & 92.18  & 16.85 (16.85) & 0.0 (0.0) & 0.0 (0.0) & 2.15 (2.15) & 75.79  \\
 & Gemini-2.5-Flash & 95.19  & 87.48  & 15.82 (14.38) & 0.0 (0.0) & 0.0 (0.0) & 2.09 (2.08) & 65.82  \\
 & GPT-4.1 & 77.22  & 53.66  & 0.0 (0.0) & N/A (N/A) & 0.0 (0.0) & 1.75 (1.75) & N/A  \\
 & Pixtral-Large-Instruct-2411 & 81.81  & 39.5  & 21.14 (0.0) & 0.0 (0.0) & 0.0 (0.0) & 1.47 (1.38) & 0.0  \\
 & Llama-3.2-90B-Vision-Instruct & N/A  & N/A  & N/A (N/A) & N/A (N/A) & N/A (N/A) & N/A (N/A) & N/A  \\
Projection & Qwen2.5-VL-72B-Instruct & N/A  & N/A  & N/A (N/A) & N/A (N/A) & N/A (N/A) & N/A (N/A) & N/A  \\
 & Pixtral 12B & N/A  & N/A  & N/A (N/A) & N/A (N/A) & N/A (N/A) & N/A (N/A) & N/A  \\
 & Qwen2.5-VL-7B-Instruct & N/A  & N/A  & N/A (N/A) & N/A (N/A) & N/A (N/A) & N/A (N/A) & N/A  \\
 & HealthGPT-L14 & N/A  & N/A  & N/A (N/A) & N/A (N/A) & N/A (N/A) & N/A (N/A) & N/A  \\
 & RadVLM & N/A  & N/A  & N/A (N/A) & N/A (N/A) & N/A (N/A) & N/A (N/A) & N/A  \\
\midrule
 & Gemini-2.5-Pro & 72.32  & 52.65  & 26.06 (13.03) & 0.0 (0.0) & 0.0 (0.0) & 2.0 (1.89) & 29.73  \\
 & Gemini-2.5-Flash & 37.69  & 64.77  & 0.0 (0.0) & N/A (N/A) & 0.0 (0.0) & 1.0 (1.0) & N/A  \\
 & GPT-4.1 & 61.27  & 29.9  & 0.0 (0.0) & N/A (N/A) & 0.0 (0.0) & 1.4 (1.4) & N/A  \\
 & Pixtral-Large-Instruct-2411 & 45.05  & 45.05  & 29.73 (0.0) & 0.0 (0.0) & 0.0 (0.0) & 2.5 (2.0) & 0.0  \\
 & Llama-3.2-90B-Vision-Instruct & N/A  & N/A  & N/A (N/A) & N/A (N/A) & N/A (N/A) & N/A (N/A) & N/A  \\
Aortic Knob Enlargement & Qwen2.5-VL-72B-Instruct & 36.39  & 0.0  & N/A (N/A) & N/A (N/A) & 0.0 (0.0) & 1.0 (1.0) & N/A  \\
 & Pixtral 12B & N/A  & N/A  & N/A (N/A) & N/A (N/A) & N/A (N/A) & N/A (N/A) & N/A  \\
 & Qwen2.5-VL-7B-Instruct & N/A  & N/A  & N/A (N/A) & N/A (N/A) & N/A (N/A) & N/A (N/A) & N/A  \\
 & HealthGPT-L14 & N/A  & N/A  & N/A (N/A) & N/A (N/A) & N/A (N/A) & N/A (N/A) & N/A  \\
 & RadVLM & N/A  & N/A  & N/A (N/A) & N/A (N/A) & N/A (N/A) & N/A (N/A) & N/A  \\
\midrule
 & Gemini-2.5-Pro & 61.02  & 68.58  & 28.81 (28.81) & 0.0 (0.0) & 0.0 (0.0) & 1.68 (1.68) & 61.27  \\
 & Gemini-2.5-Flash & 58.44  & 31.97  & 0.0 (0.0) & N/A (N/A) & 0.0 (0.0) & 1.25 (1.25) & N/A  \\
 & GPT-4.1 & 57.01  & 42.24  & 51.73 (51.73) & 0.0 (0.0) & 0.0 (0.0) & 2.5 (2.5) & 51.73  \\
 & Pixtral-Large-Instruct-2411 & 36.39  & 36.39  & 0.0 (0.0) & N/A (N/A) & 0.0 (0.0) & 2.0 (2.0) & N/A  \\
 & Llama-3.2-90B-Vision-Instruct & 36.39  & 0.0  & N/A (N/A) & N/A (N/A) & 0.0 (0.0) & 1.0 (1.0) & N/A  \\
Descending Aorta Enlargement & Qwen2.5-VL-72B-Instruct & N/A  & N/A  & N/A (N/A) & N/A (N/A) & N/A (N/A) & N/A (N/A) & N/A  \\
 & Pixtral 12B & 36.39  & 0.0  & N/A (N/A) & N/A (N/A) & 0.0 (0.0) & 1.0 (1.0) & N/A  \\
 & Qwen2.5-VL-7B-Instruct & N/A  & N/A  & N/A (N/A) & N/A (N/A) & N/A (N/A) & N/A (N/A) & N/A  \\
 & HealthGPT-L14 & N/A  & N/A  & N/A (N/A) & N/A (N/A) & N/A (N/A) & N/A (N/A) & N/A  \\
 & RadVLM & N/A  & N/A  & N/A (N/A) & N/A (N/A) & N/A (N/A) & N/A (N/A) & N/A  \\
\midrule
 & Gemini-2.5-Pro & 55.43  & 58.44  & 38.47 (19.23) & 0.0 (0.0) & 0.0 (0.0) & 1.9 (1.7) & 32.5  \\
 & Gemini-2.5-Flash & 62.18  & 51.2  & 24.51 (12.26) & 0.0 (0.0) & 0.0 (0.0) & 1.62 (1.54) & 29.73  \\
 & GPT-4.1 & 64.77  & 42.02  & 32.5 (32.5) & 29.73 (29.73) & 21.76 (21.76) & 2.17 (2.17) & 45.05  \\
 & Pixtral-Large-Instruct-2411 & 57.01  & 25.07  & 0.0 (0.0) & N/A (N/A) & 0.0 (0.0) & 1.25 (1.25) & N/A  \\
 & Llama-3.2-90B-Vision-Instruct & 36.39  & 0.0  & N/A (N/A) & N/A (N/A) & 0.0 (0.0) & 1.0 (1.0) & N/A  \\
Descending Aorta Tortuous & Qwen2.5-VL-72B-Instruct & 36.39  & 36.39  & 0.0 (0.0) & N/A (N/A) & 0.0 (0.0) & 2.0 (2.0) & N/A  \\
 & Pixtral 12B & N/A  & N/A  & N/A (N/A) & N/A (N/A) & N/A (N/A) & N/A (N/A) & N/A  \\
 & Qwen2.5-VL-7B-Instruct & N/A  & N/A  & N/A (N/A) & N/A (N/A) & N/A (N/A) & N/A (N/A) & N/A  \\
 & HealthGPT-L14 & N/A  & N/A  & N/A (N/A) & N/A (N/A) & N/A (N/A) & N/A (N/A) & N/A  \\
 & RadVLM & 0.0  & N/A  & N/A (N/A) & N/A (N/A) & 0.0 (0.0) & 0.0 (0.0) & N/A  \\
    \bottomrule
  \end{tabular}
  \end{adjustbox}
  }
\end{table}

%% file: Tables/1.Reasoning/Result_Supple_3runs_ALL.tex
\begin{table}[htb!]
  \caption{Path 1 evaluation results mean and standard deviation over three seeds for overall tasks.
  Completion: Percentage of cases completing all reasoning stages; 
  Depth: Average number of reasoning stages reached;
  Consistency: Percentage of cases where the value returned at Stage 4 matches the Stage 3 response;
  Alignment: Percentage of agreement between initial and final decisions;
  `N/A' indicates that the model did not reach the required stage to compute the corresponding metric.
  }
  \label{table:reasoning_3run_overall}
  \centering
  {\begin{adjustbox}{width=0.9\linewidth}
  \begin{tabular}{ccccc}
    \toprule
    \textbf{Models} & \textbf{Completion (0-100) $\uparrow$} & \textbf{Depth (0-4) $\uparrow$} & \textbf{Consistency (0-100) $\uparrow$} & \textbf{Alignment (0-100)$\uparrow$} \\
    \midrule

 Gemini-2.5-Pro  & 17.32($\pm$1.38) & 1.85($\pm$0.04) & 68.14($\pm$6.13) & 61.72($\pm$2.25) \\
Gemini-2.5-Flash  & 12.07($\pm$0.73) & 1.4($\pm$0.02) & 42.6($\pm$4.18) & 47.98($\pm$3.74) \\
GPT-4.1  & 8.62($\pm$0.32)&	1.11($\pm$0.02)&59.55($\pm$0.61)&39.26($\pm$1.18) \\
\midrule
Pixtral-Large-Instruct-2411  & 4.55($\pm$1.08)&1.06($\pm$0.03)&39.97($\pm$2.85)&	39.07($\pm$4.82)\\
Llama-3.2-90B-Vision-Instruct  & 1.86($\pm$0.89) &0.59($\pm$0.04) &24.7($\pm$4.09) &24.7($\pm$9.22) \\
Qwen2.5-VL-72B-Instruct  & 1.51($\pm$0.36) &0.49($\pm$0.04) & N/A & 13.26($\pm$11.63) \\
Pixtral 12B  & 0.23($\pm$0.4) & 0.34($\pm$0.03) & 18.2($\pm$0) & 4.04($\pm$7) \\
Qwen2.5-VL-7B-Instruct  & 0.94($\pm$1.18) & 0.37($\pm$0.05) & 12.13($\pm$N/A) & 26.34($\pm$14.21) \\
HealthGPT-L14  & 1.27($\pm$0.14) & 0.33($\pm$0.01) & 33.21($\pm$13.7) & 31.45($\pm$2.06) \\
RadVLM  & 0.29($\pm$0.09) & 0.22($\pm$0.01) & 36.39($\pm$N/A) & 35.44($\pm$16.79) \\
 
    \bottomrule
  \end{tabular}
  \end{adjustbox}
  }
\end{table}

%% file: Tables/2.Guidance/Result_Supple_3runs_ALL.tex
\begin{table}[htb!]
  \caption{Path 2 evaluation results mean and standard deviation over three seeds for overall tasks.
  Completion: Percentage of cases completing all reasoning stages; 
  Depth: Average number of reasoning stages reached;
  Consistency: Percentage of cases where the value returned at Stage 3 matches the Stage 2 response;
  `N/A' indicates that the model did not reach the required stage to compute the corresponding metric.
  }
  \label{table:guidance_3runs_overall}
  \centering
  {\begin{adjustbox}{width=0.9\linewidth}
  \begin{tabular}{cccc}
    \toprule
    \textbf{Models} & \textbf{Completion (0-100) $\uparrow$} & \textbf{Depth (0-3) $\uparrow$} & \textbf{Consistency (0-100) $\uparrow$} \\
    \midrule
Gemini-2.5-Pro&49.31($\pm$3.39)&2.11($\pm$0.2)&72.68($\pm$1.43)\\
Gemini-2.5-Flash&39.83($\pm$1.14)&1.84($\pm$0.04)&62.53($\pm$2.37)\\
GPT-4.1&25.35($\pm$0.84)&1.21($\pm$0.1)&70.37($\pm$1.51)\\
Pixtral-Large&28.85($\pm$3.83)&1.55($\pm$0.15)&61.63($\pm$6.68)\\
Llama-3.2-90B-Vision&7.02($\pm$0.72)&0.24($\pm$0.02)&25.35($\pm$9.93)\\
Qwen2.5-VL-72B&11.33($\pm$0.24)&0.67($\pm$0.01)&36.93($\pm$2.31)\\
Pixtral12B&9.2($\pm$2.68)&0.52($\pm$0.09)&34.1($\pm$9.54)\\
Qwen2.5-VL-7B&5.87($\pm$0.27)&0.41($\pm$0.02)&27.47($\pm$N/A)\\
HealthGPT-L14&1.38($\pm$0.57)&0.1($\pm$0.03)&20.79($\pm$13.19)\\
RadVLM&0.67($\pm$1.16)&0.07($\pm$0.04)&27.09($\pm$N/A)\\

    \bottomrule
  \end{tabular}
  \end{adjustbox}
  }
\end{table}

%% file: Tables/3.Review/Result_Supple_3runs_ALL.tex
\begin{table}[htb!]
  \caption{Re-evaluated Path 1 evaluation results mean and standard deviation over three seeds for overall tasks.
  Completion: Percentage of cases completing all reasoning stages; 
  Depth: Average number of reasoning stages reached;
  Consistency: Percentage of cases where the value returned at Stage 4 matches the Stage 3 response;
  Alignment: Percentage of agreement between initial and final decisions;
  `N/A' indicates that the model did not reach the required stage to compute the corresponding metric.
  }
  \label{table:review_3runs_overall}
  \centering
  {\begin{adjustbox}{width=0.9\linewidth}
  \begin{tabular}{ccccc}
    \toprule
    \textbf{Models} & \textbf{Completion (0-100) $\uparrow$} & \textbf{Depth (0-4) $\uparrow$} & \textbf{Consistency (0-100) $\uparrow$} & \textbf{Alignment (0-100)$\uparrow$} \\
    \midrule
Gemini-2.5-Pro&6.28($\pm$1.8)&1.77($\pm$0.19)&53.35($\pm$2.99)&13.04($\pm$1.96)\\
Gemini-2.5-Flash&2.21($\pm$1.94)&1.37($\pm$0.13)&27.8($\pm$13.57)&15.57($\pm$9.15)\\
GPT-4.1&5.34($\pm$0.66)&1.3($\pm$0.1)&38.83($\pm$3.27)&21.53($\pm$12.06)\\
Pixtral-Large&0($\pm$0)&1.31($\pm$0.28)&33.21($\pm$13.7)&0($\pm$0)\\
Llama-3.2-90B-Vision&1.11($\pm$1.92)&0.92($\pm$0.12)&18.2($\pm$25.73)&18.2($\pm$25.73)\\
Qwen2.5-VL-72B&0.8($\pm$1.38)&1.25($\pm$0.15)&0($\pm$-)&9.91($\pm$-)\\
Pixtral12B&0($\pm$0)&0.44($\pm$0.12)&-($\pm$-)&-($\pm$-)\\
Qwen2.5-VL-7B&0($\pm$0)&0.61($\pm$0.24)&-($\pm$-)&-($\pm$-)\\
HealthGPT-L14&0($\pm$-)&0($\pm$-)&-($\pm$-)&-($\pm$-)\\
RadVLM&0($\pm$-)&0($\pm$-)&-($\pm$-)&-($\pm$-)\\

    \bottomrule
  \end{tabular}
  \end{adjustbox}
  }
\end{table}